\documentclass[runningheads]{llncs}

\usepackage{eccv}

\usepackage{eccvabbrv}

\usepackage{comment}
\usepackage{amsmath,amssymb}
\usepackage{color}
\usepackage{url}
\usepackage{booktabs}
\usepackage{multirow}
\usepackage{makecell}
\usepackage{wrapfig}
\usepackage{tabularx}
\usepackage{subcaption}
\usepackage{graphicx}
\usepackage{xcolor}
\usepackage{placeins}
\usepackage{marvosym}

\usepackage[accsupp]{axessibility}  

\usepackage{hyperref}

\usepackage{orcidlink}

\usepackage{hyperref}
\usepackage{tcolorbox}
\tcbuselibrary{skins}

\newcommand{\corrauth}{\textsuperscript{\Letter}}

\begin{document}

\title{X-MULTI: VLM-based Imaging Factor Disentanglement for Factor-Aware Image Synthesis}

\titlerunning{\textbf{X-MULTI}: VLM-based Imaging Factor Disentanglement}

\author{
Sonali Godavarthy\inst{1}\textsuperscript{*}
\and
Matthias Neuwirth-Trapp\inst{2,3}\corrauth
\and\\
Tim-Felix Faasch\inst{3}
\and
Maarten Bieshaar\inst{3}
\and
Michael Moeller\inst{1}
\and\\
Kristof Van Laerhoven\inst{1}
\and
Danda Pani Paudel\inst{4}
}

\authorrunning{S. Godavarthy et al.}

\institute{
University of Siegen, Siegen, Germany \and
ETH Zürich, Zürich, Switzerland \and
Bosch Research, Hildesheim, Germany \and
INSAIT, Sofia University “St.~Kliment Ohridski”, Sofia, Bulgaria}

\maketitle

\begingroup
\renewcommand\thefootnote{}
\footnotetext{\textsuperscript{*}Work conducted while at Bosch Research.\\
\textsuperscript{\Letter}Corresponding author: 
\href{mailto:mneuwirth@ethz.ch}{mneuwirth@ethz.ch}}
\endgroup


\begin{abstract}
Imaging factor disentanglement in text-to-image generation aims to independently control image acquisition properties such as types of camera lenses, sensor types, viewpoints, and domains to enable combinatorial generalization. This should let the model synthesize novel factor combinations unobserved in the training data, such as pairing a fisheye lens with an event sensor never observed in training data. 
Recent work, MULTI, introduced learnable, factor-specific embeddings to disentangle imaging factors, along with the Factor Alignment Accuracy (FAA) metric to evaluate disentanglement quality. We identify and address two independent limitations.
First, MULTI's pixel-level reconstruction objective supervises the model only on observed imaging factor combinations, providing no direct training signal for novel combinations. We therefore propose X-MULTI, which uses a pretrained vision-language model (VLM) to supervise novel factor combinations synthesized during training.
Second, we show the FAA metric exhibits severe cross-factor correlation leakage, misrepresenting true disentanglement quality. We therefore propose Improved-FAA (I-FAA), which employs factor-specific augmentation strategies to break these correlations and enables more rigorous evaluation. 
Experiments demonstrate that X-MULTI achieves improved factor alignment on novel combinations compared to MULTI. Moreover, we show that correlation leakage in FAA distorts the evaluation of true factor disentanglement and I-FAA reduces this leakage and therefore provides a more robust assessment of factor alignment.

\end{abstract}

\section{Introduction} \label{sec:intro}


Text-to-image (T2I) diffusion models have achieved remarkable visual fidelity in synthesizing images from natural language 
descriptions~\cite{esser2024scaling,nichol2022glide,podell2023sdxl,rombach2022high,saharia2022photorealistic}. However, precise control over low-level image acquisition properties such as types of camera lens, sensor type, viewpoint, and domain remains challenging~\cite{godavarthy2026multi}. These factors define different aspects of the image formation process, but in real datasets they often appear only in limited and correlated combinations. Therefore, a model can be considered factor-disentangled only if it can modify each factor independently and synthesize valid novel combinations that were not observed during training. This problem, termed \textit{imaging factor disentanglement}, requires learning representations that capture the semantic and physical identity of each factor while allowing them to be modified independently~\cite{godavarthy2026multi} to generate images with novel factor combinations that are not present during training. 
Controlling these imaging factors is important because collecting real data for all acquisition conditions is expensive, while vision systems still need to remain robust across sensor types, lenses, viewpoints, and domains.

Recent work, MULTI~\cite{godavarthy2026multi}, introduced learnable factor-specific text embeddings optimized using Textual Inversion (TI)~\cite{gal2022image}. These embeddings are inserted into the text prompt to represent imaging factors, and are optimized with the standard diffusion reconstruction loss while keeping the T2I diffusion model frozen.
However, MULTI's reconstruction-based training objective is applied only for factor combinations that are represented in the training data, so the training does not receive any signal for novel, unseen combinations. This leaves the model unable to reliably distinguish semantically distinct factors (e.g., \texttt{thermal sensor} from \texttt{rgb sensor}), resulting in factor confusion at inference time. To encourage disentanglement and thereby allow the synthesis of images for novel factor combinations, a supervisory signal is required that can provide meaningful signal during training. We therefore propose \textbf{X-MULTI} (eXtended MULTI) which introduces supervision from a zero-shot Vision-Language Model (VLM)~\cite{bai2025qwen3,luo2025dual}. During training, images with novel factor combinations are generated and the VLM acts as external semantic classifier, independently predicting factors for these generated images and thereby providing factor-level supervisory signals.

To evaluate whether generated images contain the specified factors, authors of MULTI~\cite{godavarthy2026multi} proposed Factor Alignment Accuracy (FAA), a novel classifier-based metric measuring factor correctness. We demonstrate that this metric is contaminated by severe correlations between the factor classifiers, enabling the metric to exploit these correlations rather than independently taking each one into account. This evaluation misrepresents the true quality of imaging factor disentanglement. To address this critical limitation, we introduce \textbf{I-FAA} (Improved Factor Alignment Accuracy), a novel evaluation metric, which employs class-balanced under-sampling and diverse factor-specific augmentation strategies during training of the imaging factor classifiers to break correlations between factors and mitigate shortcut learning of dataset-specific cues. This design provides more accurate factor disentanglement evaluation by eliminating the reliance on inter-factor patterns.

To study imaging factor disentanglement, we conduct experiments on DF-RICO~\cite{godavarthy2026multi} benchmark. First, we show improvements of our method over MULTI and other baselines, particularly for novel factor combinations. Second, we show cross-factor leakage in FAA, and I-FAA through correlation analysis.

Our main contributions are:
\begin{itemize}
\item \textbf{X-MULTI Approach:} Introduces zero-shot VLM-based supervision for synthetically generated unseen imaging-factor combinations, providing explicit factor-level supervision beyond the reconstruction-only training.
\item \textbf{I-FAA Metric:} Introduces an improved factor-alignment evaluation metric that reduces cross-factor shortcut learning through class-balanced training and factor-specific augmentations, enabling more robust assessment of imaging factor disentanglement.
\end{itemize}

\section{Related Work}


\subsubsection{Disentanglement in Generative Models:}
T2I models such as SDXL~\cite{podell2023sdxl} provide strong generative backbones, while DreamBooth~\cite{ruiz2023dreambooth} adapts these models to specific visual concepts. Disentanglement in T2I models typically focuses on separating objects from their attributes, such as color or texture. Early approaches utilized Low-Rank Adaptation (LoRA) blocks to decouple style from content~\cite{frenkel2024implicit,liu2025unziplora}. Textual Inversion (TI)\cite{gal2022image} optimizes a new token embedding for a target concept while keeping the model weights frozen, which is then inserted into prompts to generate that concept.
This idea has been applied to learn embeddings for visual concepts such as color, shape, style, texture and object appearance\cite{vinker2023concept,sohn2023styledrop,xu2024dreamanime,motamed2023lego,butt2024colorpeel}. To prevent attribute leakage, several methods steer cross-attention maps by constraining concept tokens to attend to their corresponding image regions~\cite{hertz2022prompt,jin2024image,chefer2023attend,zhang2024attention,shentu2024attencraft} or employ explicit masks to isolate foreground objects from their backgrounds~\cite{avrahami2023break,kwon2024concept,li2023layerdiffusion}. More recently, transformer-based architectures have leveraged the modulation space for multi-concept disentanglement~\cite{garibi2025tokenverse,zhong2025mod}. However, these methods mainly disentangle semantic or personalized concepts rather than imaging acquisition factors, making direct comparison with X-MULTI difficult.

\subsubsection{Imaging-Factor Disentanglement:}
The image synthesis conditioned on imaging factors has received less attention. MULTI~\cite{godavarthy2026multi} addressed this by utilizing TI to optimize imaging factor embeddings. While MULTI improved over baselines for novel factor combinations, its factor alignment remained limited. We improve upon MULTI by introducing VLM-guided supervision to encourage disentanglement of imaging factors.

\subsubsection{Discriminator-Based Guidance and VLM Supervision:}
Large-Language Models (LLMs) and Vision-Language Models (VLMs) have emerged as versatile zero-shot discriminators~\cite{luo2025dual}. LLMs and VLMs are increasingly repurposed for complex tasks such as object detection~\cite{bai2025qwen3}, optical character recognition~\cite{laurenccon2024matters}, and more~\cite{cho2023dall,wen2023improving}. Recent frameworks have integrated these models as frozen discriminators to provide guidance, ensuring that generative models follow high-level conceptual constraints~\cite{bai2025qwen3,wu2024v,cai2024vip}.
X-MULTI leverages this paradigm by utilizing a VLM as an external classifier to enforce factor identity, improving image synthesis of novel factor combinations.

\subsubsection{Disentanglement Evaluation Metrics:}
Evaluating factor disentanglement requires metrics that assess each factor in isolation. \cite{godavarthy2026multi} introduced FAA, which evaluates imaging-factor disentanglement using factor classifiers trained on the DF-RICO benchmark~\cite{godavarthy2026multi}. While classifier-based evaluation metrics like FAA often rely on highly correlated, co-occurring dataset shortcuts, they fail to isolate individual target attributes. To address this, I-FAA introduces factor-specific augmentations that break these correlations to ensure reliable disentanglement assessment.
\section{Background}

\noindent\textbf{Imaging Factor Disentanglement:}
Following MULTI~\cite{godavarthy2026multi}, imaging factor disentanglement aims to decompose image formation into $\mathcal{K}$ factors such as lens, sensor, viewpoint, and domain. We consider a collection of datasets $\mathcal{D}={D^{(1)},\dots,D^{(N)}}$ with shared factor annotations, where each image is associated with a factor tuple $\mathbf{f}=(f^{(k)})_{k\in\mathcal{K}}$. 
Each factor $k$ takes values from a discrete category $\mathcal{F}^{(k)}$, defining the full factor space $\mathcal{F}=\mathcal{F}^{(1)}\times\cdots\times\mathcal{F}^{(\mathcal{K})}$. In practice, training data usually only covers a sparse subset $\mathcal{F}_{\mathrm{obs}}\subset \mathcal{F}$. The challenge is to learn factor representations that produce images which reflect any target tuple $\mathbf{f}$, including combinations absent from training.

\noindent\textbf{Learnable Embeddings:}
Each factor value $f^{(k)}$ is represented by a learnable token $t^{(k)}$, implemented as a sequence of $n$ embedding vectors,
\(
    \mathcal{V}(t^{(k)})=\{v_1^{(k)},\ldots,v_n^{(k)}\}\in\mathbb{R}^{n\times d}.
\)
Following MULTI~\cite{godavarthy2026multi}, these tokens are inserted into a structured prompt together with an image caption and optimized using the standard diffusion loss, 
\begin{equation}
    \mathcal{L}_{\mathrm{diff}} = \mathbb{E}_{z_t, x, \epsilon, t} \left[ \left\| \epsilon - \mathcal{U}(z_t, t, \mathbf{c}) \right\|_2^2 \right]
\end{equation}
where $\epsilon$ is the sampled noise, $z_t$ is the noisy latent at timestep $t$, $\mathcal{U}$ is the denoising network, and $\mathbf{c}$ is the text-conditioning embedding from the structured prompt. The diffusion backbone remains frozen.

\noindent\textbf{Factor Alignment Accuracy (FAA):}
To evaluate whether the generated images faithfully represent the intended factors, MULTI introduced Factor Alignment Accuracy (FAA)~\cite{godavarthy2026multi}, where for each factor category, a classifier is trained.
During evaluation, a generated image is passed through these classifiers, and each predicted factor is compared against the corresponding factor label used to condition the generation. FAA measures the fraction of predictions that match the target factor labels, thereby estimating whether the generated image visually contains the requested imaging factors.

\section{Methodology}

\subsection{X-MULTI: VLM-based Factor Disentanglement}

\begin{figure*}[t]
    \centering
    \includegraphics[width=1\textwidth]{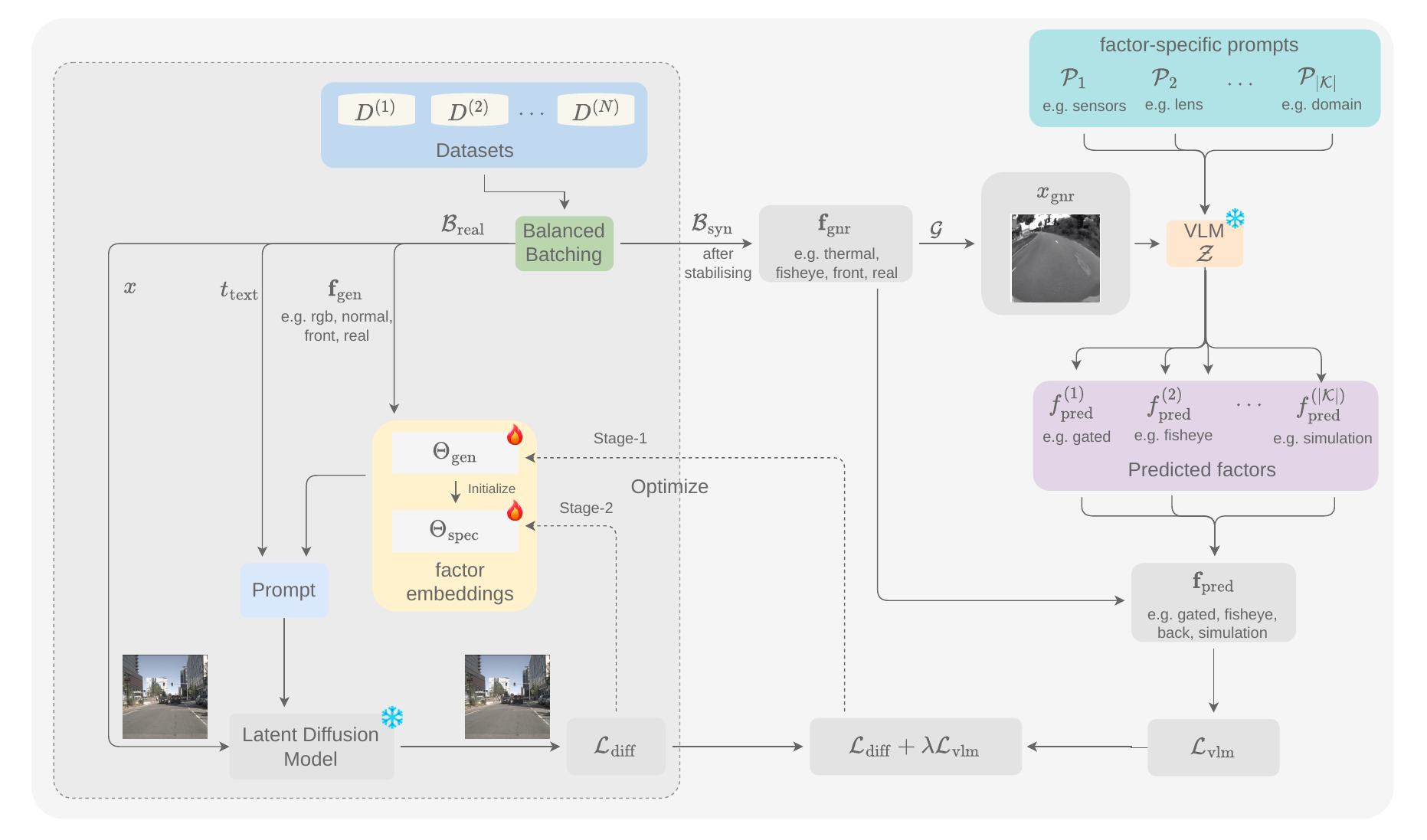}
    \caption{\textbf{X-MULTI Architecture}. Overview of our framework, which augments MULTI (components enclosed within the dark grey dashed box) with zero-shot VLM supervision to optimize factor embeddings via joint diffusion and VLM-based factor alignment loss.}
    \label{fig:xmulti}
\end{figure*}

To strengthen factor disentanglement and synthesize images with novel factor combinations, we introduce X-MULTI, which augments MULTI with zero-shot VLM-based supervision. MULTI learns factor embeddings in two stages: (1), it learns general embeddings $\Theta_{\text{gen}}$ for broad imaging factors such as lens, sensor, viewpoint, and domain (e.g. fisheye lens); (2), it adapts these embeddings to the specific factor values of a target dataset $\Theta_{\text{spec}}$ (e.g., fisheye lens of Woodscape dataset). X-MULTI retains this two-stage structure, and adds VLM-based supervision only at Stage-1. Fig.~\ref{fig:xmulti} provides an overview of the proposed pipeline. 

\noindent\textbf{VLM-based Supervision:} Let $x_{\mathrm{gnr}}$ be an image generated with the target factor tuple $\mathbf{f}_{\mathrm{gnr}}=(f_{\mathrm{gnr}}^{(k)})_{k\in\mathcal{K}}$. We employ a frozen zero-shot VLM $\mathcal{Z}$ and factor-specific prompts $\mathcal{P}=\{\mathcal{P}_1,\ldots,\mathcal{P}_{|\mathcal{K}|}\}$. For each factor category $k$, the VLM predicts
\(
    f_{\mathrm{pred}}^{(k)}
    =
    \mathcal{Z}(x_{\mathrm{gnr}},\mathcal{P}_k).
\)
The VLM-based factor alignment loss is defined as
\begin{equation}
    \mathcal{L}_{\mathrm{vlm}}
    =
    \sum_{k\in\mathcal{K}}
    \ell_{\mathrm{CE}}
    \left(
    f_{\mathrm{pred}}^{(k)},
    f_{\mathrm{gnr}}^{(k)}
    \right),
\end{equation}
where $\ell_{\mathrm{CE}}$ denotes cross-entropy loss. Factor-wise VLM reliability is determined by evaluating $\mathcal{Z}$ on ground-truth images and comparing its predictions against true labels using the factor-specific prompts. 

\noindent\textbf{Training Objective:} Following MULTI~\cite{godavarthy2026multi}, we optimize the general factor embeddings $\Theta_{\mathrm{gen}}$ with the diffusion objective. After an initial diffusion-only warm-up, in each training step, a real batch $\mathcal{B}_{\mathrm{real}}$ is sampled from the training datasets. From the factor values present in $\mathcal{B}_{\mathrm{real}}$, we additionally construct a synthetic factor tuple $\mathbf{f}_{\mathrm{gnr}}$ by resampling factor values across categories. The current diffusion model $\mathcal{G}$ then produces an image $x_{\mathrm{gnr}} = \mathcal{G}(\mathbf{f}_{\mathrm{gnr}})$, which is used only for VLM-based supervision. The training objective is
\begin{equation}
\hat{\Theta}_{\mathrm{gen}}
=
\arg\min_{\Theta_{\mathrm{gen}}}
\mathbb{E}
\left[
\mathcal{L}_{\mathrm{diff}}
+
\lambda \mathcal{L}_{\mathrm{vlm}}
\right],
\end{equation}
where $\lambda$ controls the strength of VLM-based supervision.

\subsection{Improved Factor Alignment Accuracy (I-FAA)}

To evaluate imaging factor disentanglement, we propose Improved Factor Alignment Accuracy (I-FAA) as a new metric. Given a set of $N$ generated images $\{x_i\}_{i=1}^N$, where each $x_i$ is conditioned on the factor tuple $\mathbf{f}_i = (f_i^{(1)}, \ldots, f_i^{(\mathcal{K})})$, we use a set of factor classifiers \(
    \mathcal{S} = \{\mathcal{S}_1, \mathcal{S}_2, \ldots, \mathcal{S}_{\mathcal{K}}\}
\), where $\mathcal{S}_k$ predicts factor values in $\mathcal{F}^{(k)}$. The per-category alignment accuracy is defined as
\begin{equation}
    \mathrm{I\text{-}FAA}_k = 
    \frac{1}{N} \sum_{i=1}^{N} 
    \mathbf{1}\!\left[\mathcal{S}_k(x_i) ==f_i^{(k)}\right].
    \label{eq:ifaa_k}
\end{equation}
The overall I-FAA score is obtained by averaging over all factor categories
\begin{equation}
    \mathrm{I\text{-}FAA}_{\mathrm{avg}} = 
    \frac{1}{\mathcal{K}} \sum_{k=1}^{\mathcal{K}} 
    \mathrm{I\text{-}FAA}_k.
    \label{eq:ifaa_avg}
\end{equation}

\noindent\textbf{Classifier Architecture:} Each factor classifier $\mathcal{S}_k$ consists of the shared frozen visual backbone $\phi$ with parameters $\Theta_{\mathrm{bb}}$, followed by a factor-specific trainable linear head $h_k$, with parameters $\Theta_{h_k}$. Thus, $\mathcal{S}_k$ is parameterized by the shared frozen backbone parameters $\Theta_{\mathrm{bb}}$ and the factor-specific head parameters $\Theta_{h_k}$. During training, only $\Theta_{h_k}$ is optimized, while $\Theta_{\mathrm{bb}}$ remains fixed.

\noindent\textbf{Class-Balanced Sampling:} To mitigate bias in highly skewed data, we adopt class-balanced under-sampling during training of the classifiers. Let $D_k = \{(x_i,\allowbreak f_i^{(k)})\}$ denote the training dataset for factor category $k$, where $f_i^{(k)} \in \mathcal{F}^{(k)}$ is the factor label associated with image $x_i$. 
Let \(n_{\min}^{(k)}\) denote the number of the smallest class in category $k$. Samples from every class $c \in \mathcal{F}^{(k)}$ are randomly under-sampled to  $n_{\min}^{(k)}$, ensuring balanced class representation.

\noindent\textbf{Factor-Specific Augmentation:} Beyond class imbalance, classifiers may acquire shortcuts by learning dataset-level features that co-occur with a target factor. To reduce such shortcut learning, we employ each classifier $\mathcal{S}_k$ with dedicated augmentations. Formally, for category $k$ we define
\(
    \mathcal{A}^{(k)} = \{a_1^{(k)},\, a_2^{(k)},\, \ldots,\, a_{M_k}^{(k)}\},
\)
where $M_k$ denotes the number of augmentation operations assigned to category $k$. During training, augmentations are sampled uniformly from $\mathcal{A}^{(k)}$ and applied to the image $x$ to produce the augmented image $\tilde{x}^{(k)}$.

\noindent\textbf{Optimization:} Given augmented inputs and balanced mini-batches, each classifier $\mathcal{S}_k$ is optimized via the standard cross-entropy objective
\begin{equation}
    \hat{\Theta}_{h_k} = 
    \arg\min_{\Theta_{h_k}} \; 
    \mathbb{E}_{(x,\, f^{(k)}) \,\sim\, D_k} 
    \left[ 
        \ell_{\mathrm{CE}}\!\left( \mathcal{S}_k\!\left(\tilde{x}^{(k)}\right),\, f^{(k)} 
        \right) 
    \right].
    \label{eq:ifaa_loss}
\end{equation}
\begin{table}[t]
\caption{\textbf{Factor-specific Prompts for VLM-based supervision}. Our method uses structured prompts and JSON output constraints to independently query the zero-shot VLM for each imaging factor.}
\label{tab:factor_prompts}
\centering
\scriptsize 
\renewcommand{\arraystretch}{1.2} 
\begin{tabularx}{\linewidth}{@{} >{\raggedright\arraybackslash}p{1.5cm} >{\raggedright\arraybackslash}X @{} }
\toprule
\textbf{Factor} & \textbf{Prompt} \\
\midrule

Sensor &
Classify the SENSOR type of this image (0-4): \newline
0 = RGB: normal color camera \newline
1 = Thermal: grayscale images with bright/warm colors/white \newline
2 = Gated: high contrast black and white, harsh lighting, foggy/smoky \newline
3 = Event: motion changes, red/blue motion capture \newline
4 = RGB Thermal (Fusion): Color images containing an ``overlay'' or ``glow'' of thermal data. Look for heat signatures (bright orange/yellow/white highlights) mapped onto real-world objects and textures. \newline
\texttt{Return only JSON: \{"sensor": <int>\}} \\

\midrule

Lens &
Classify the LENS type of this image (0-1): \newline
0 = Normal: standard camera lens \newline
1 = Fisheye: distorted edges, wide field of view \newline
\texttt{Return only JSON: \{"lens": <int>\}} \\

\midrule

Viewpoint &
Classify the VIEWPOINT of this image (0-4): \newline
0 = Drone: camera placed on a flying drone \newline
1 = Pole: camera mounted on a pole \newline
2 = Front: camera faces forward \newline
3 = Back: camera faces backward \newline
4 = Side: camera placed on left/right (typically on car) \newline
\texttt{Return only JSON: \{"viewpoint": <int>\}} \\

\midrule

Domain &
Classify the DOMAIN of this image (0-2): \newline
0 = Real: real-world images \newline
1 = Simulation: rendered or simulated images \newline
2 = Video Game: captured from video-game \newline
\texttt{Return only JSON: \{"domain": <int>\}} \\

\bottomrule
\end{tabularx}
\end{table}


\begin{table}[t]
\centering
\caption{\textbf{Data augmentations for I-FAA classifier training.} Category-specific, non-overlapping image augmentations applied during I-FAA (ours) training.}
\label{tab:classifier_augmentations}
\tiny
\setlength{\tabcolsep}{2pt}
\renewcommand{\arraystretch}{1.2}
\begin{tabularx}{\linewidth}{@{} l *{4}{>{\centering\arraybackslash}X} @{}}
\toprule
\textbf{Augmentation Category \& Type} & \textbf{Lens} & \textbf{Sensor} & \textbf{Domain} & \textbf{Viewpoint} \\
\midrule
\textbf{Target Classes} & normal, fisheye & rgb, thermal, rgb-thermal, event, gated & real, simulation, video-game & front, back, side, pole, drone \\
\midrule

\multicolumn{5}{@{}l}{\textbf{Geometric}} \\
\quad Horizontal flip & \checkmark & \checkmark & \checkmark & Mild \\
\quad Vertical flip   & \checkmark & \checkmark & \checkmark & --- \\
\quad Random rotation \& translation & Mild & Moderate & Moderate & Mild \\
\midrule

\multicolumn{5}{@{}l}{\textbf{Color / Appearance}} \\
\quad Color jitter, Grayscale, \& Noise & \checkmark & --- & \checkmark & \checkmark \\
\quad Gaussian blur & Moderate & Moderate & Strong & Mild \\
\midrule

\multicolumn{5}{@{}l}{\textbf{Occlusion / Region}} \\
\quad Cutout, Random erasing, \& Hide-and-seek & Moderate & Mild & Moderate & --- \\
\midrule

\multicolumn{5}{@{}l}{\textbf{Additional}} \\
\quad Radial distortion & --- & Moderate & Moderate & Mild \\
\quad Perspective warp & --- & \checkmark & \checkmark & --- \\
\bottomrule
\end{tabularx}
\end{table}

\begin{table*}[t]
\centering
\caption{\textbf{Qualitative comparison of novel combinations.} Image generation with unseen factor combinations. Our method, X-MULTI shows better factor adherence.}
\scriptsize
\setlength{\tabcolsep}{2pt}
\renewcommand{\arraystretch}{1.05}
\resizebox{\textwidth}{!}{
\begin{tabular}{lcccccccc}
\toprule

\textbf{Domain} 
& real & video-game & real & real & video-game & simulation & video-game & simulation(shift) \\
\textbf{Sensor} 
& thermal & rgb & gated & rgb-thermal & rgb & thermal & thermal & thermal(timo) \\
\textbf{Viewpoint} 
& fisheye(loaf) & fisheye & normal & fisheye & normal & normal & fisheye & normal \\
\textbf{Lens} 
& front & drone(visdrone) & pole & front & pole & drone(visdrone) & pole & pole(fisheye8k) \\

\midrule

SDXL Zeroshot~\cite{podell2023sdxl}
& \includegraphics[width=1.7cm,height=1.7cm]{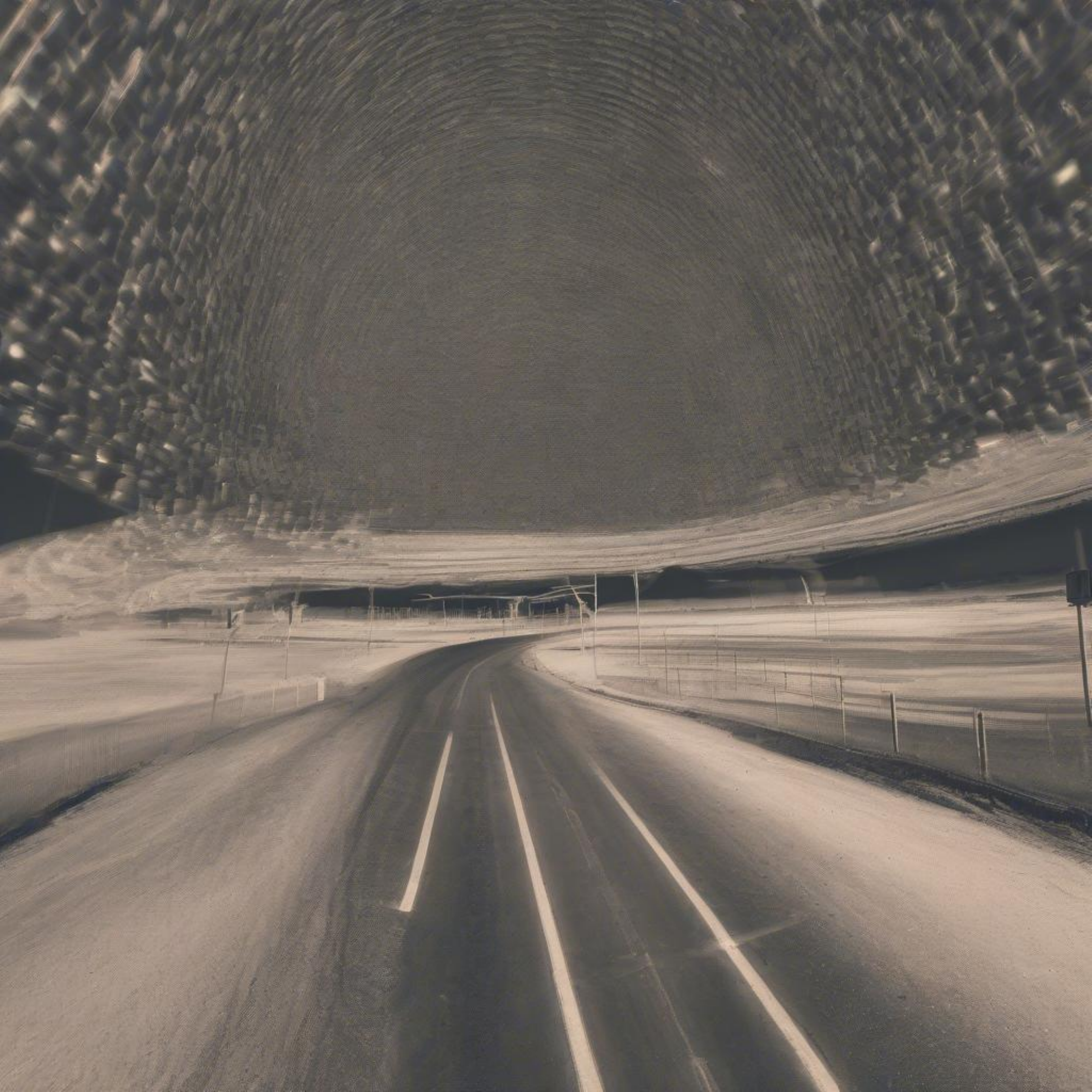}
& \includegraphics[width=1.7cm,height=1.7cm]{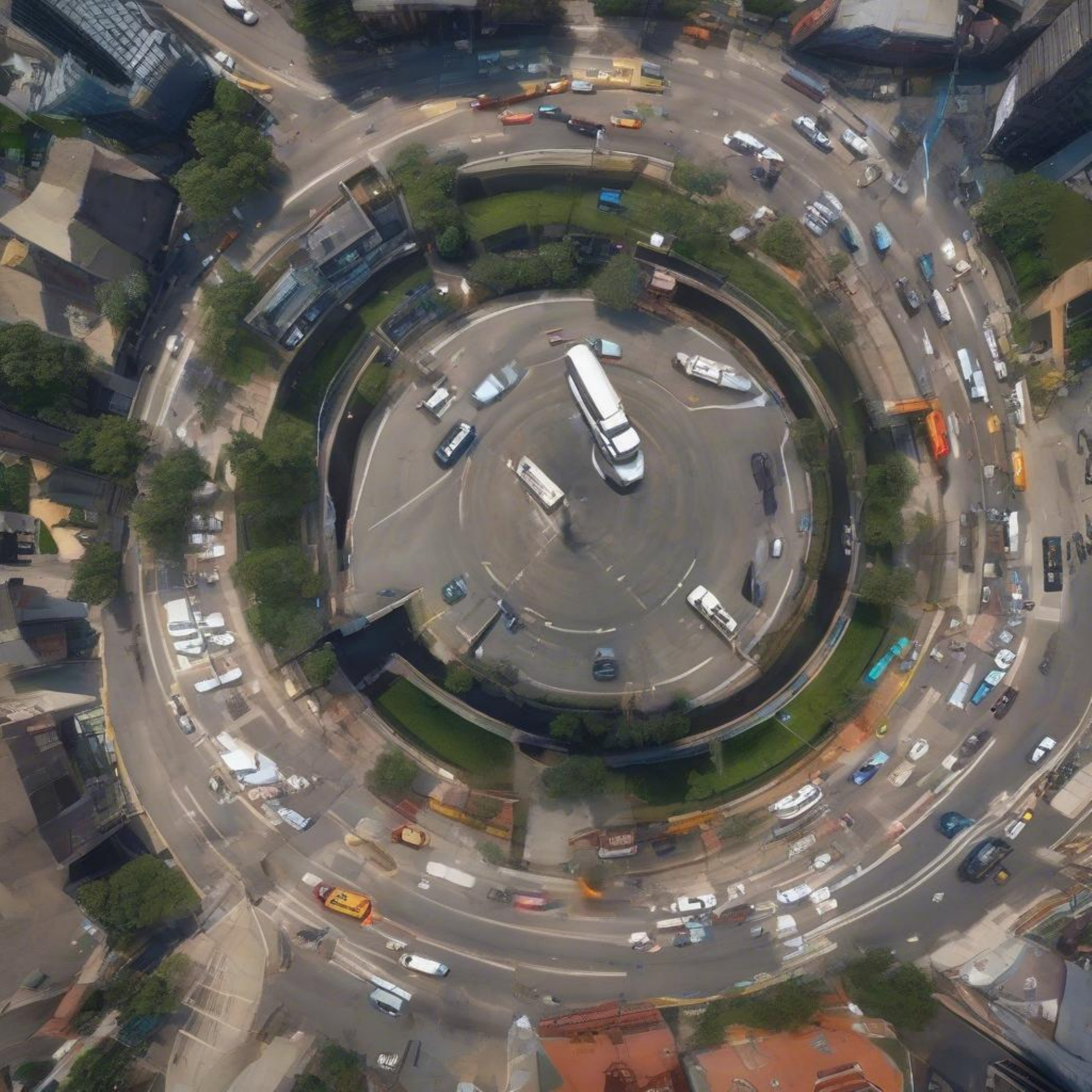}
& \includegraphics[width=1.7cm,height=1.7cm]{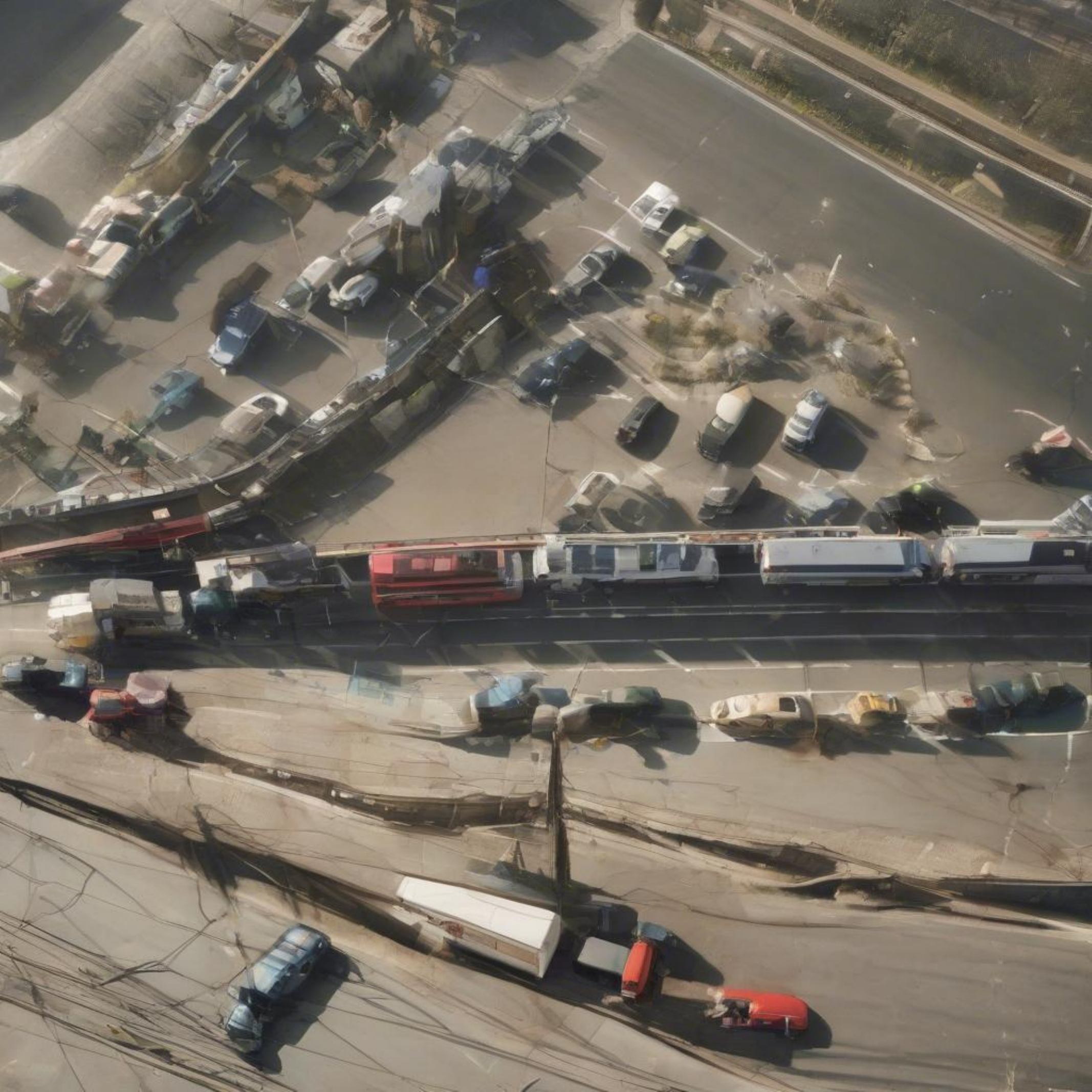}
& \includegraphics[width=1.7cm,height=1.7cm]{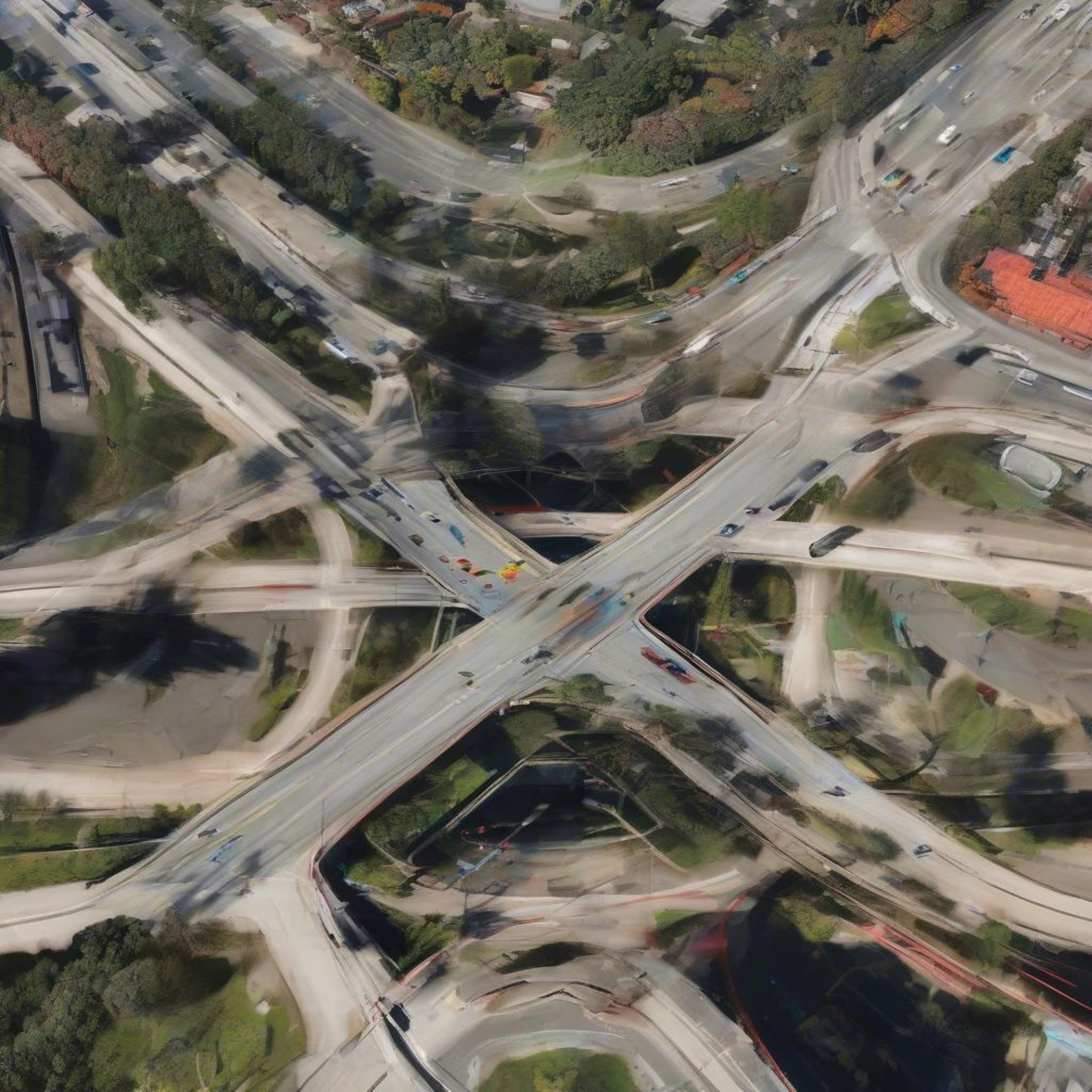}
& \includegraphics[width=1.7cm,height=1.7cm]{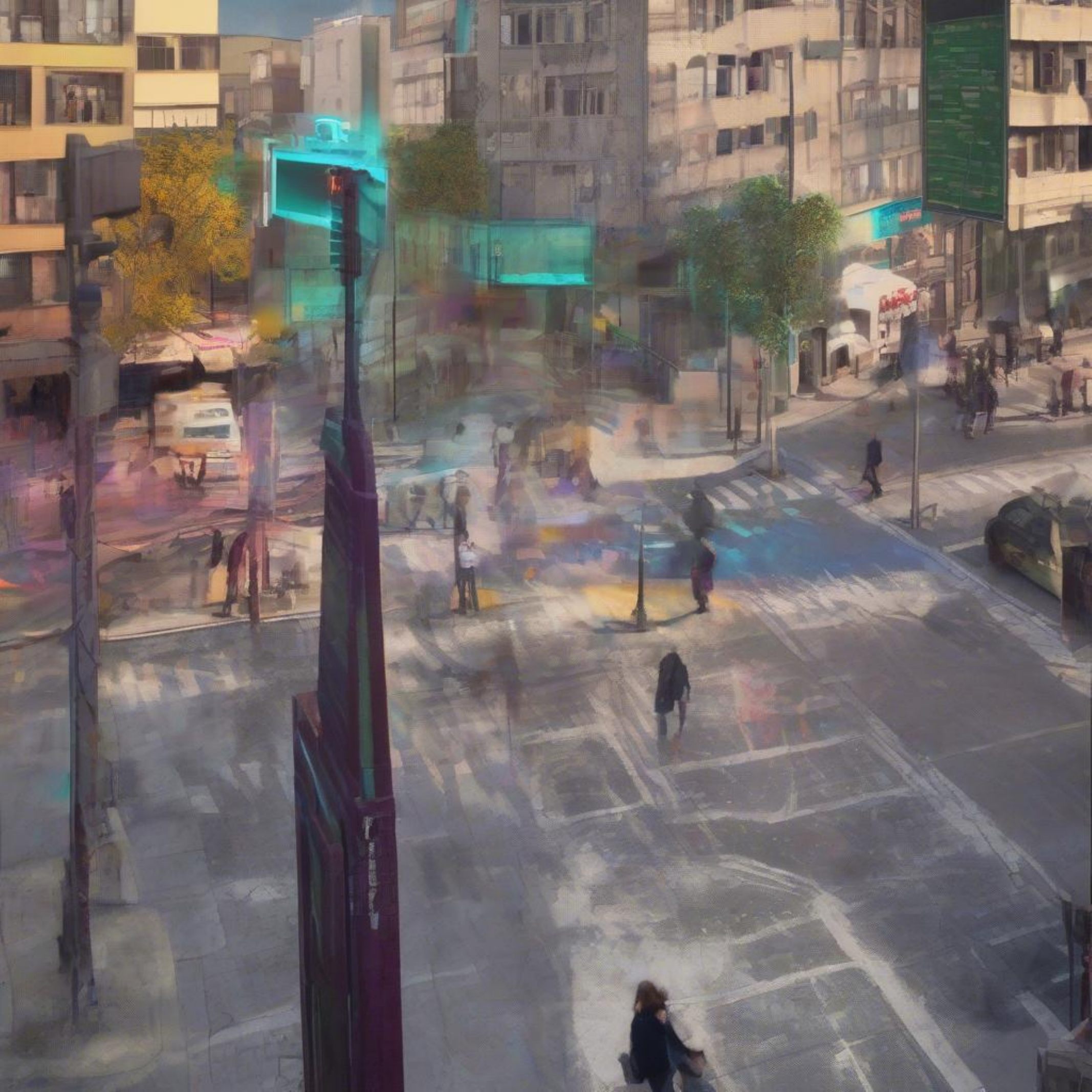}
& \includegraphics[width=1.7cm,height=1.7cm]{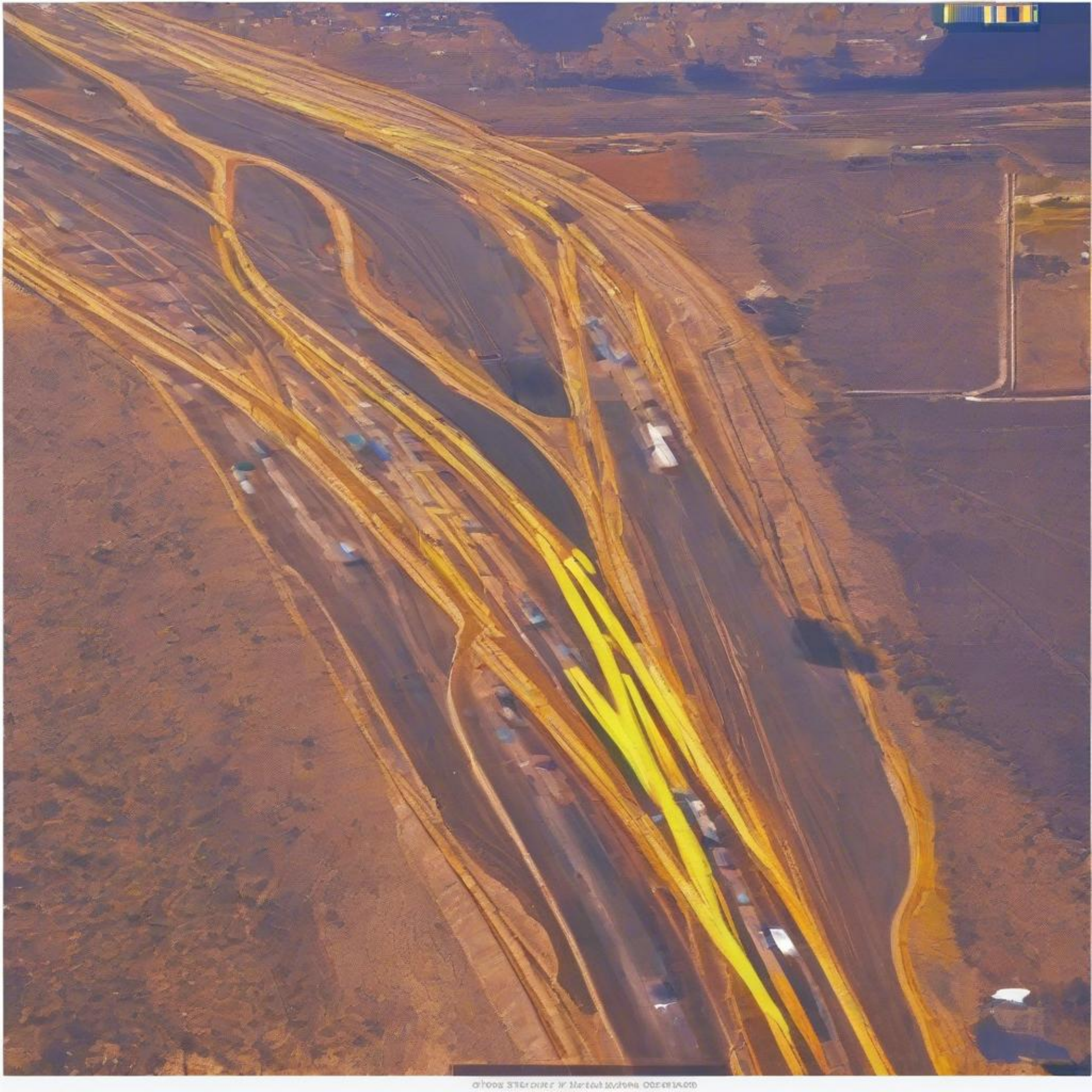}
& \includegraphics[width=1.7cm,height=1.7cm]{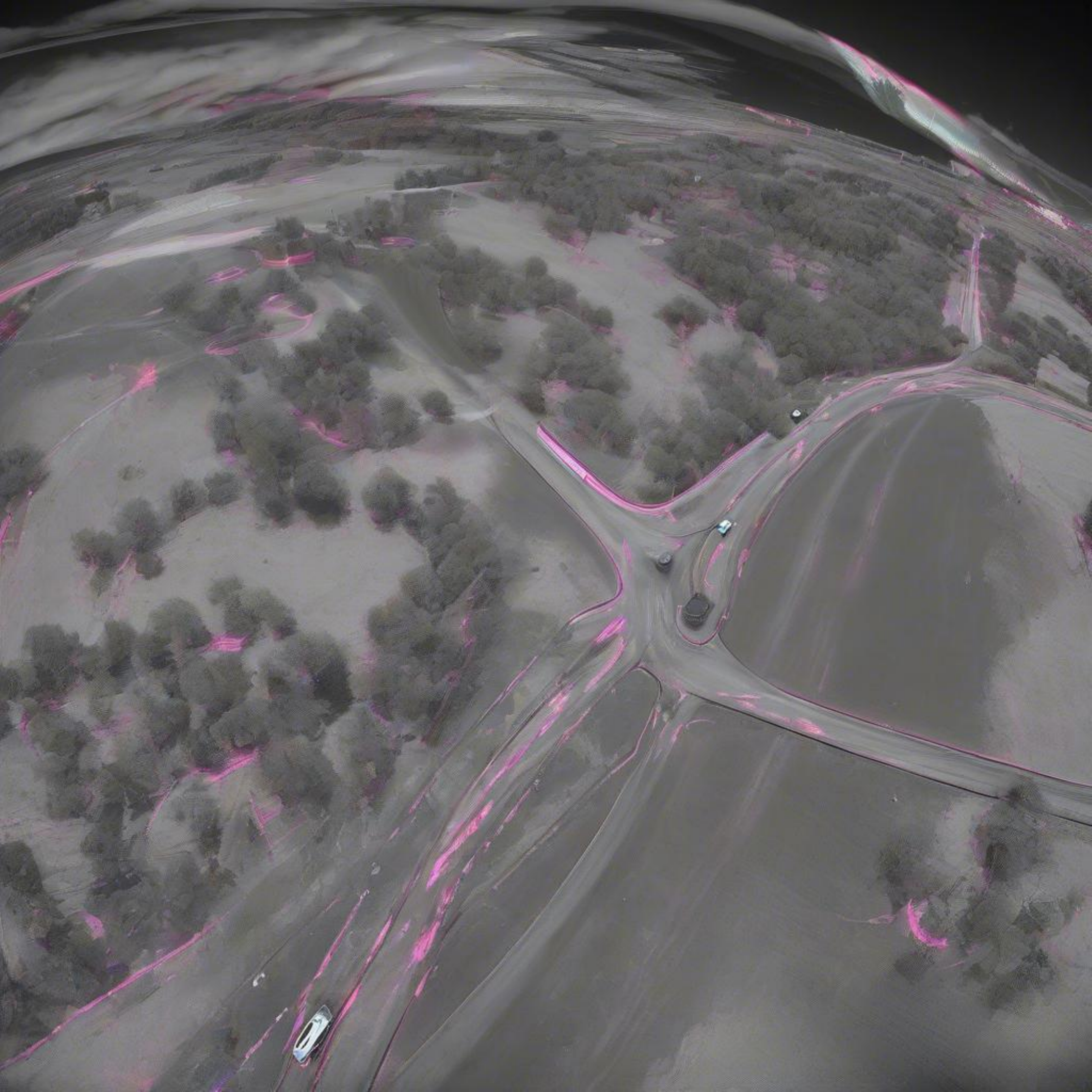}
& \includegraphics[width=1.7cm,height=1.7cm]{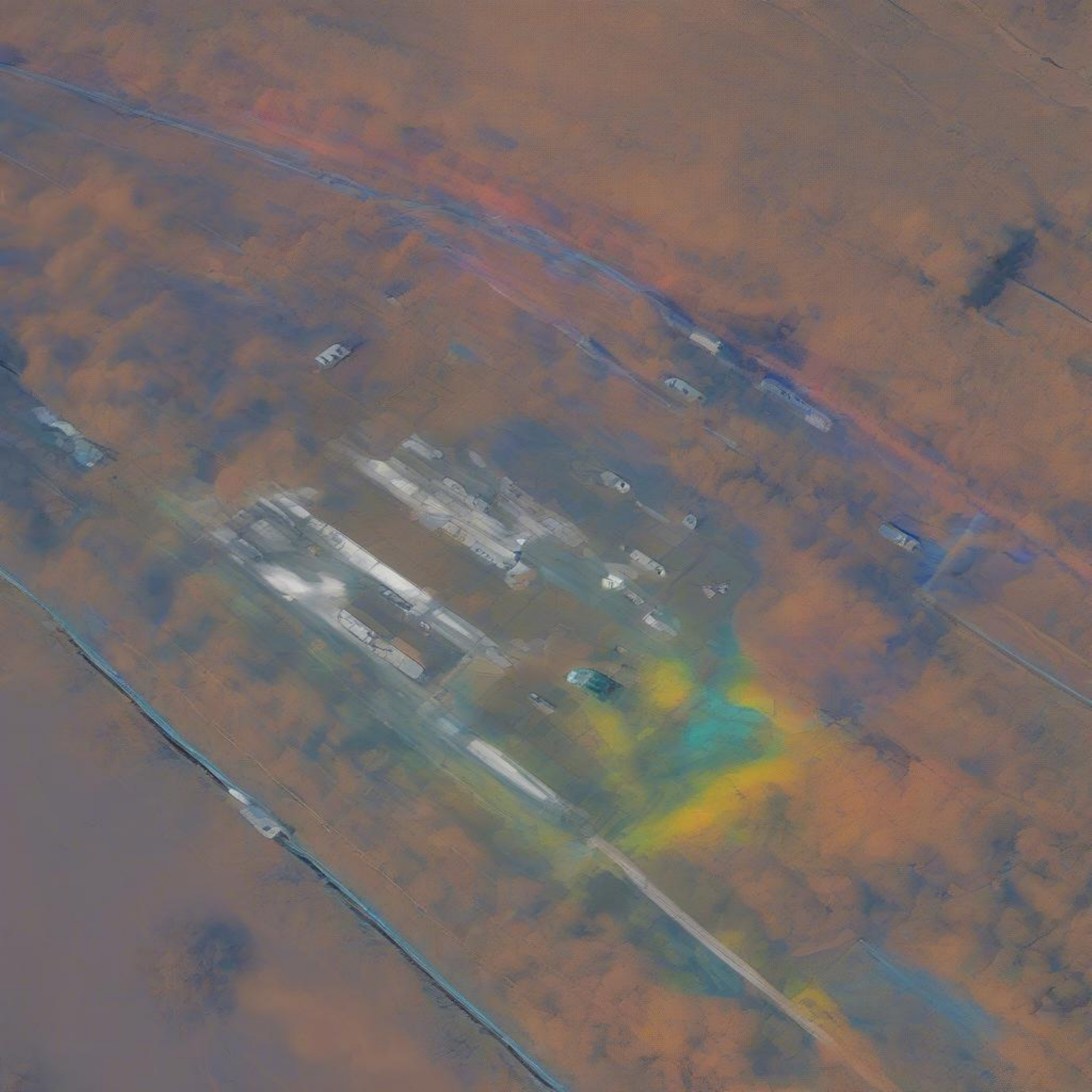} \\

DreamBooth~\cite{ruiz2023dreambooth}
& \includegraphics[width=1.7cm,height=1.7cm]{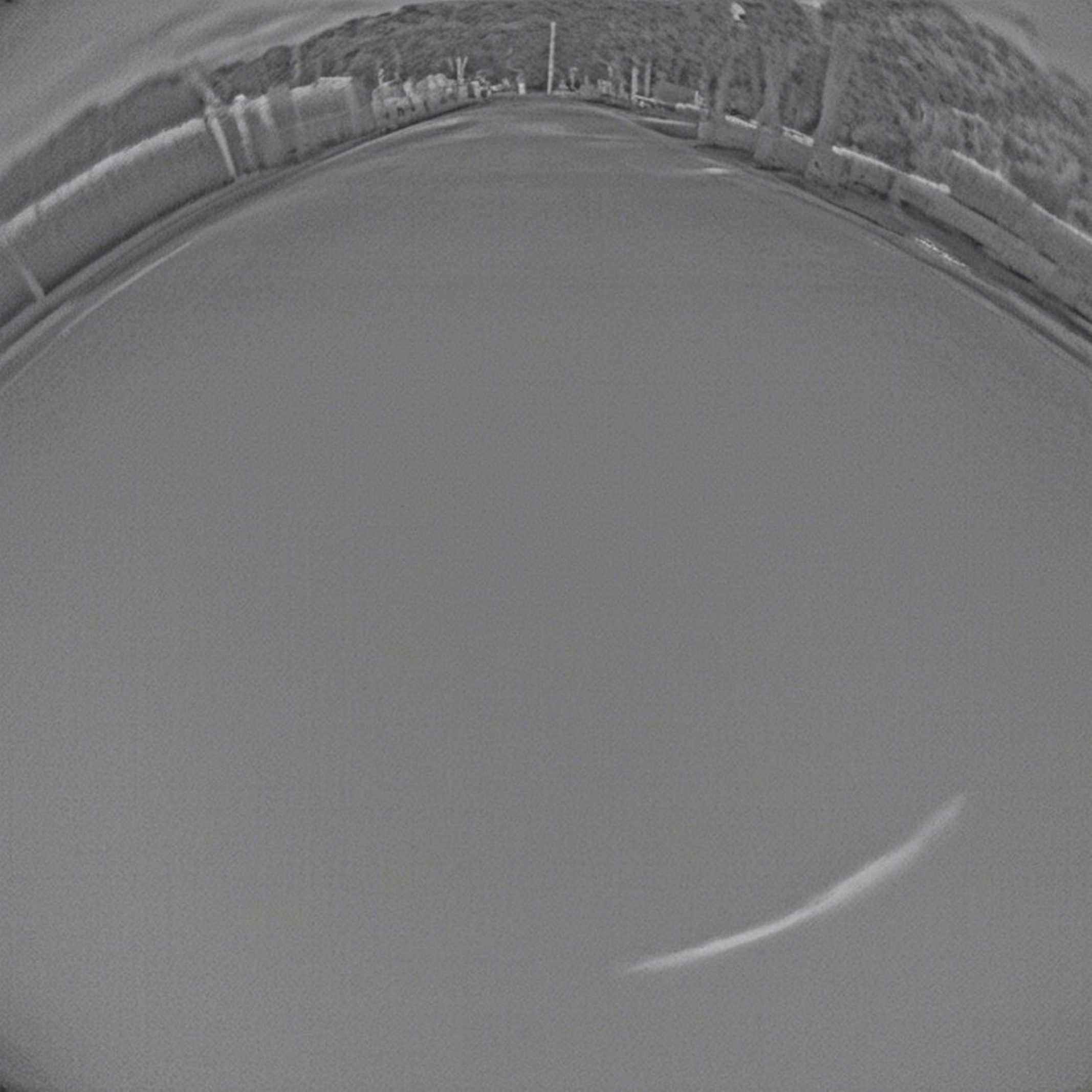}
& \includegraphics[width=1.7cm,height=1.7cm]{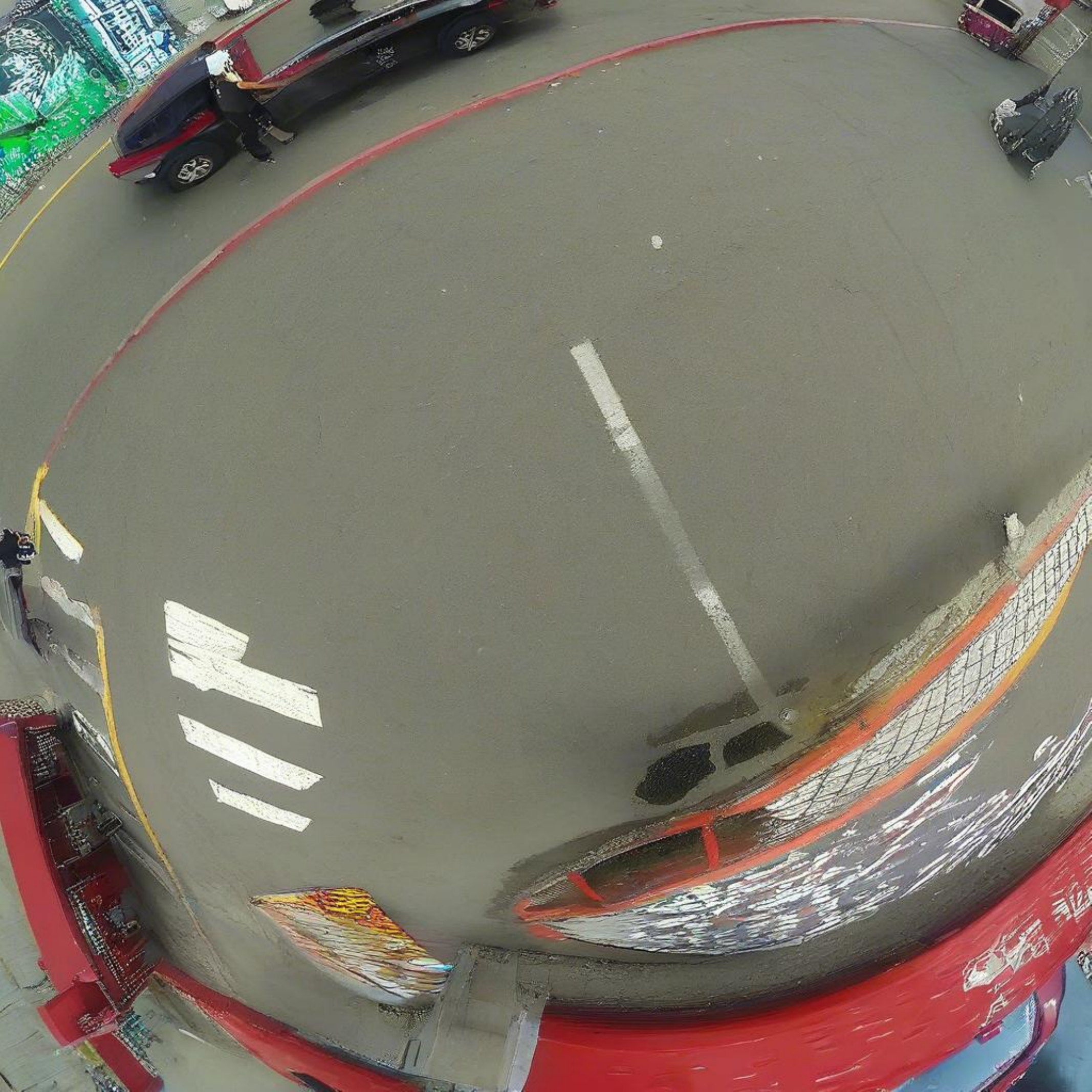}
& \includegraphics[width=1.7cm,height=1.7cm]{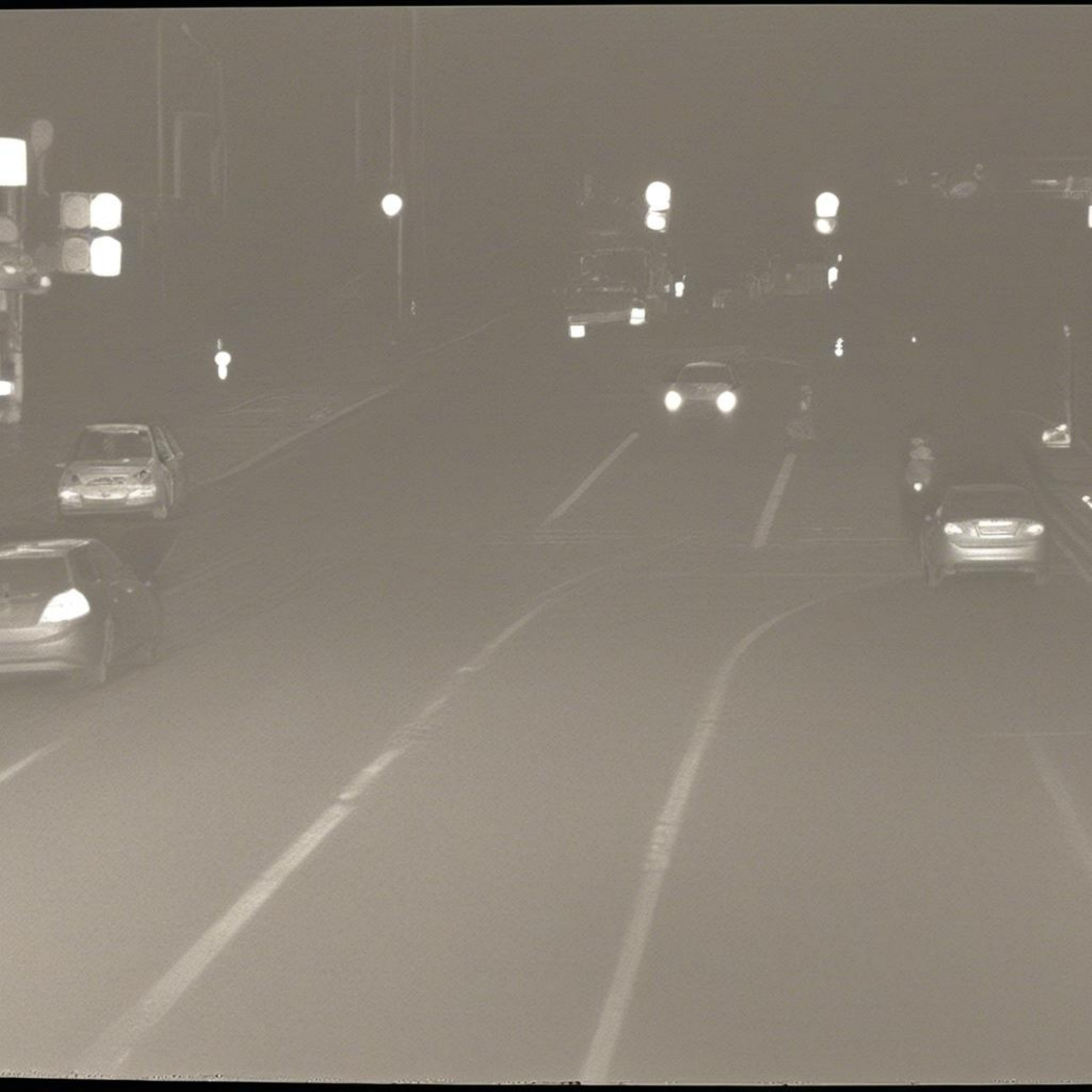}
& \includegraphics[width=1.7cm,height=1.7cm]{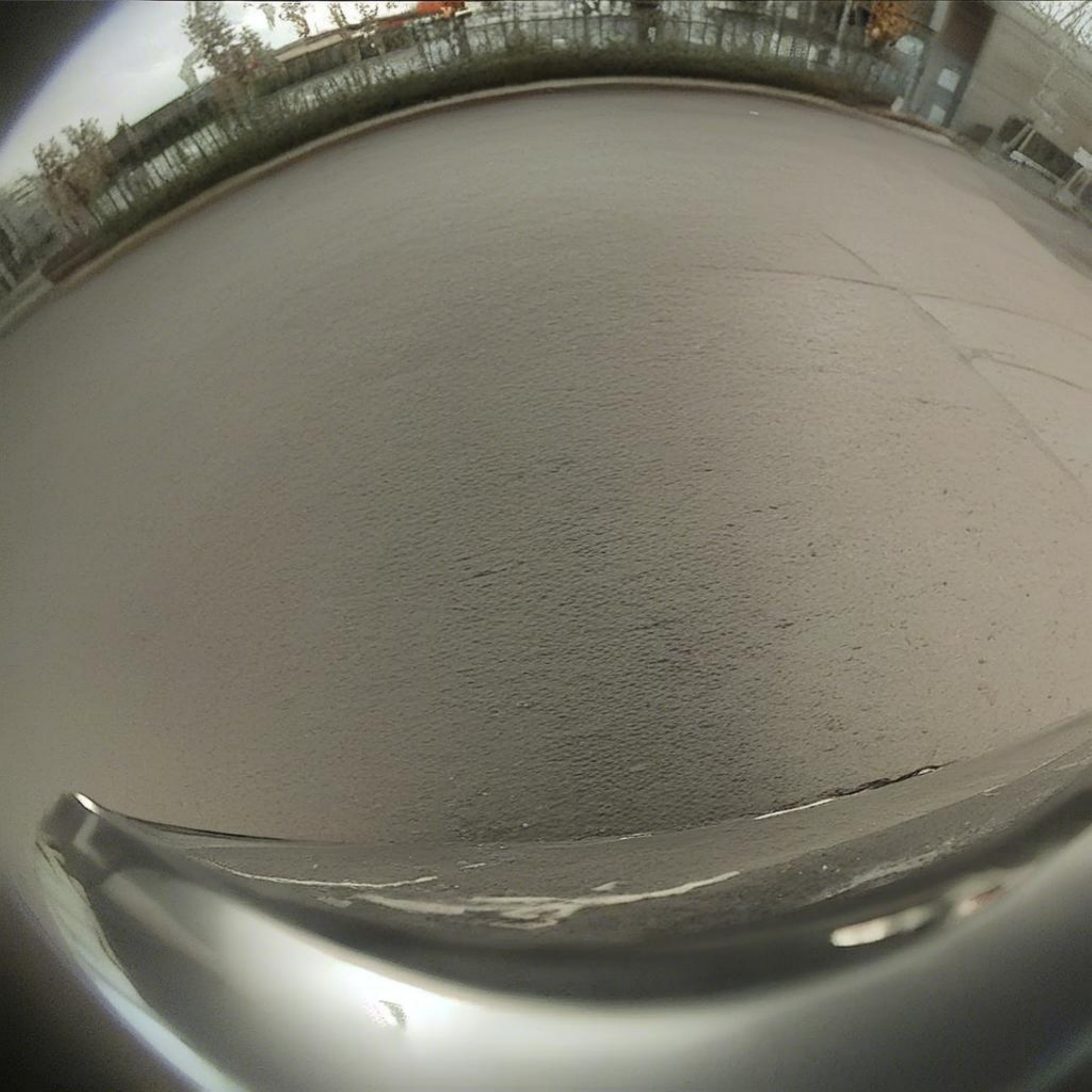}
& \includegraphics[width=1.7cm,height=1.7cm]{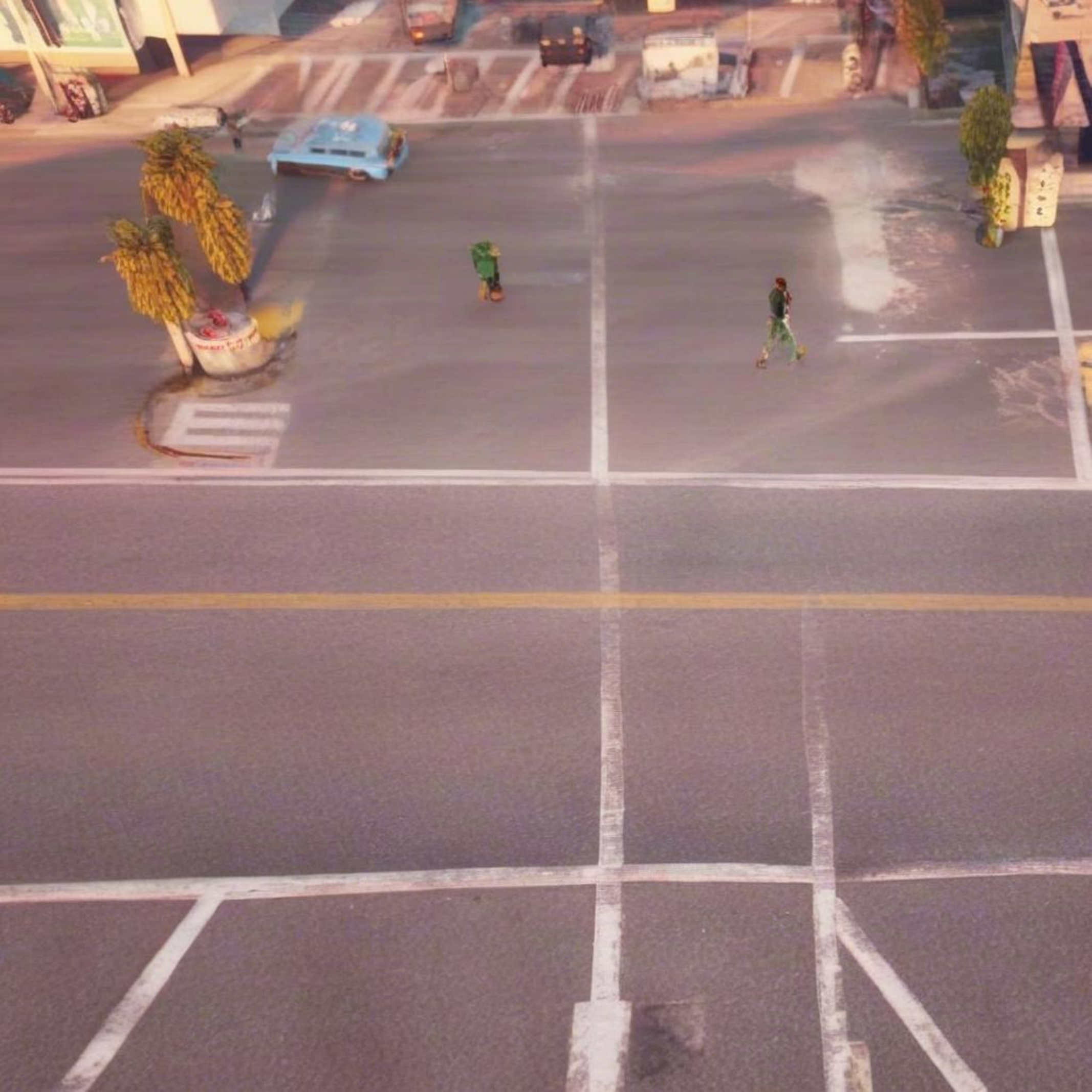}
& \includegraphics[width=1.7cm,height=1.7cm]{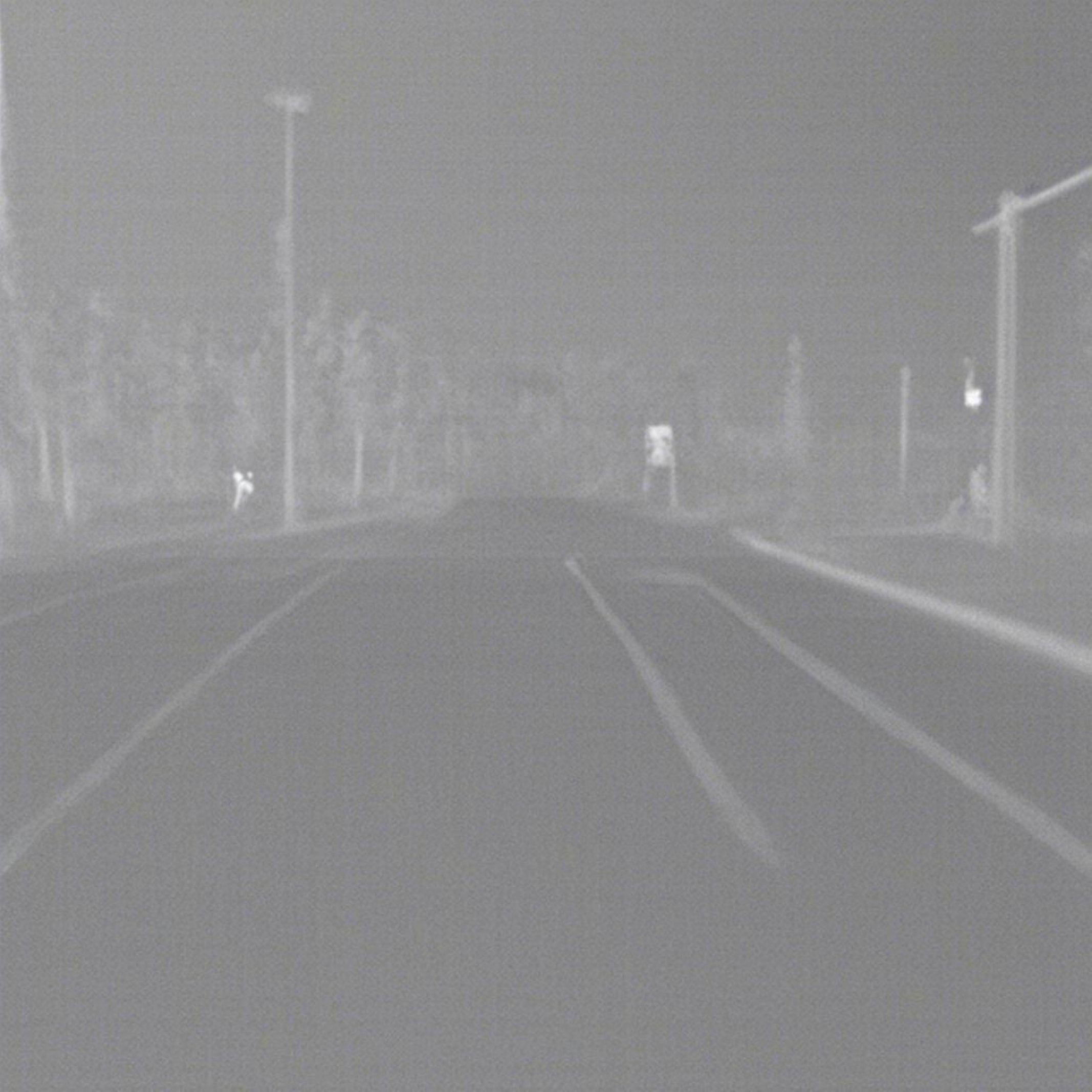}
& \includegraphics[width=1.7cm,height=1.7cm]{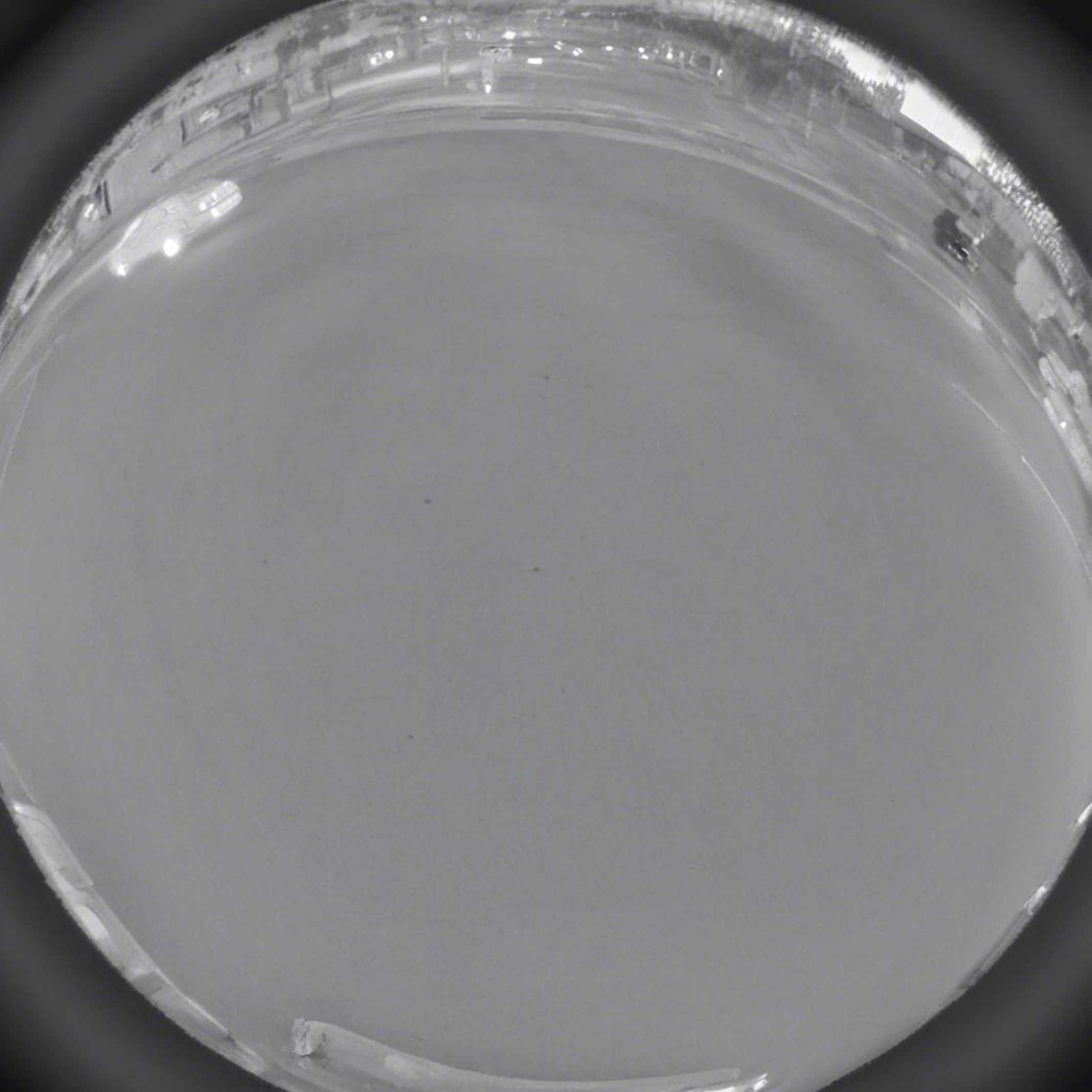}
& \includegraphics[width=1.7cm,height=1.7cm]{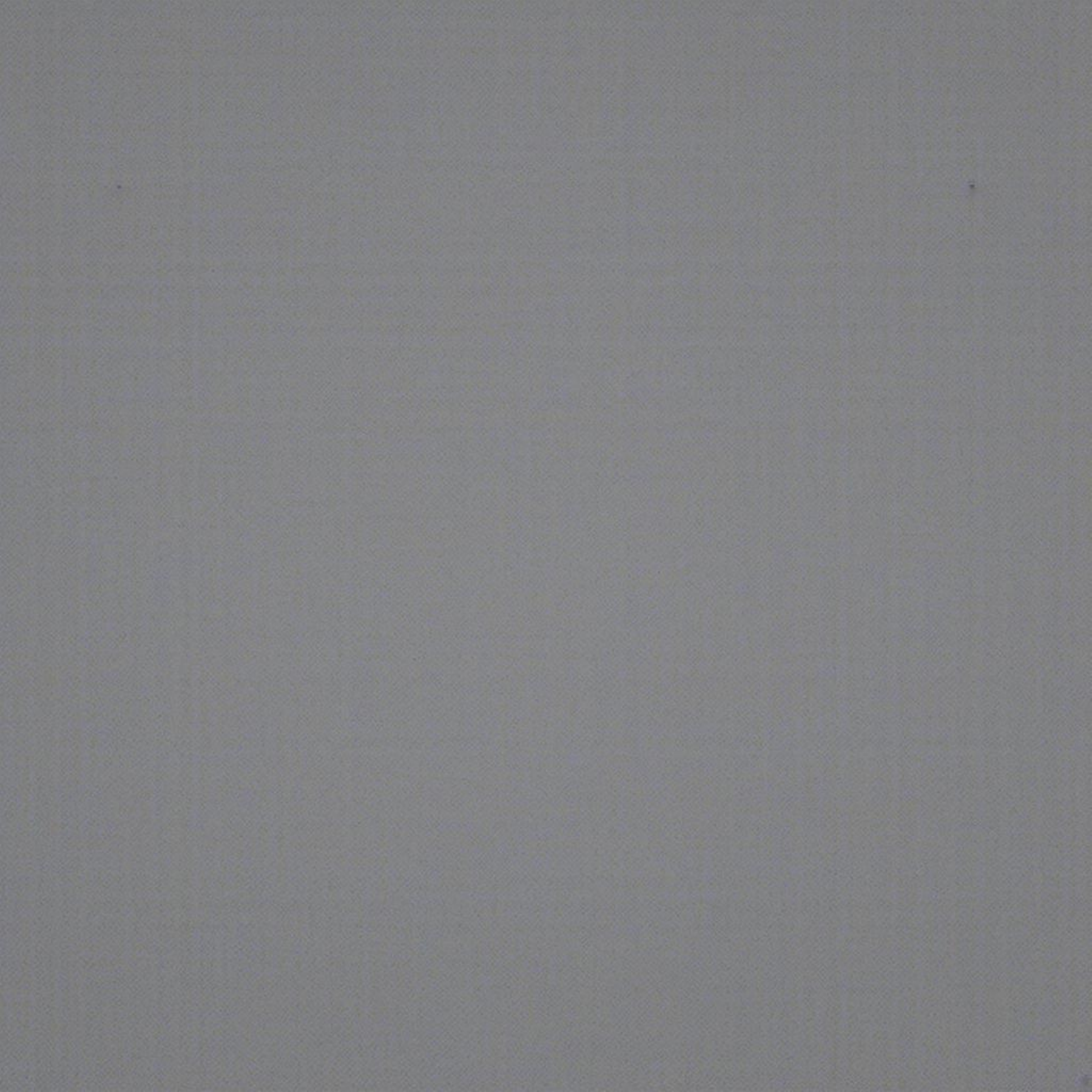} \\

MULTI~\cite{godavarthy2026multi}
& \includegraphics[width=1.7cm,height=1.7cm]{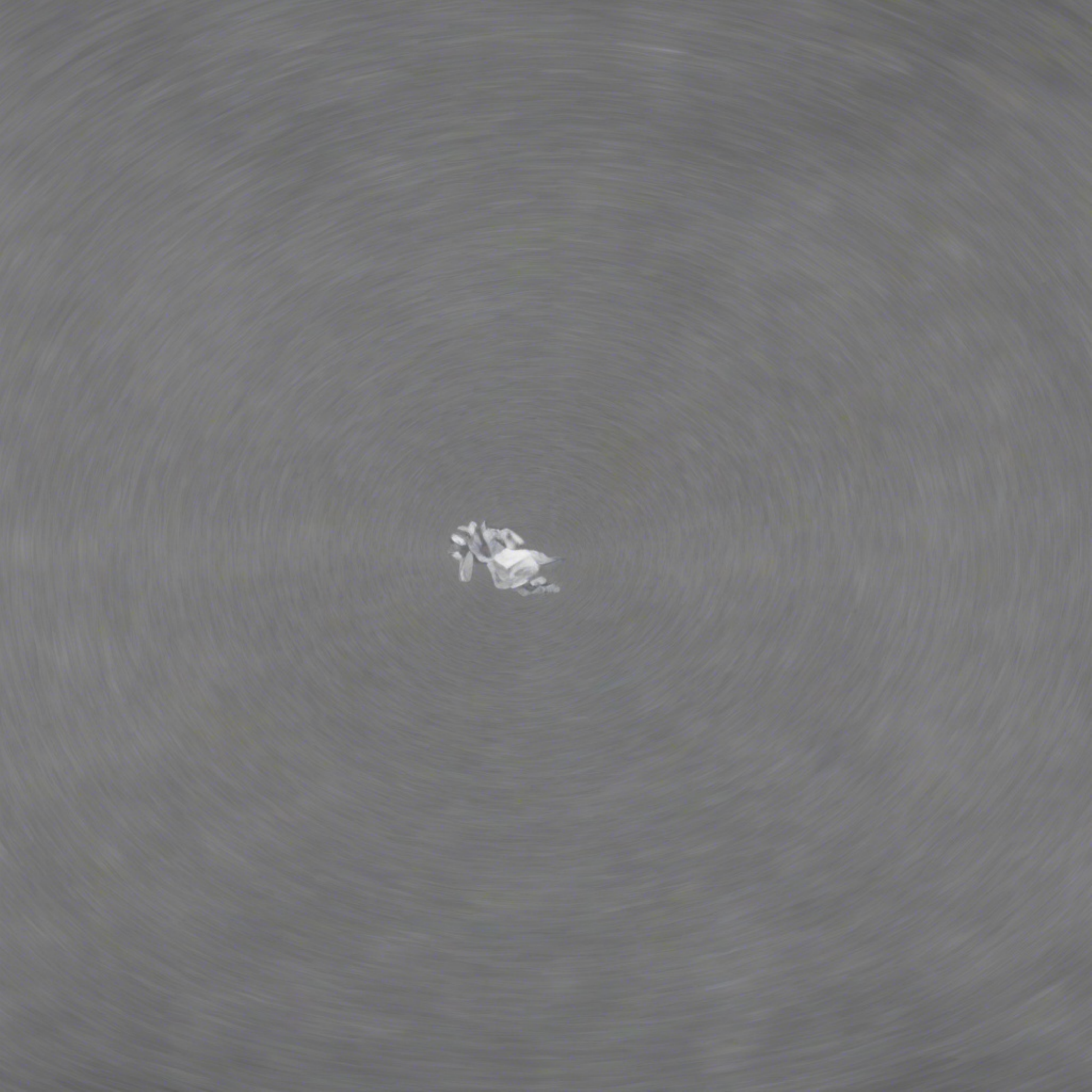}
& \includegraphics[width=1.7cm,height=1.7cm]{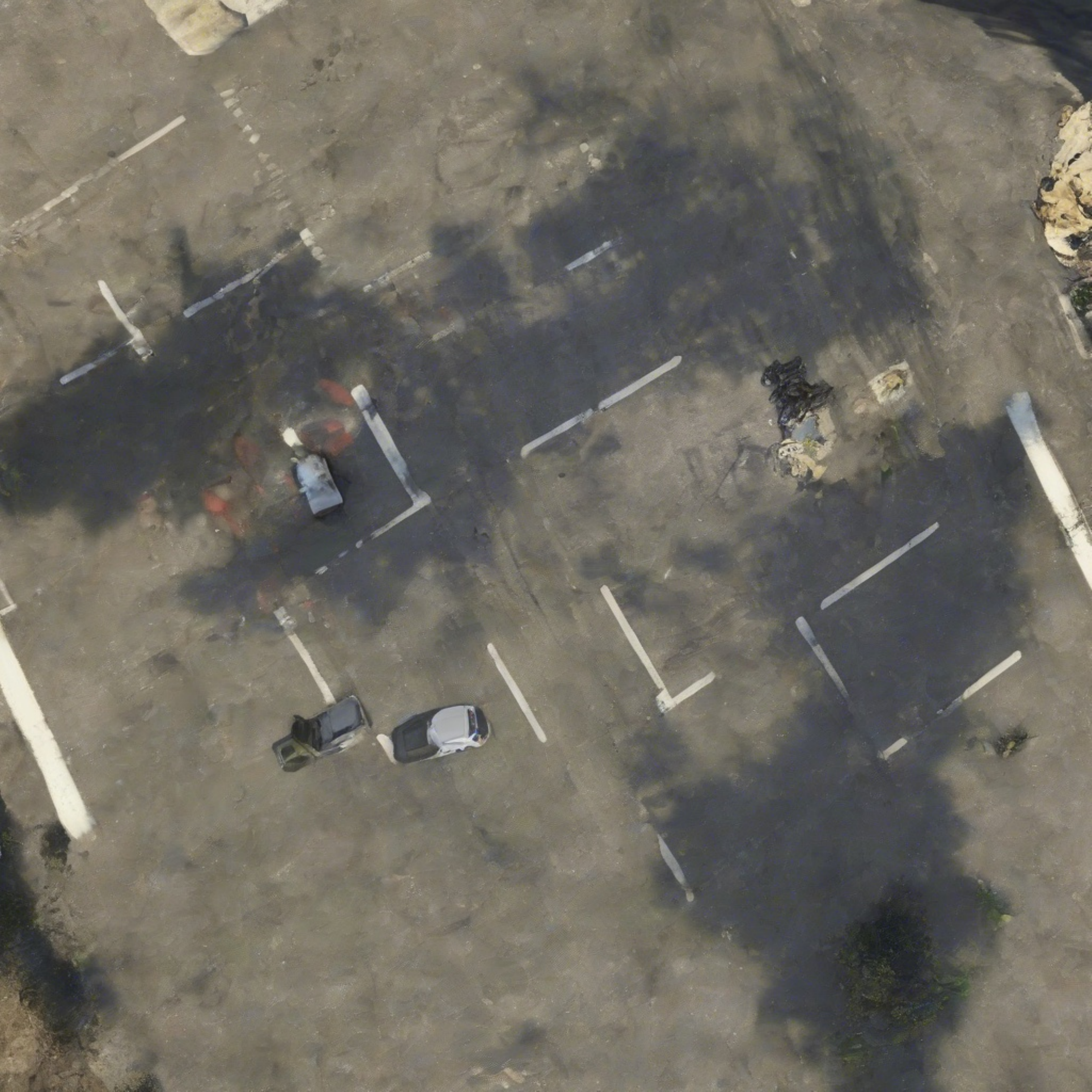}
& \includegraphics[width=1.7cm,height=1.7cm]{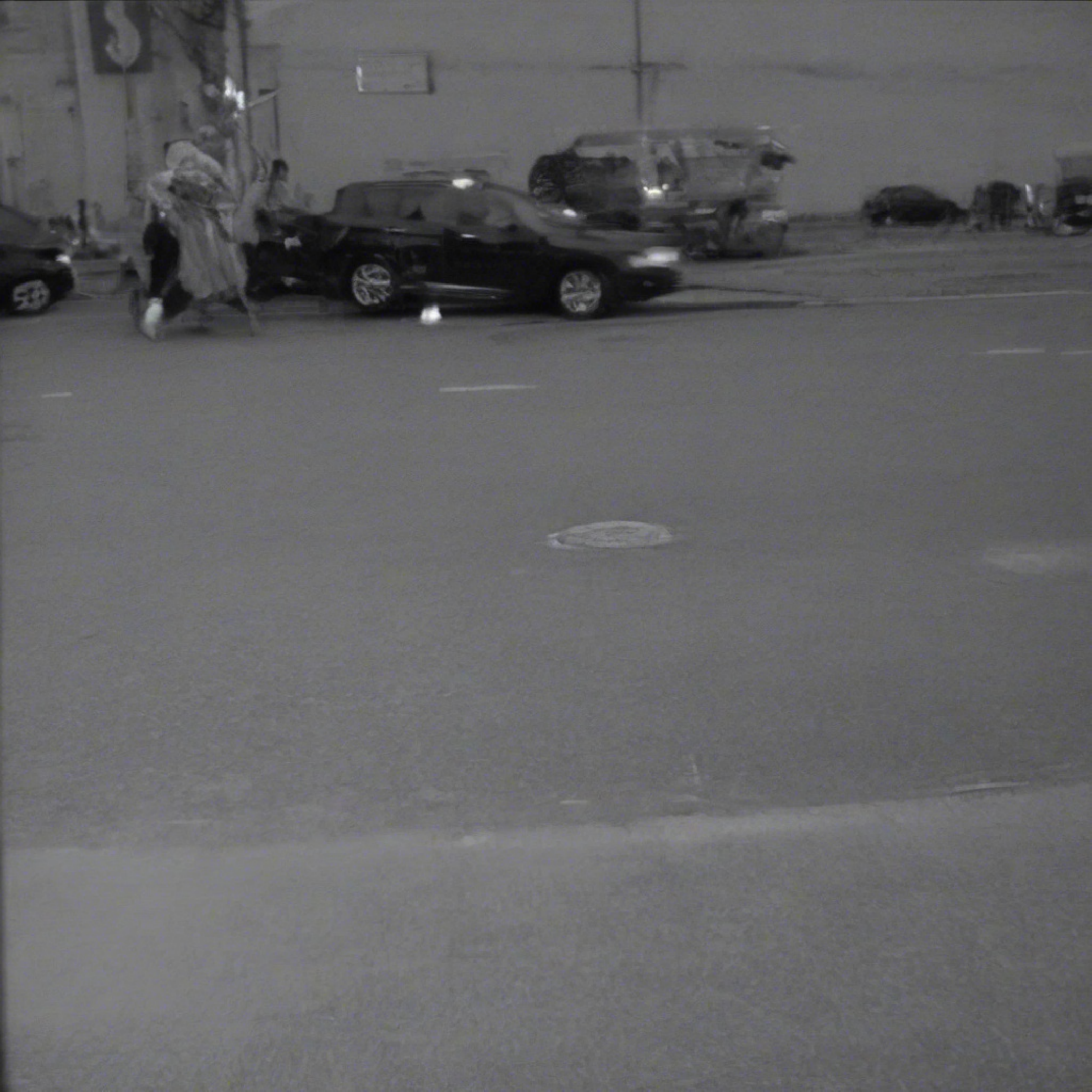}
& \includegraphics[width=1.7cm,height=1.7cm]{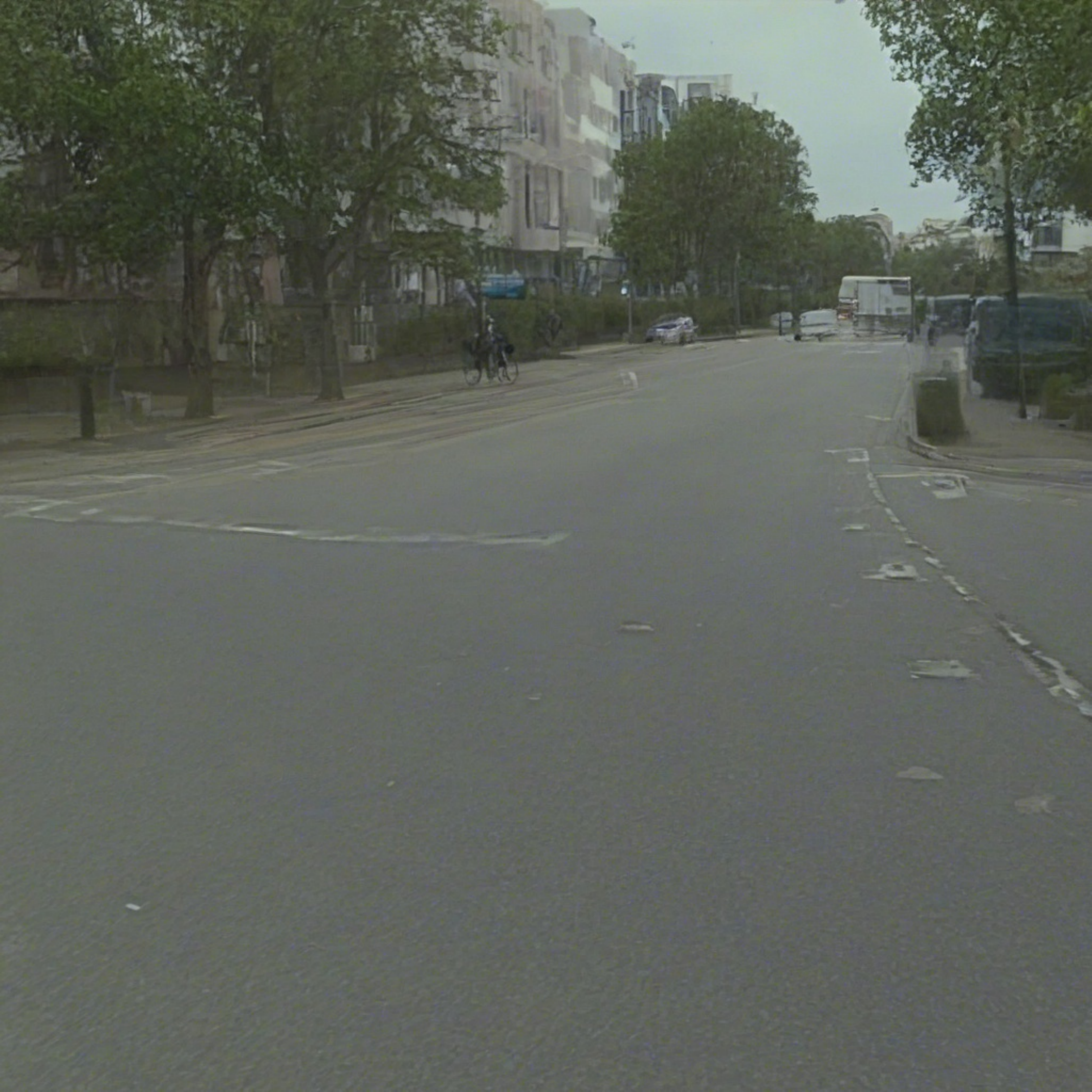}
& \includegraphics[width=1.7cm,height=1.7cm]{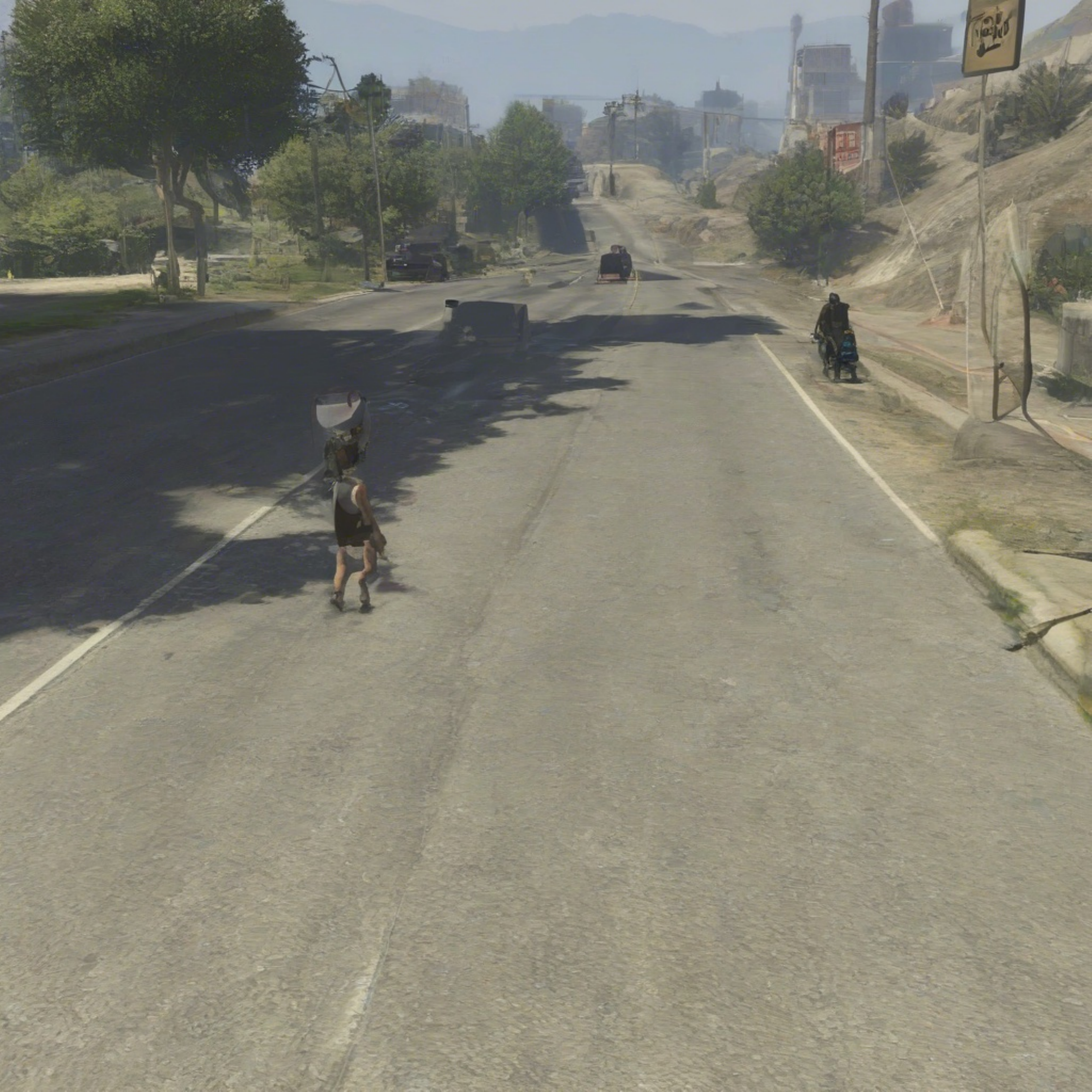}
& \includegraphics[width=1.7cm,height=1.7cm]{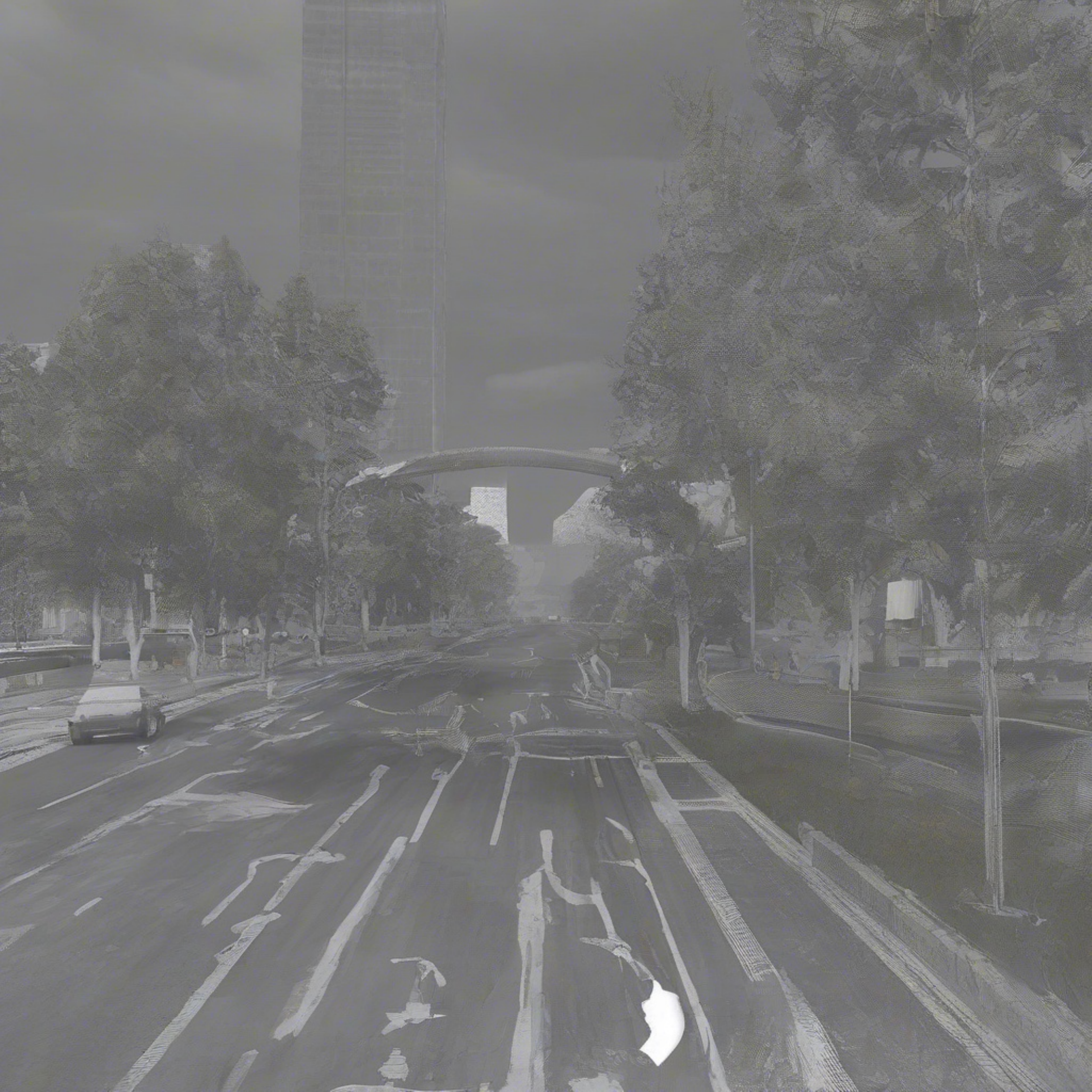}
& \includegraphics[width=1.7cm,height=1.7cm]{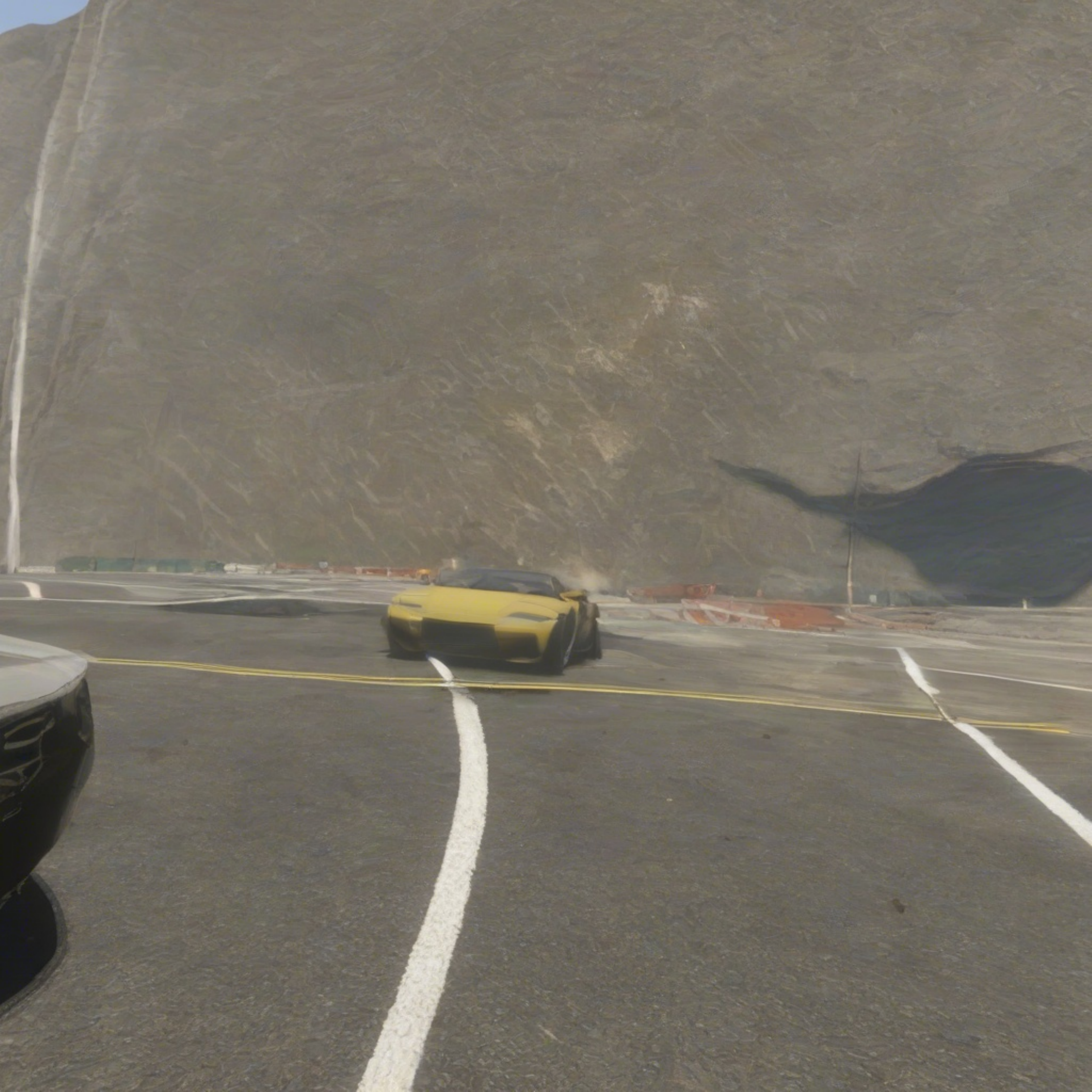}
& \includegraphics[width=1.7cm,height=1.7cm]{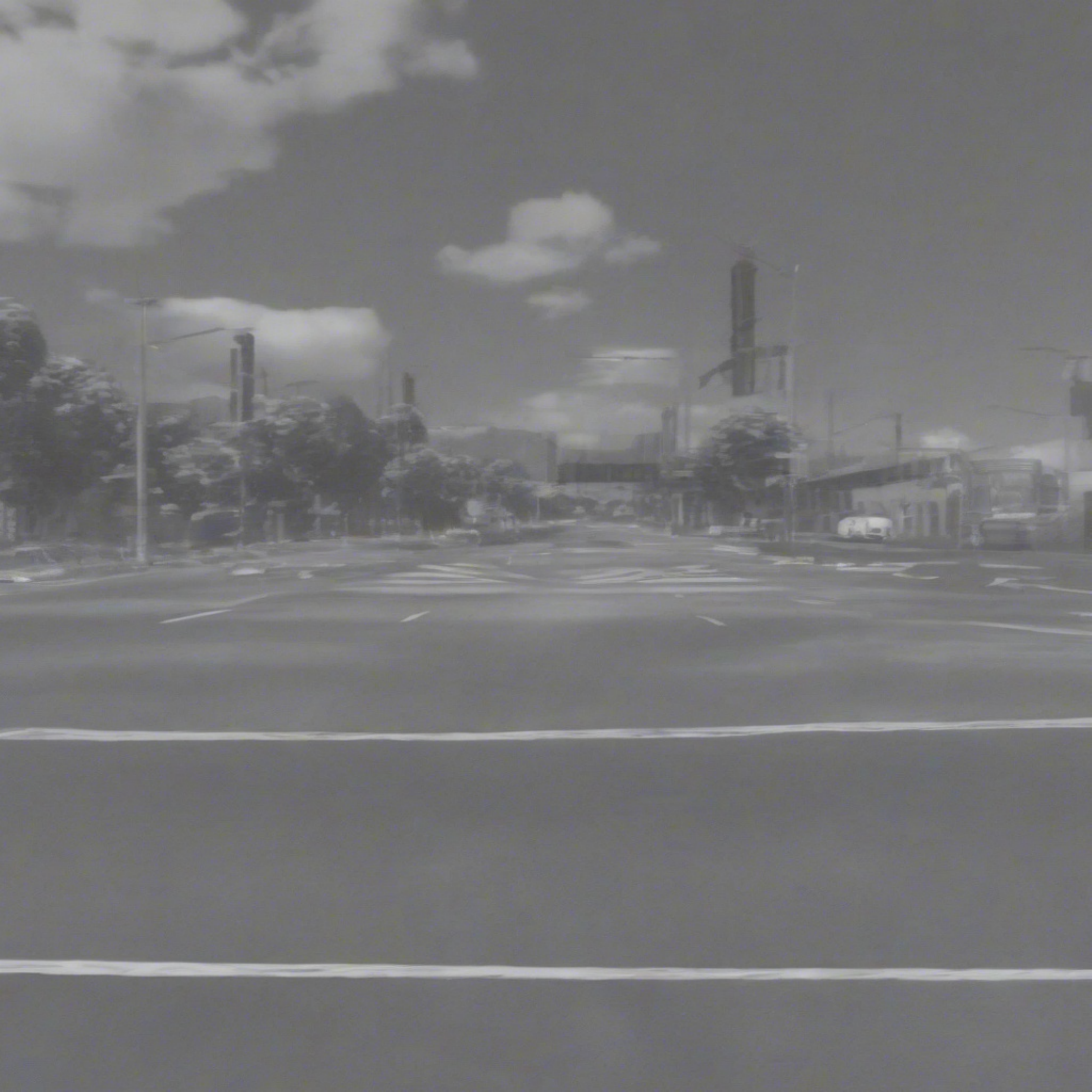} \\

X-MULTI
& \includegraphics[width=1.7cm,height=1.7cm]{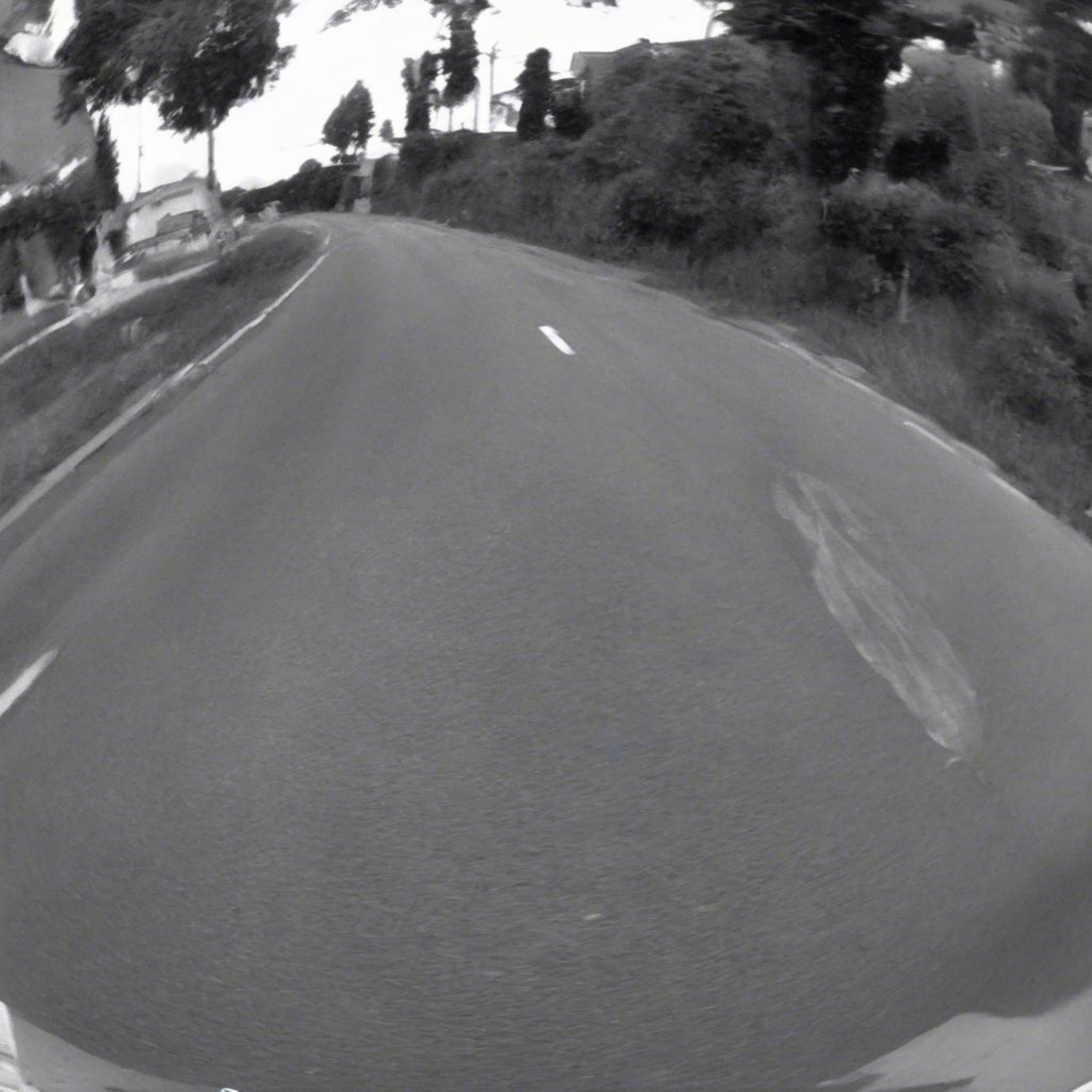}
& \includegraphics[width=1.7cm,height=1.7cm]{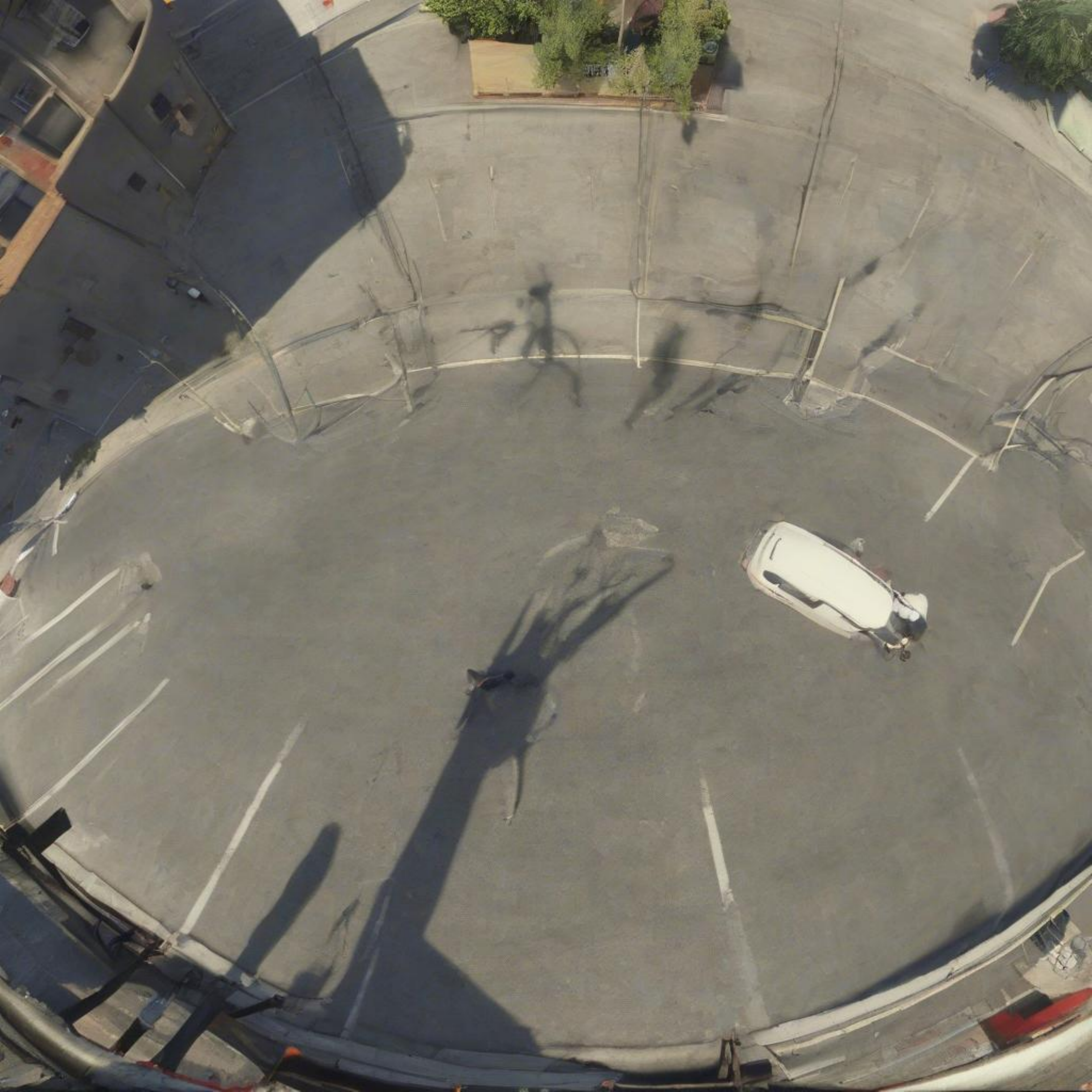}
& \includegraphics[width=1.7cm,height=1.7cm]{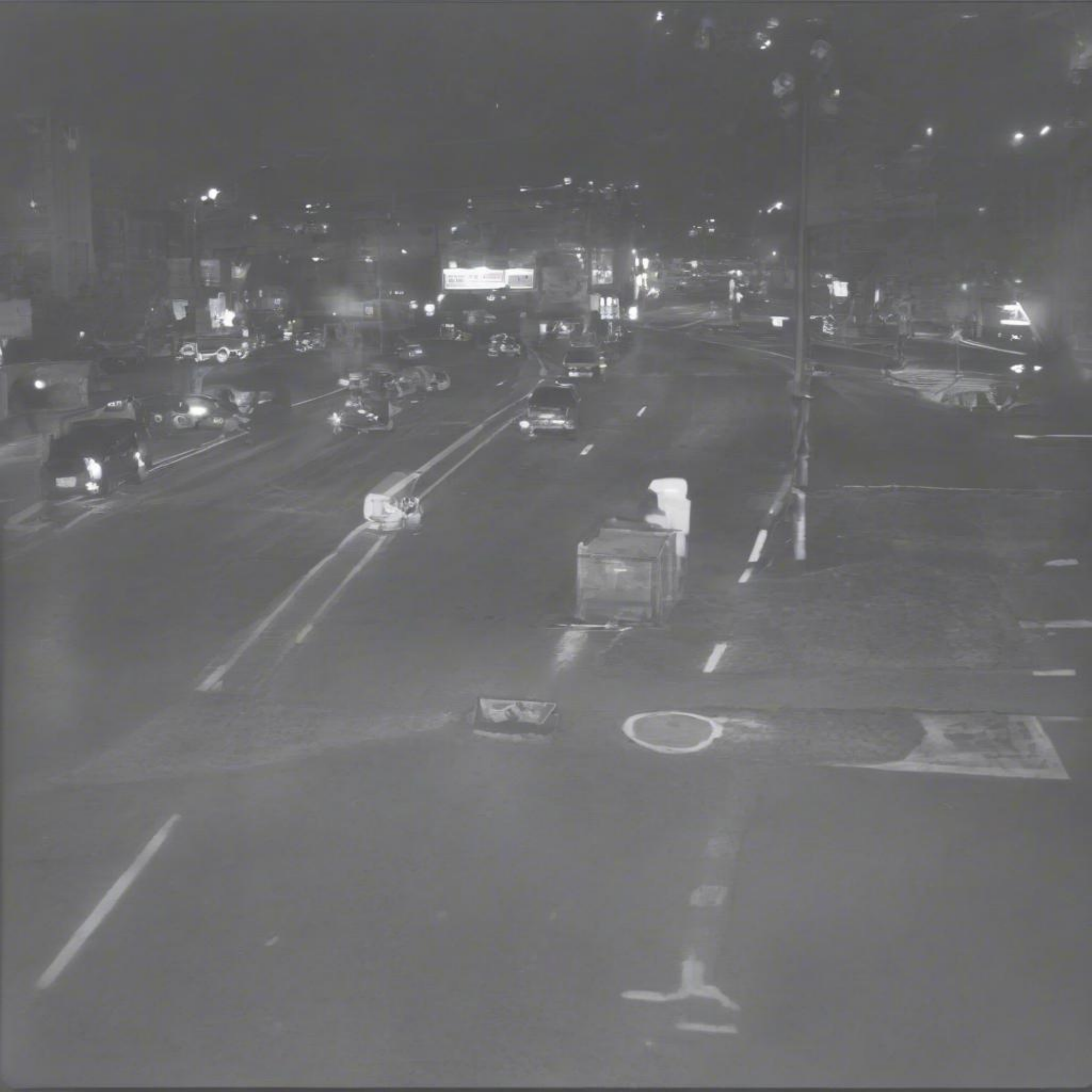}
& \includegraphics[width=1.7cm,height=1.7cm]{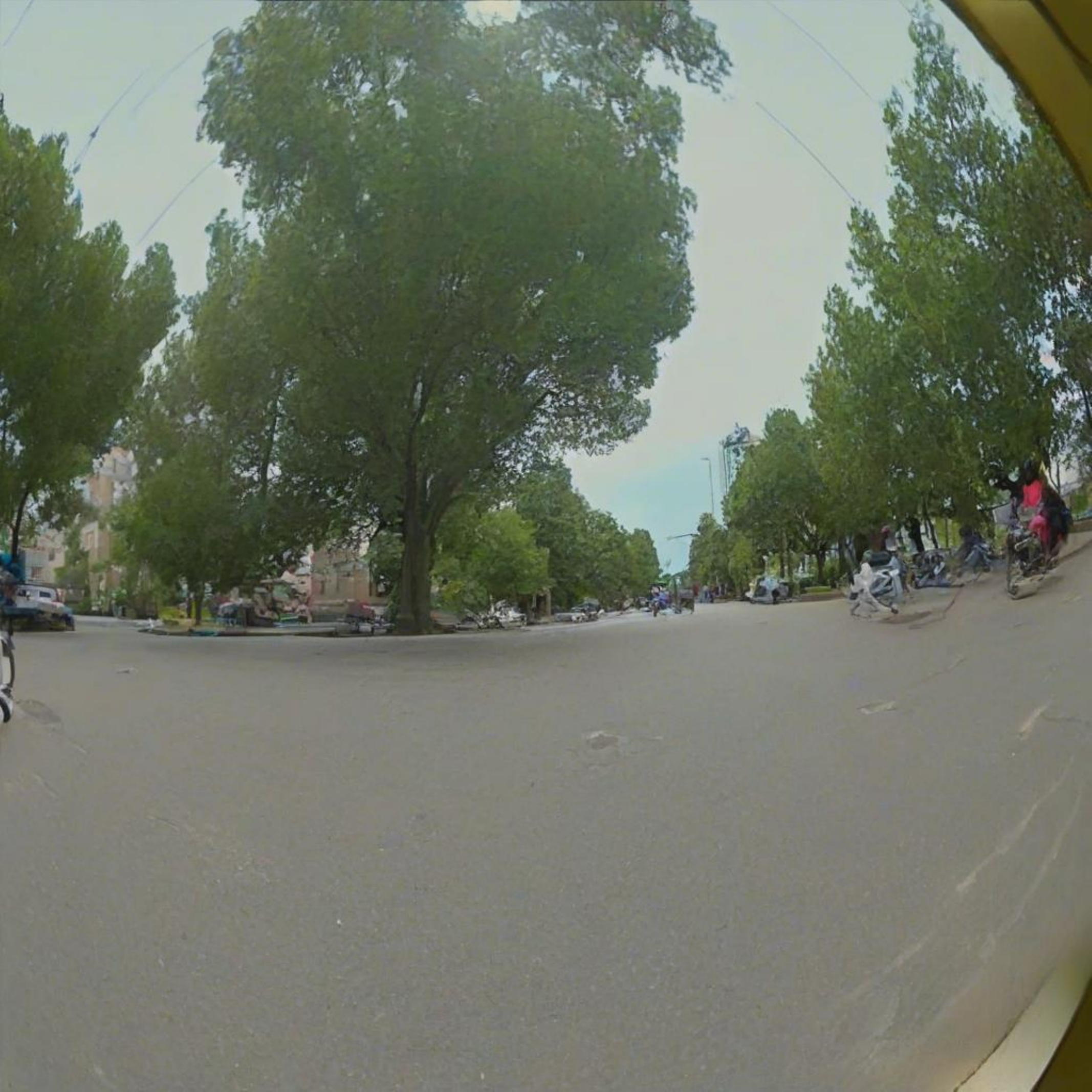}
& \includegraphics[width=1.7cm,height=1.7cm]{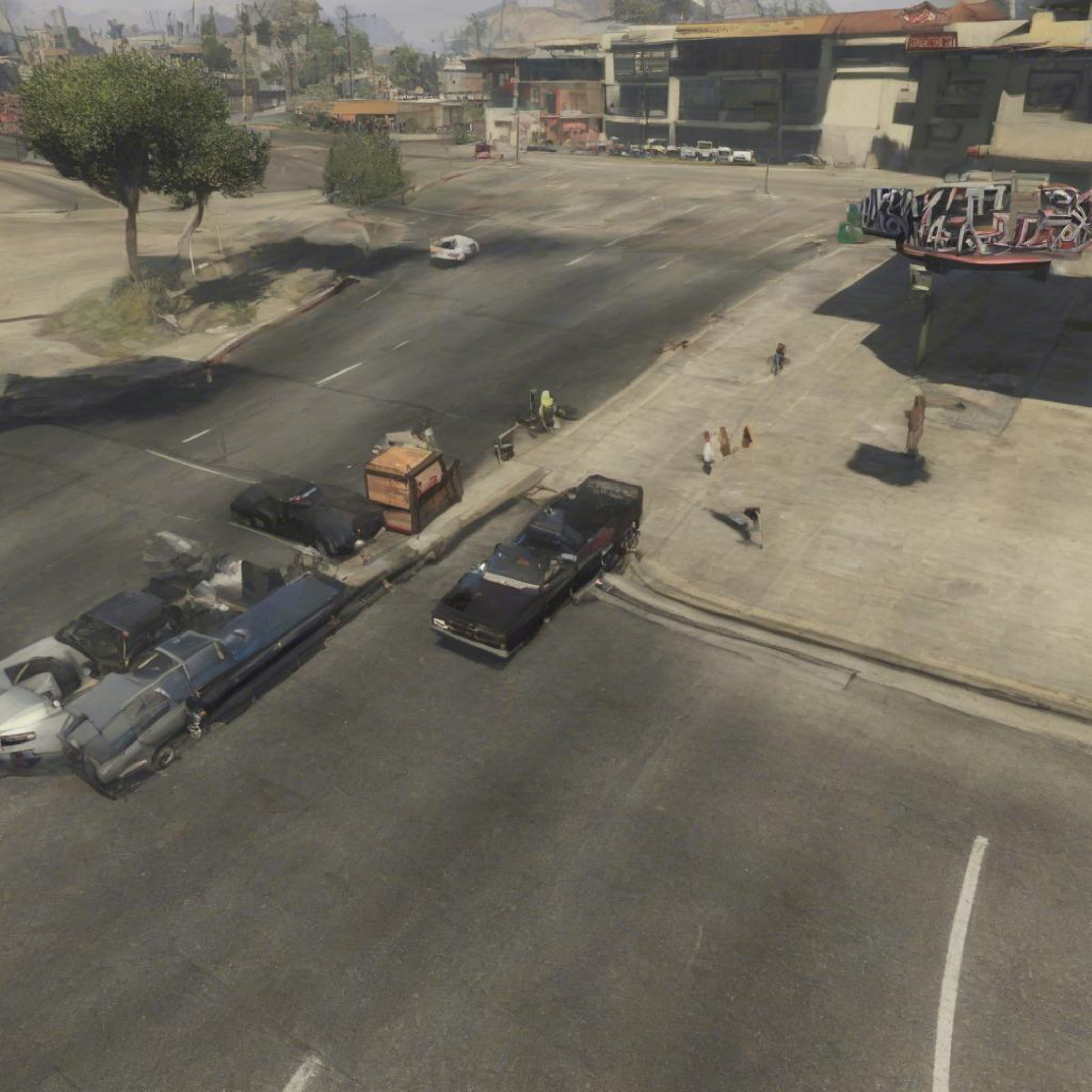}
& \includegraphics[width=1.7cm,height=1.7cm]{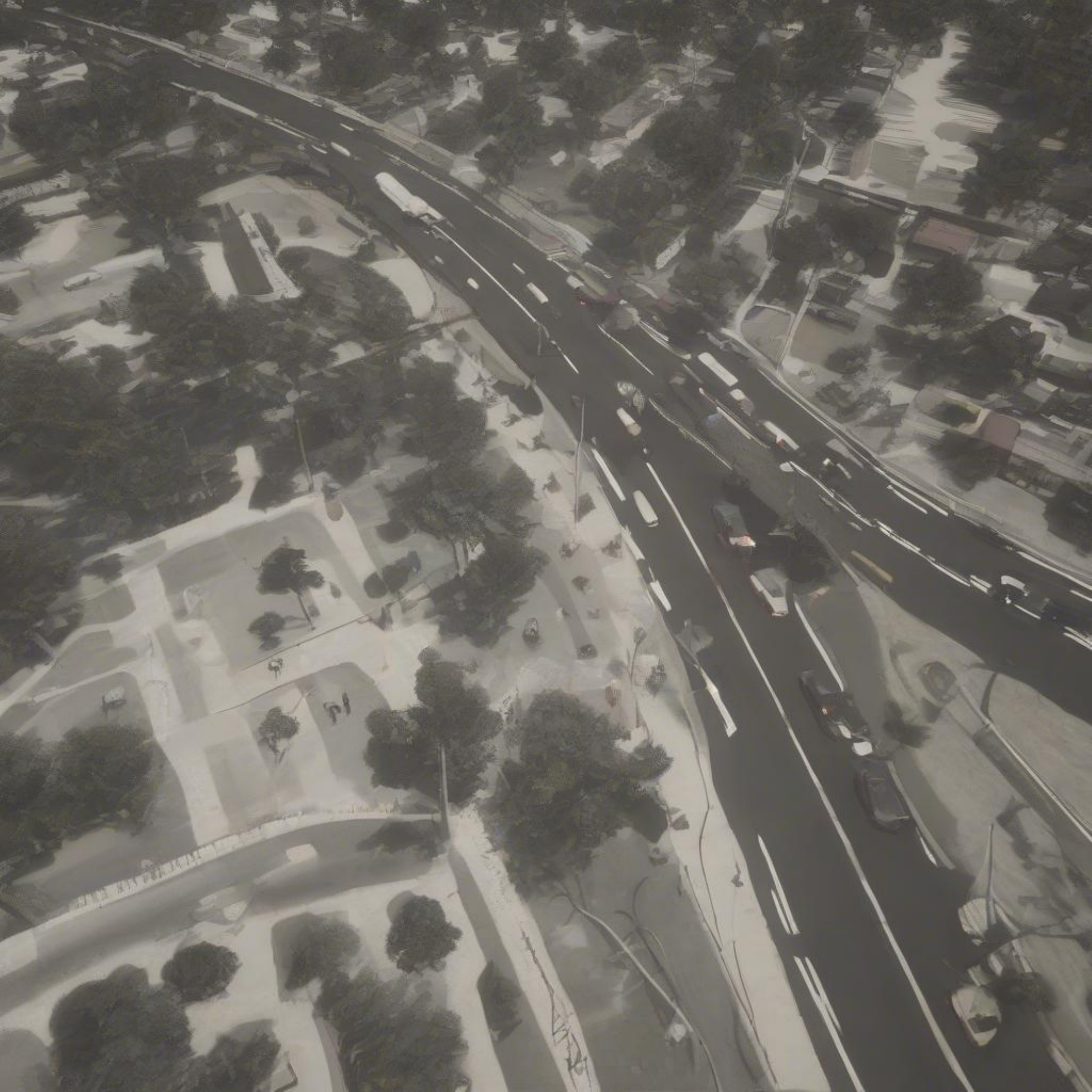}
& \includegraphics[width=1.7cm,height=1.7cm]{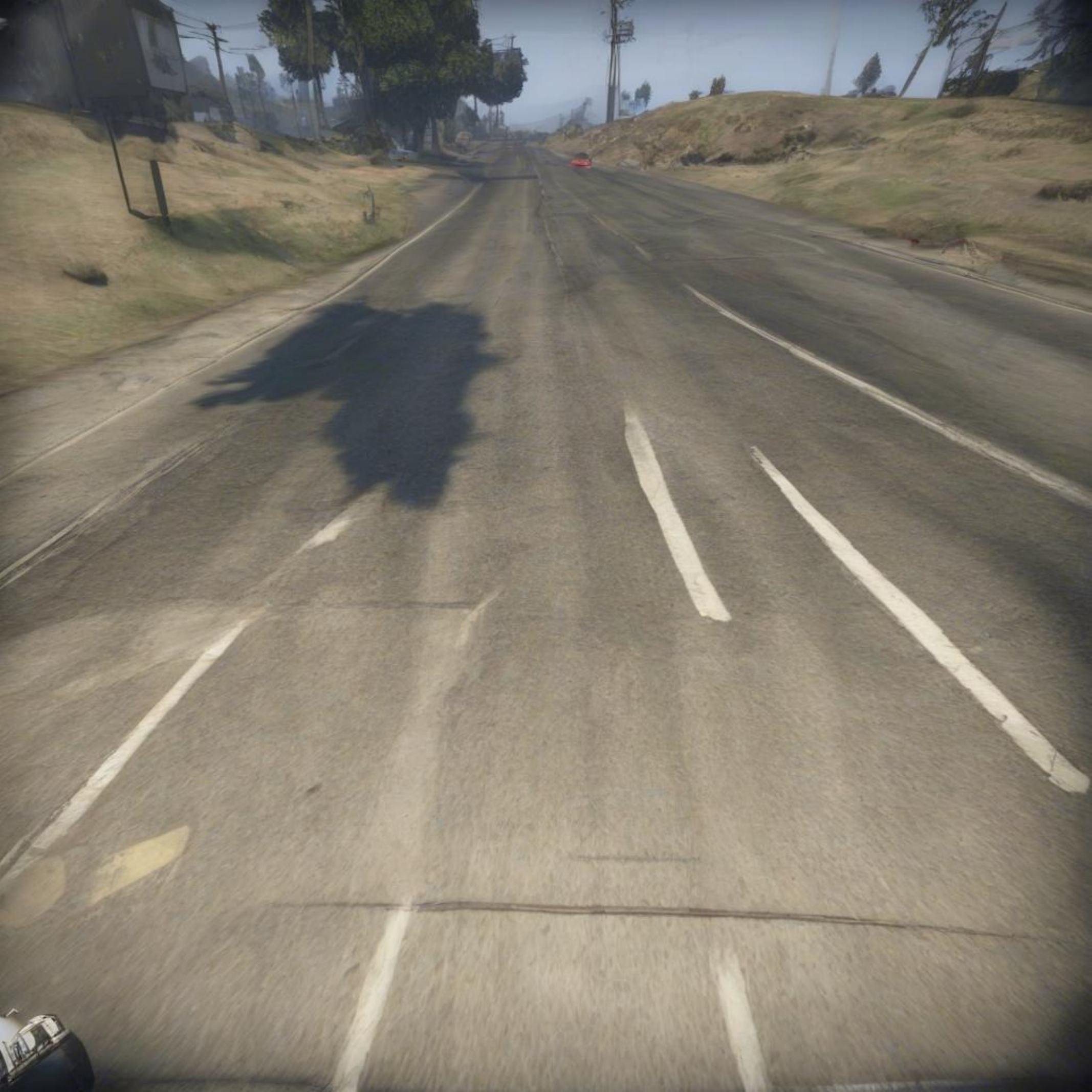}
& \includegraphics[width=1.7cm,height=1.7cm]{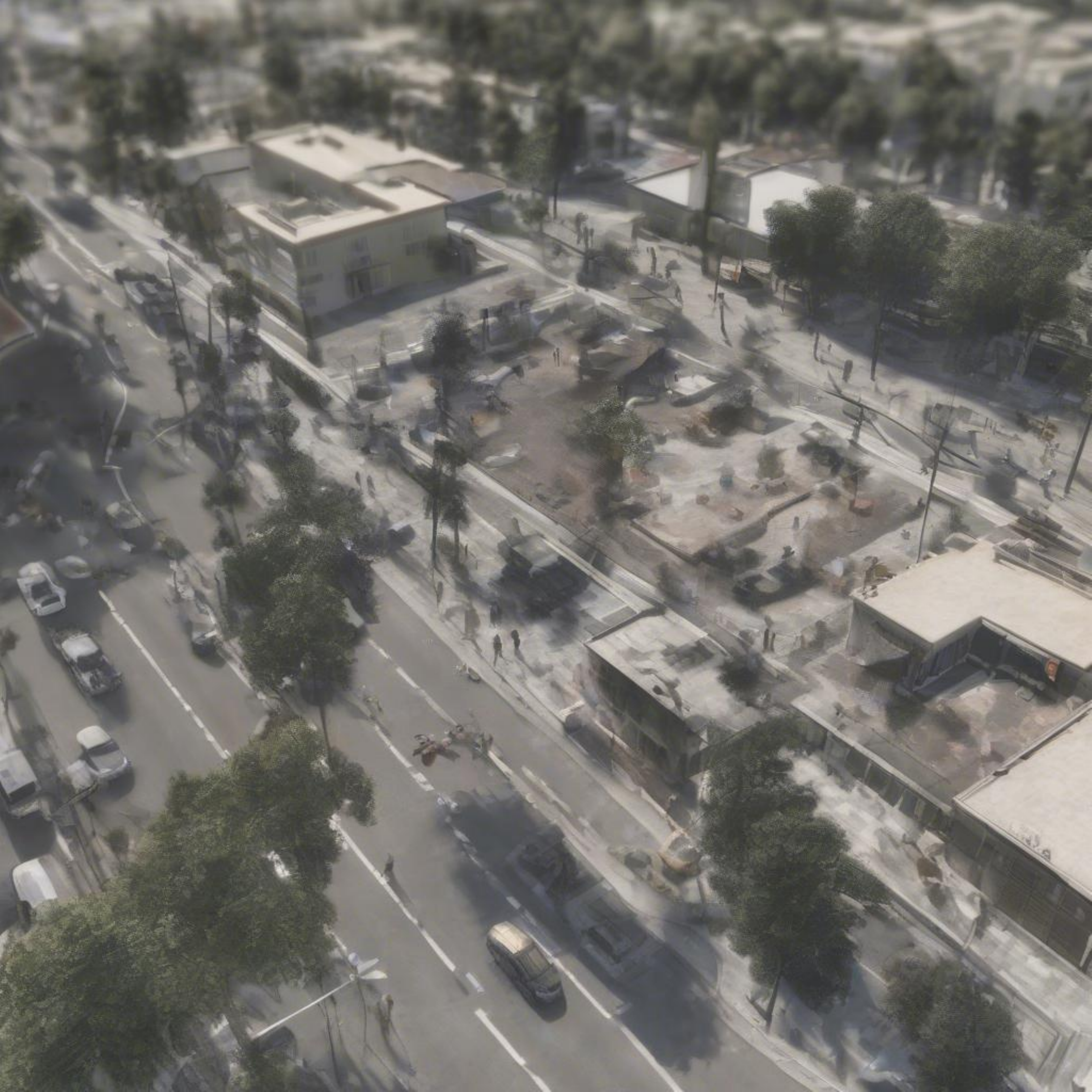} \\

\bottomrule
\end{tabular}}
\label{tab:qual_novel}
\end{table*}

\begin{table*}[t]
\centering
\caption{\textbf{Qualitative comparison of novel factor combinations with ControlNets.}
Image-to-image generation using ControlNet guidance with altered factors, showing X-MULTI (ours) achieves best factor adherence.}
\scriptsize
\setlength{\tabcolsep}{2pt}
\renewcommand{\arraystretch}{1.05}
\resizebox{\textwidth}{!}{
\begin{tabular}{lcccccccc}
\toprule

\textbf{Domain}
& simulation & real & real & real & real & real & real & simulation \\
\textbf{Sensor}
& rgb & rgb-thermal & rgb & event & rgb-thermal & rgb & event & rgb \\
\textbf{Lens}
& normal & normal & normal & normal & normal & fisheye & normal & normal \\
\textbf{Viewpoint}
& front & front & back & front & front & front & front & front \\

\midrule

Original
& \includegraphics[width=1.7cm,height=1.7cm]{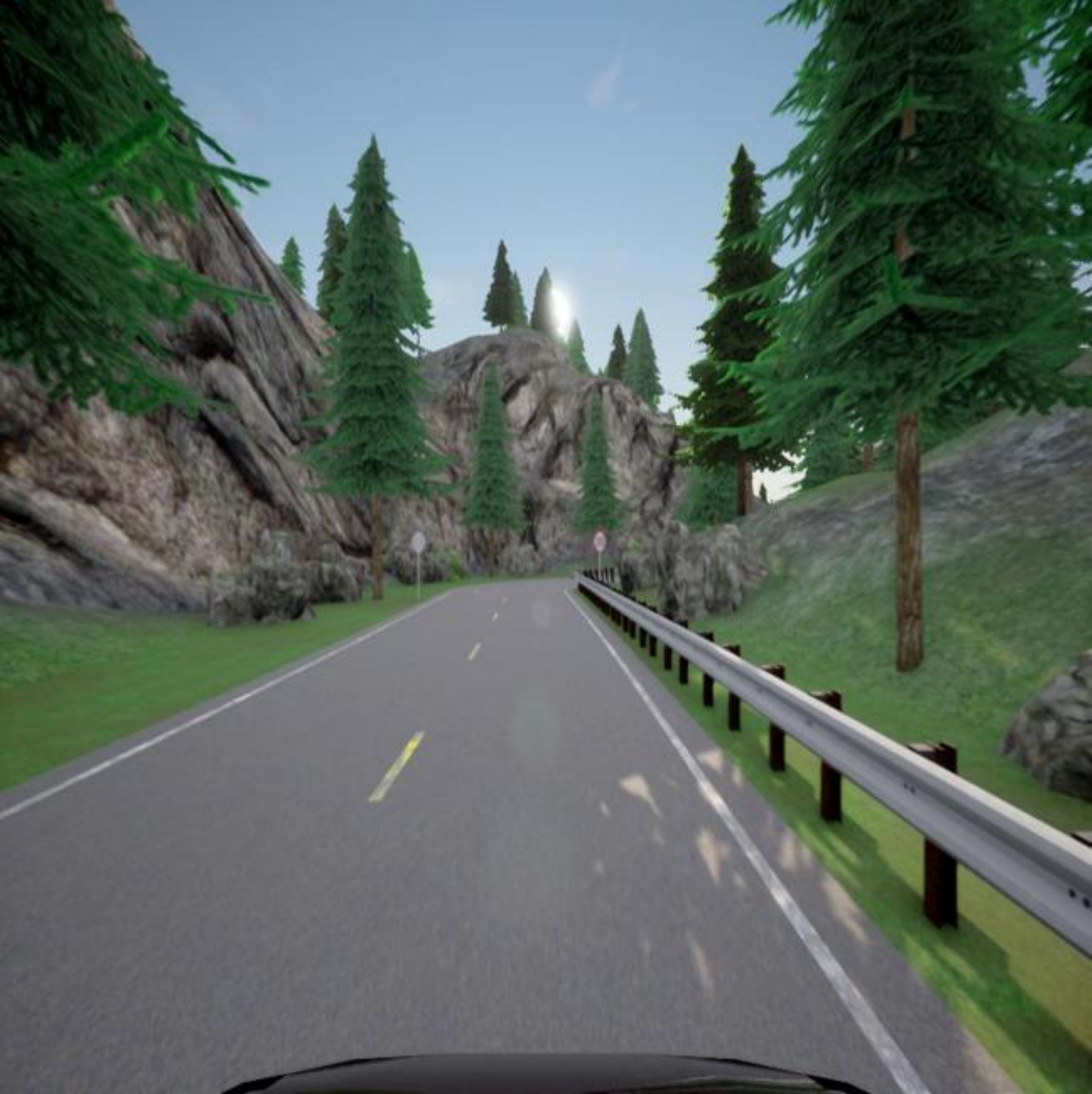}
& \includegraphics[width=1.7cm,height=1.7cm]{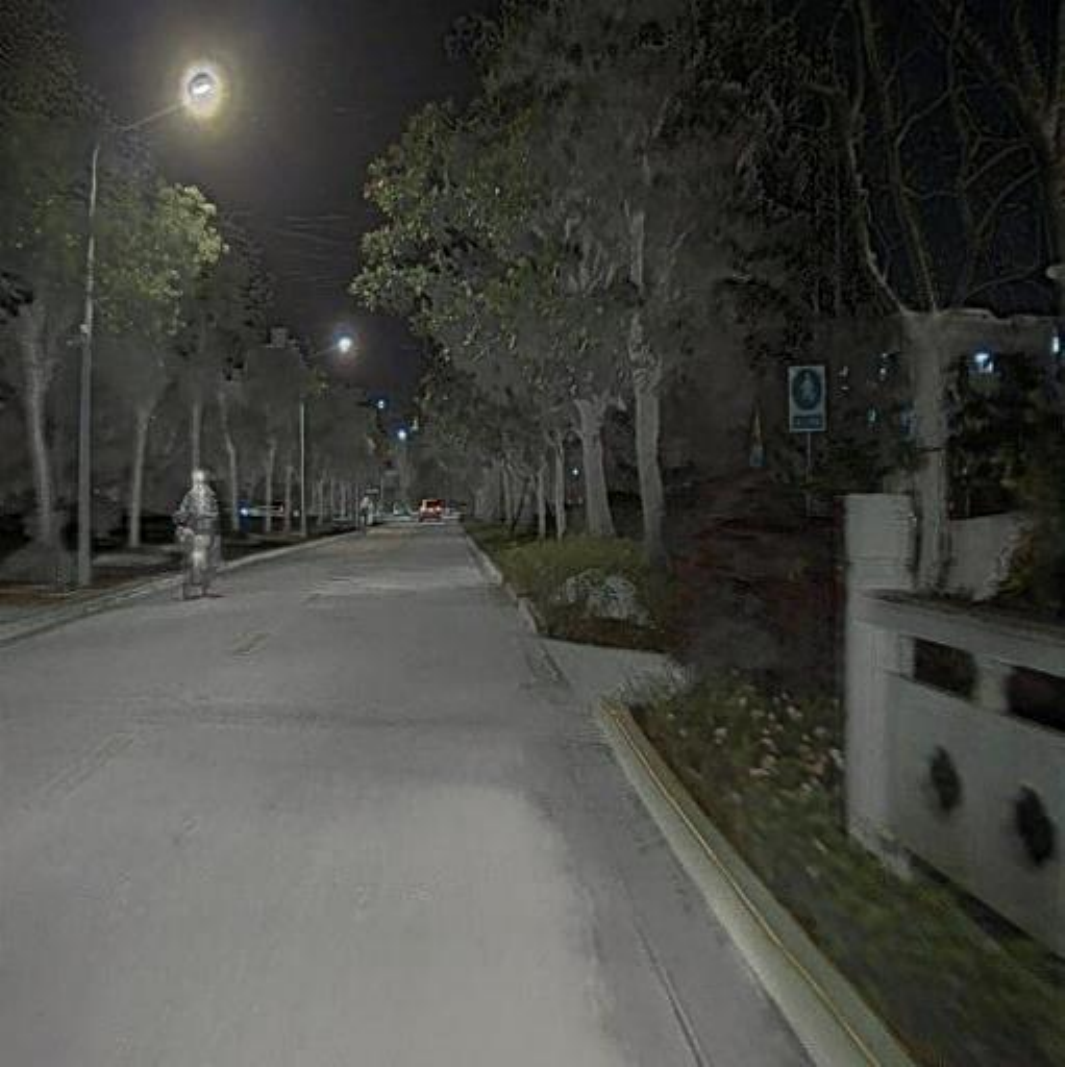}
& \includegraphics[width=1.7cm,height=1.7cm]{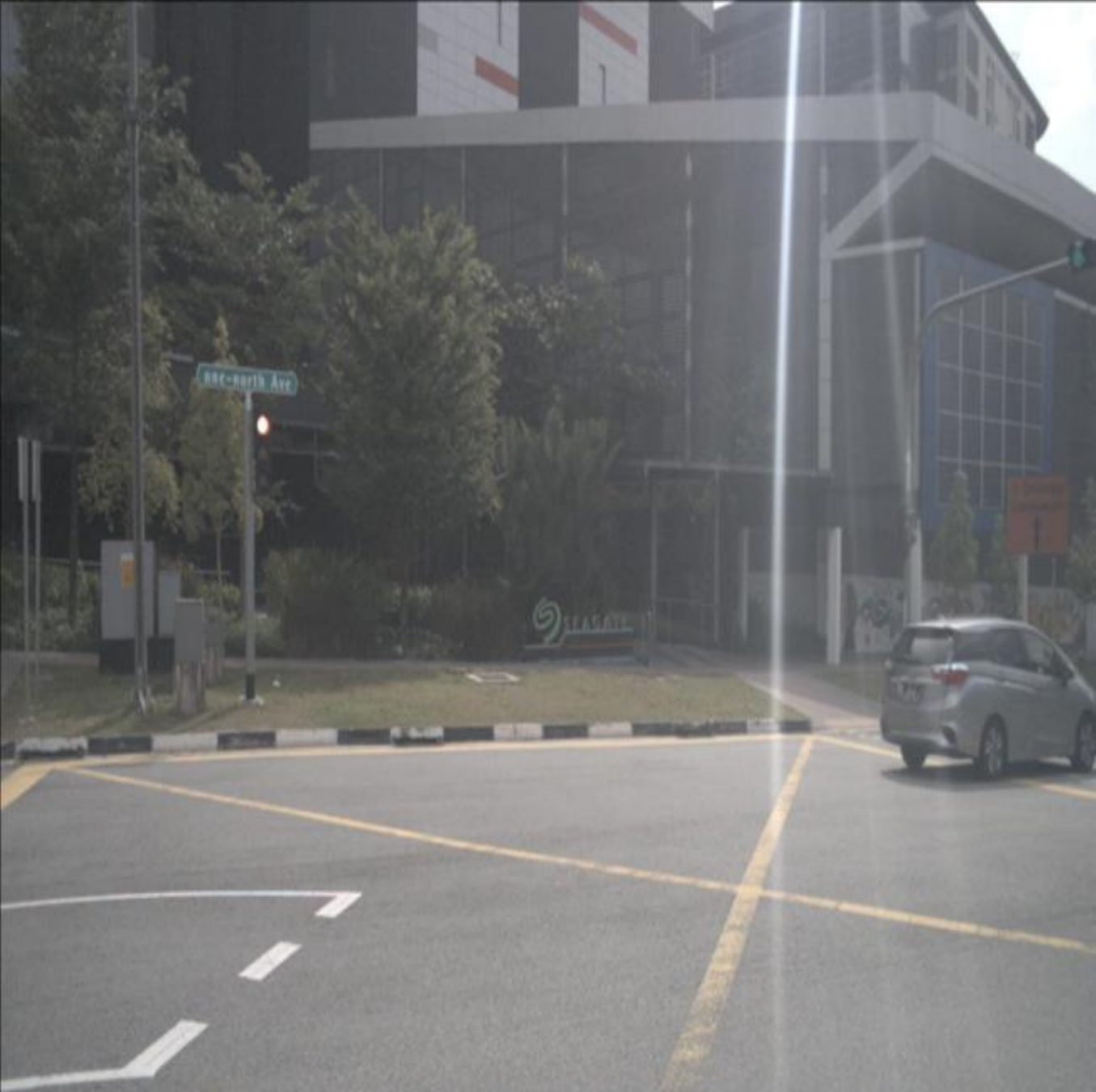}
& \includegraphics[width=1.7cm,height=1.7cm]{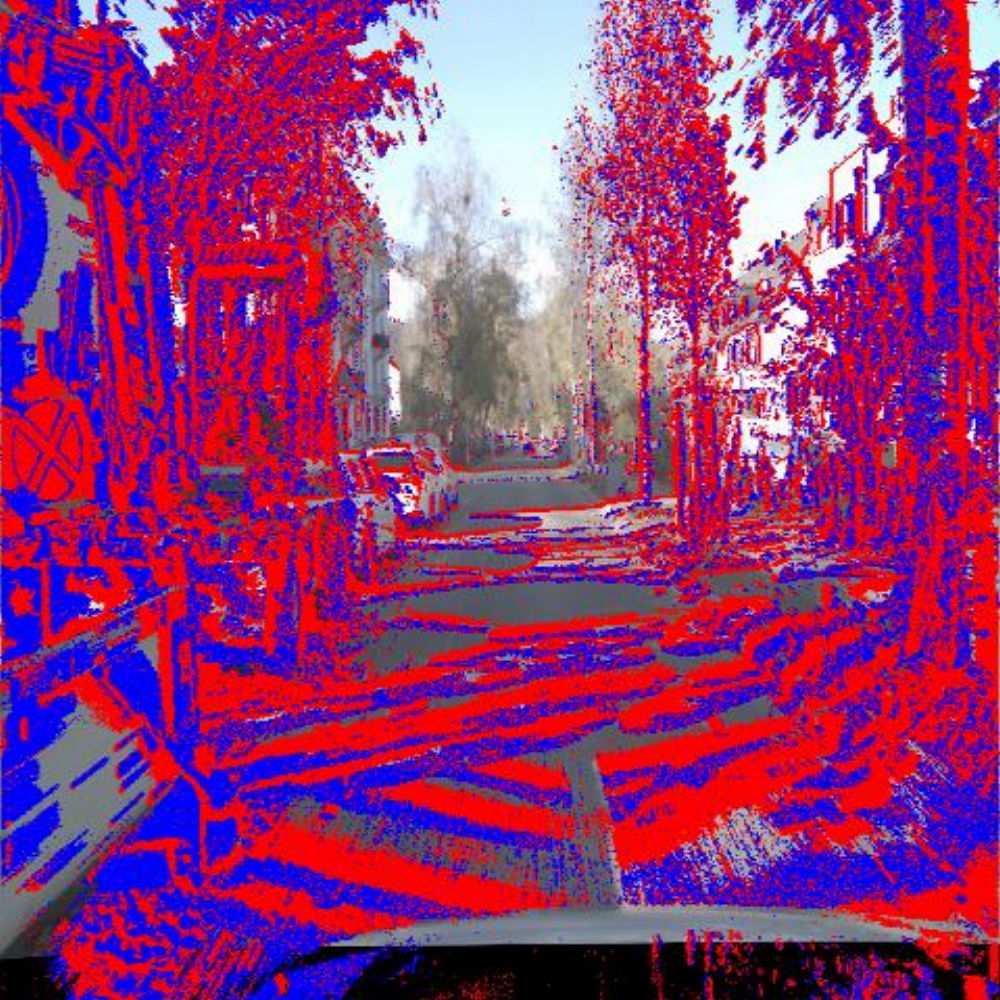}
& \includegraphics[width=1.7cm,height=1.7cm]{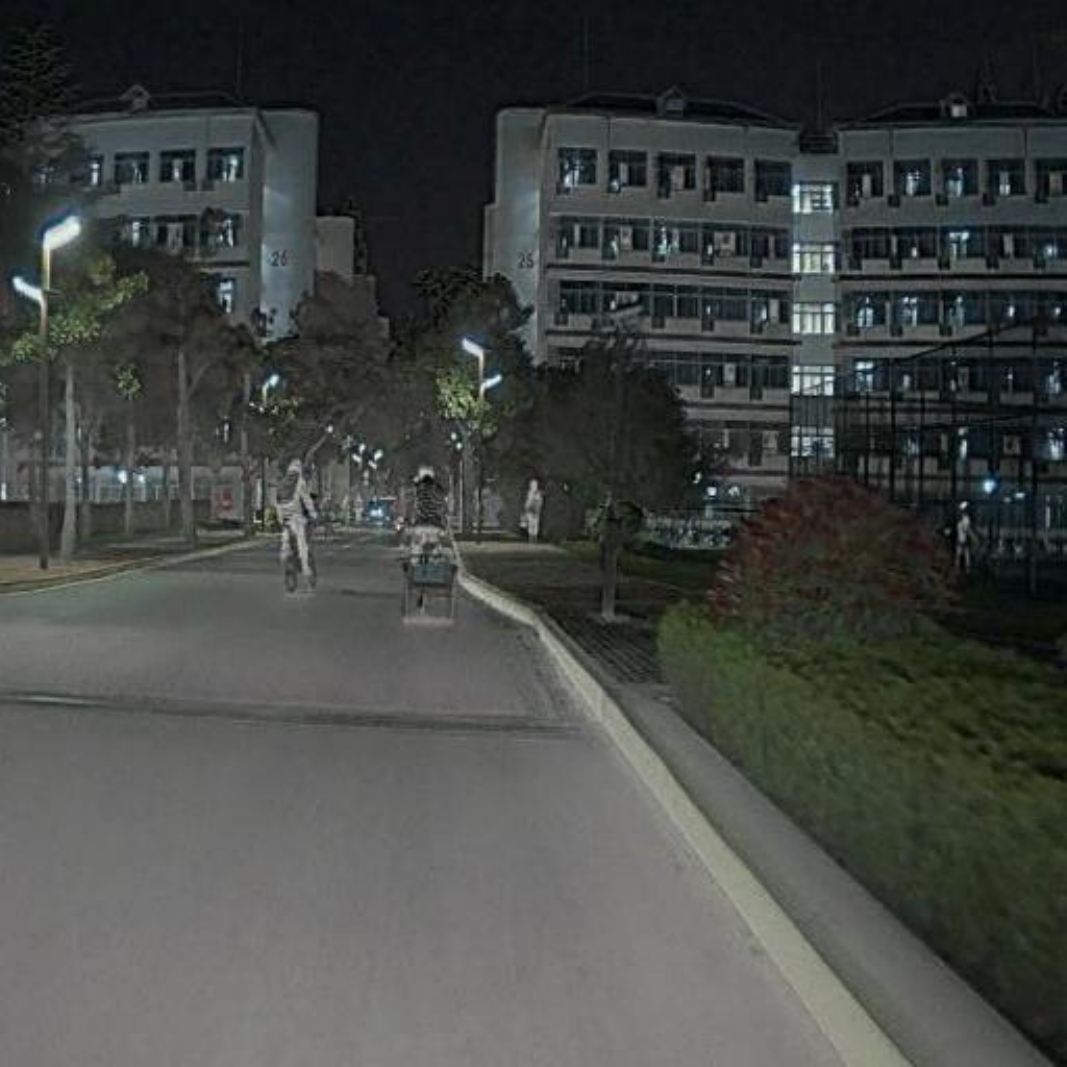}
& \includegraphics[width=1.7cm,height=1.7cm]{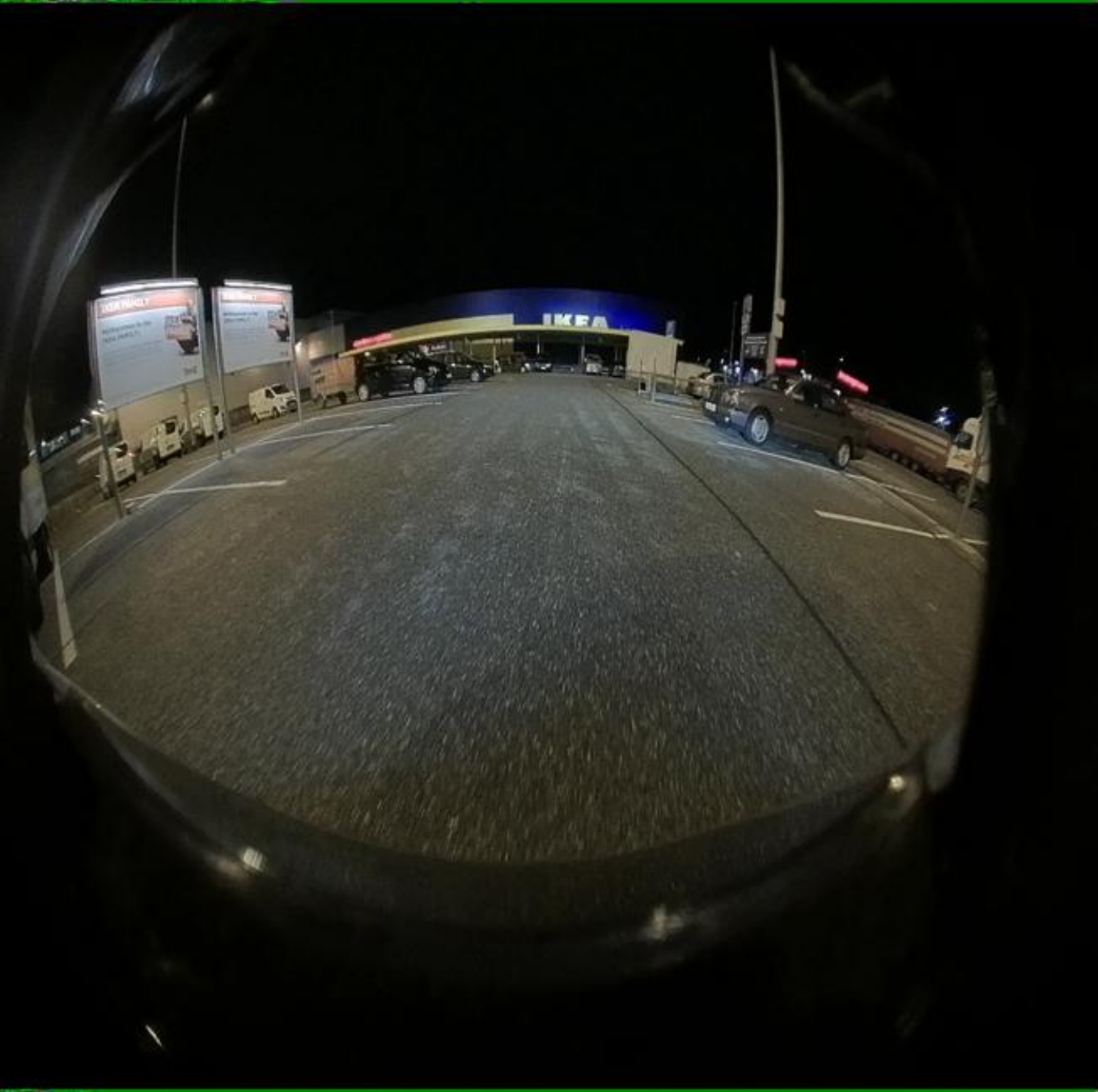}
& \includegraphics[width=1.7cm,height=1.7cm]{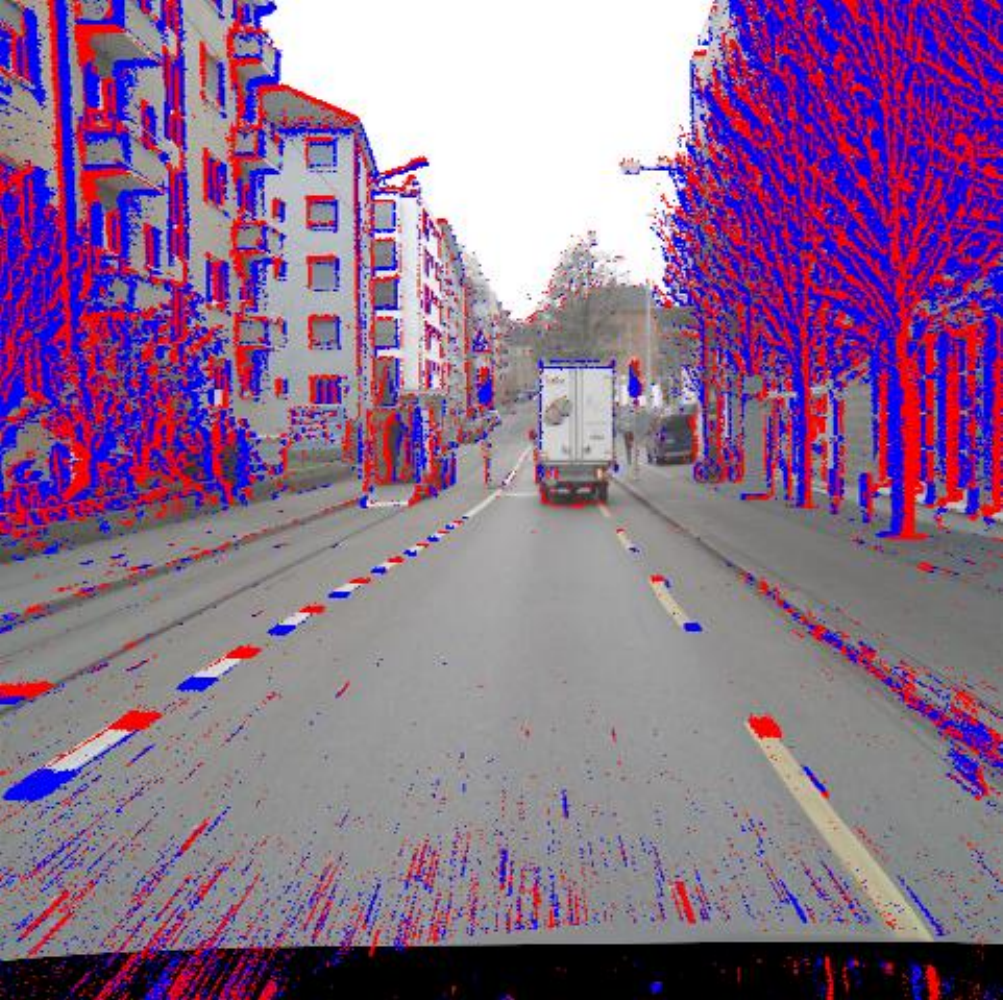}
& \includegraphics[width=1.7cm,height=1.7cm]{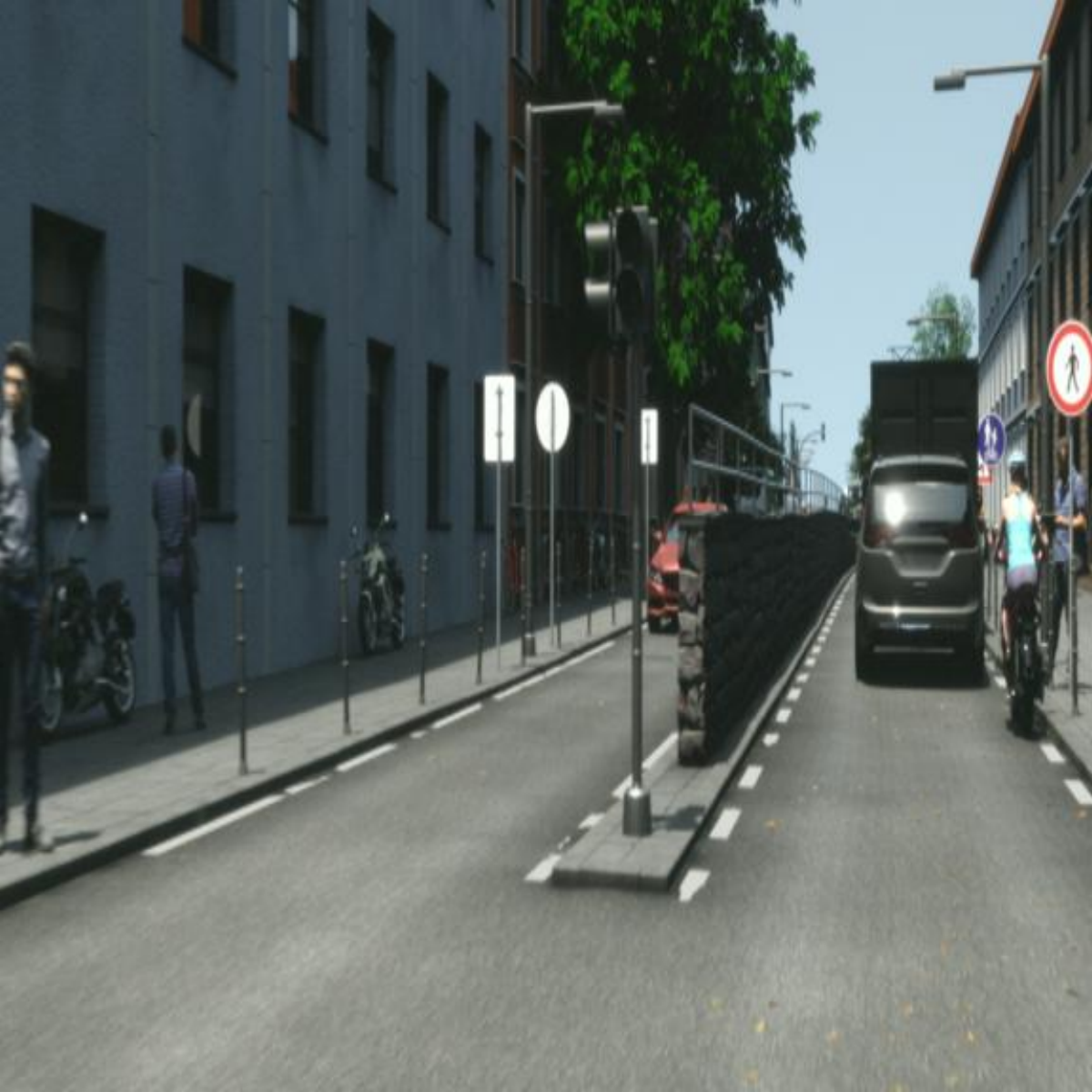} \\

ControlNet Input
& \includegraphics[width=1.7cm,height=1.7cm]{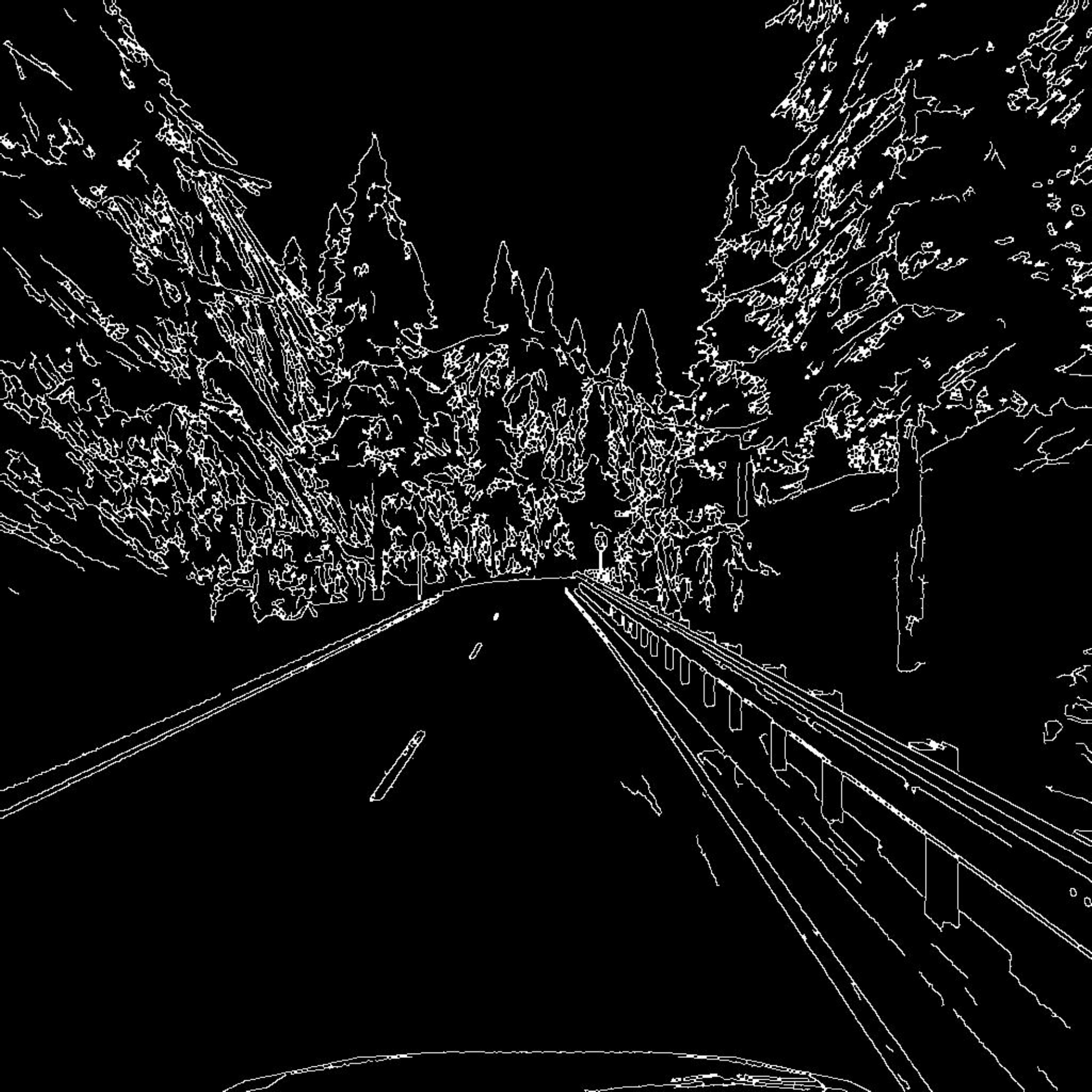}
& \includegraphics[width=1.7cm,height=1.7cm]{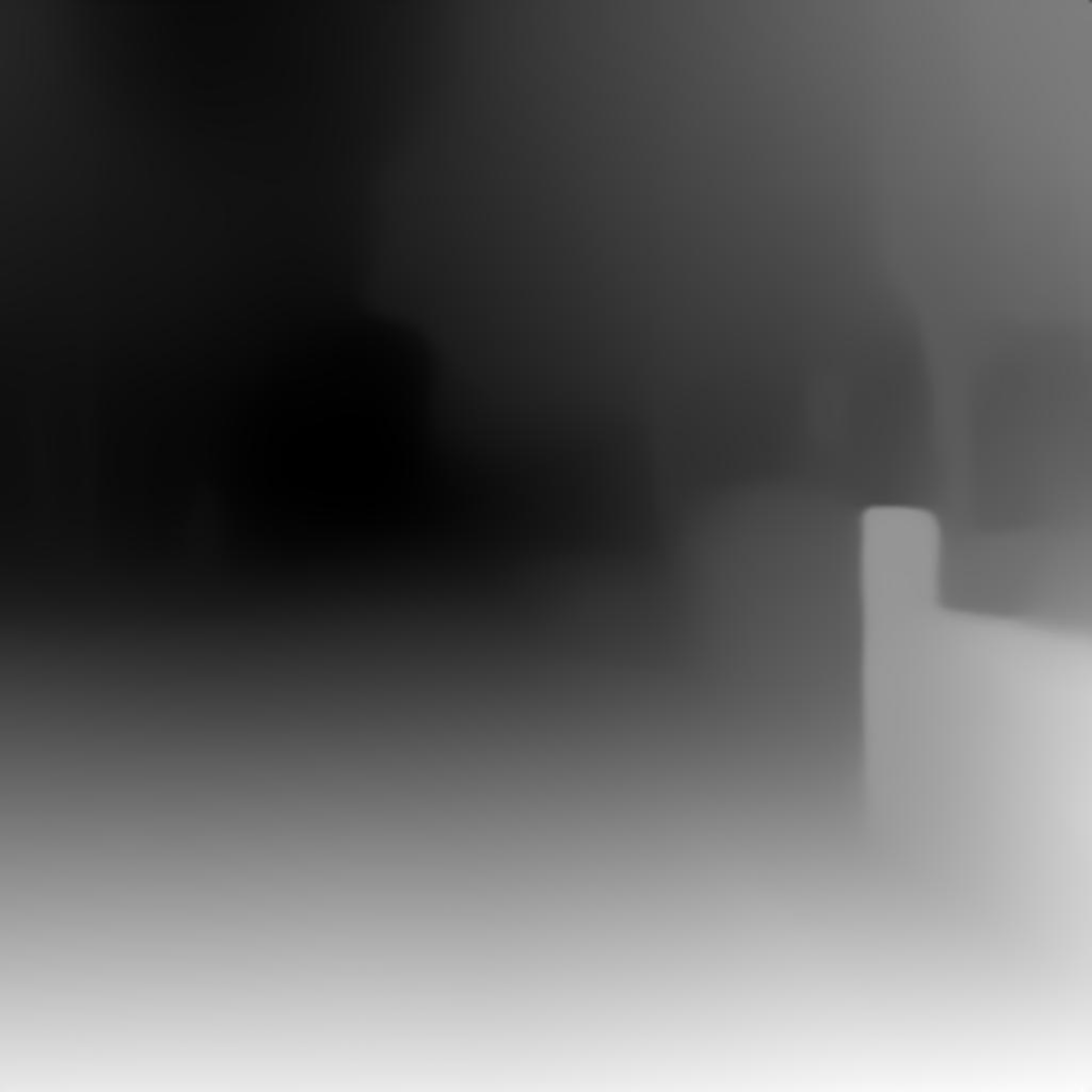}
& \includegraphics[width=1.7cm,height=1.7cm]{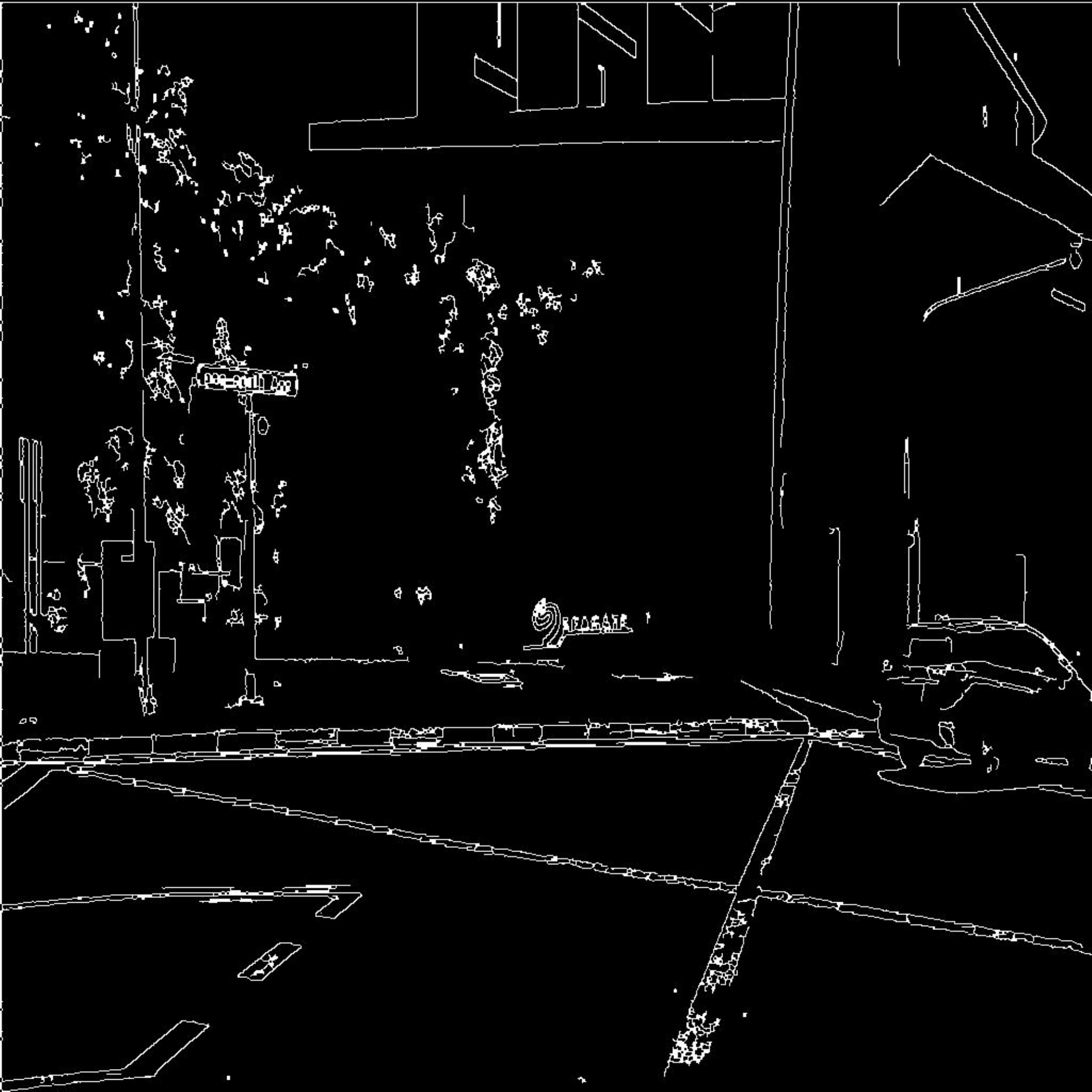}
& \includegraphics[width=1.7cm,height=1.7cm]{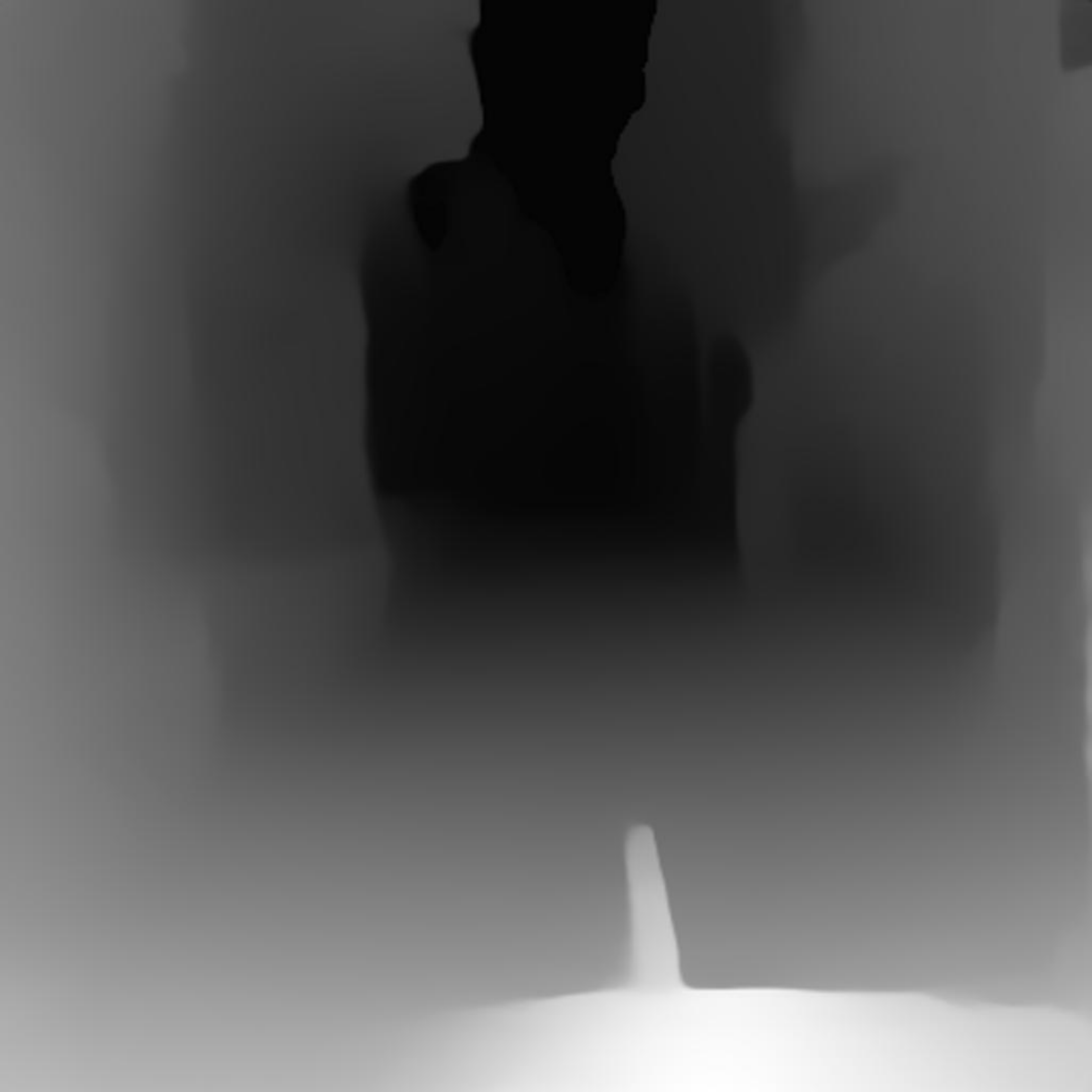}
& \includegraphics[width=1.7cm,height=1.7cm]{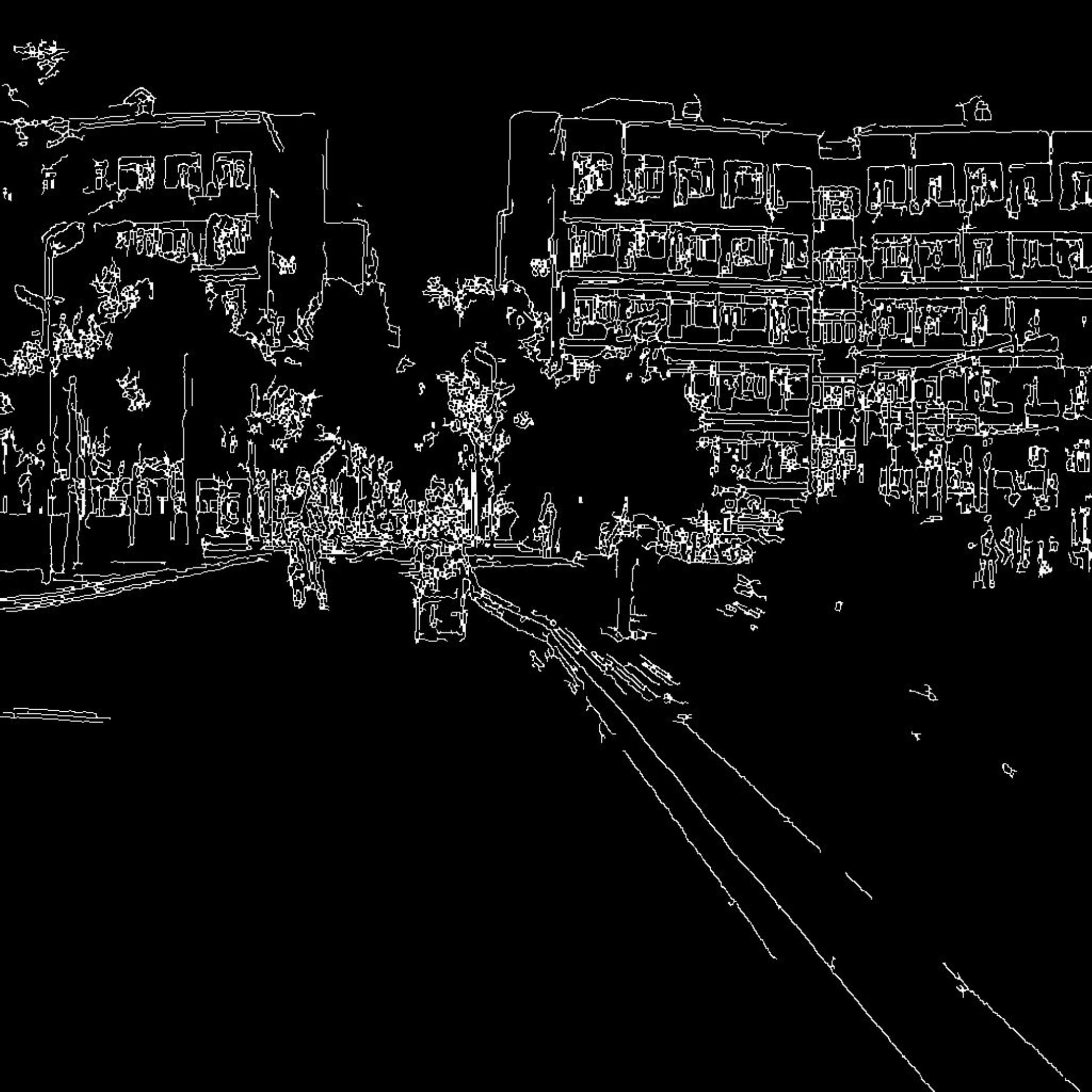}
& \includegraphics[width=1.7cm,height=1.7cm]{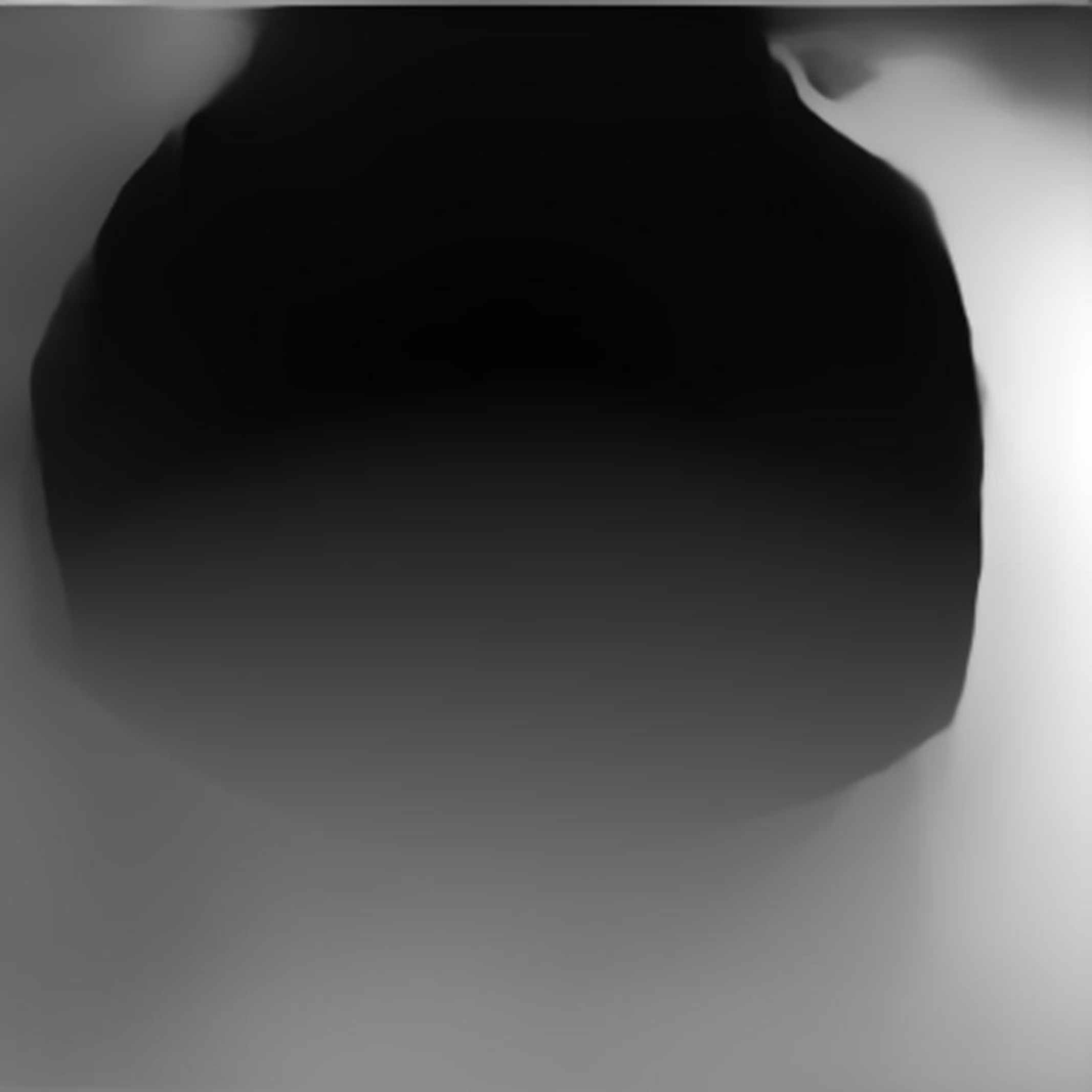}
& \includegraphics[width=1.7cm,height=1.7cm]{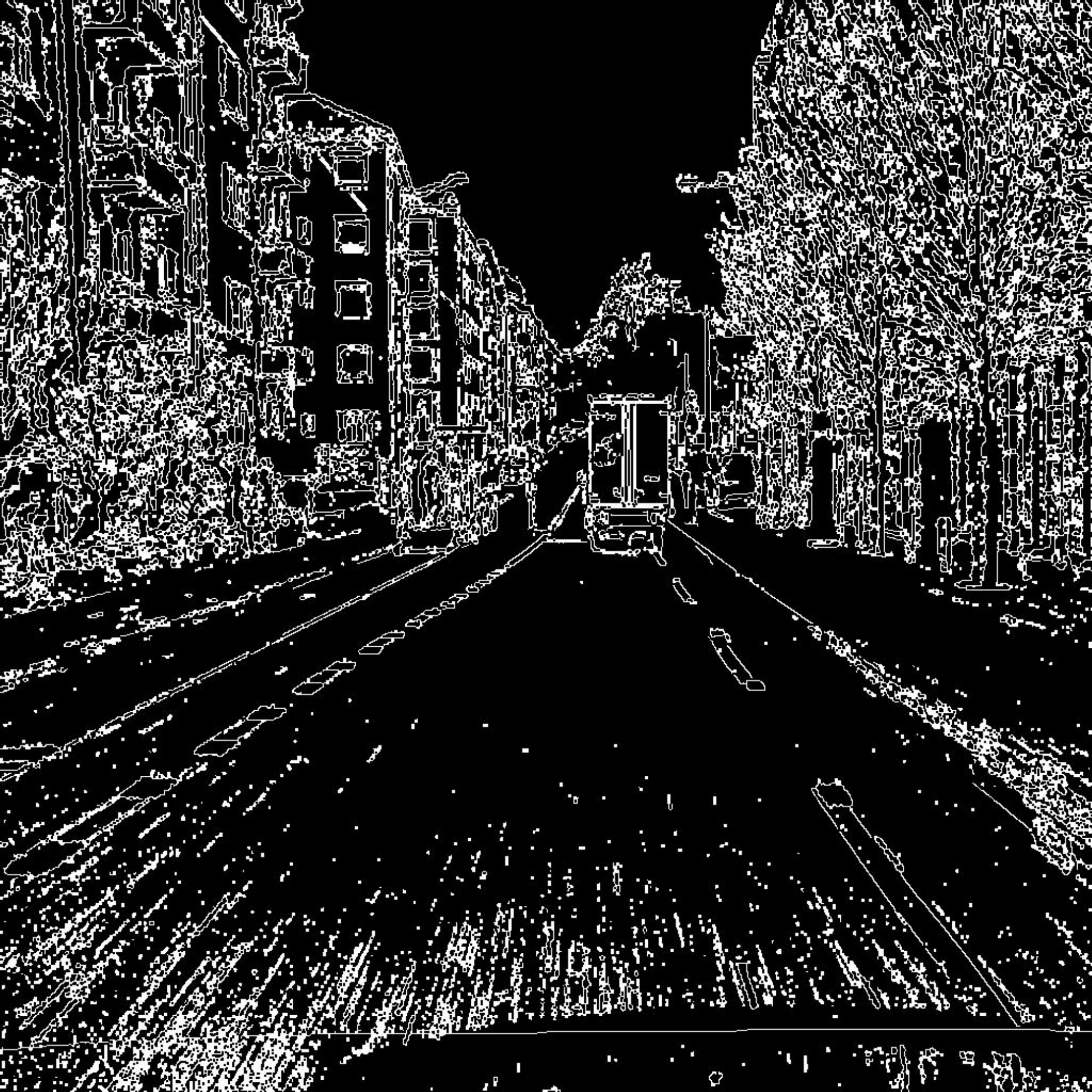}
& \includegraphics[width=1.7cm,height=1.7cm]{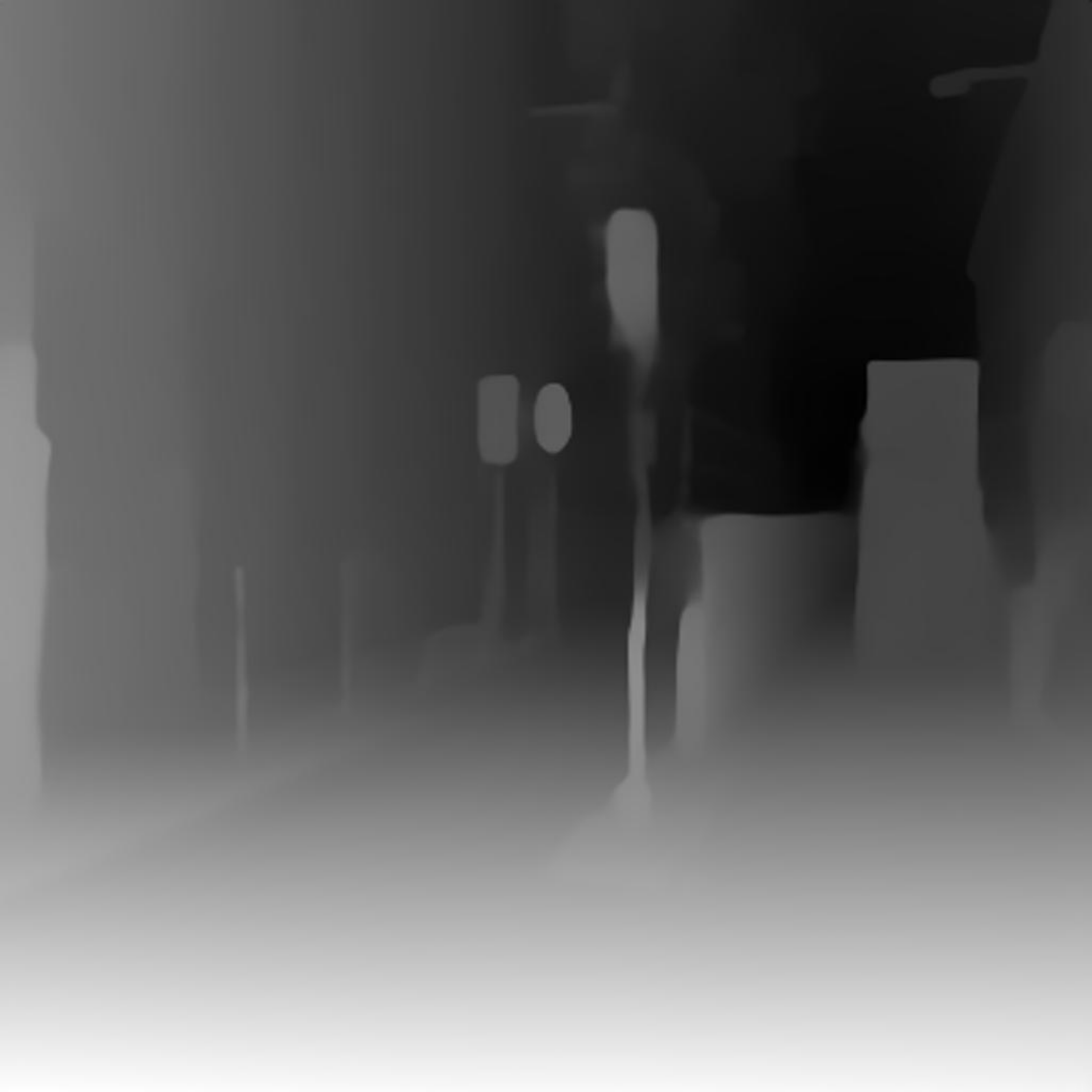} \\

\midrule

Altered Factor
& \makecell[c]{simulation\\$\downarrow$\\real}
& \makecell[c]{rgb-thermal\\$\downarrow$\\rgb}
& \makecell[c]{real\\$\downarrow$\\video-game}
& \makecell[c]{event\\$\downarrow$\\gated}
& \makecell[c]{rgb-thermal\\$\downarrow$\\gated}
& \makecell[c]{rgb\\$\downarrow$\\thermal}
& \makecell[c]{event\\$\downarrow$\\event(desc)}
& \makecell[c]{rgb\\$\downarrow$\\rgb(leaf)} \\

\midrule

SDXL Zeroshot~\cite{podell2023sdxl}
& \includegraphics[width=1.7cm,height=1.7cm]{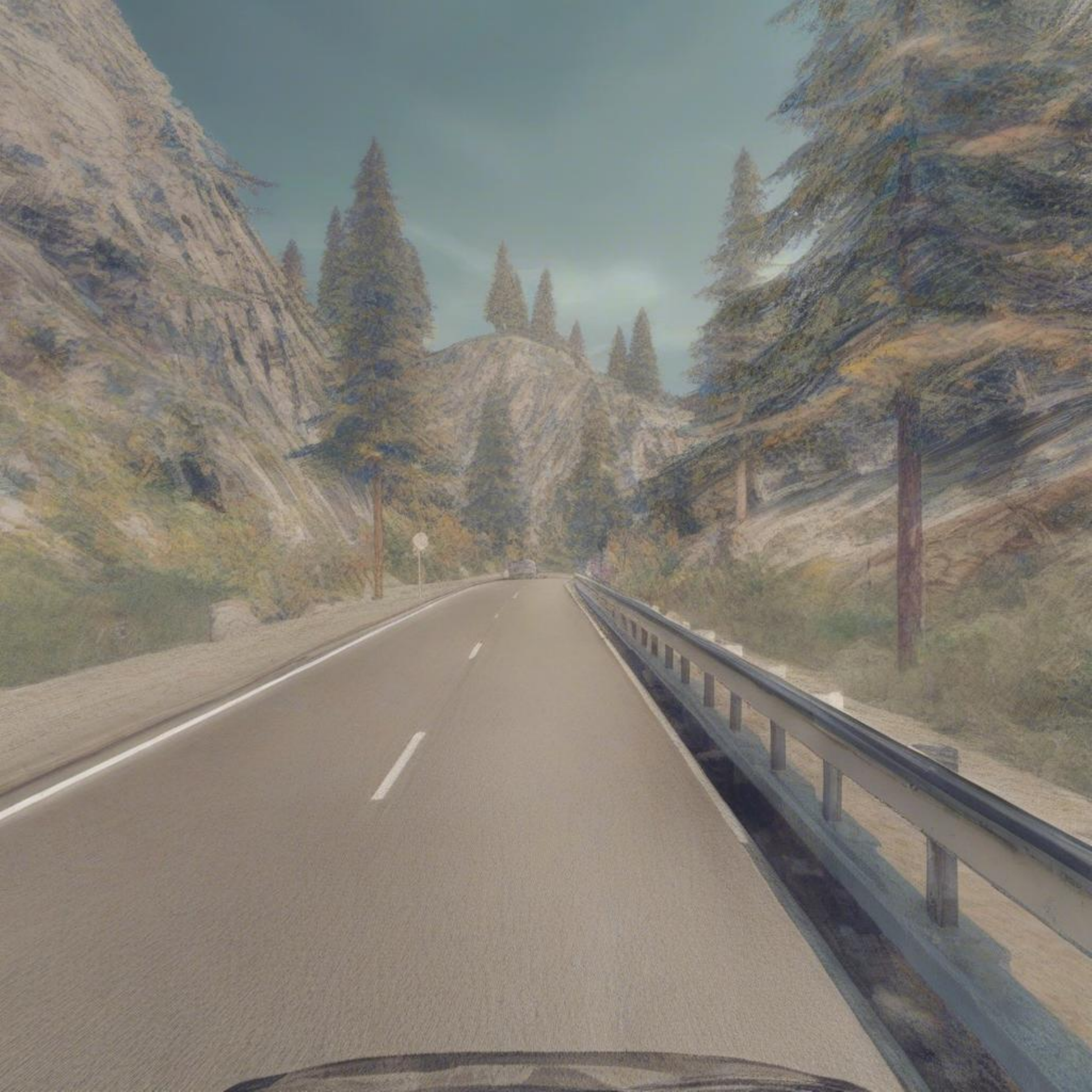}
& \includegraphics[width=1.7cm,height=1.7cm]{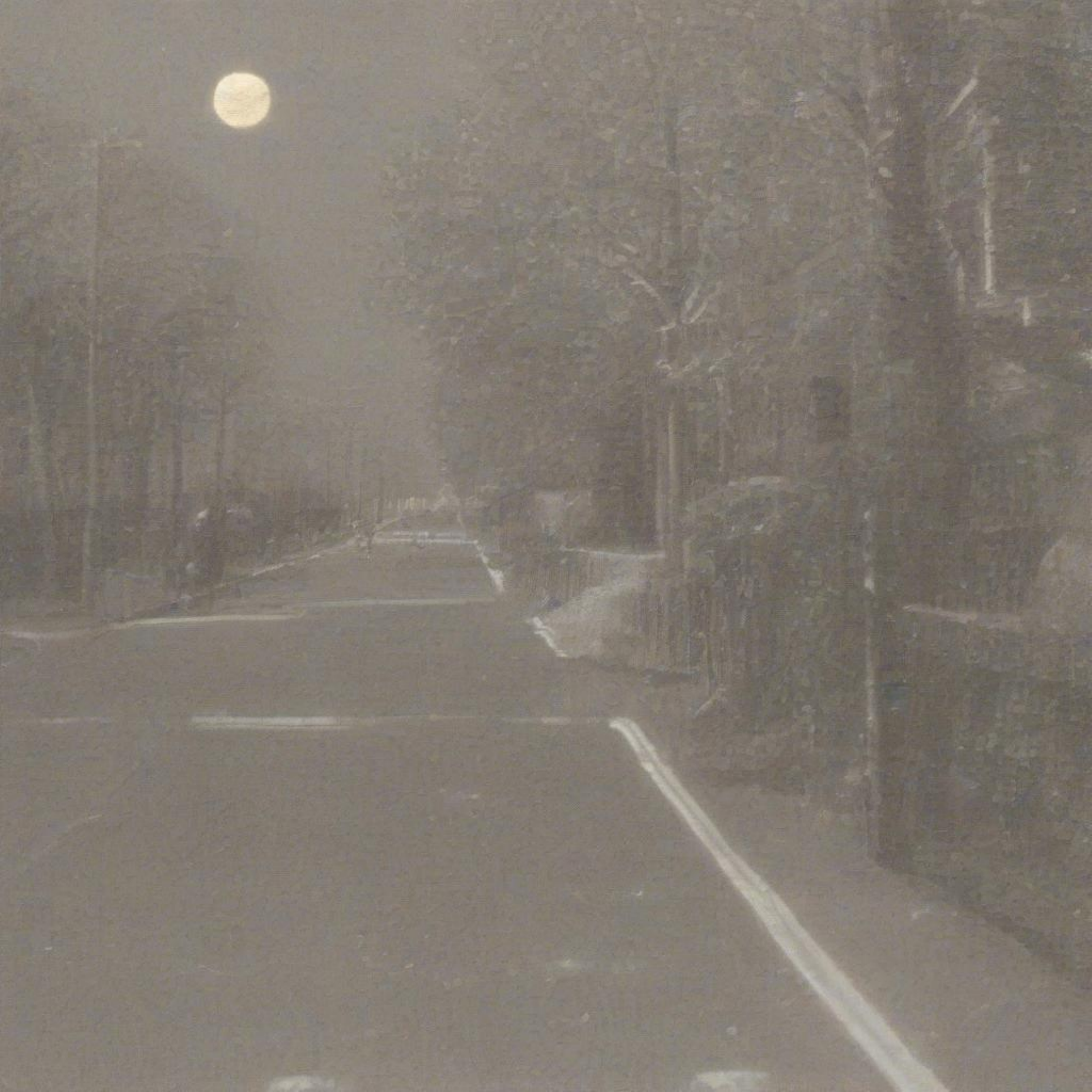}
& \includegraphics[width=1.7cm,height=1.7cm]{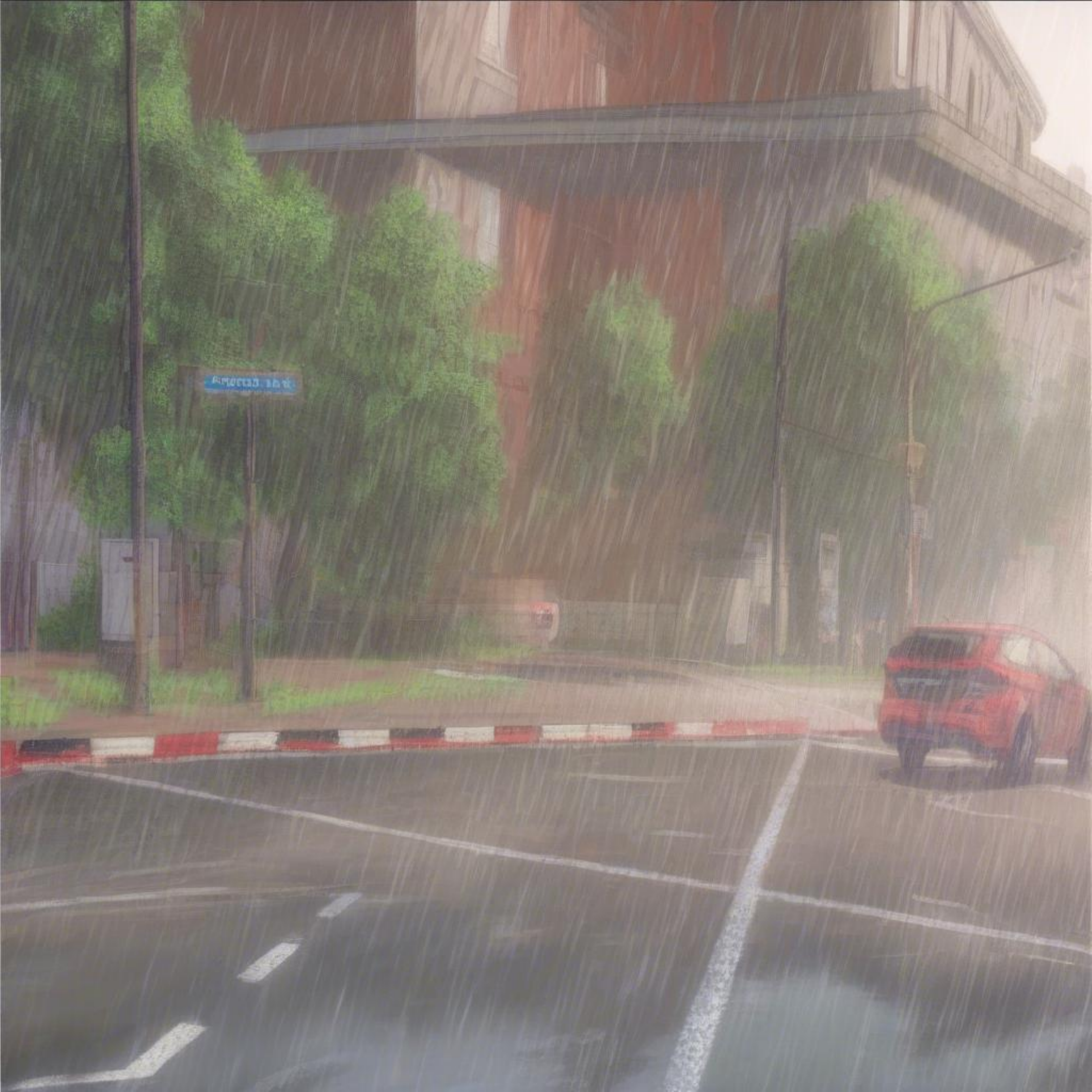}
& \includegraphics[width=1.7cm,height=1.7cm]{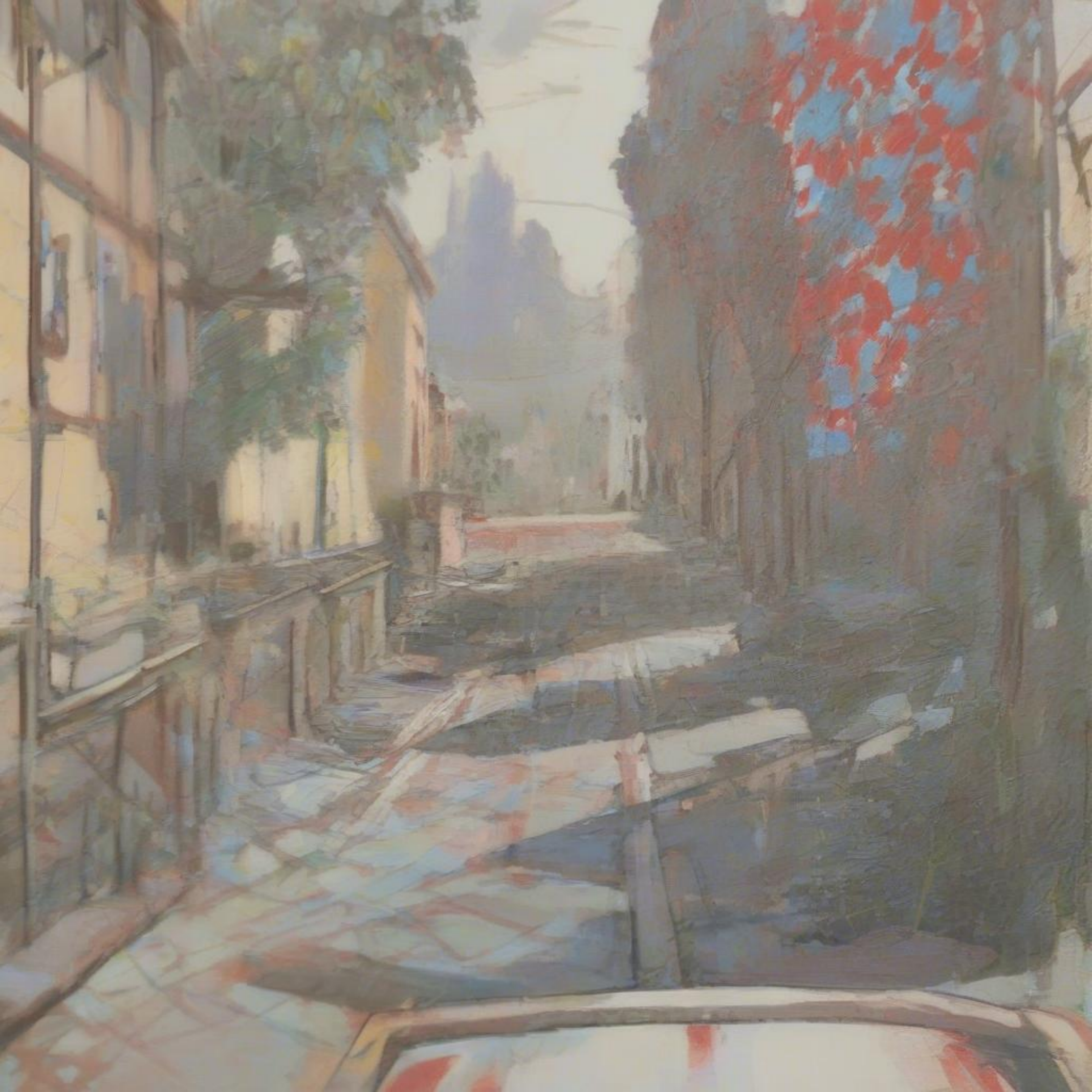}
& \includegraphics[width=1.7cm,height=1.7cm]{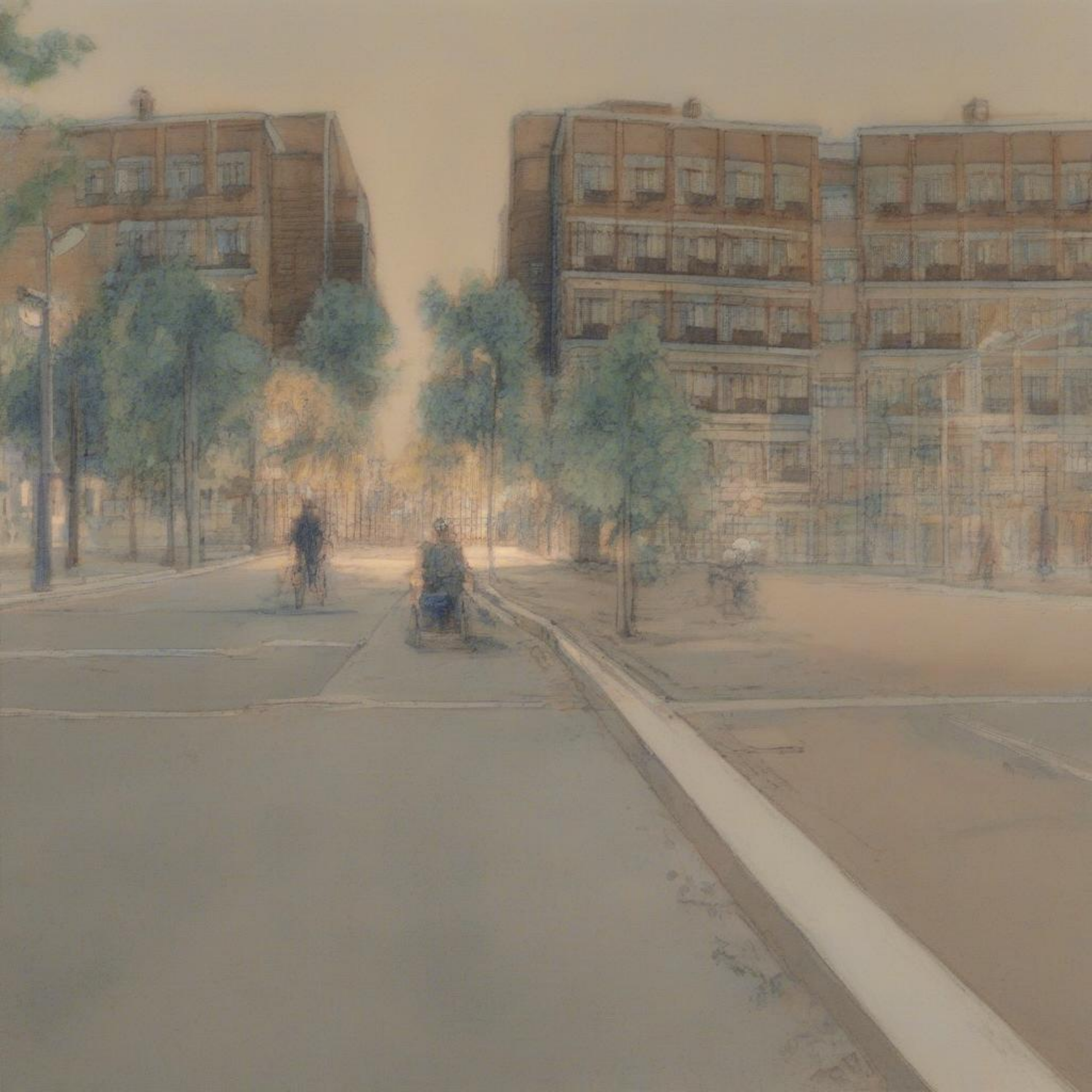}
& \includegraphics[width=1.7cm,height=1.7cm]{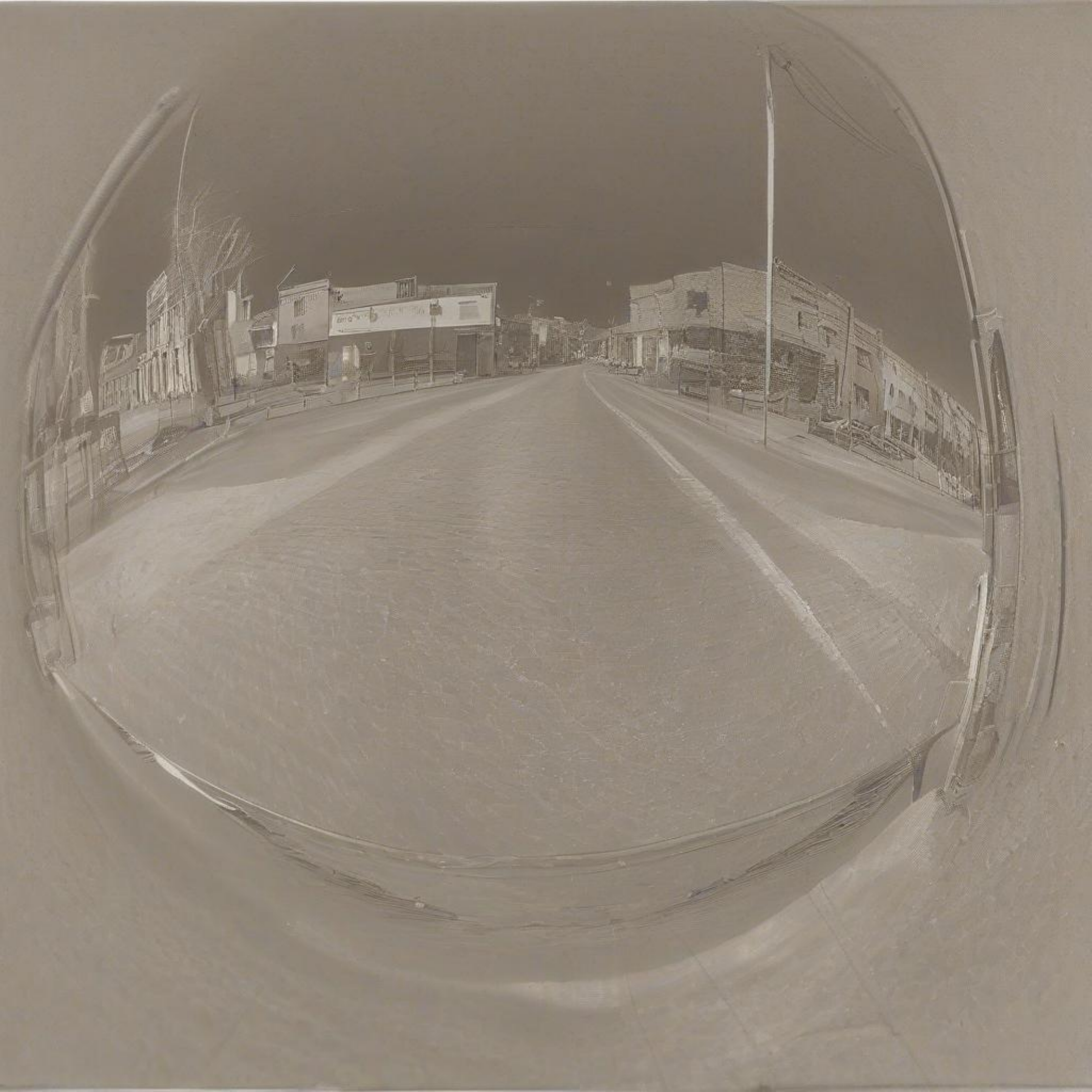}
& \includegraphics[width=1.7cm,height=1.7cm]{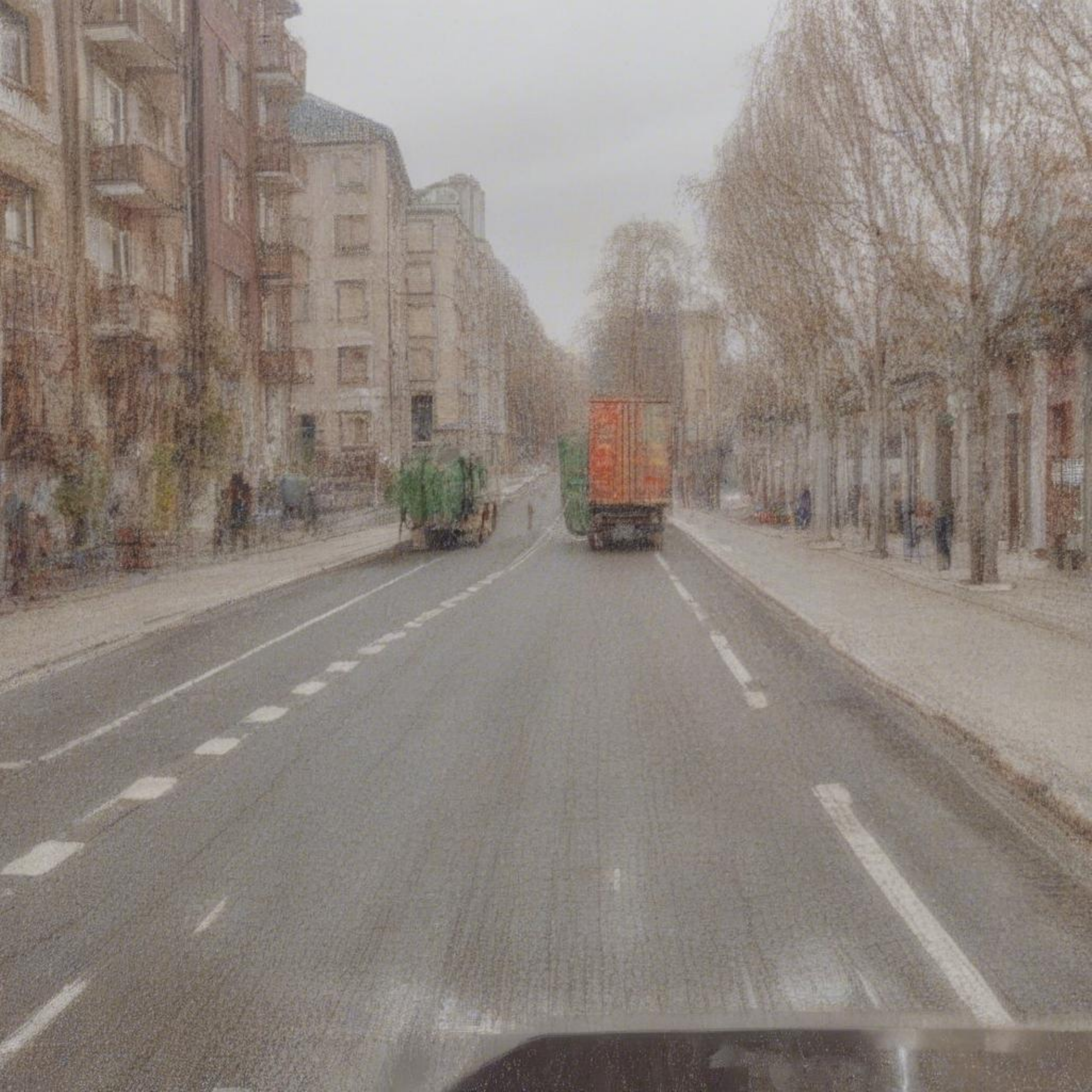}
& \includegraphics[width=1.7cm,height=1.7cm]{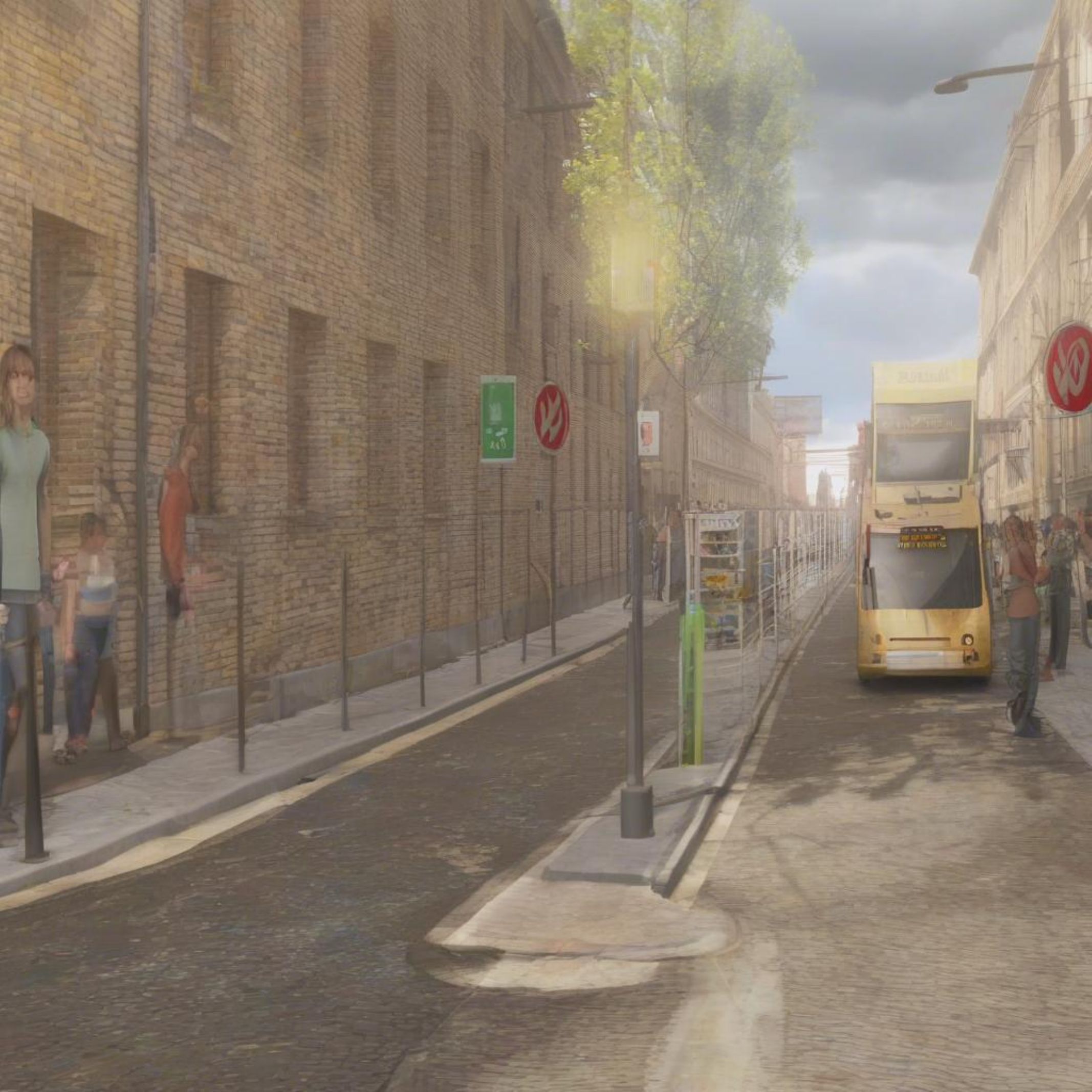}  \\

DreamBooth~\cite{ruiz2023dreambooth}
& \includegraphics[width=1.7cm,height=1.7cm]{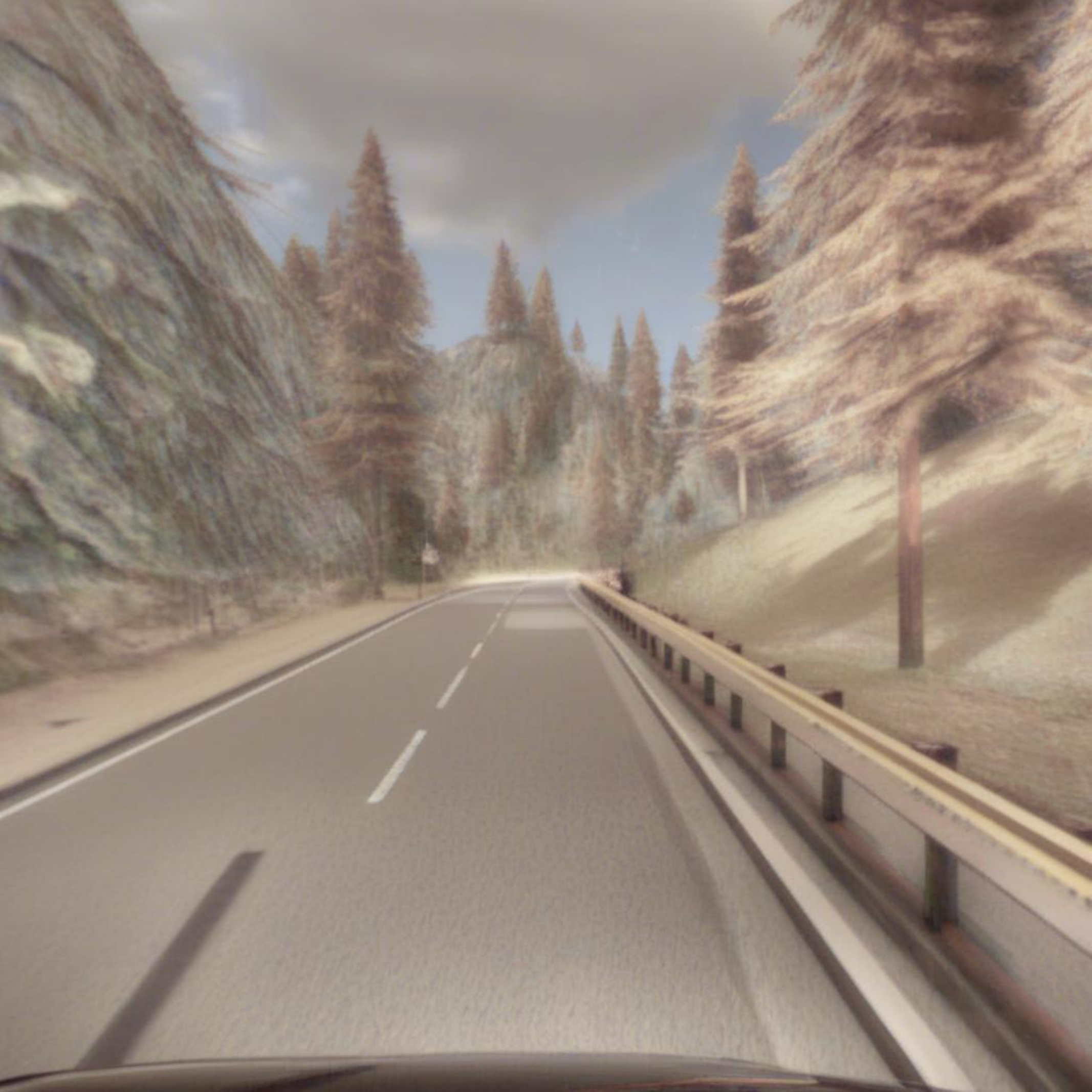}
& \includegraphics[width=1.7cm,height=1.7cm]{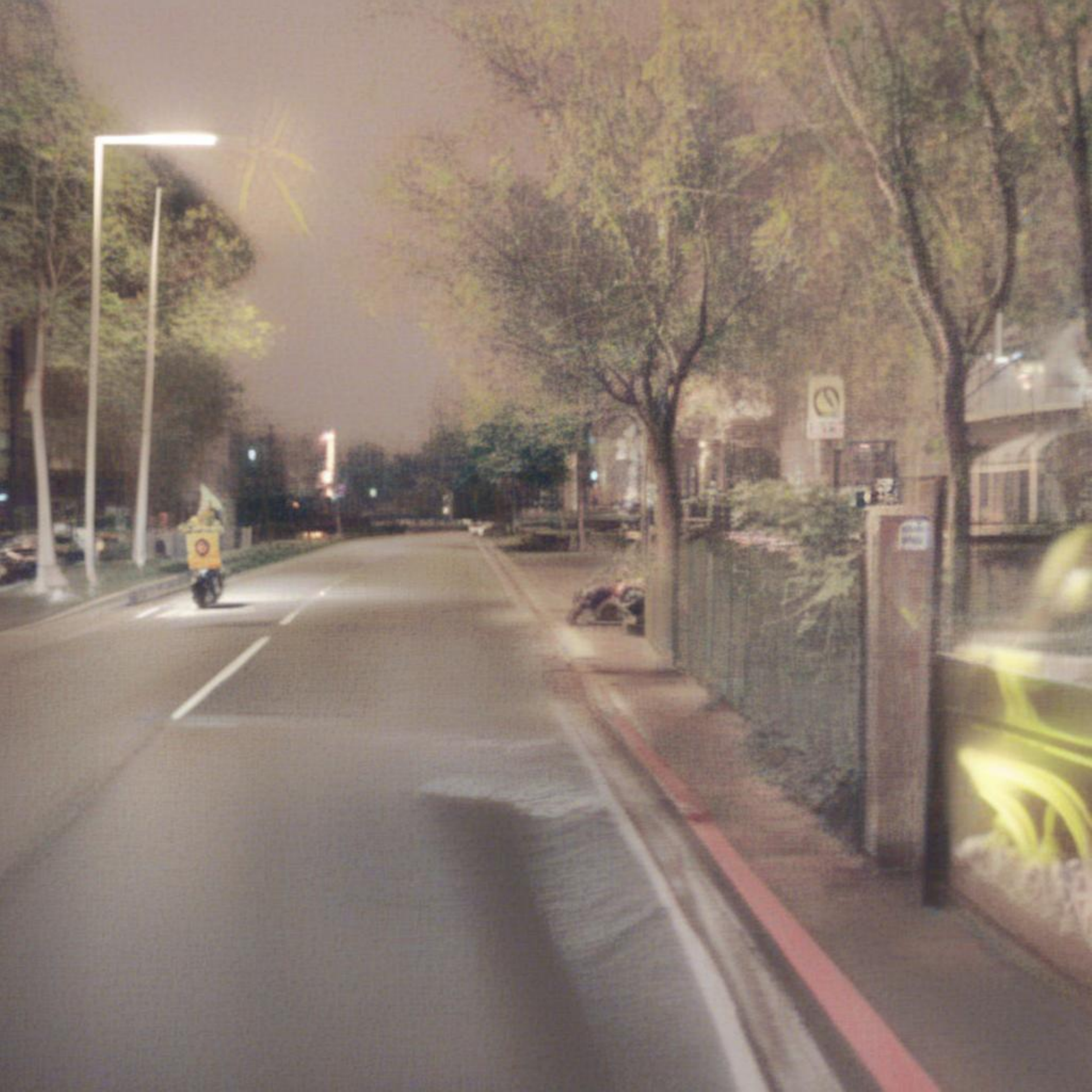}
& \includegraphics[width=1.7cm,height=1.7cm]{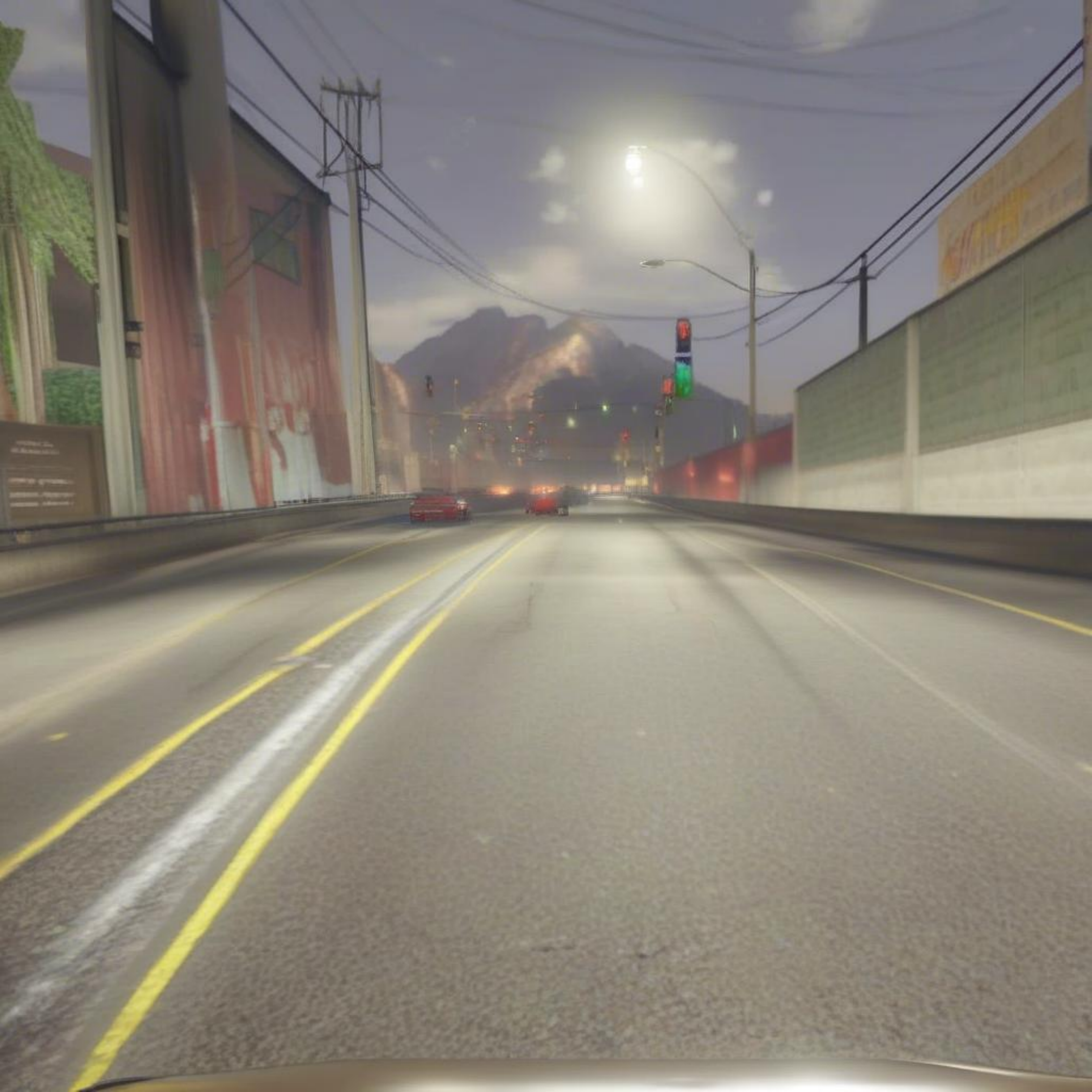}
& \includegraphics[width=1.7cm,height=1.7cm]{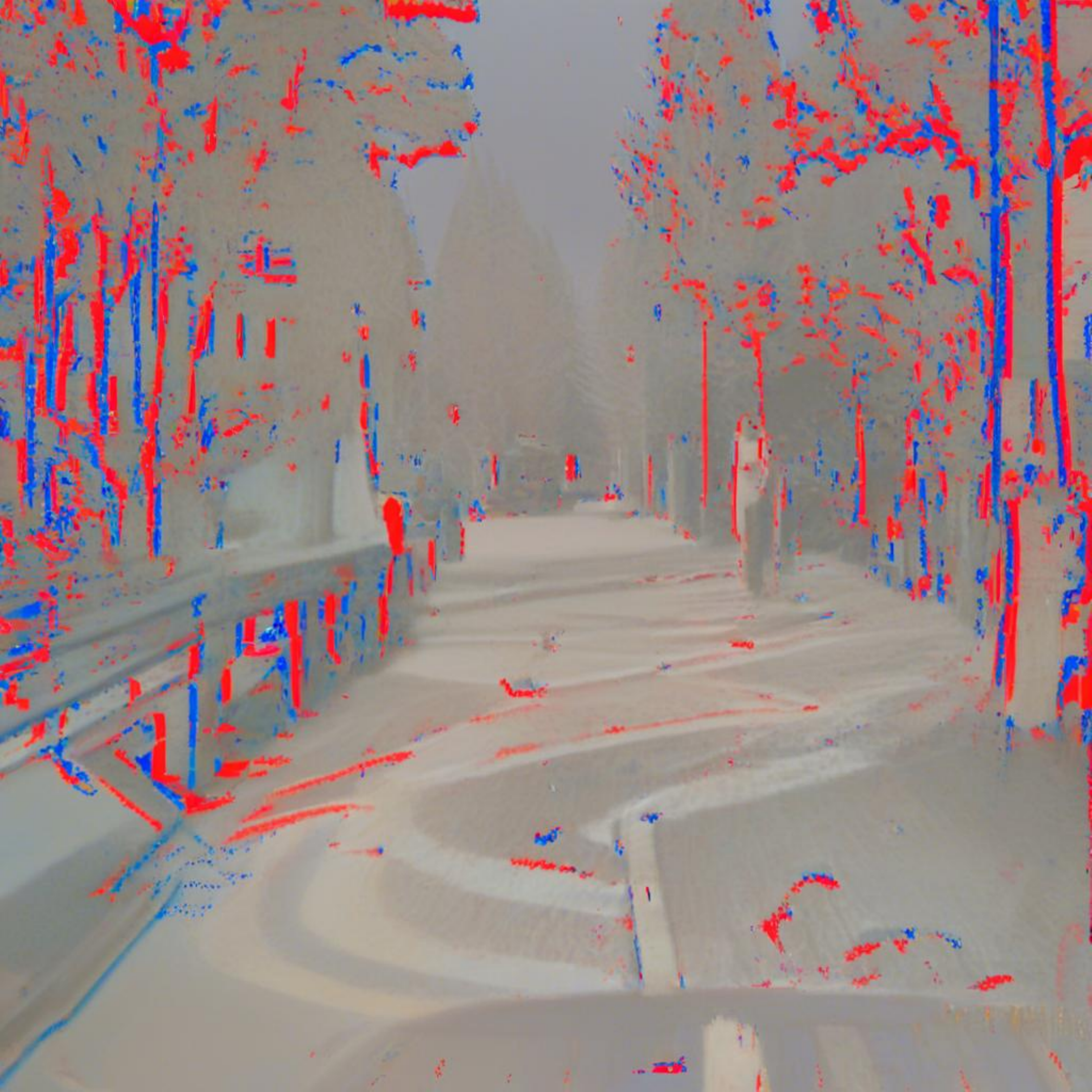}
& \includegraphics[width=1.7cm,height=1.7cm]{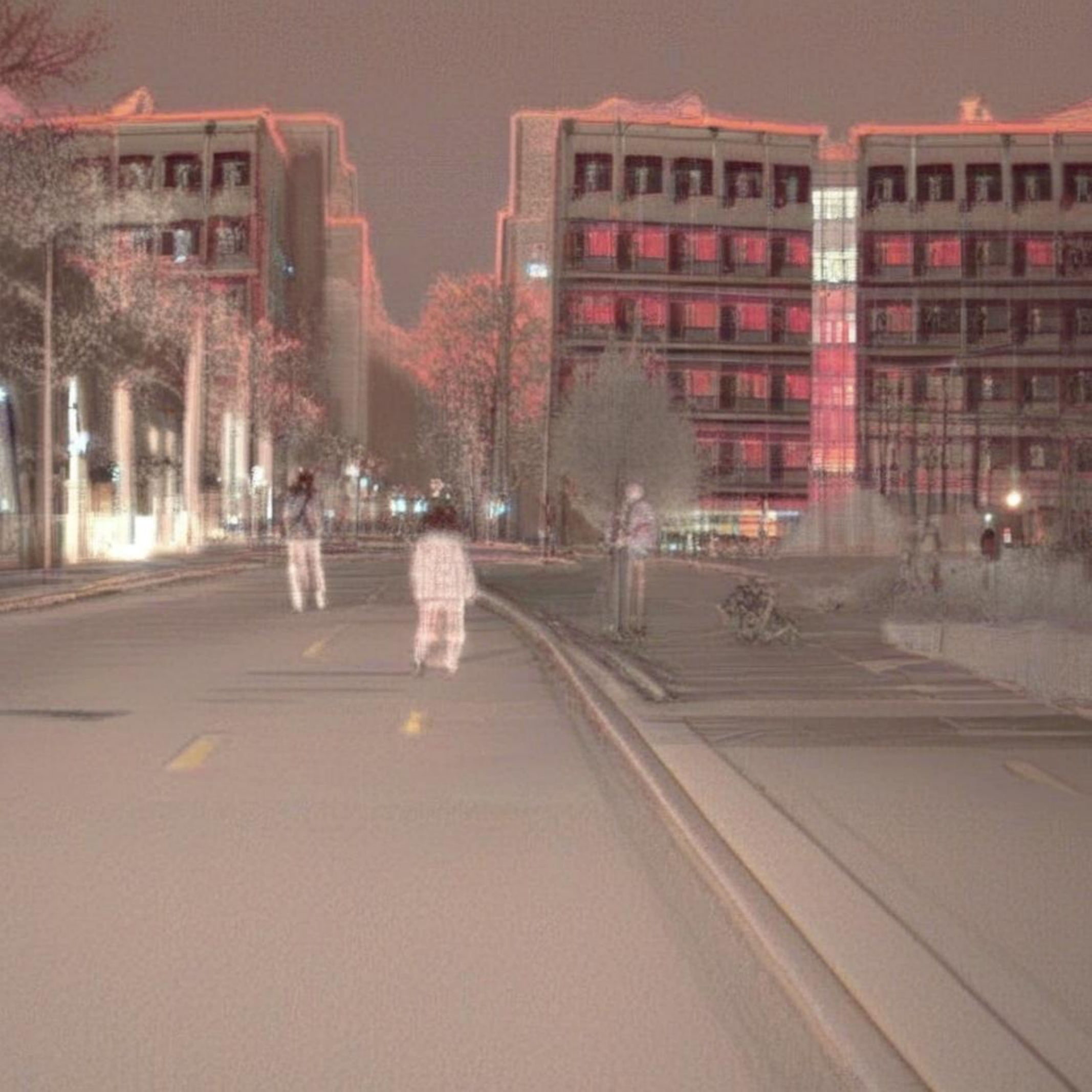}
& \includegraphics[width=1.7cm,height=1.7cm]{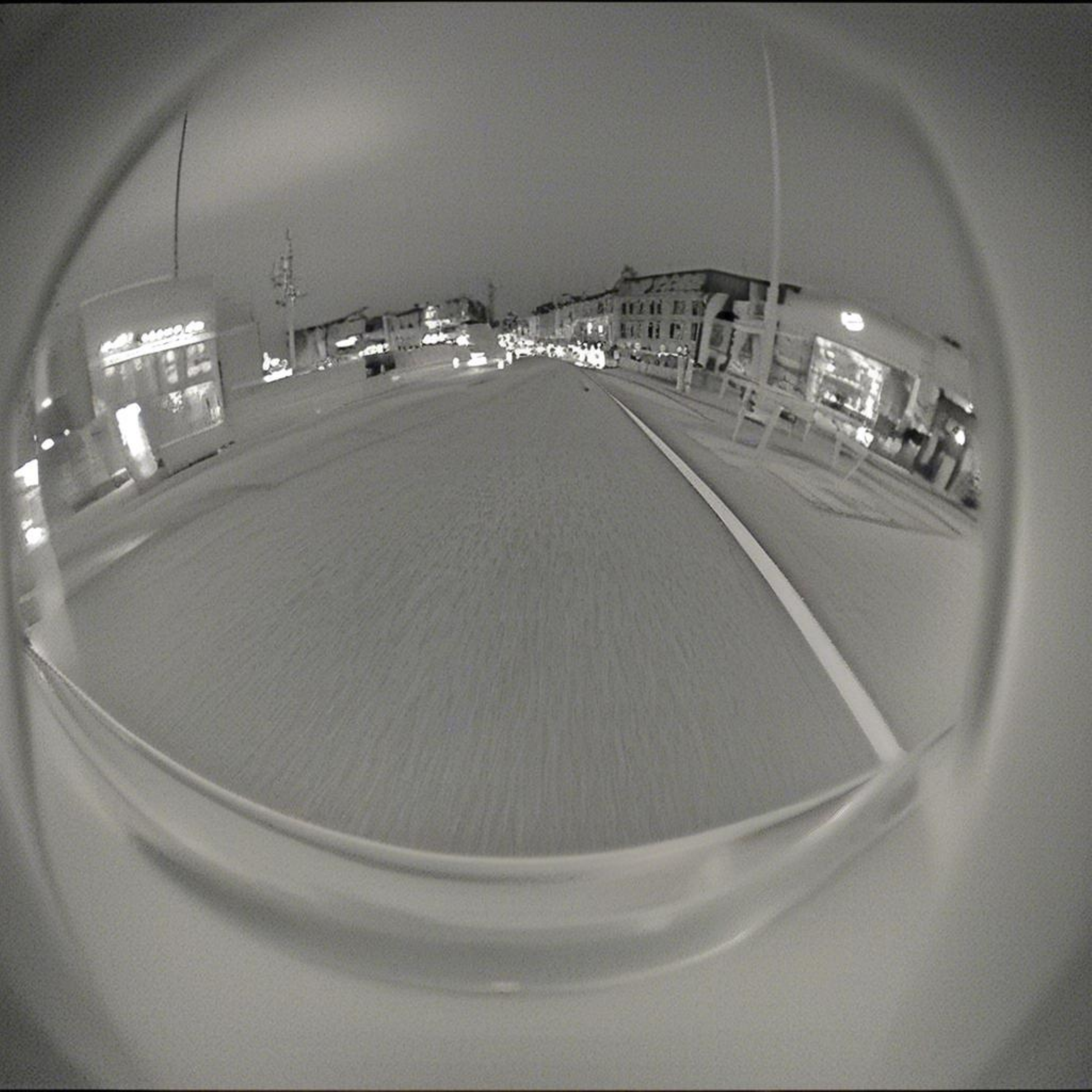}
& \includegraphics[width=1.7cm,height=1.7cm]{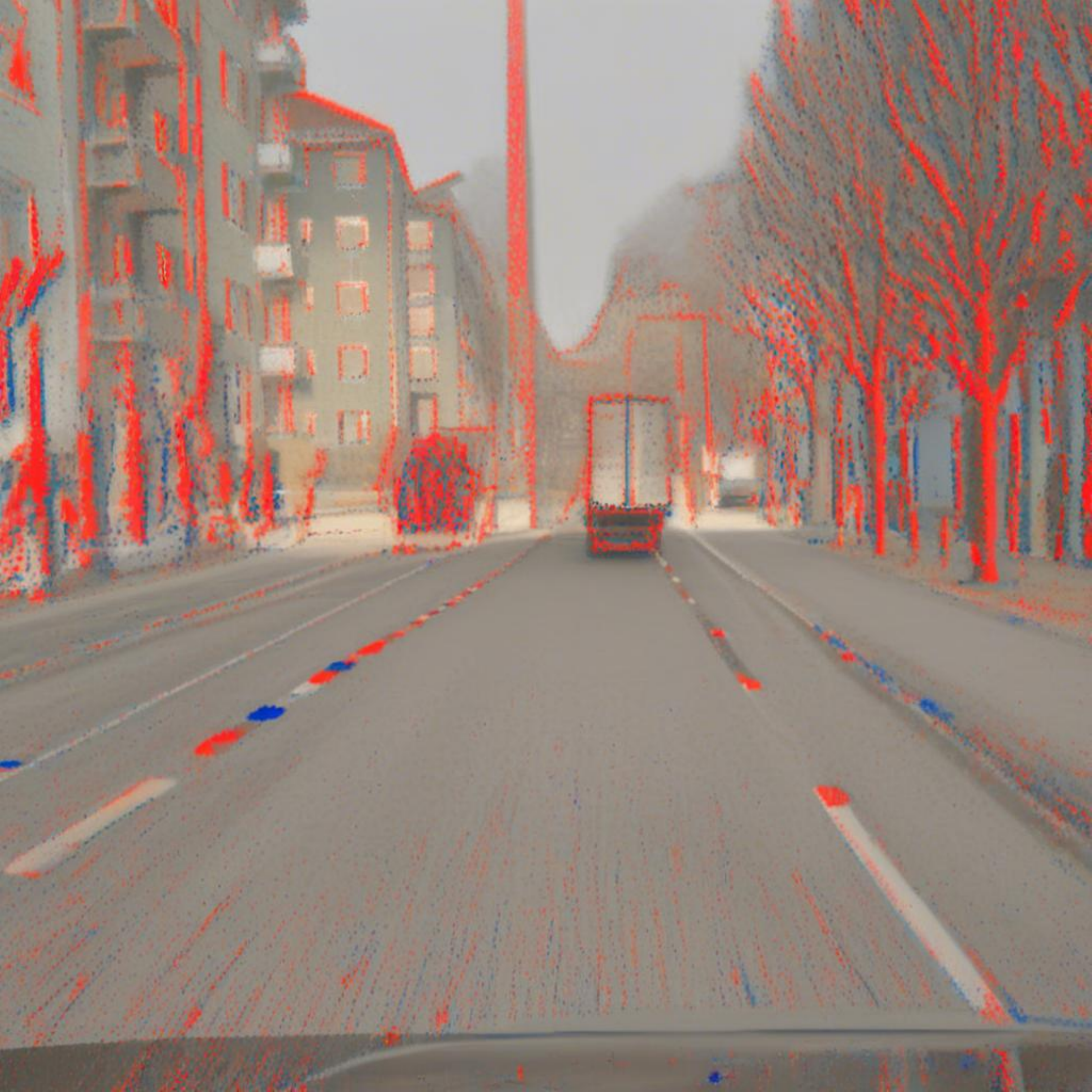}
& \includegraphics[width=1.7cm,height=1.7cm]{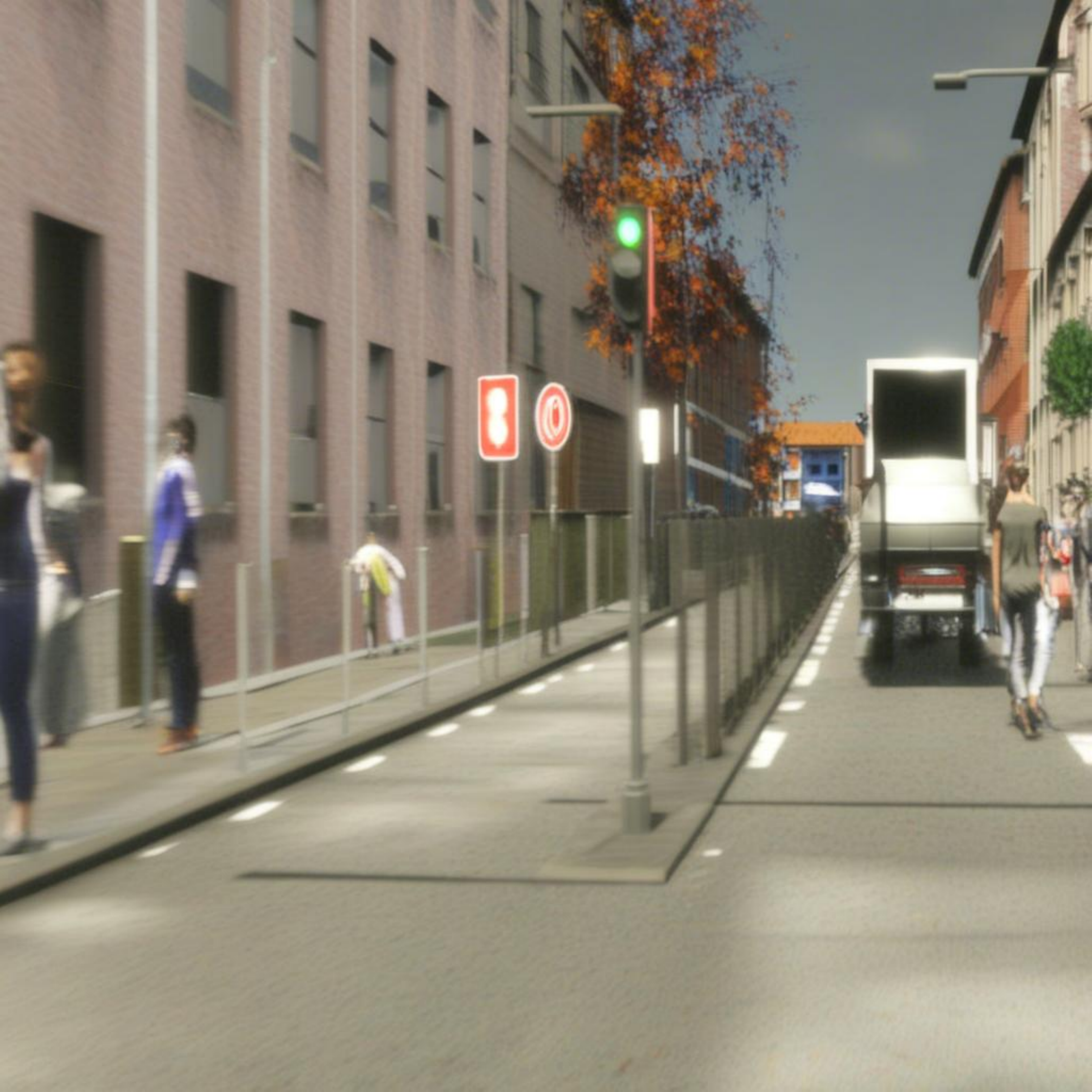}  \\

MULTI~\cite{godavarthy2026multi}
& \includegraphics[width=1.7cm,height=1.7cm]{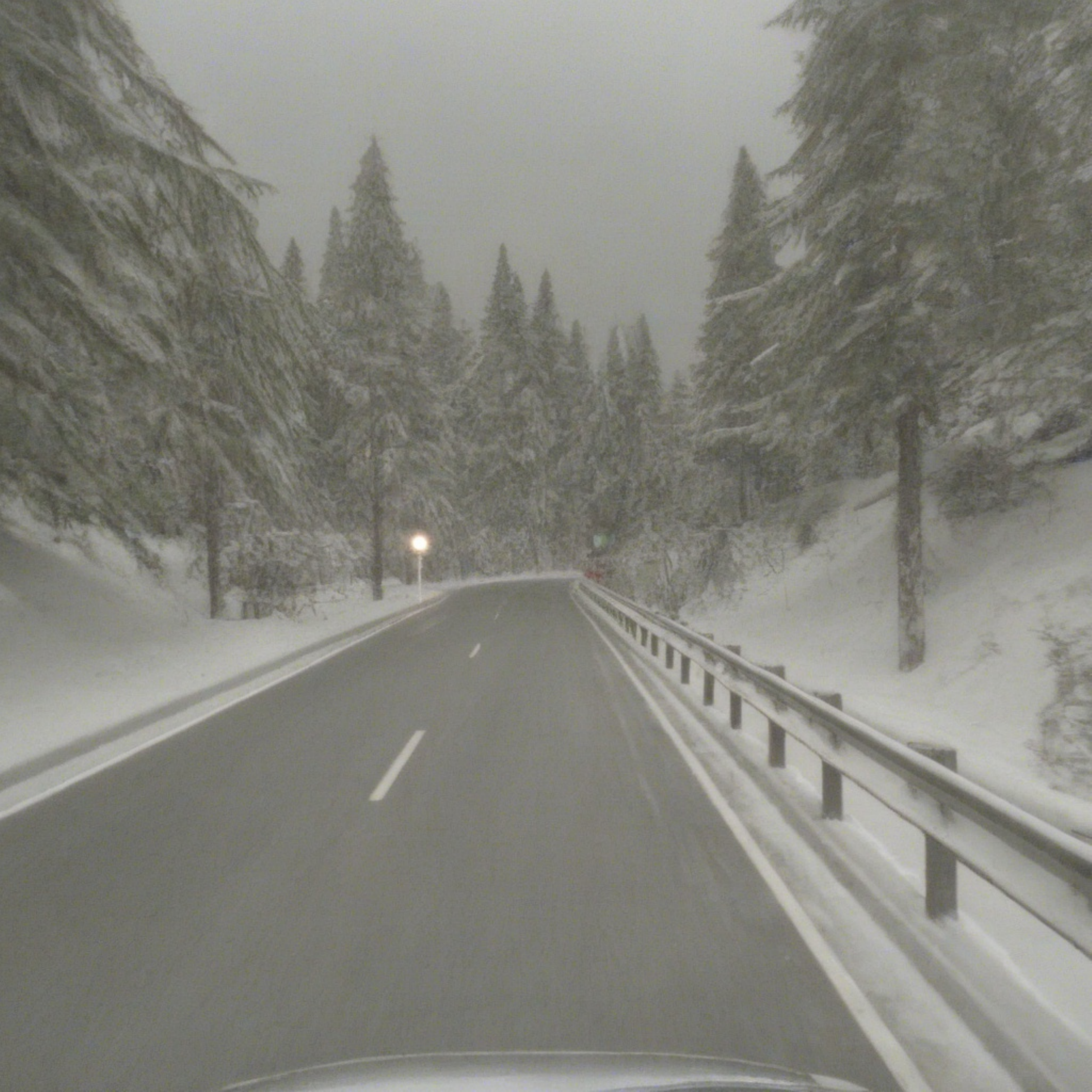}
& \includegraphics[width=1.7cm,height=1.7cm]{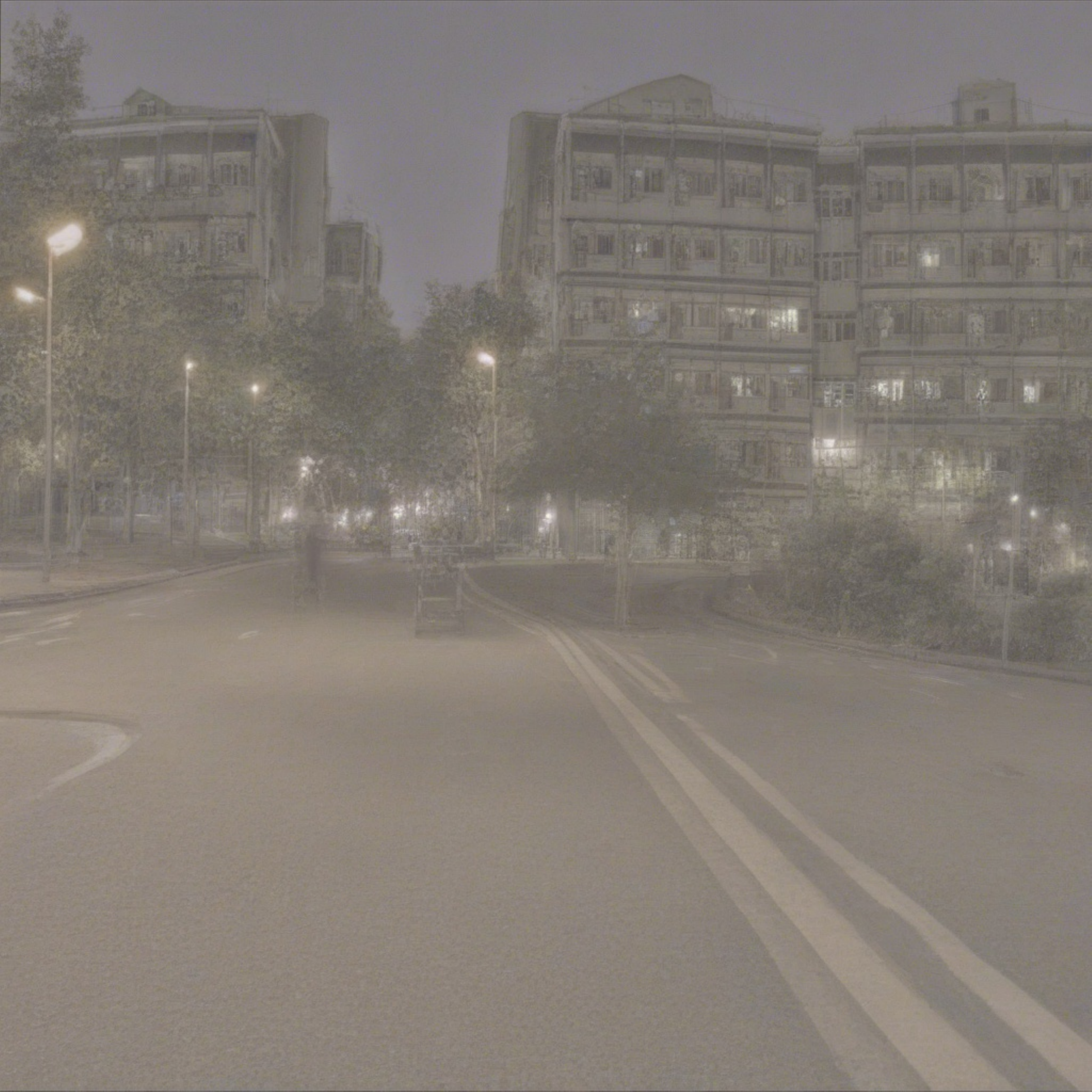}
& \includegraphics[width=1.7cm,height=1.7cm]{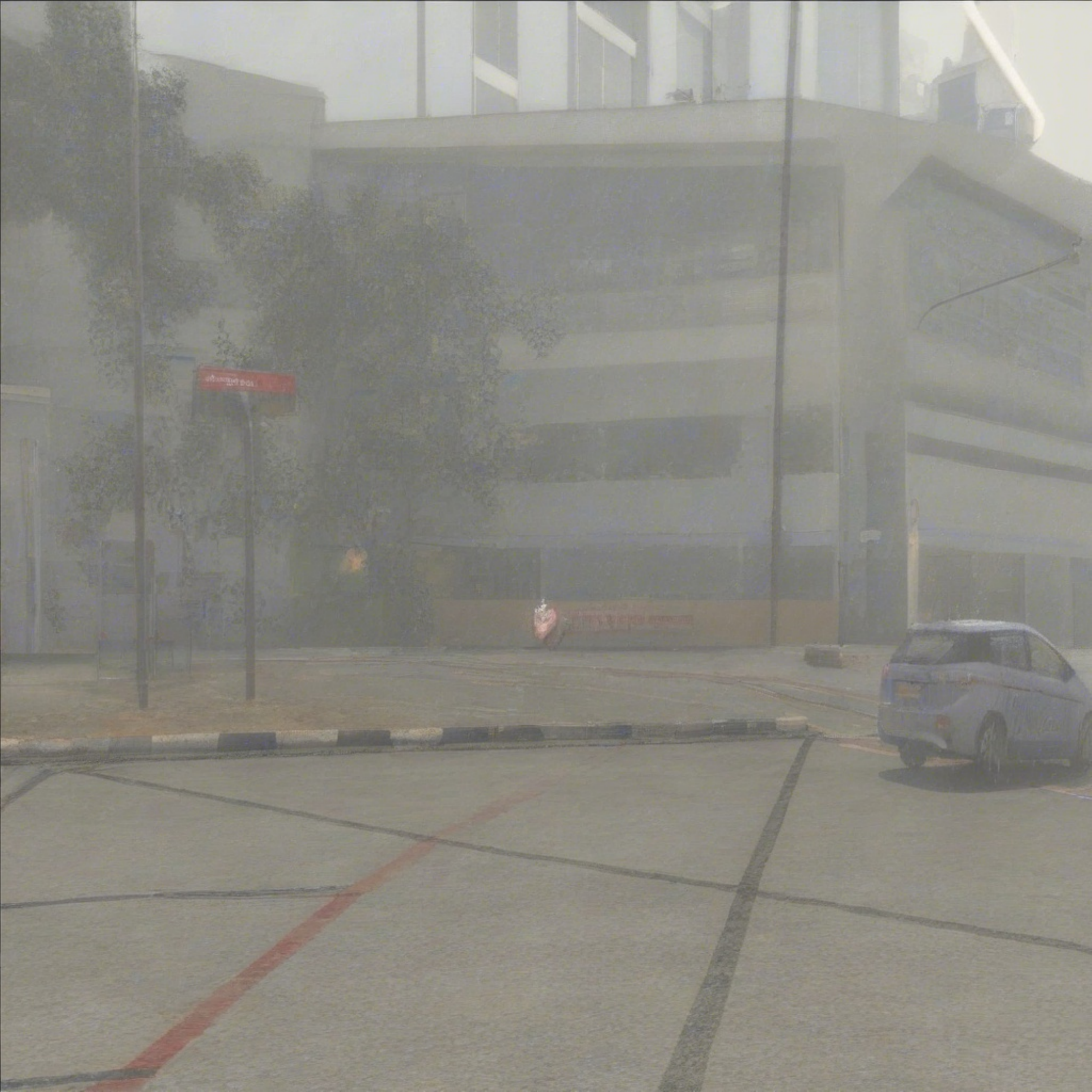}
& \includegraphics[width=1.7cm,height=1.7cm]{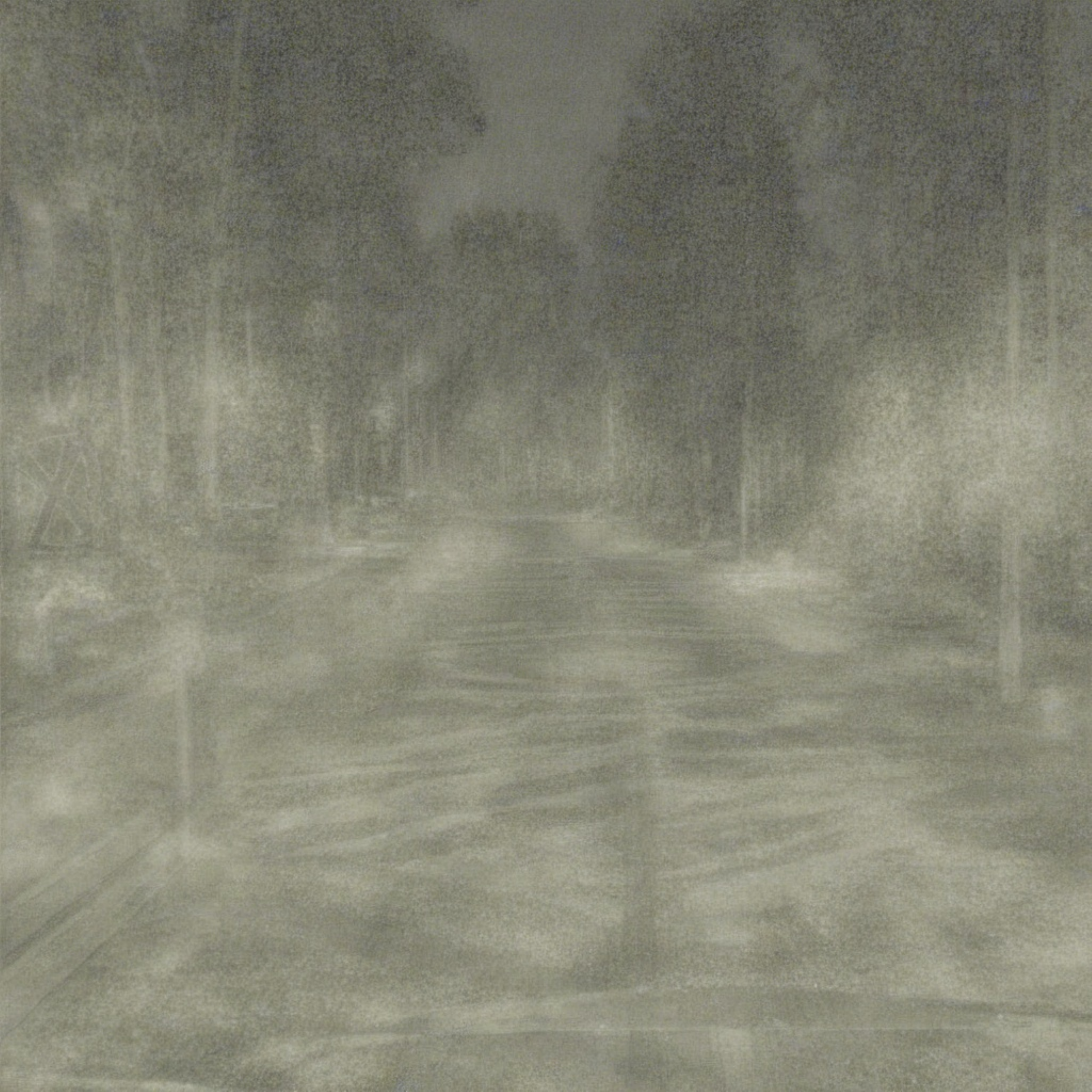}
& \includegraphics[width=1.7cm,height=1.7cm]{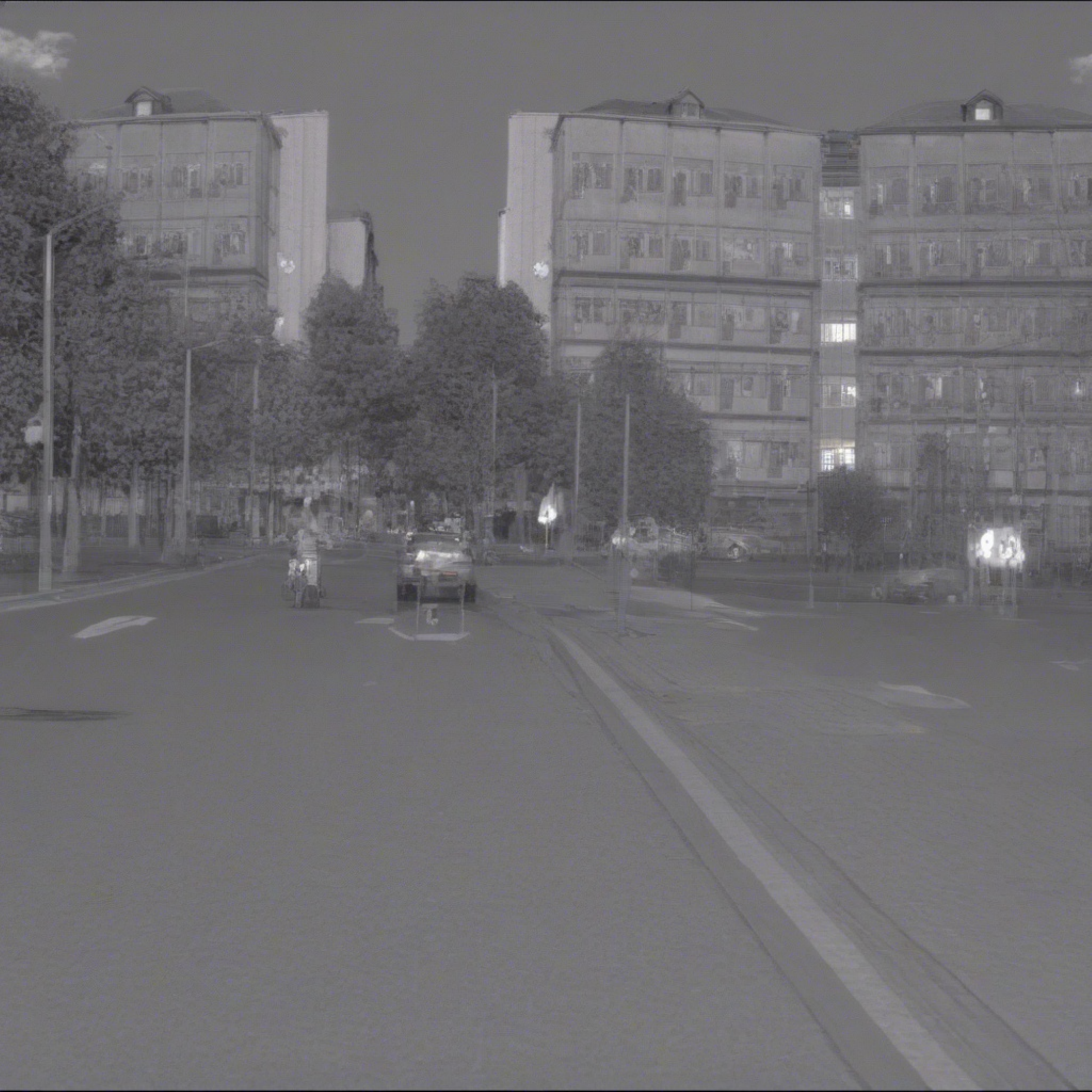}
& \includegraphics[width=1.7cm,height=1.7cm]{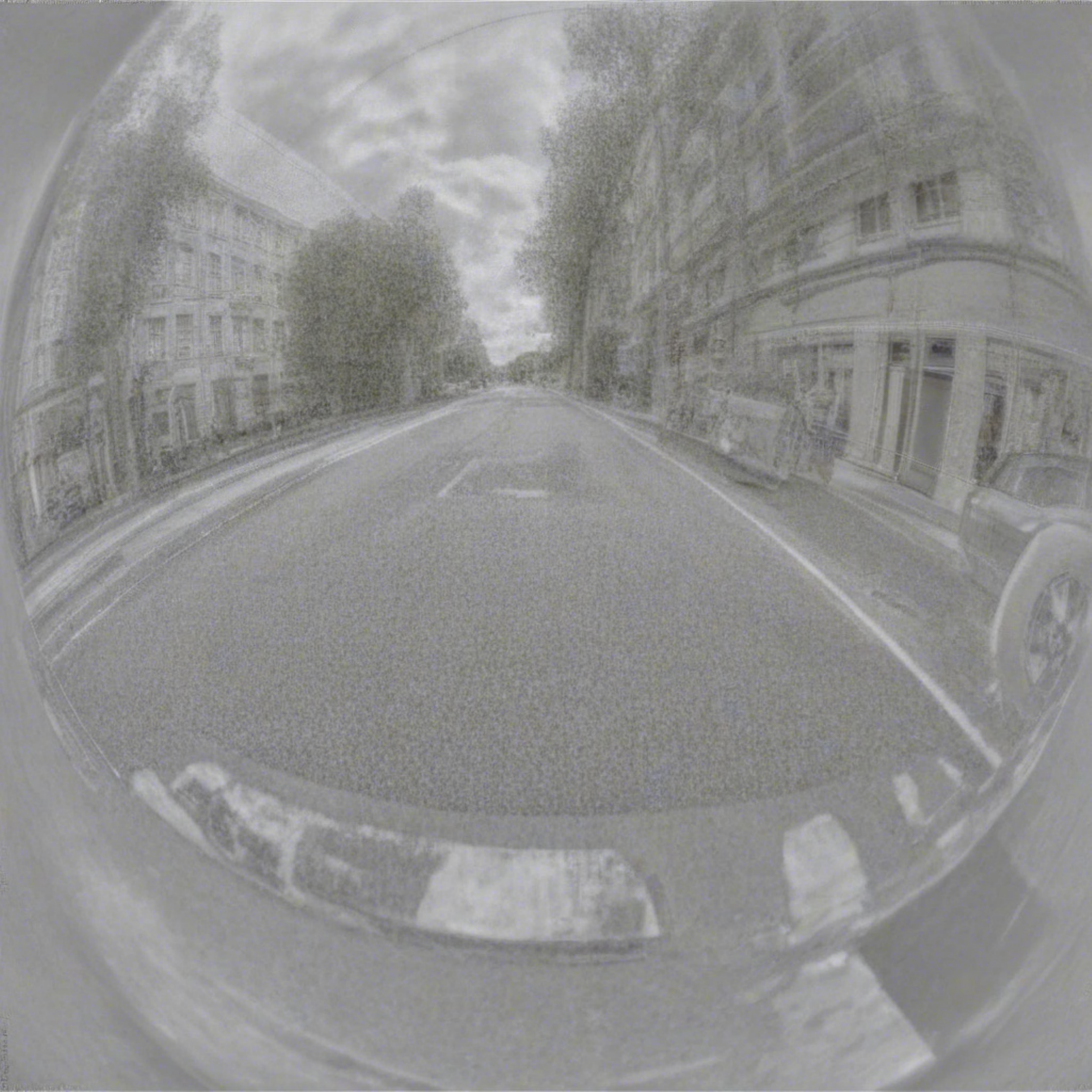}
& \includegraphics[width=1.7cm,height=1.7cm]{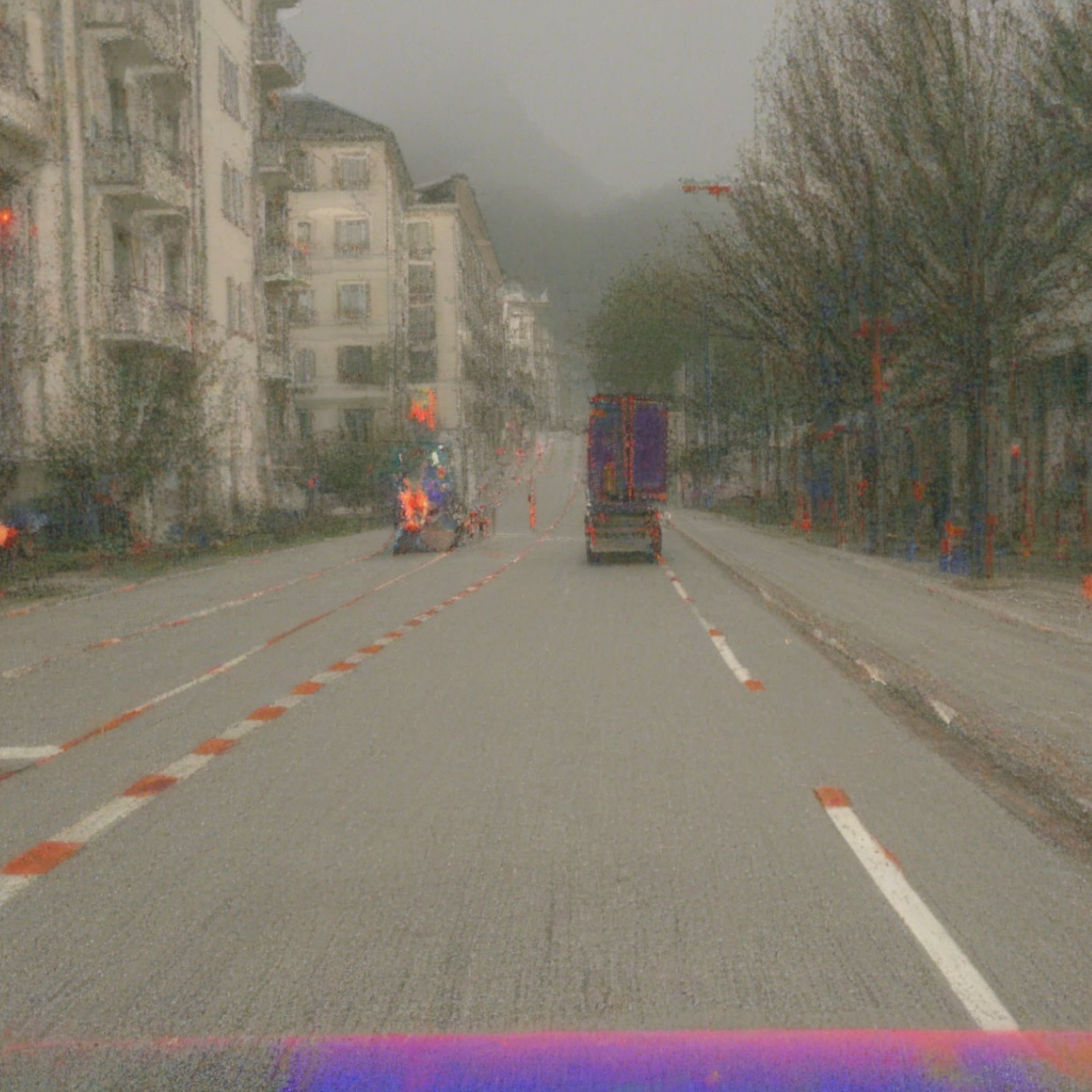}
& \includegraphics[width=1.7cm,height=1.7cm]{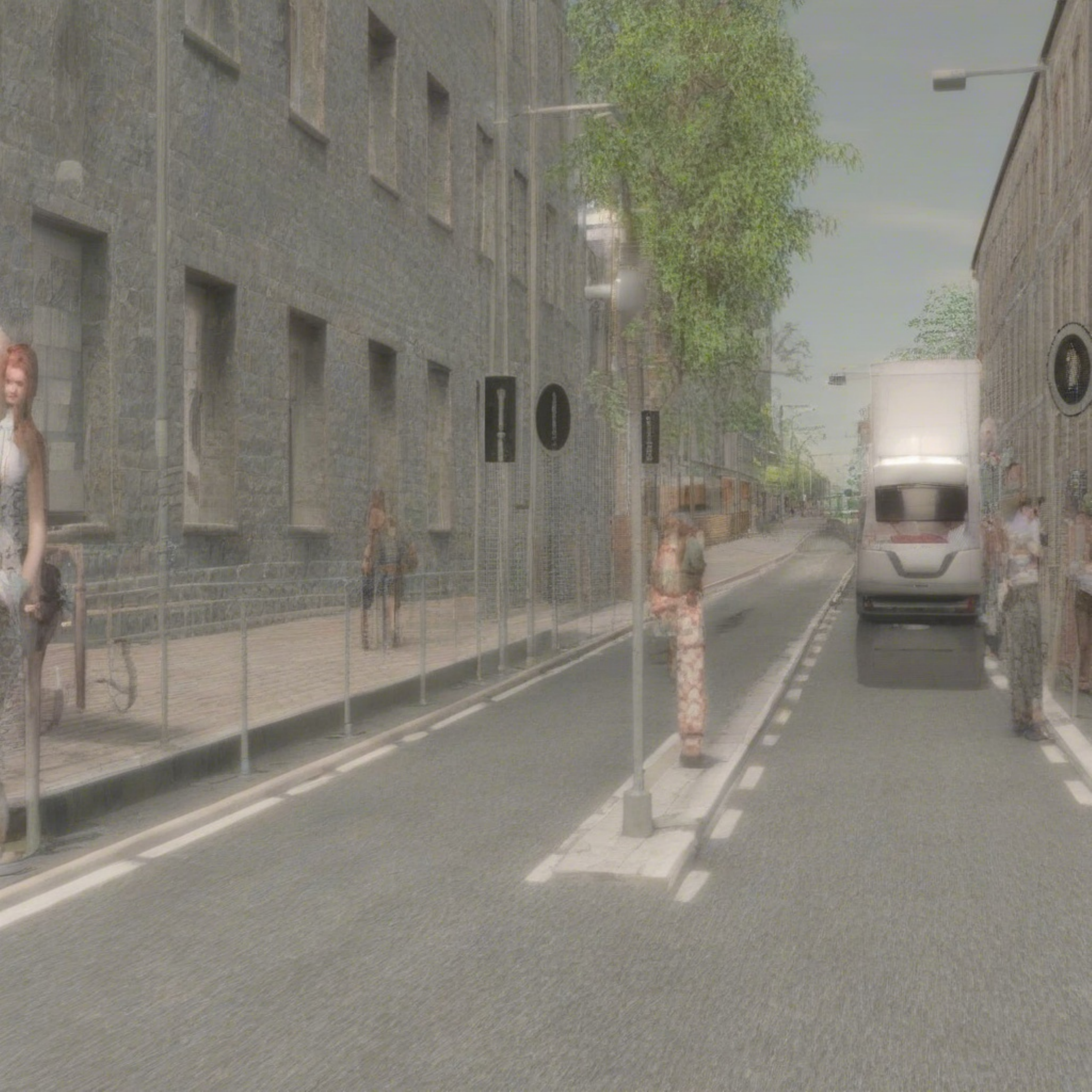} \\

X-MULTI
& \includegraphics[width=1.7cm,height=1.7cm]{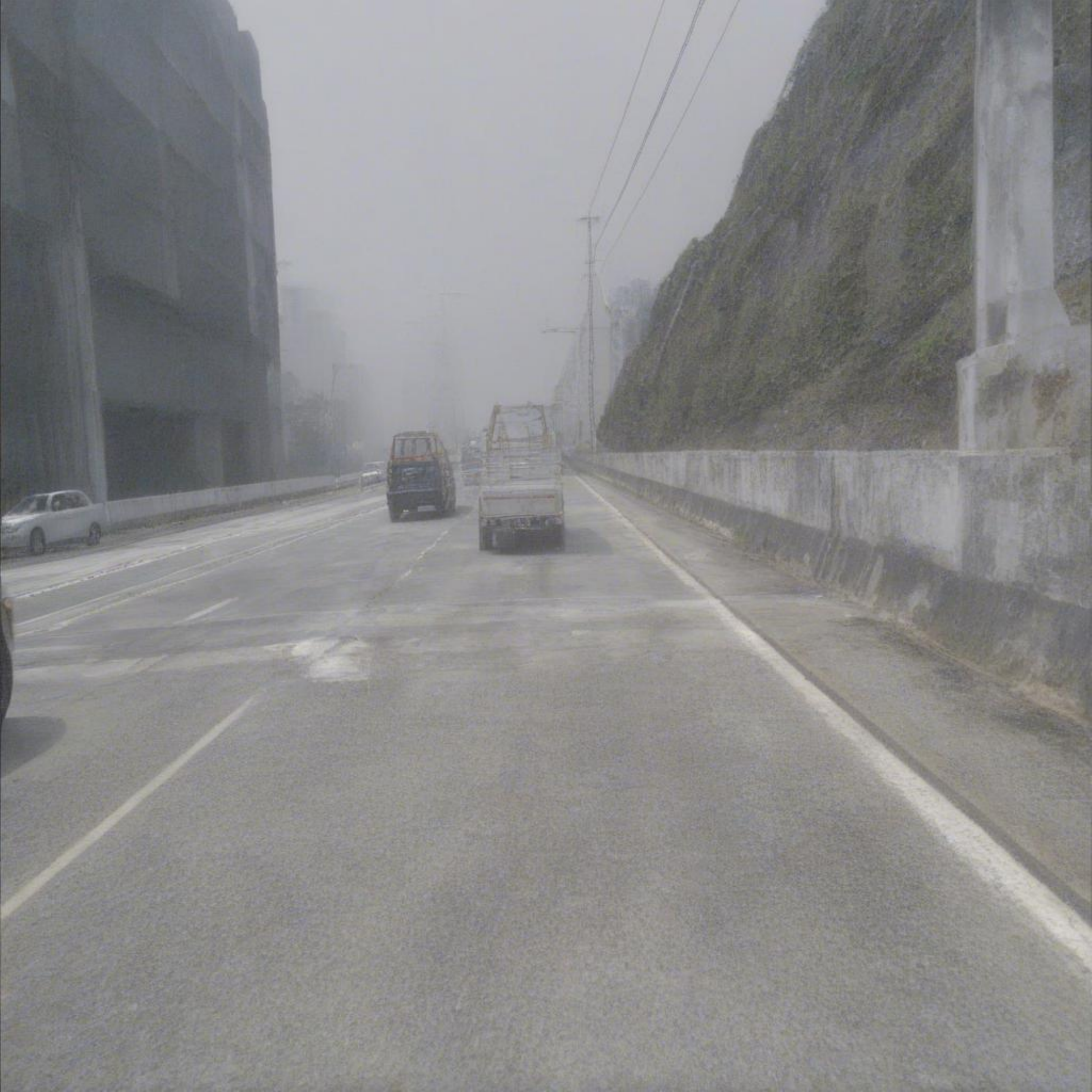}
& \includegraphics[width=1.7cm,height=1.7cm]{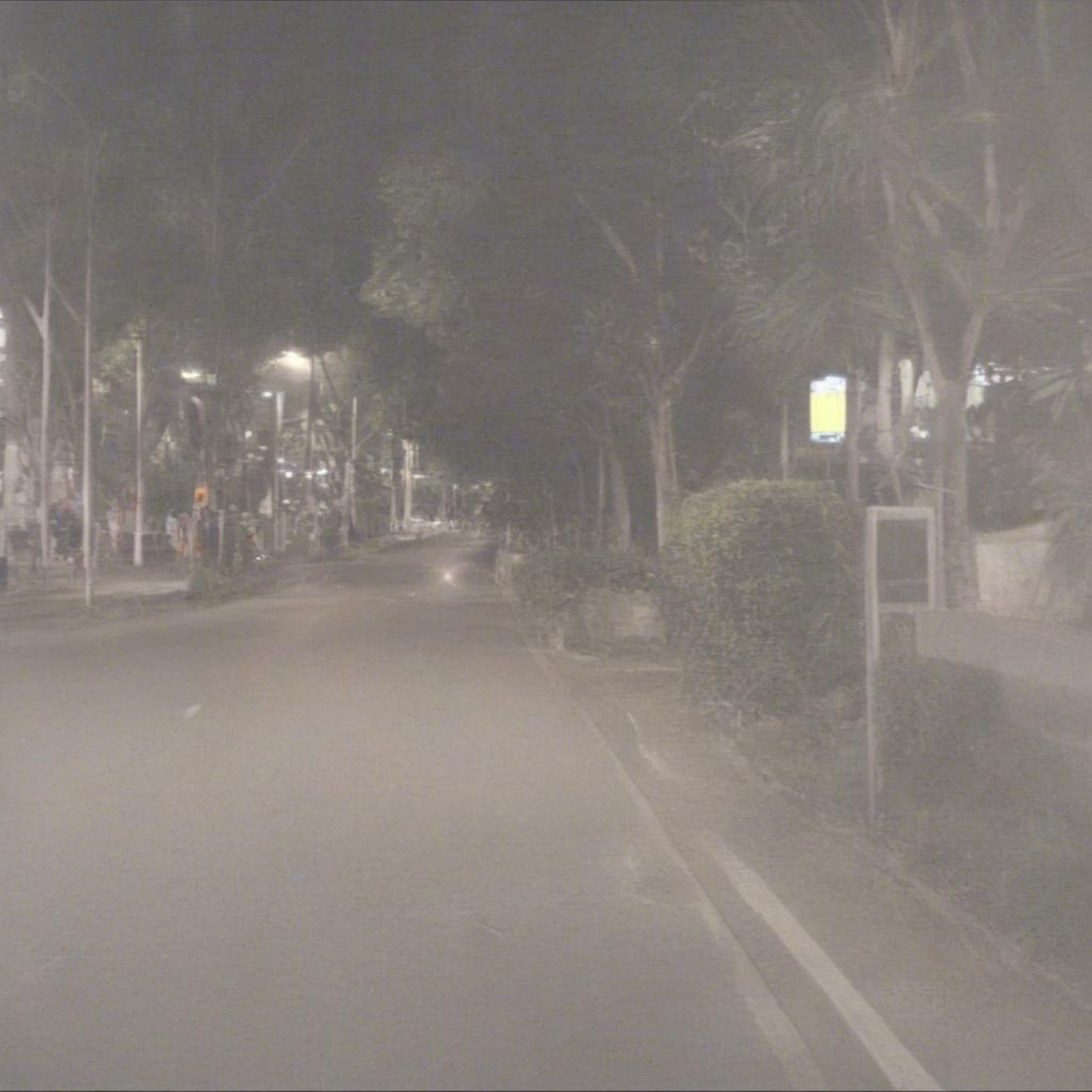}
& \includegraphics[width=1.7cm,height=1.7cm]{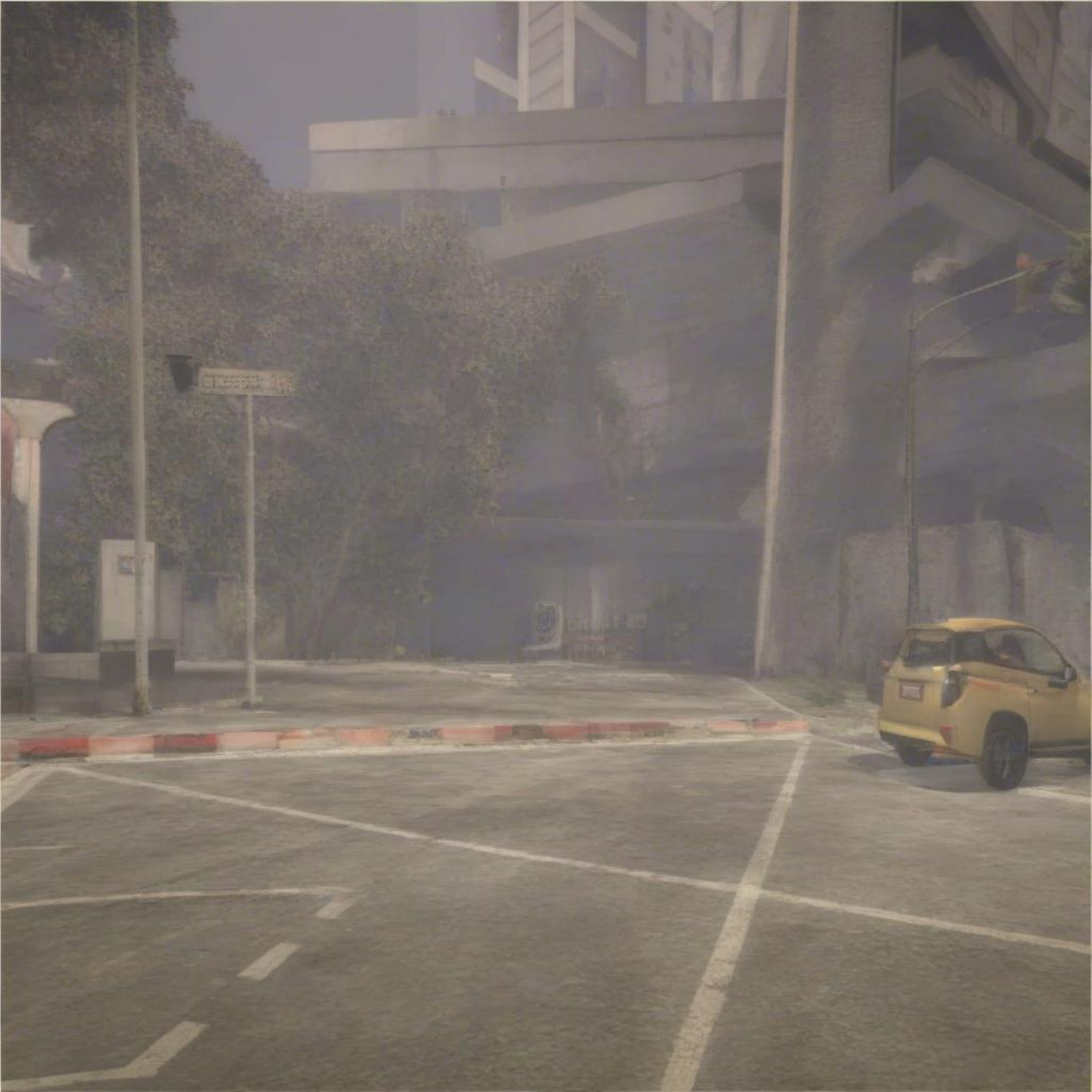}
& \includegraphics[width=1.7cm,height=1.7cm]{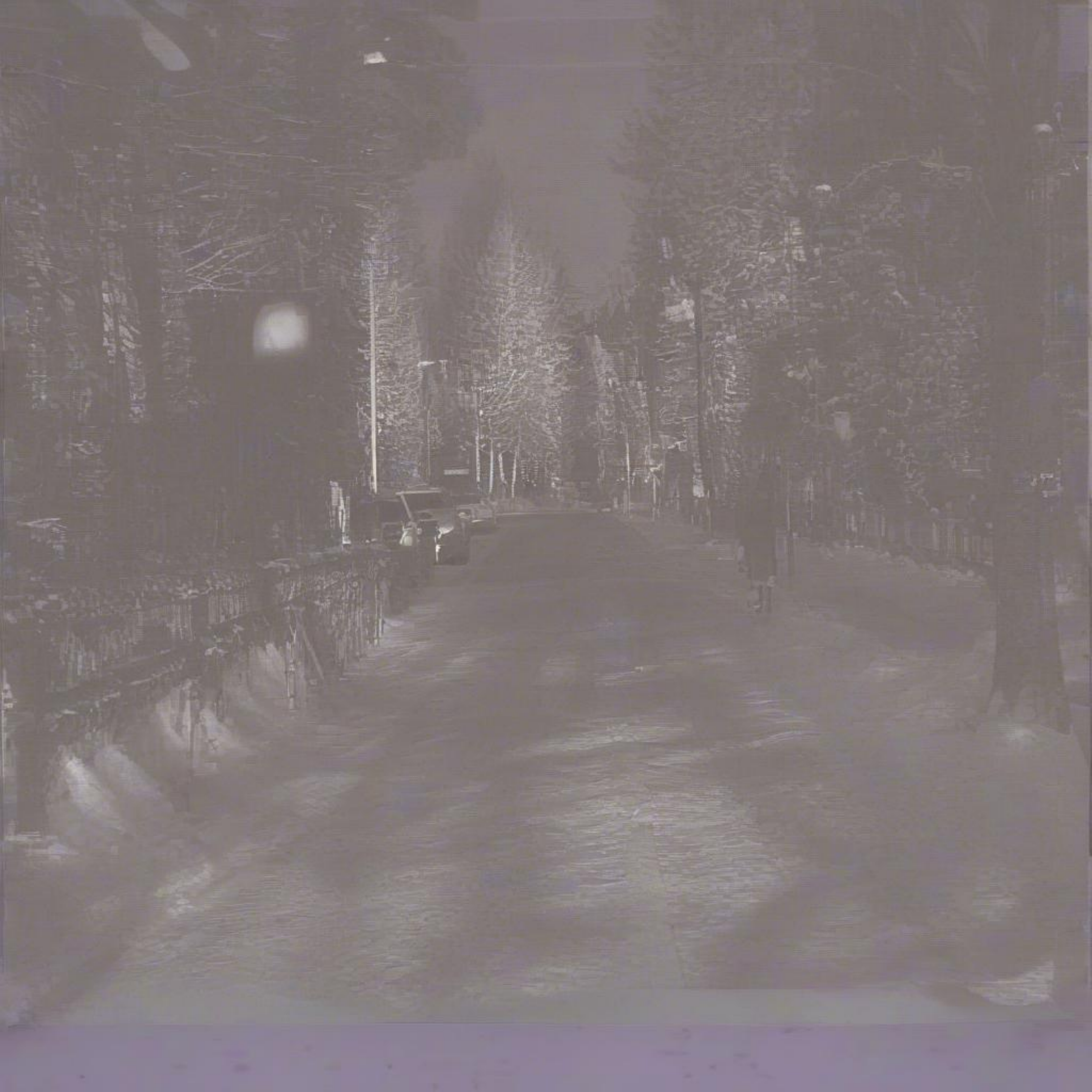}
& \includegraphics[width=1.7cm,height=1.7cm]{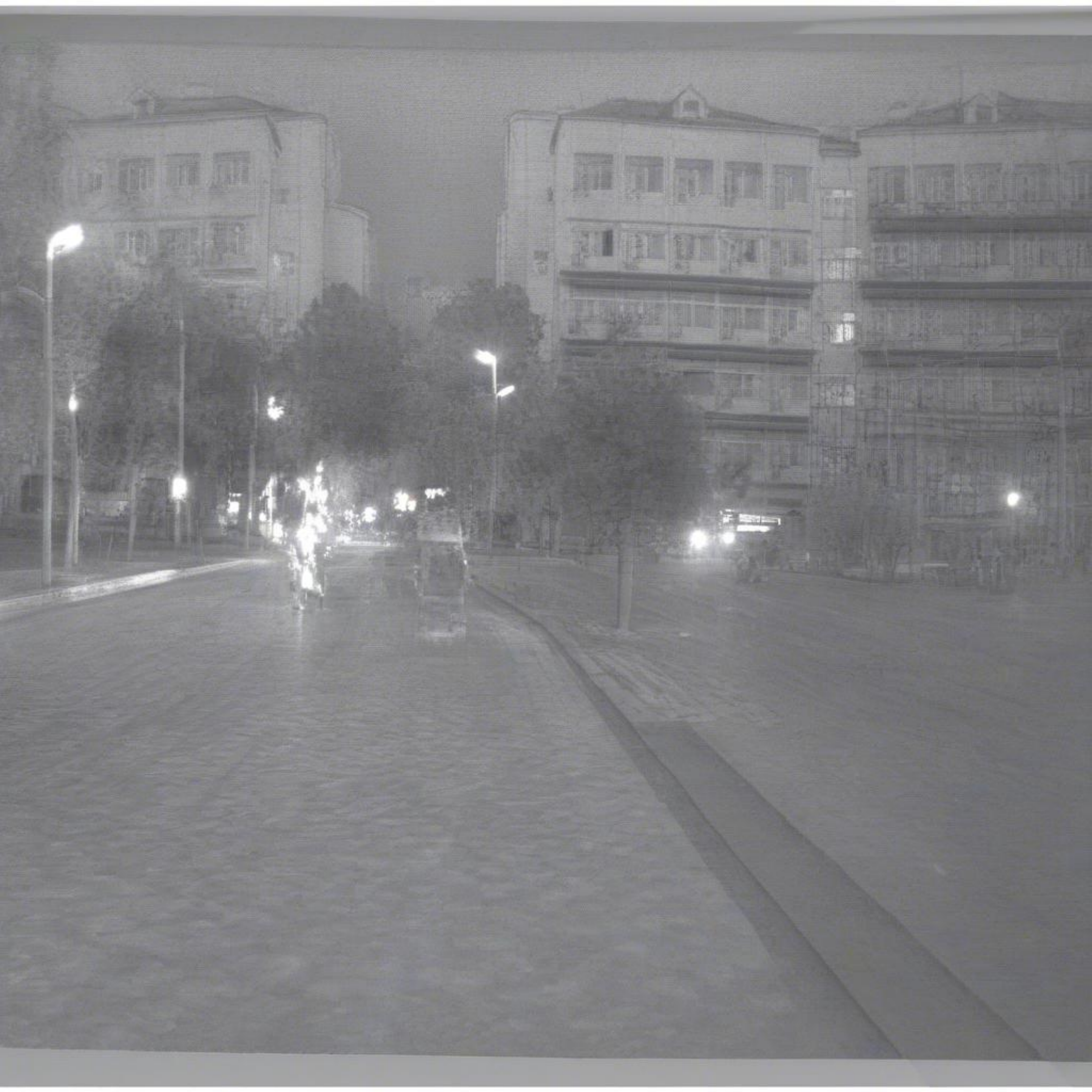}
& \includegraphics[width=1.7cm,height=1.7cm]{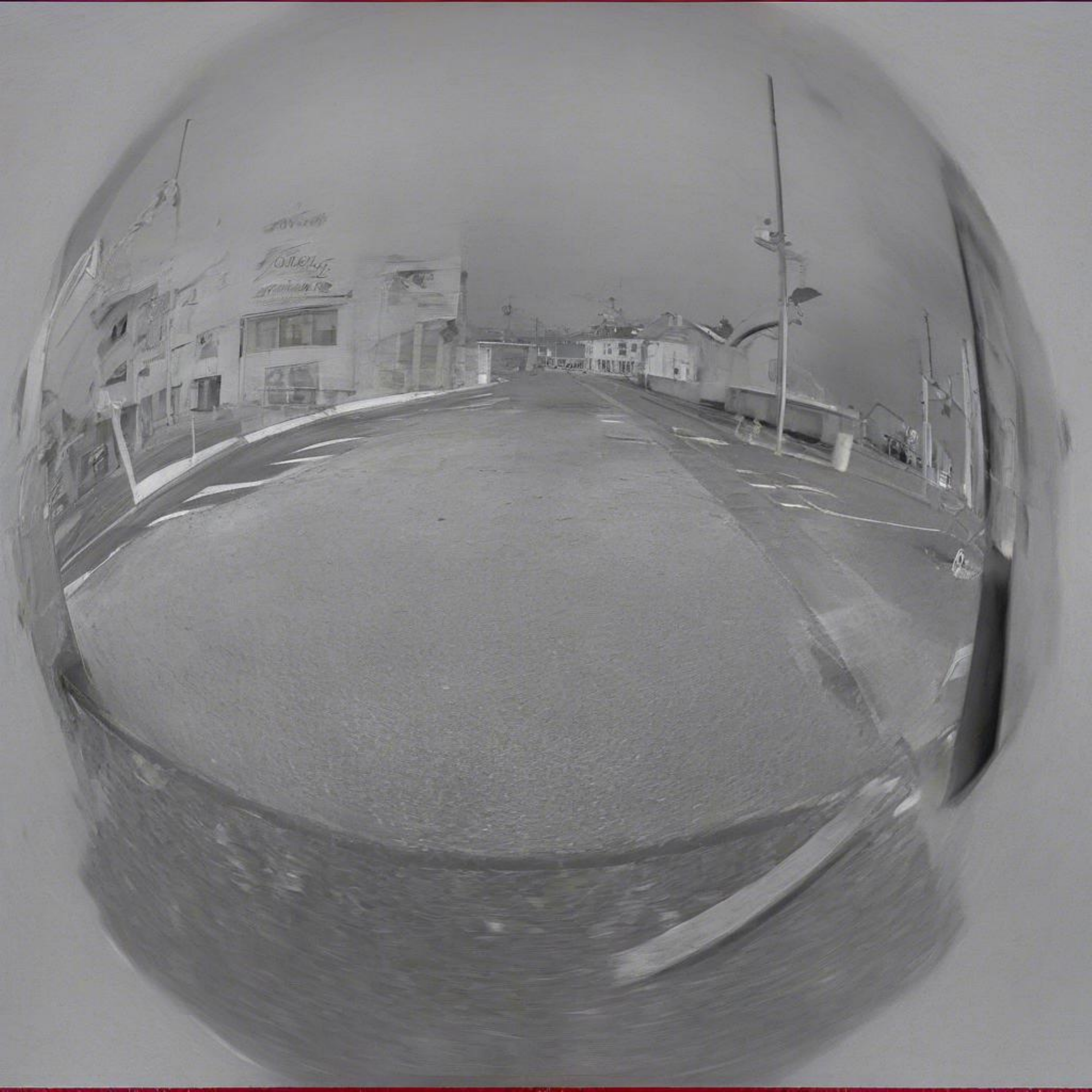}
& \includegraphics[width=1.7cm,height=1.7cm]{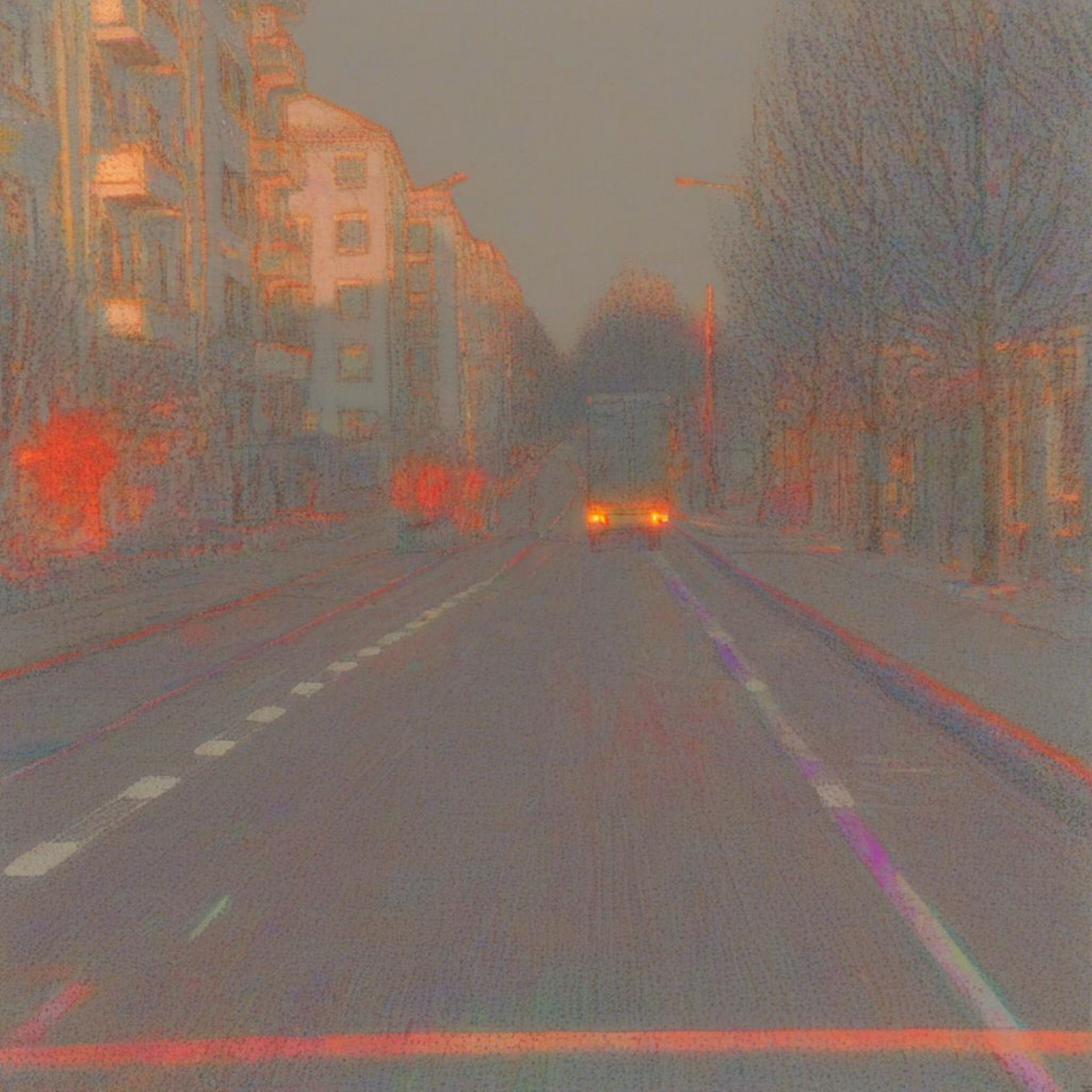}
& \includegraphics[width=1.7cm,height=1.7cm]{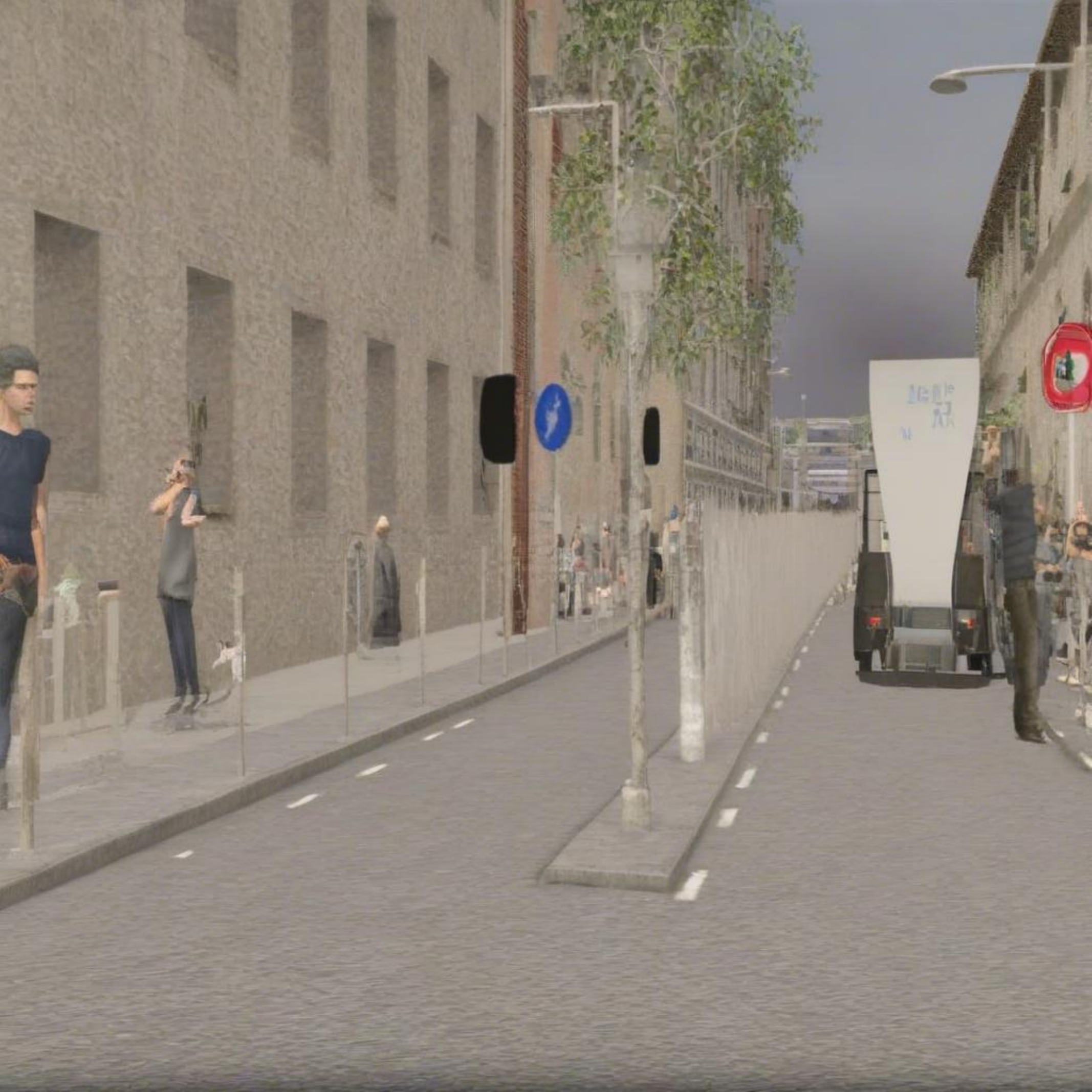} \\

\bottomrule
\end{tabular}}
\label{tab:qual_novel_controlnet}
\end{table*}


\begin{table*}[t]
\centering
\caption{\textbf{Quantitative comparison on novel factor combinations.} Performance metrics across methods with and without ControlNet structural guidance, with best values highlighted in bold. X-MULTI (ours) achieves best overall disentanglement.}
\scriptsize
\setlength{\tabcolsep}{3.2pt}
\renewcommand{\arraystretch}{1.08}
\resizebox{\textwidth}{!}{
\begin{tabular}{lccccccccccc}
\toprule
\multirow{3}{*}{Method}
& \multicolumn{6}{c}{Image Generation Metrics}
& \multicolumn{5}{c}{Improved-Factor Alignment Accuracy (I-FAA)} \\
\cmidrule(lr){2-7} \cmidrule(lr){8-12}
& \multirow{2}{*}{IS~\cite{salimans2016improved} $\uparrow$}
& \multicolumn{4}{c}{CLIP Score~\cite{hessel2021clipscore} $\uparrow$}
& \multirow{2}{*}{DS~\cite{zhang2018unreasonable} $\uparrow$}
& \multirow{2}{*}{Lens $\uparrow$}
& \multirow{2}{*}{Sensor $\uparrow$}
& \multirow{2}{*}{Domain $\uparrow$}
& \multirow{2}{*}{View $\uparrow$}
& \multirow{2}{*}{Avg. $\uparrow$} \\
\cmidrule(lr){3-6}
& & factor & context & average & full-prompt & & & & & & \\
\midrule

DreamBooth~\cite{ruiz2023dreambooth} & \textbf{3.68} & 23.96 & 23.35 & 23.66 & 24.67 & 0.55 & \textbf{0.78} & 0.24 & 0.48 & 0.27 & 0.44 \\
SDXL Zeroshot~\cite{podell2023sdxl} & 3.11 & 22.31 & \textbf{25.59} & 23.95 & 27.80 & \textbf{0.59} & 0.72 & 0.25 & 0.29 & 0.29 & 0.39 \\
MULTI~\cite{godavarthy2026multi} & 3.28 & 24.20 & 23.43 & 23.82 & 27.32 & 0.65 & 0.63 & 0.29 & 0.64 & 0.34 & 0.47 \\
X-MULTI (ours)& 3.23 & \textbf{24.62} & 23.57 & \textbf{24.10} & \textbf{28.22} & \textbf{0.59} & 0.76 & \textbf{0.30} & \textbf{0.66} & \textbf{0.40} & \textbf{0.53} \\

\midrule
DreamBooth-Canny & \textbf{5.78} 
& 23.67 & 25.79 & 25.23 & 29.12 
& 0.50 
& 0.99 & 0.71 & \textbf{0.89} & 0.80 & 0.84 \\
SDXL Zeroshot-Canny & 5.74 
& 23.88 & \textbf{26.75} & \textbf{25.31} & 28.95 
& 0.51 
& 0.99 & 0.63 & 0.84 & 0.75 & 0.78 \\
MULTI-Canny & 4.88 & 22.82 & 23.08 & 22.95 & 26.72 & 0.55 & 0.99 & 0.73 & 0.84 & 0.86 & 0.85 \\
X-MULTI-Canny (ours) & 4.64
& \textbf{23.91} & 23.12 & 23.51 & \textbf{29.60} 
& \textbf{0.56} 
& \textbf{0.99} & \textbf{0.81} & \textbf{0.89} & \textbf{0.87} & \textbf{0.89} \\

\midrule
DreamBooth-Depth       
& \textbf{5.56}
& 23.57 & 25.40 & 24.98 & 27.95 
& \textbf{0.53}
& \textbf{1.00} & 0.69 & \textbf{0.88} & 0.77 & 0.84 \\
SDXL Zeroshot-Depth & 5.49
& 23.62 & \textbf{26.57} & \textbf{25.10} & \textbf{29.82} 
& 0.52 
& \textbf{1.00} & 0.66 & 0.81 & 0.71 & 0.77 \\
MULTI-Depth & 4.36 & 22.12 & 23.45 & 22.78 & 26.82 & 0.52 & 0.99 & \textbf{0.76} & 0.84 & 0.87 & 0.86 \\
X-MULTI-Depth (ours)& 4.87
& \textbf{23.63} & 24.79 & 23.21 & 29.81 
& \textbf{0.53} 
& \textbf{1.00} & 0.71 & \textbf{0.88} & \textbf{0.88} & \textbf{0.87} \\

\bottomrule
\end{tabular}}
\label{tab:novel_quantitative}
\end{table*}

\begin{table}[t]
\centering
\caption{\textbf{Comparison of attention maps for FAA and I-FAA.} Grad-CAM visualizations showing that I-FAA (ours) learns more semantically meaningful attention regions.}
\label{tab:attention}
\scriptsize
\setlength{\tabcolsep}{4pt}       
\renewcommand{\arraystretch}{0.5} 

\begin{tabular}{cccccc}
\toprule
\textbf{Image} & \textbf{Metric} & \textbf{Lens} & \textbf{Sensor} & \textbf{Domain} & \textbf{View} \\
\midrule

\multirow{2}{*}{\scalebox{1.0}{\begin{tabular}[c]{@{}c@{}}\includegraphics[width=0.090\textwidth]{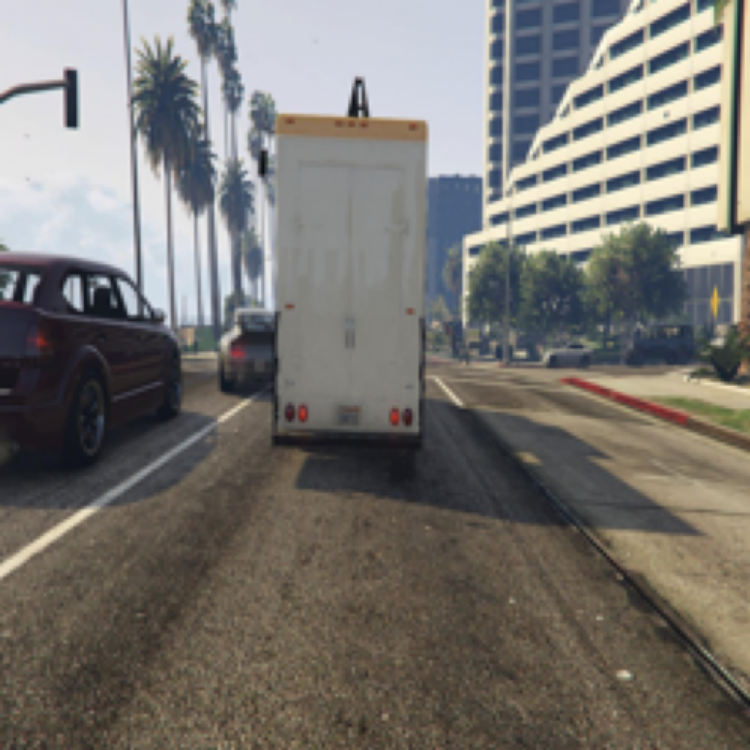}\\[-0.2ex]{\tiny normal/rgb/video-game/front}\end{tabular}}}
& \textbf{FAA}~\cite{godavarthy2026multi}
& \includegraphics[width=0.075\textwidth]{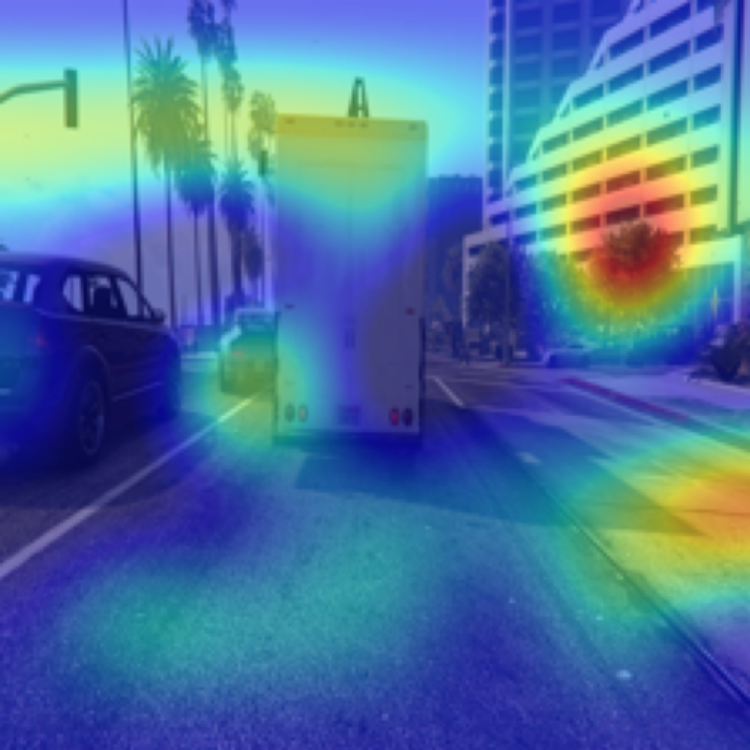}
& \includegraphics[width=0.075\textwidth]{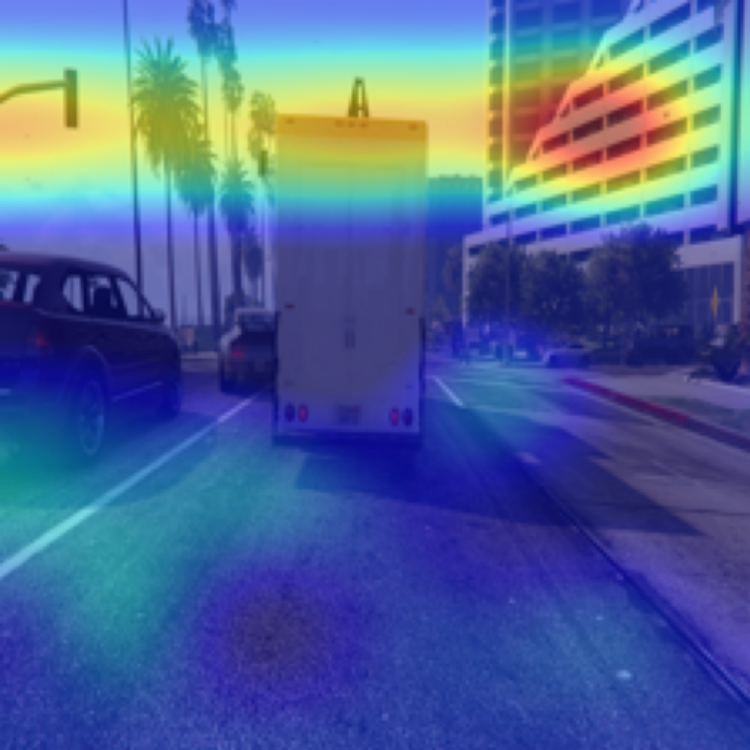}
& \includegraphics[width=0.075\textwidth]{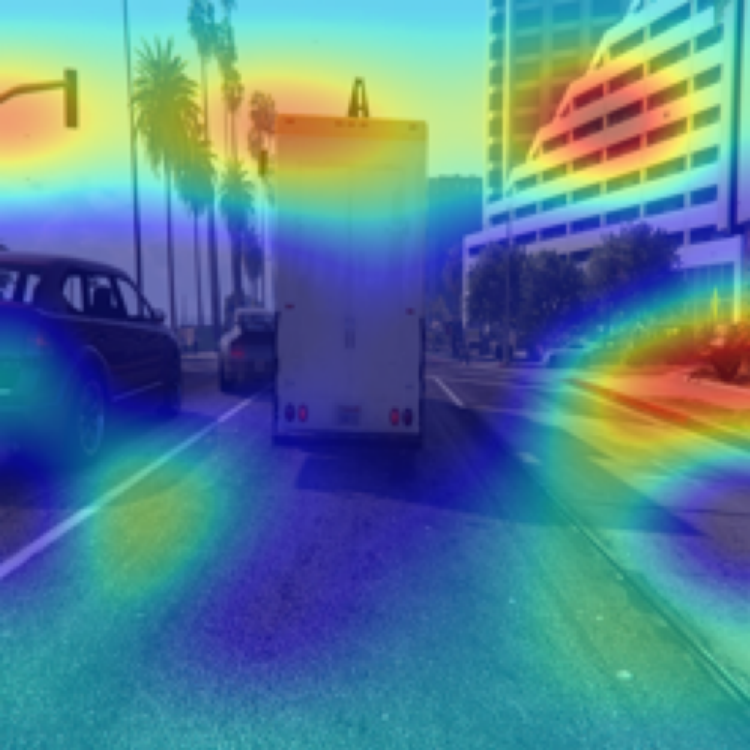}
& \includegraphics[width=0.075\textwidth]{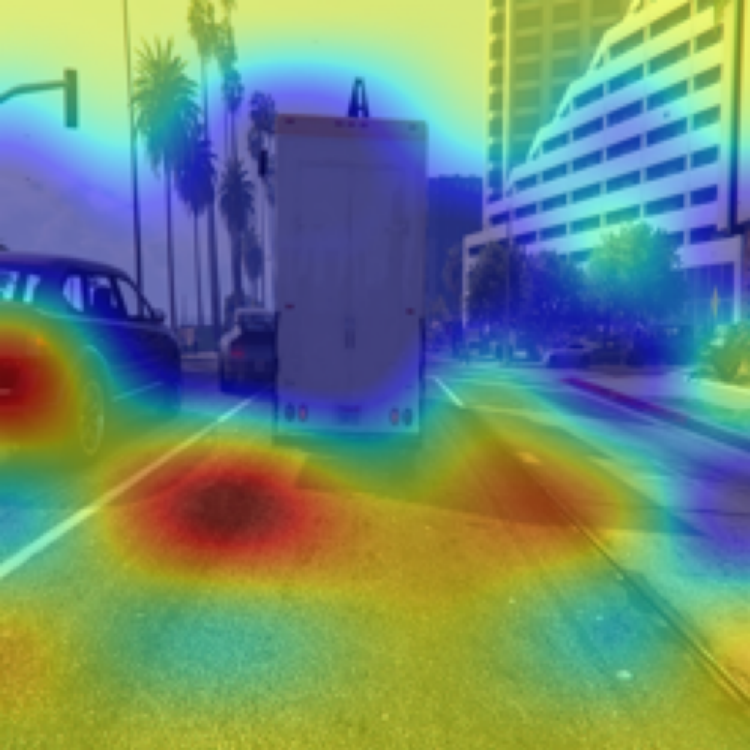} \\
\cmidrule(lr){2-6}
& \textbf{I-FAA}
& \includegraphics[width=0.075\textwidth]{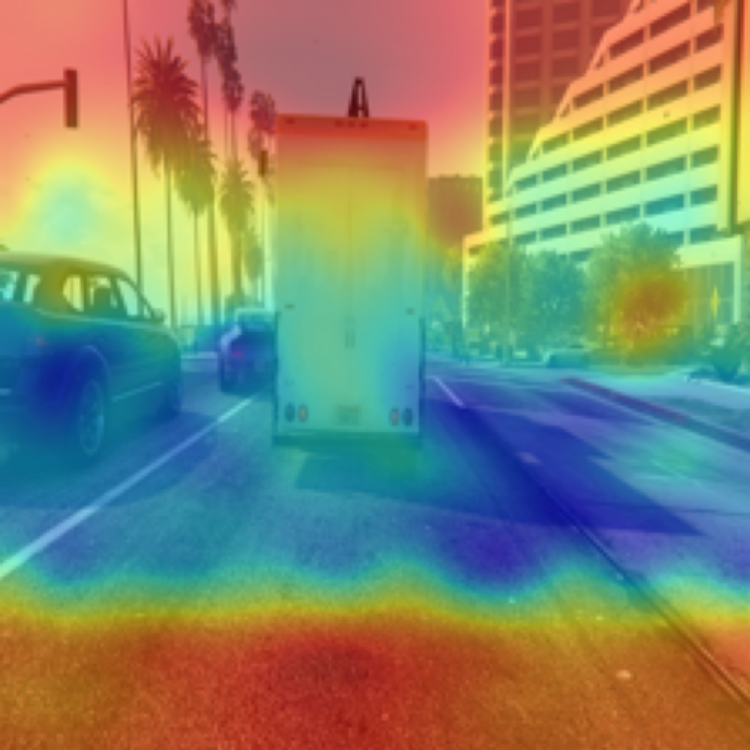}
& \includegraphics[width=0.075\textwidth]{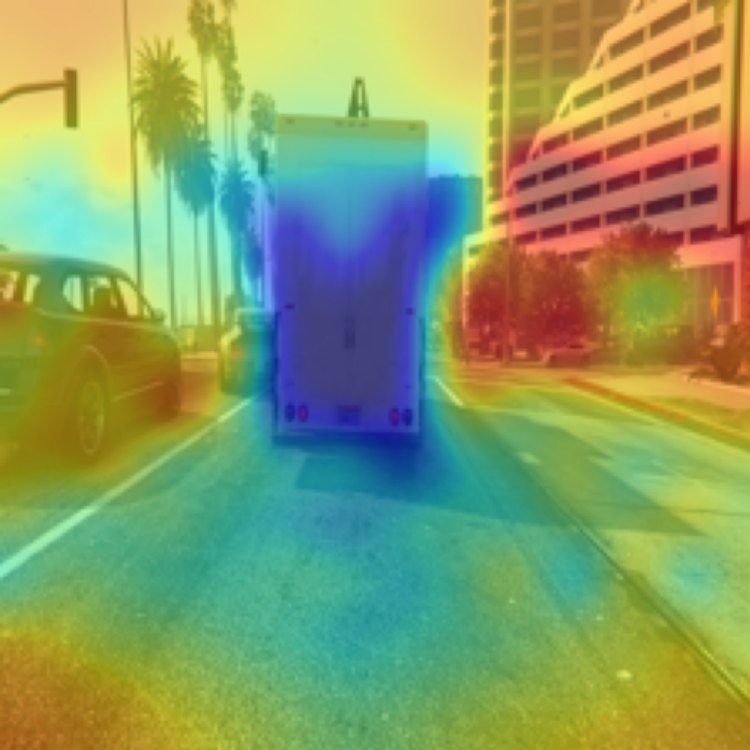}
& \includegraphics[width=0.075\textwidth]{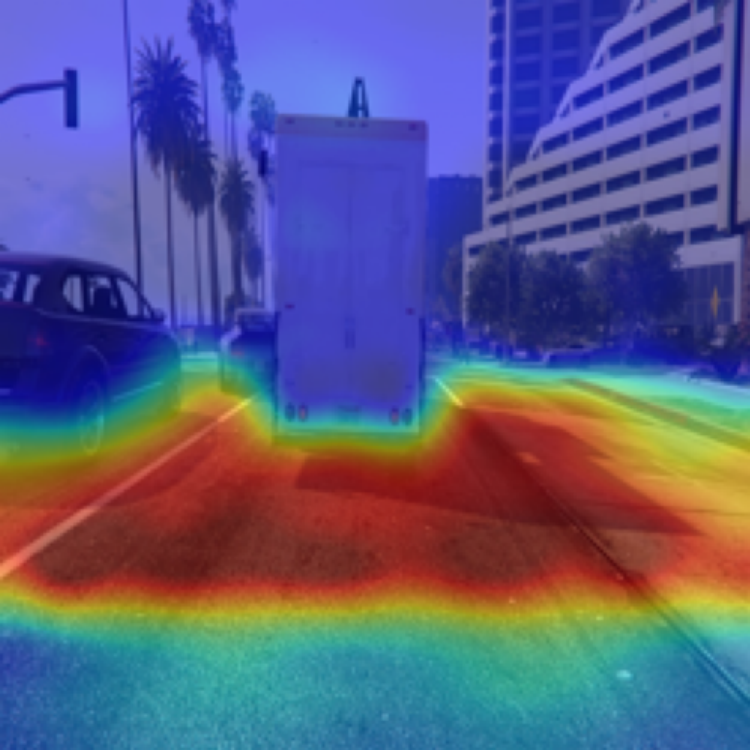}
& \includegraphics[width=0.075\textwidth]{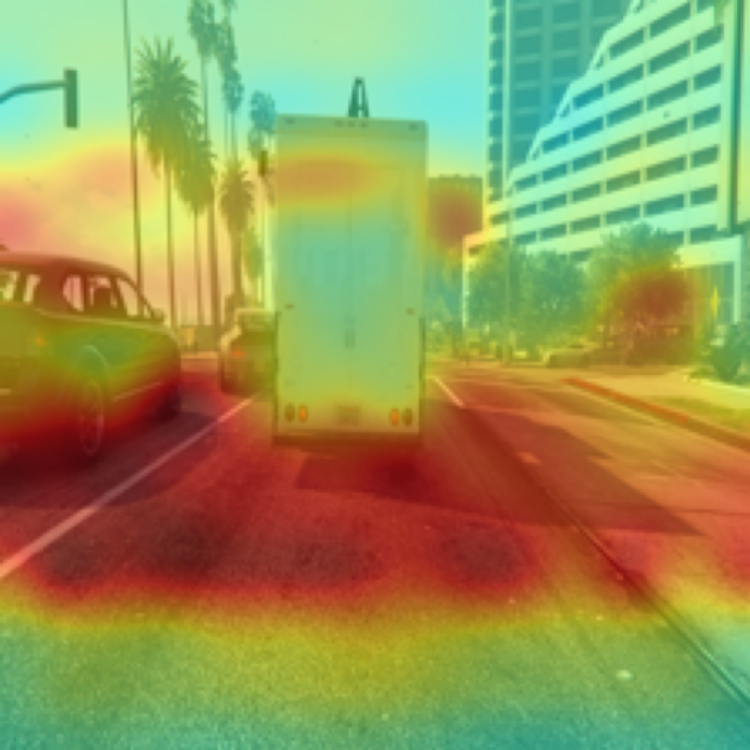} \\

\midrule

\multirow{2}{*}{\scalebox{1.0}{\begin{tabular}[c]{@{}c@{}}\includegraphics[width=0.090\textwidth]{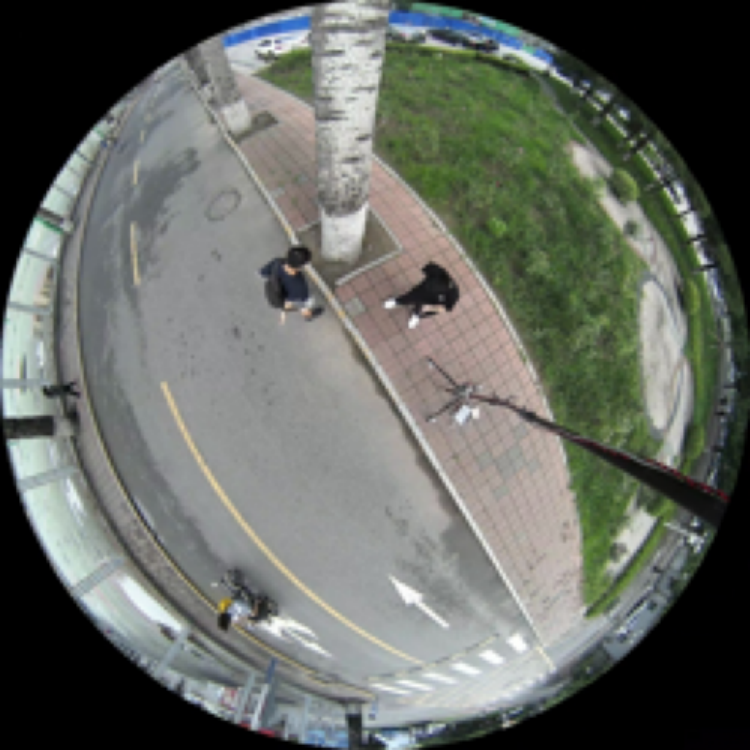}\\[-0.2ex]{\tiny fisheye/rgb/real/drone}\end{tabular}}}
& \textbf{FAA}~\cite{godavarthy2026multi}
& \includegraphics[width=0.075\textwidth]{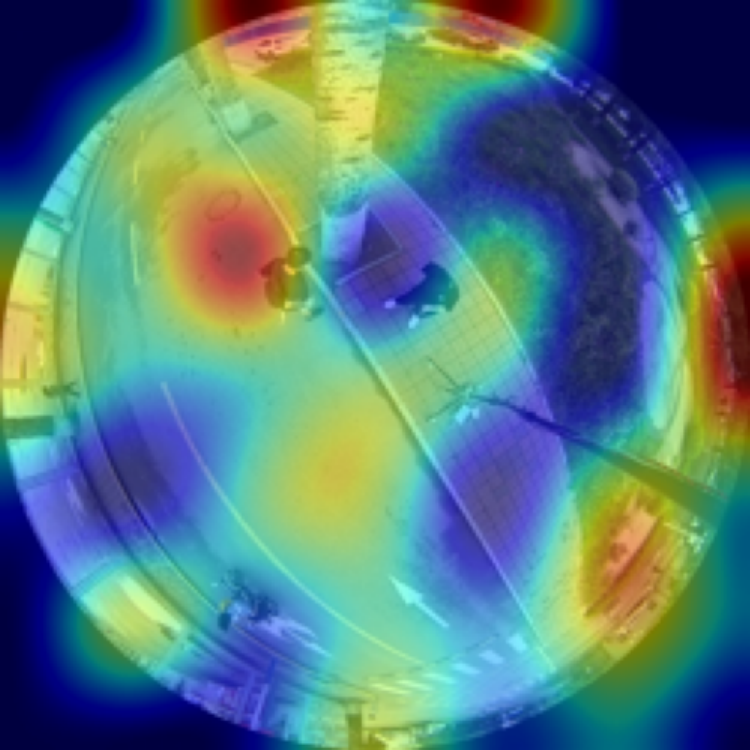}
& \includegraphics[width=0.075\textwidth]{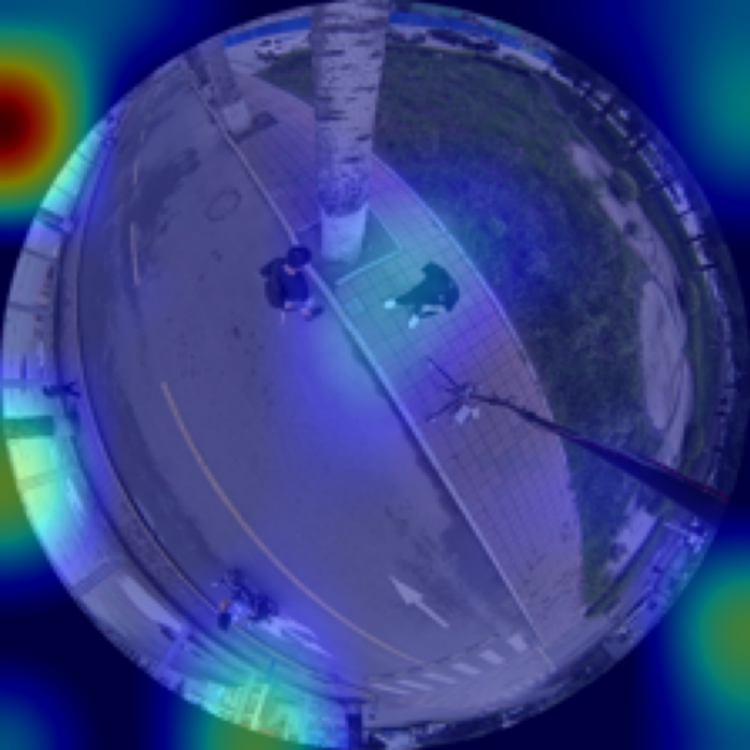}
& \includegraphics[width=0.075\textwidth]{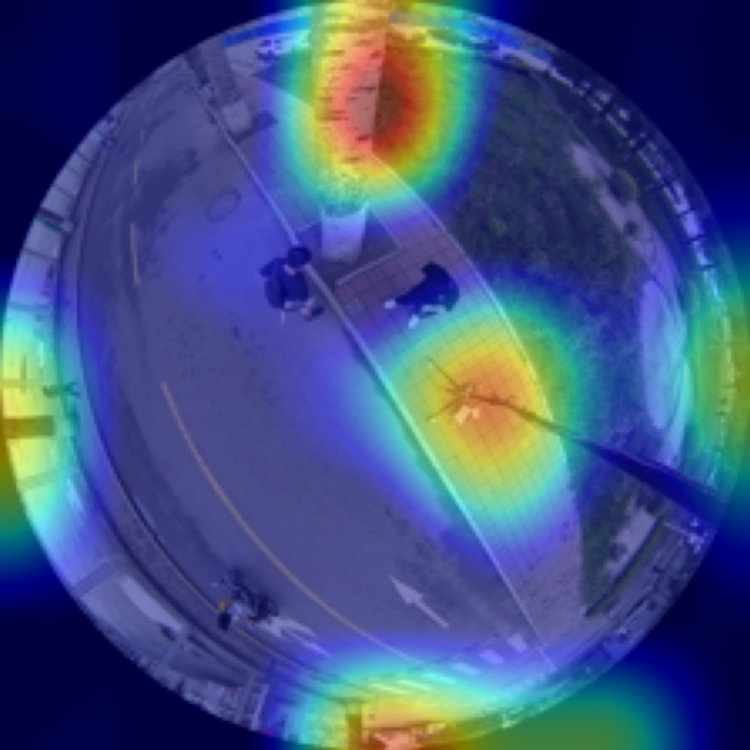}
& \includegraphics[width=0.075\textwidth]{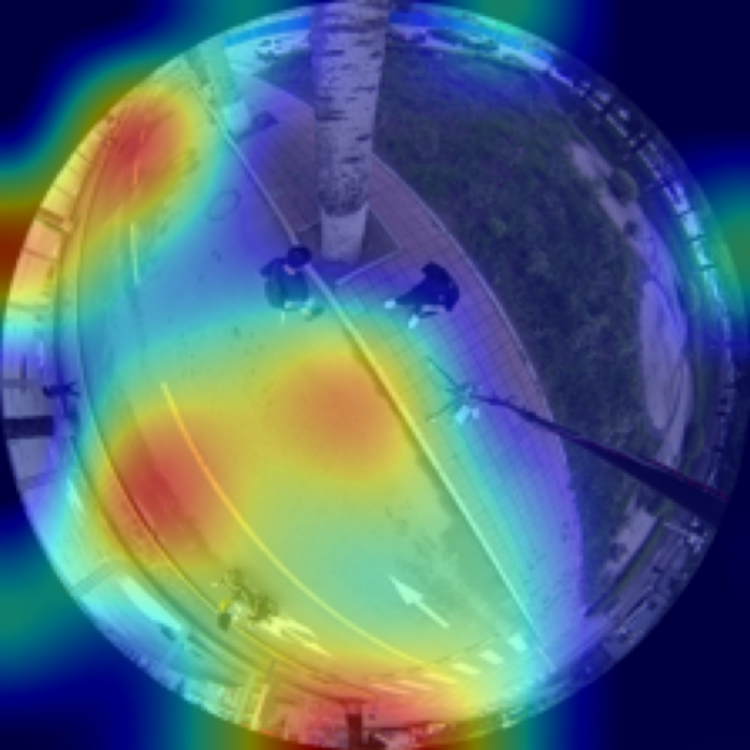} \\
\cmidrule(lr){2-6}
& \textbf{I-FAA}
& \includegraphics[width=0.075\textwidth]{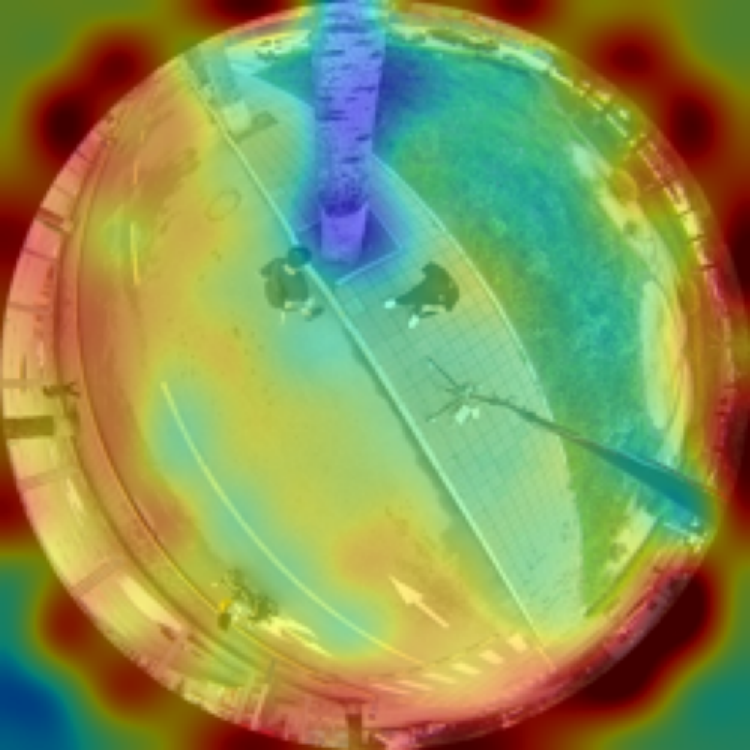}
& \includegraphics[width=0.075\textwidth]{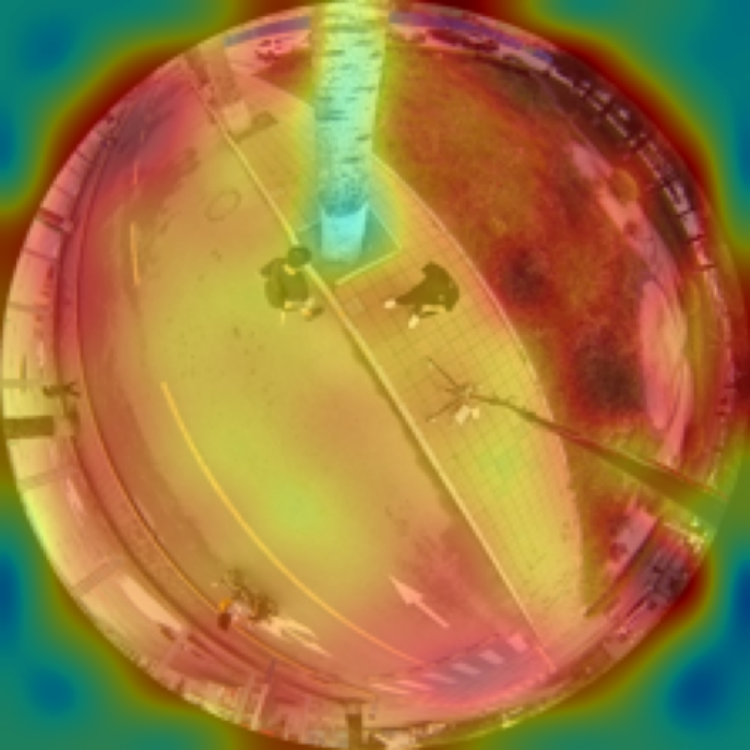}
& \includegraphics[width=0.075\textwidth]{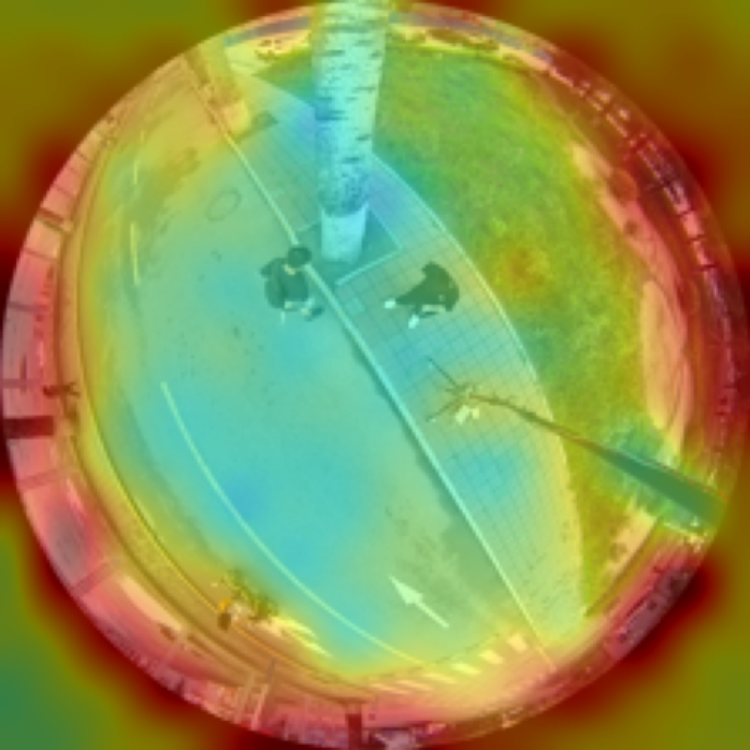}
& \includegraphics[width=0.075\textwidth]{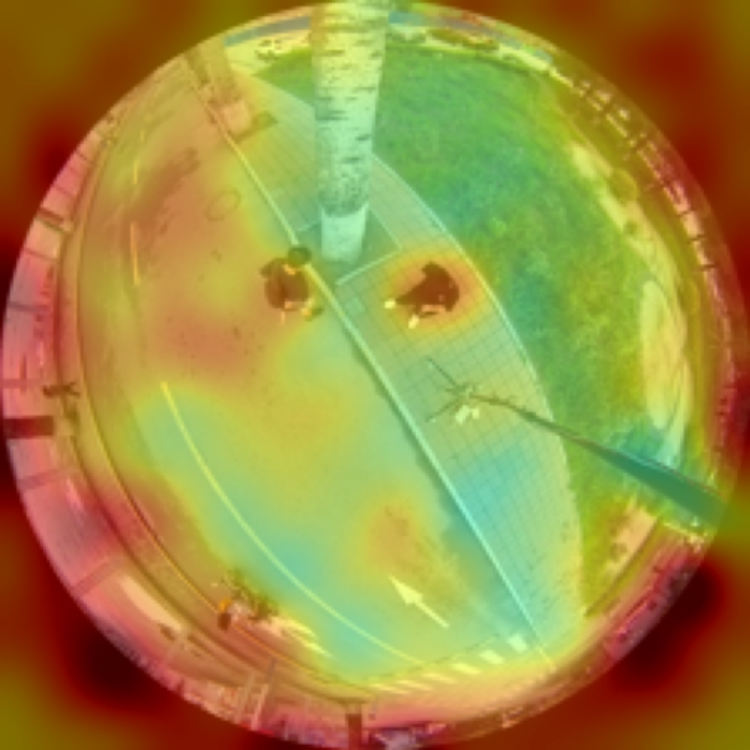} \\

\midrule

\multirow{2}{*}{\scalebox{1.0}{\begin{tabular}[c]{@{}c@{}}\includegraphics[width=0.090\textwidth]{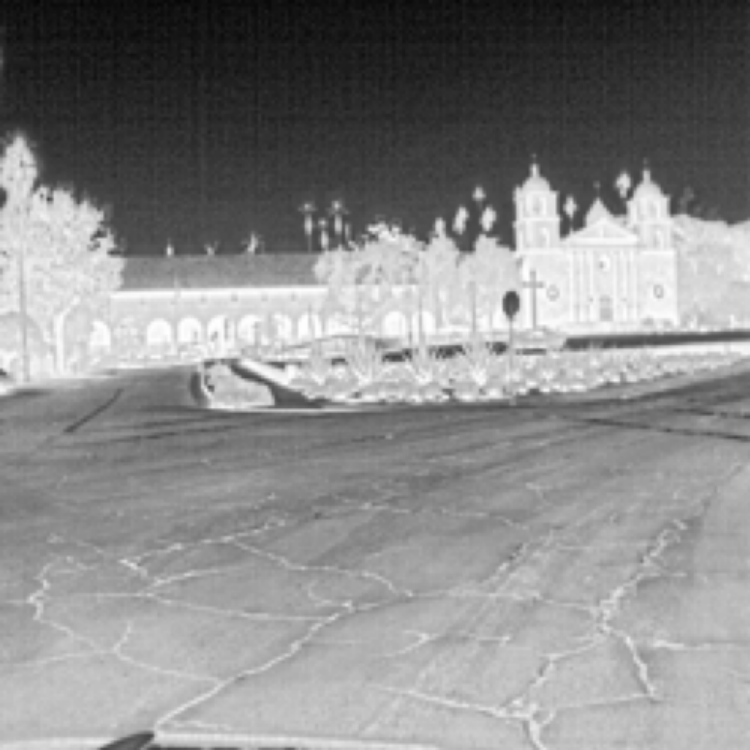}\\[-0.2ex]{\tiny normal/thermal/real/front}\end{tabular}}}
& \textbf{FAA}~\cite{godavarthy2026multi}
& \includegraphics[width=0.075\textwidth]{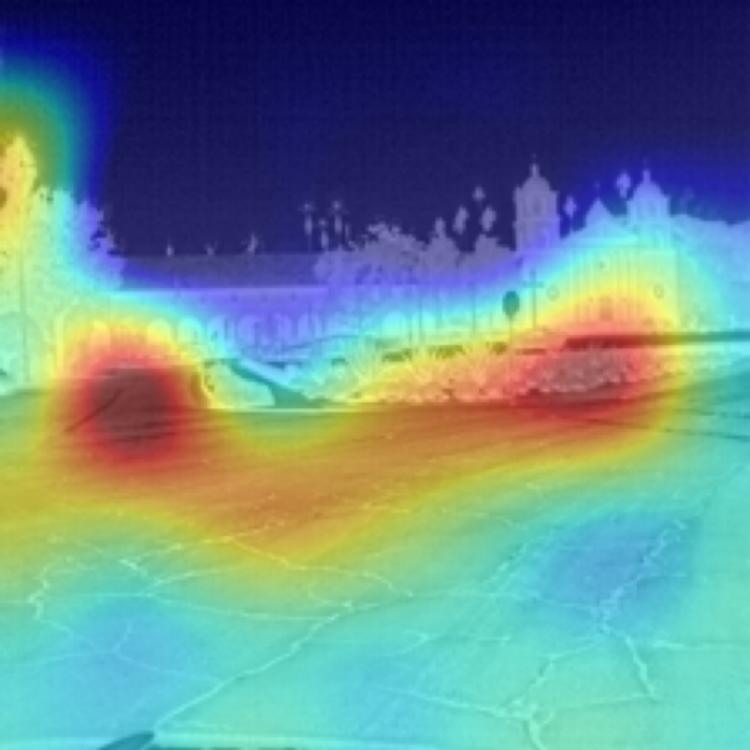}
& \includegraphics[width=0.075\textwidth]{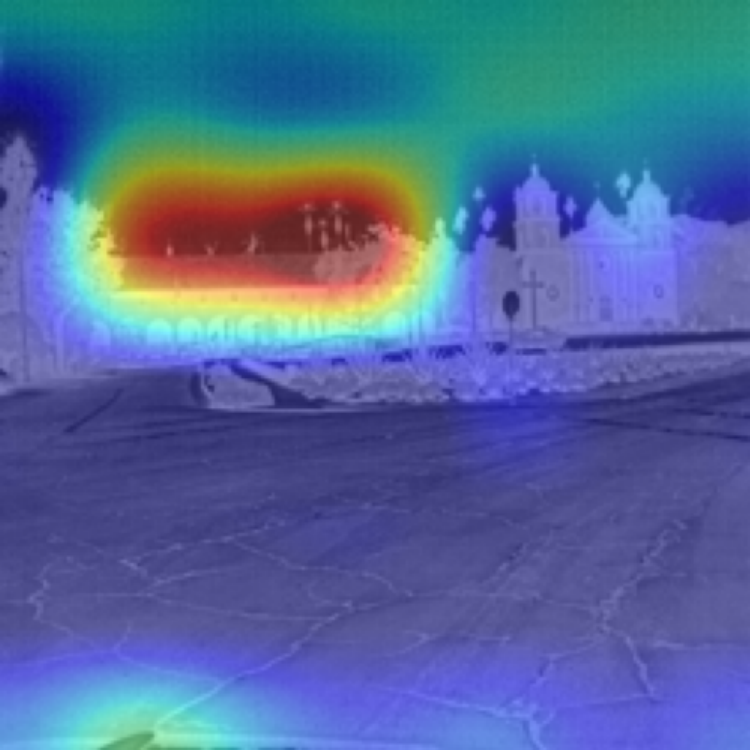}
& \includegraphics[width=0.075\textwidth]{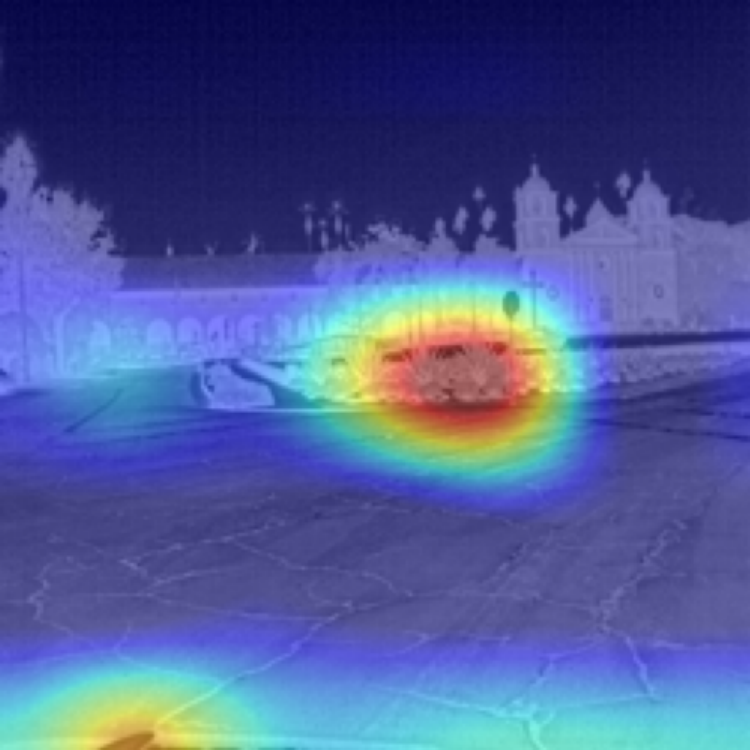}
& \includegraphics[width=0.075\textwidth]{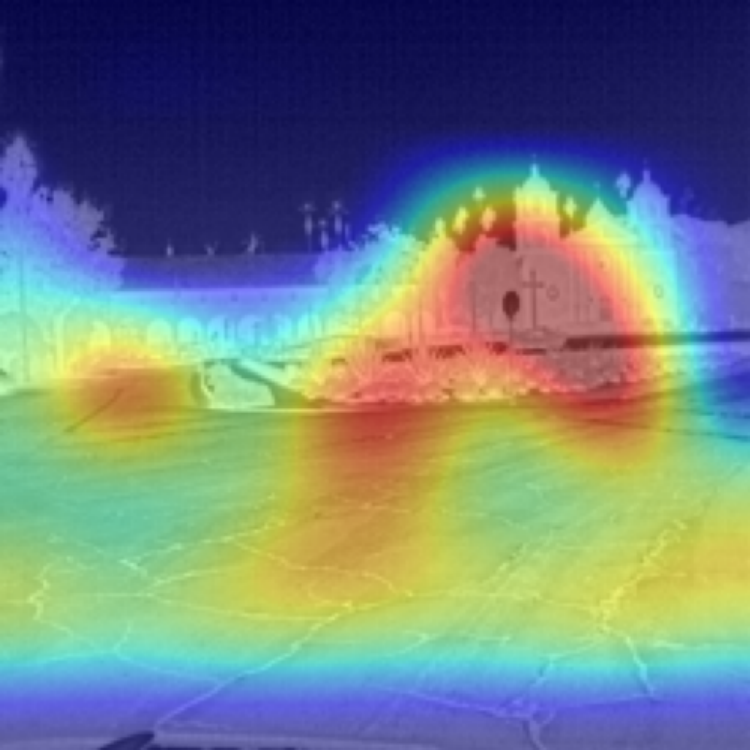} \\
\cmidrule(lr){2-6}
& \textbf{I-FAA}
& \includegraphics[width=0.075\textwidth]{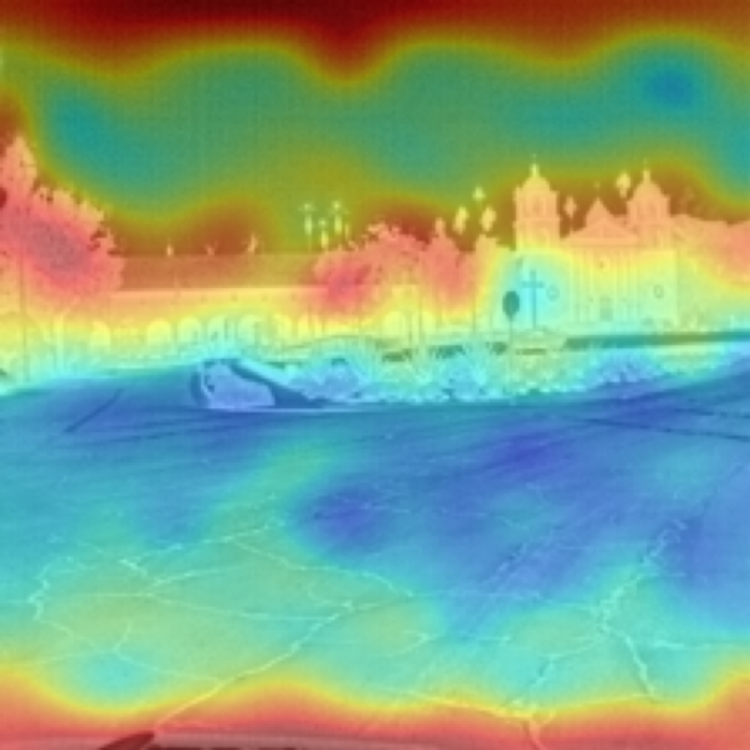}
& \includegraphics[width=0.075\textwidth]{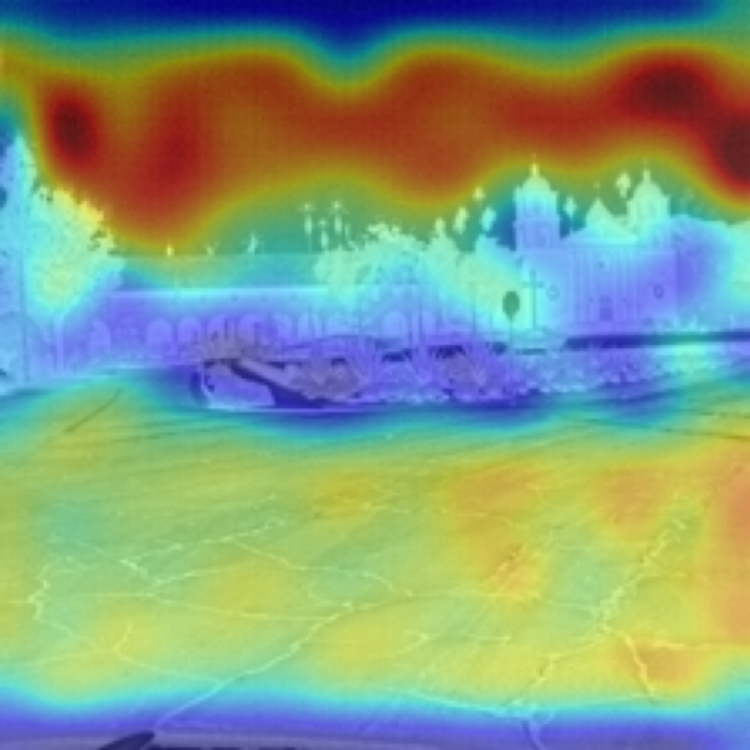}
& \includegraphics[width=0.075\textwidth]{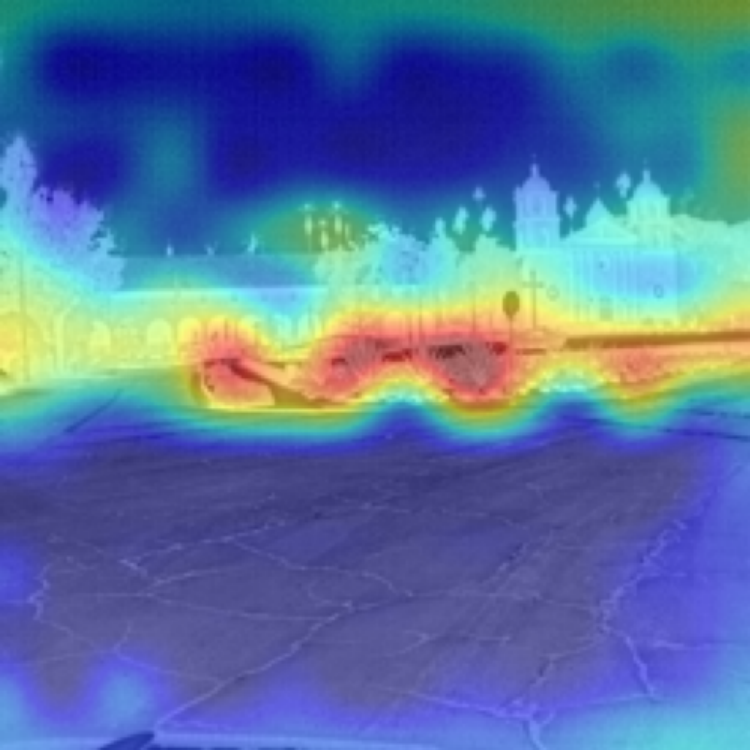}
& \includegraphics[width=0.075\textwidth]{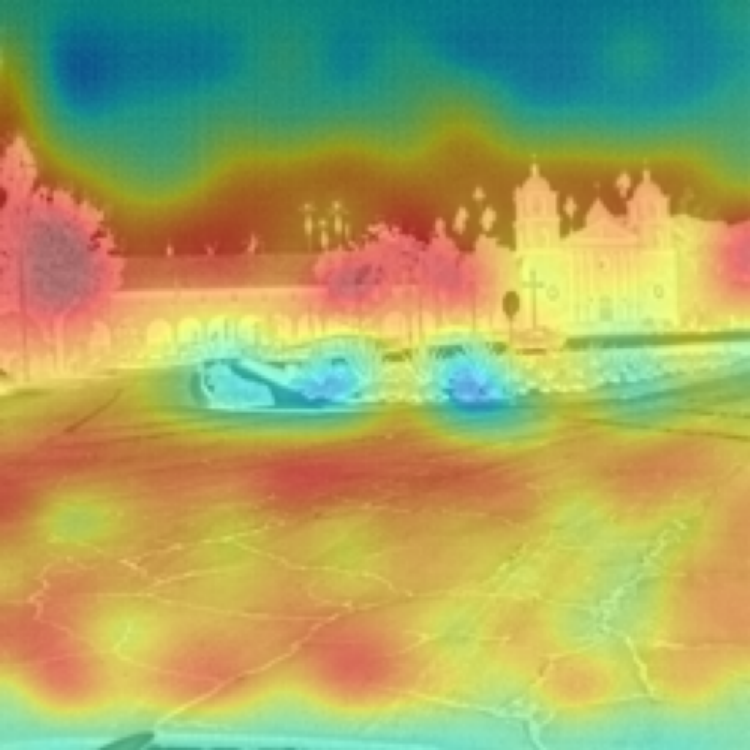} \\

\bottomrule
\end{tabular}
\end{table}


\begin{figure*}[t]
    \centering
    \includegraphics[width=0.60\textwidth]{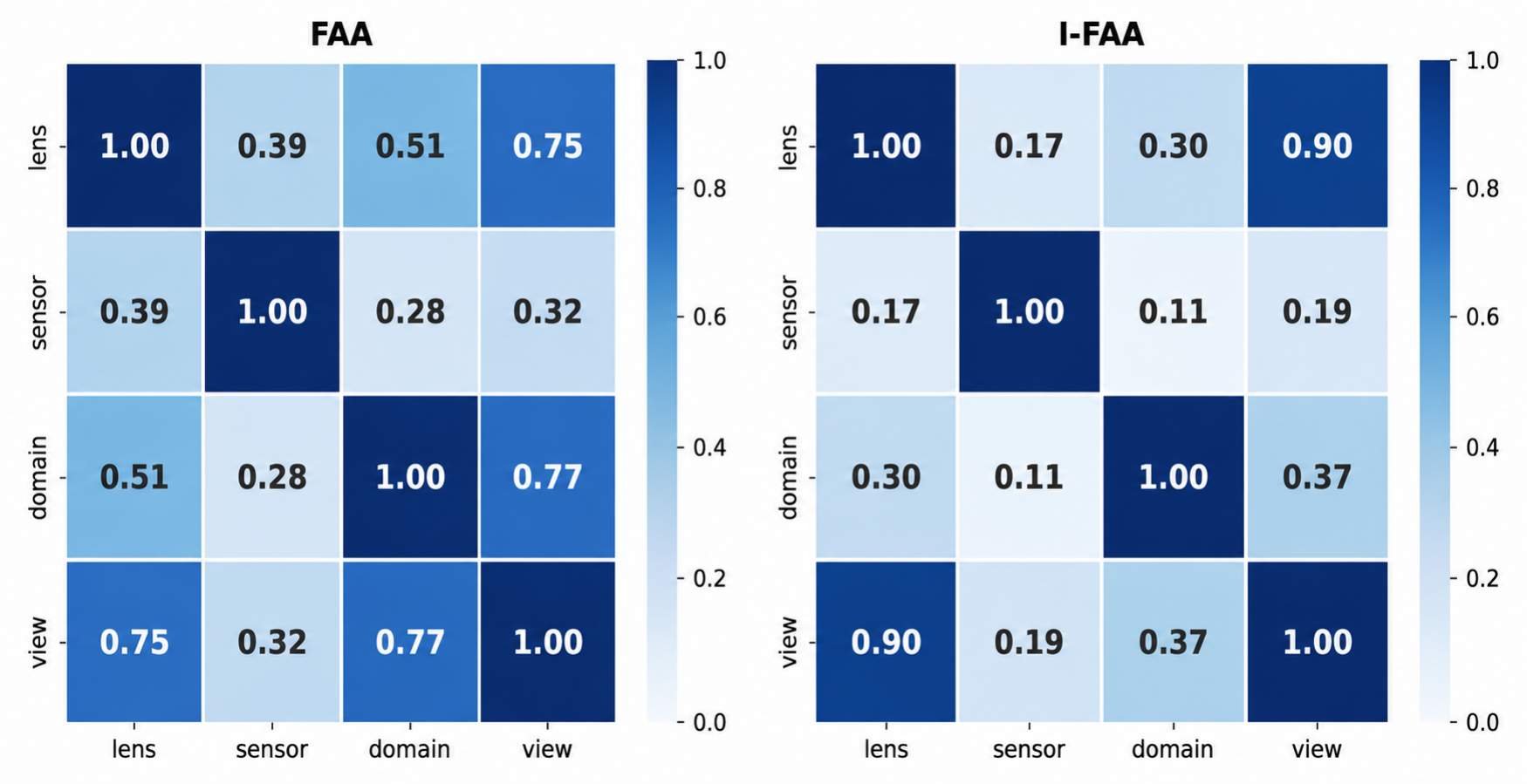}
    \caption{\textbf{Cramer's V correlations.} Pairwise correlations between predicted factors for FAA (left) and I-FAA (ours) (right), showing that I-FAA reduces cross-factor information leakage. High values show strong correlations. }
    \label{fig:cramersv}
\end{figure*}

\begin{table}[t]
\centering
\caption{\textbf{FAA and I-FAA on ground-truth DF-RICO images.} Comparison of FAA and I-FAA (ours) metric on ground-truth images.}
\label{tab:metrics_evaluation_gt}
\scriptsize
\setlength{\tabcolsep}{4pt}
\renewcommand{\arraystretch}{0.95}
\begin{tabular}{lccccc}
\toprule
\textbf{Metric} & \textbf{Lens} & \textbf{Sensor} & \textbf{Domain} & \textbf{View} & \textbf{Avg.} \\  
\midrule
FAA~\cite{godavarthy2026multi}
& 0.23 & 0.83 & 0.50 & 0.17 & 0.43 \\
I-FAA
& \textbf{1.00} & \textbf{0.97} & \textbf{0.98} & \textbf{0.84} & \textbf{0.95} \\
\bottomrule
\end{tabular}
\end{table}


\begin{table*}[t]
\centering
\caption{\textbf{Factor prediction comparison.} Per-factor classification outputs compared against ground-truth labels for ground truth images. I-FAA (ours) predicts factor correctly.}
\label{tab:metric_prediction_comparison_gt}

\providecommand{\cgood}[1]{\textcolor{green!55!black}{#1}}
\providecommand{\cbad}[1]{\textcolor{red!80!black}{#1}}

\resizebox{\textwidth}{!}{%
\begin{minipage}{\textwidth}
\centering
\tiny
\setlength{\tabcolsep}{1.8pt}
\renewcommand{\arraystretch}{0.95}

\begin{subtable}[t]{0.49\textwidth}
\centering

\begin{tabular}{clccc}
\toprule
\textbf{Image} & \textbf{Factor} & \textbf{GT} & \textbf{FAA~\cite{godavarthy2026multi}} & \textbf{I-FAA} \\
\midrule

\multirow{4}{*}{\includegraphics[width=0.8cm,height=0.8cm]{figures/ground_truth_cominations_predictions/gen1.pdf}}
& Lens   & normal     & \cbad{fisheye}     & \cgood{normal} \\
& Sensor & rgb        & \cgood{rgb}        & \cgood{rgb} \\
& Domain & real       & \cbad{simulation}  & \cgood{real} \\
& View   & front      & \cgood{front}      & \cgood{front} \\
\midrule

\multirow{4}{*}{\includegraphics[width=0.8cm,height=0.8cm]{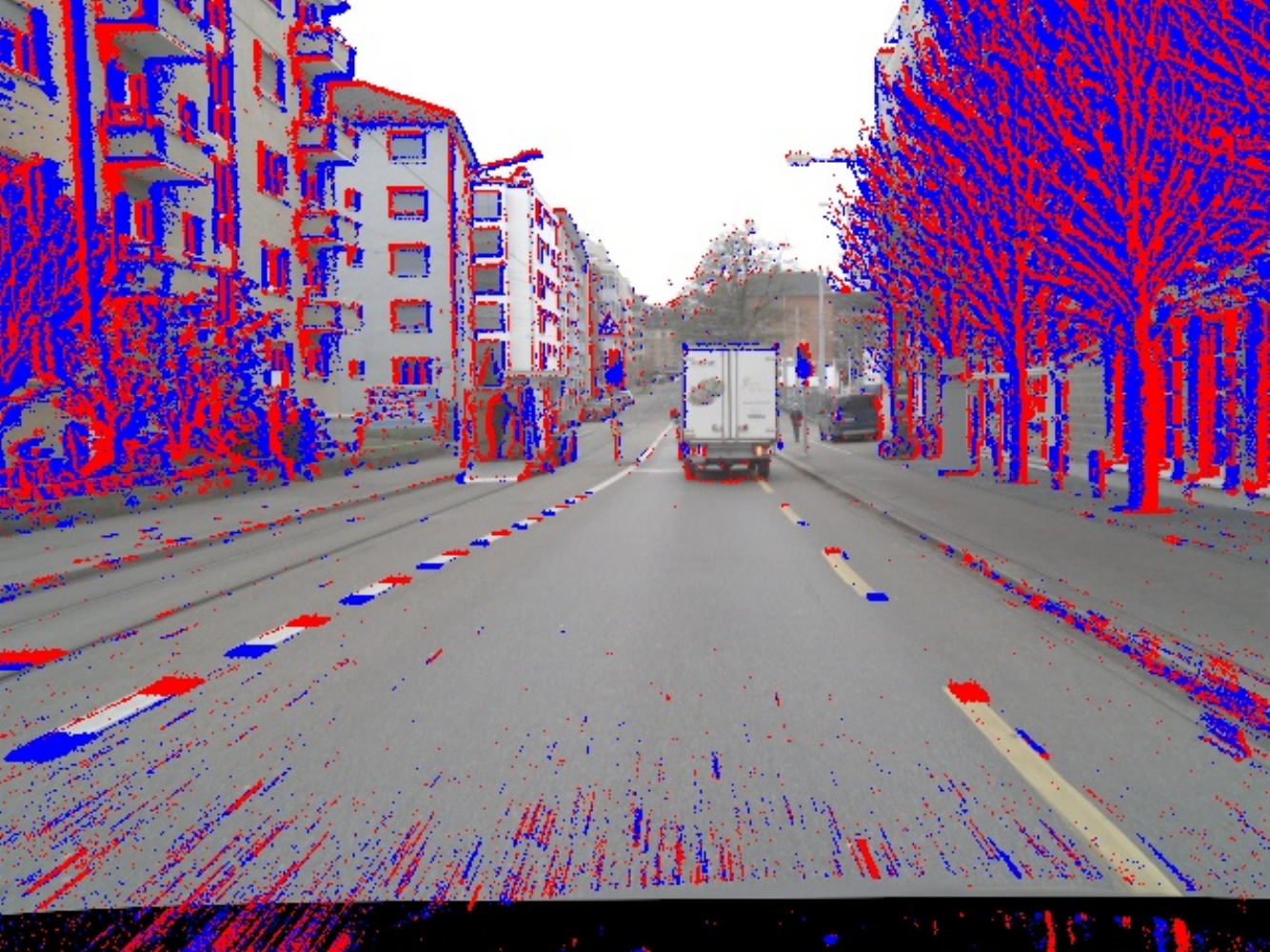}}
& Lens   & normal     & \cbad{fisheye}     & \cgood{normal} \\
& Sensor & event      & \cbad{rgb}         & \cgood{event} \\
& Domain & real       & \cgood{real}       & \cgood{real} \\
& View   & front      & \cgood{front}      & \cgood{front} \\
\midrule

\multirow{4}{*}{\includegraphics[width=0.8cm,height=0.8cm]{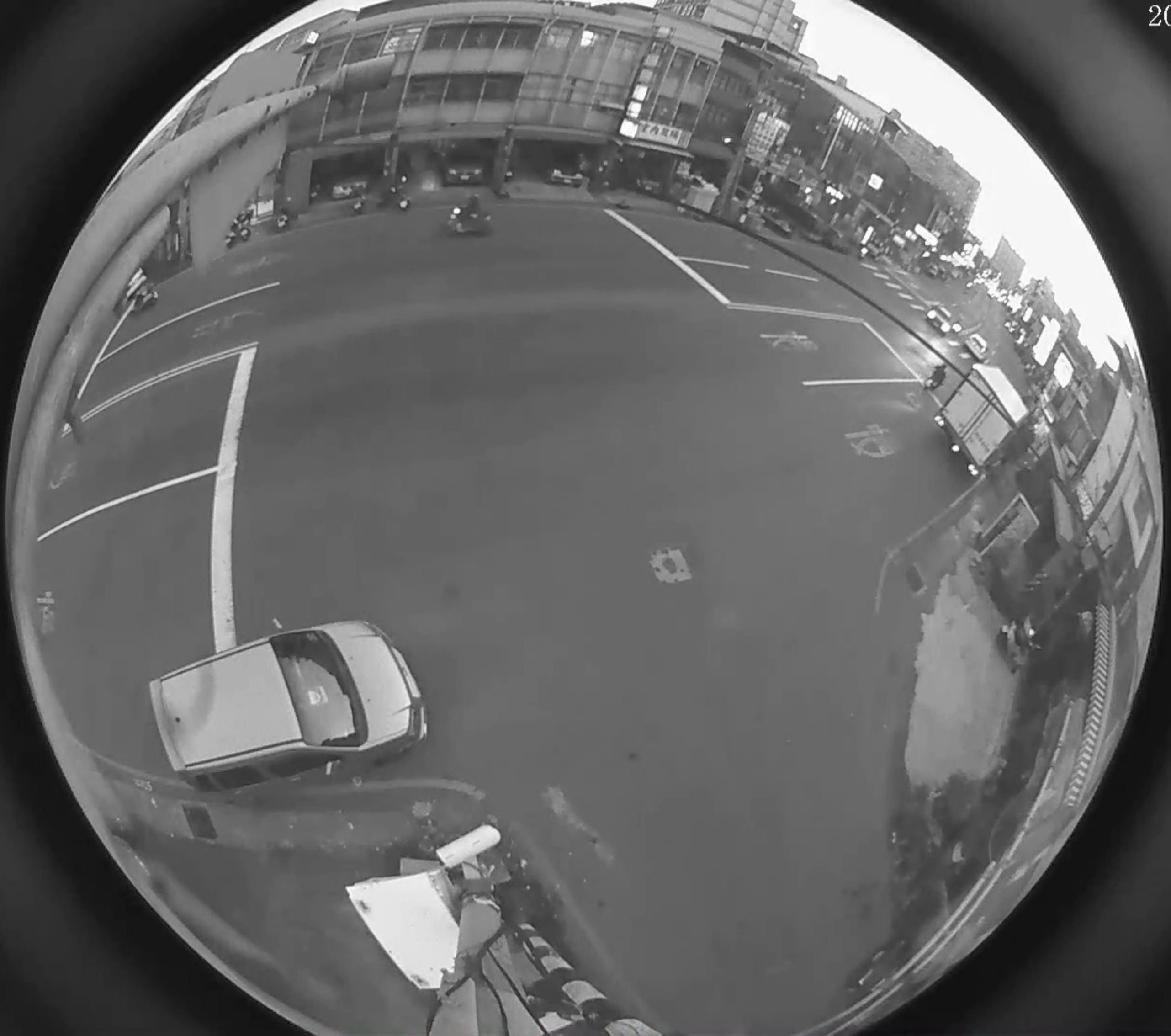}}
& Lens   & fisheye    & \cgood{fisheye}    & \cgood{fisheye} \\
& Sensor & rgb        & \cgood{rgb}        & \cgood{rgb} \\
& Domain & real       & \cgood{real}       & \cgood{real} \\
& View   & pole       & \cbad{drone}       & \cgood{pole} \\
\midrule

\multirow{4}{*}{\includegraphics[width=0.8cm,height=0.8cm]{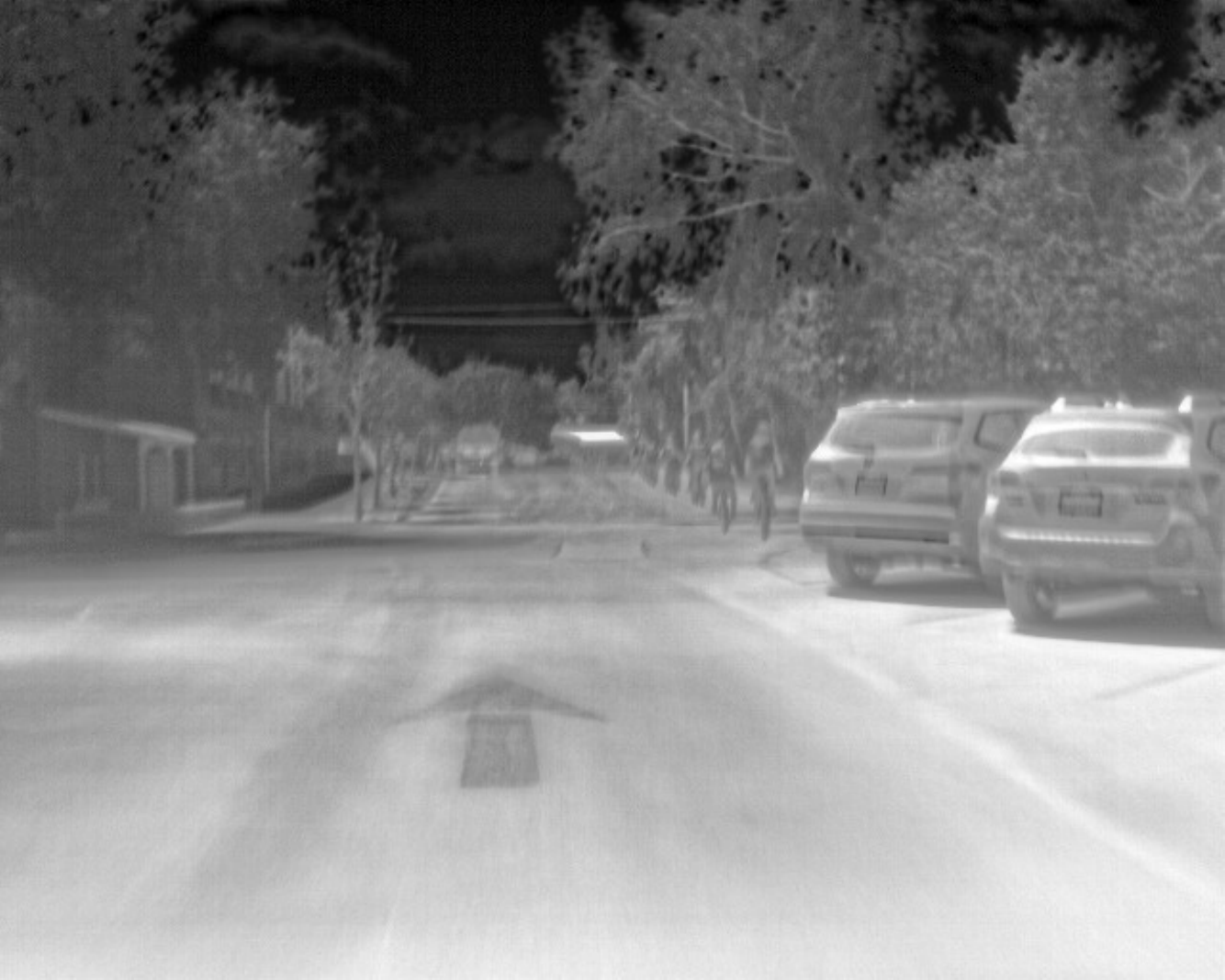}}
& Lens   & normal     & \cgood{normal}     & \cgood{normal} \\
& Sensor & thermal    & \cgood{thermal}    & \cgood{thermal} \\
& Domain & real       & \cgood{real}       & \cgood{real} \\
& View   & front      & \cbad{pole}        & \cgood{front} \\

\bottomrule
\end{tabular}
\end{subtable}
\hfill
\begin{subtable}[t]{0.49\textwidth}
\centering

\begin{tabular}{clccc}
\toprule
\textbf{Image} & \textbf{Factor} & \textbf{GT} & \textbf{FAA~\cite{godavarthy2026multi}} & \textbf{I-FAA} \\
\midrule

\multirow{4}{*}{\includegraphics[width=0.8cm,height=0.8cm]{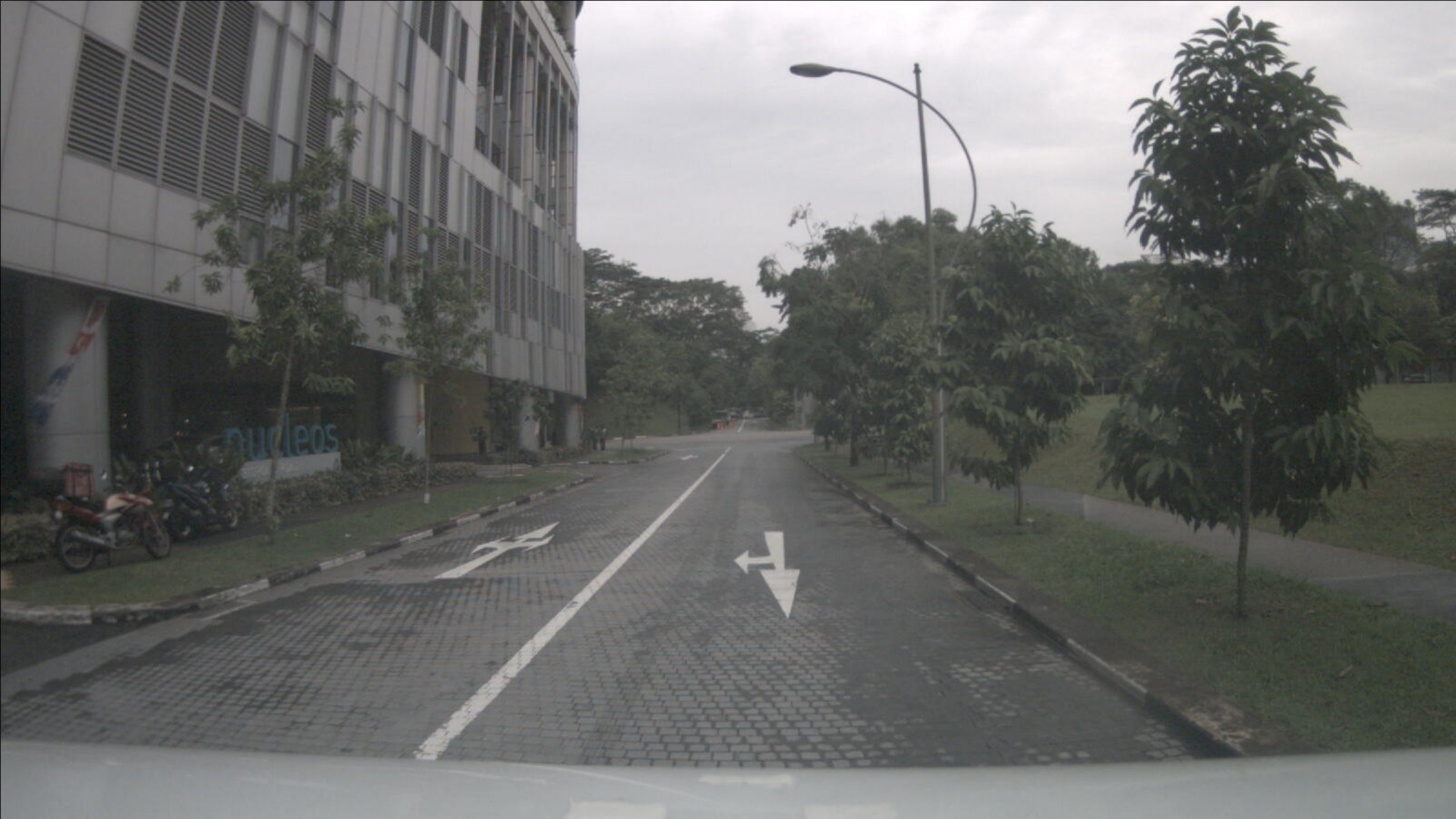}}
& Lens   & normal     & \cgood{normal}     & \cgood{normal} \\
& Sensor & rgb        & \cgood{rgb}        & \cgood{rgb} \\
& Domain & real       & \cgood{real}       & \cgood{real} \\
& View   & back       & \cgood{back}       & \cgood{back} \\
\midrule

\multirow{4}{*}{\includegraphics[width=0.8cm,height=0.8cm]{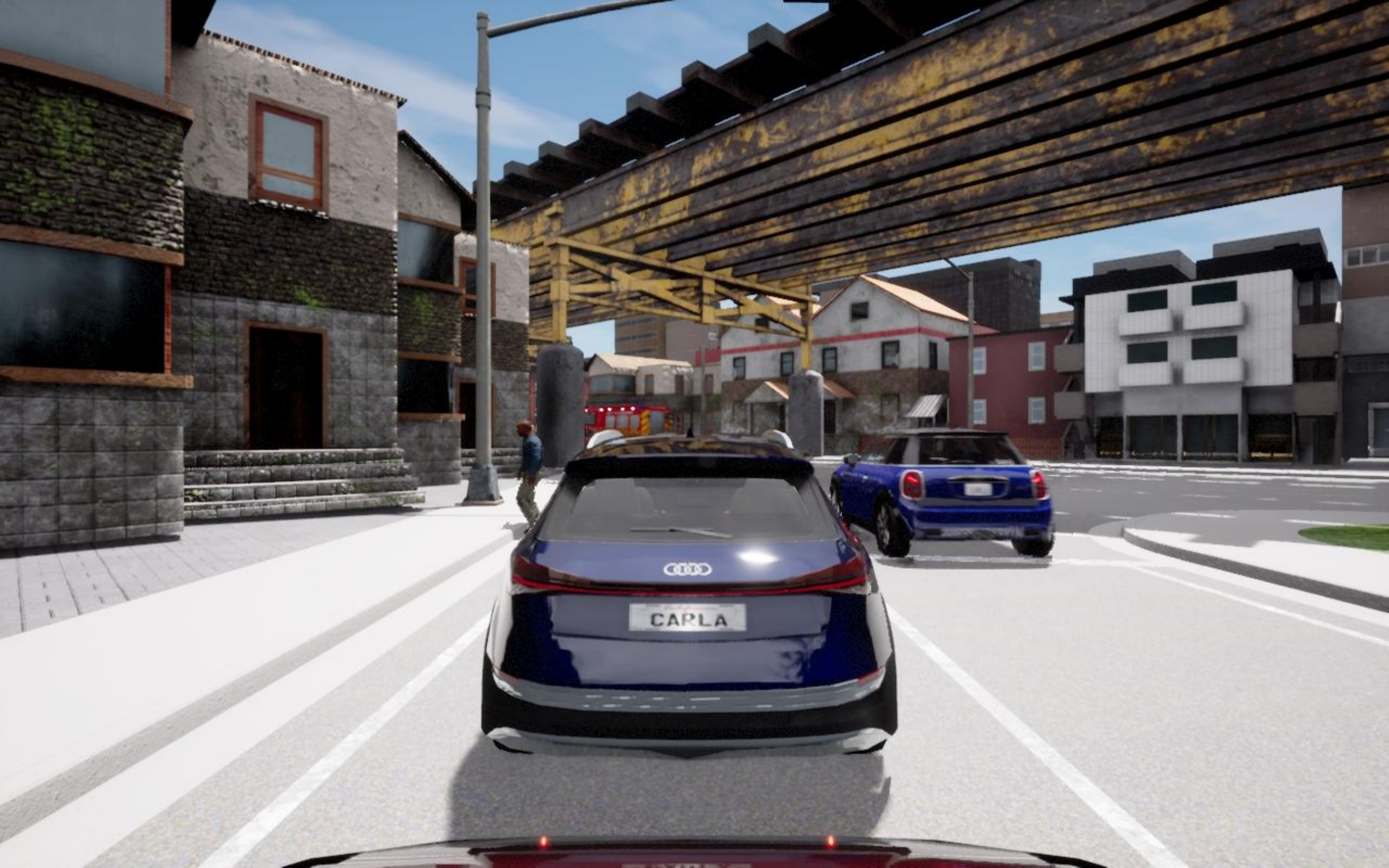}}
& Lens   & normal     & \cgood{normal}     & \cgood{normal} \\
& Sensor & rgb        & \cgood{rgb}        & \cgood{rgb} \\
& Domain & simulation & \cgood{simulation} & \cgood{simulation} \\
& View   & front      & \cbad{back}        & \cgood{front} \\
\midrule

\multirow{4}{*}{\includegraphics[width=0.8cm,height=0.8cm]{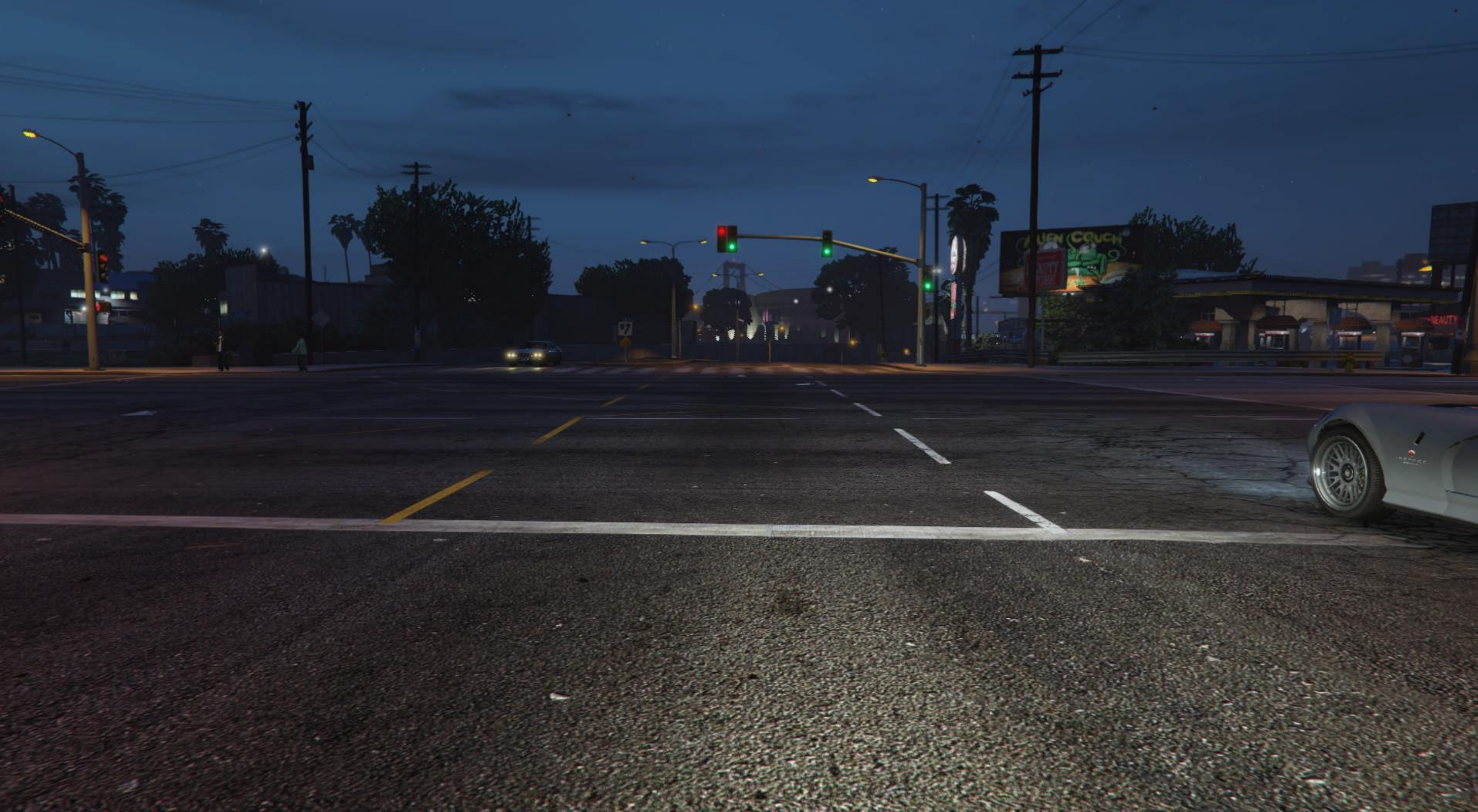}}
& Lens   & normal     & \cgood{normal}     & \cgood{normal} \\
& Sensor & rgb        & \cgood{rgb}        & \cgood{rgb} \\
& Domain & video-game & \cbad{simulation}  & \cgood{video-game} \\
& View   & front      & \cgood{front}      & \cgood{front} \\
\midrule

\multirow{4}{*}{\includegraphics[width=0.8cm,height=0.8cm]{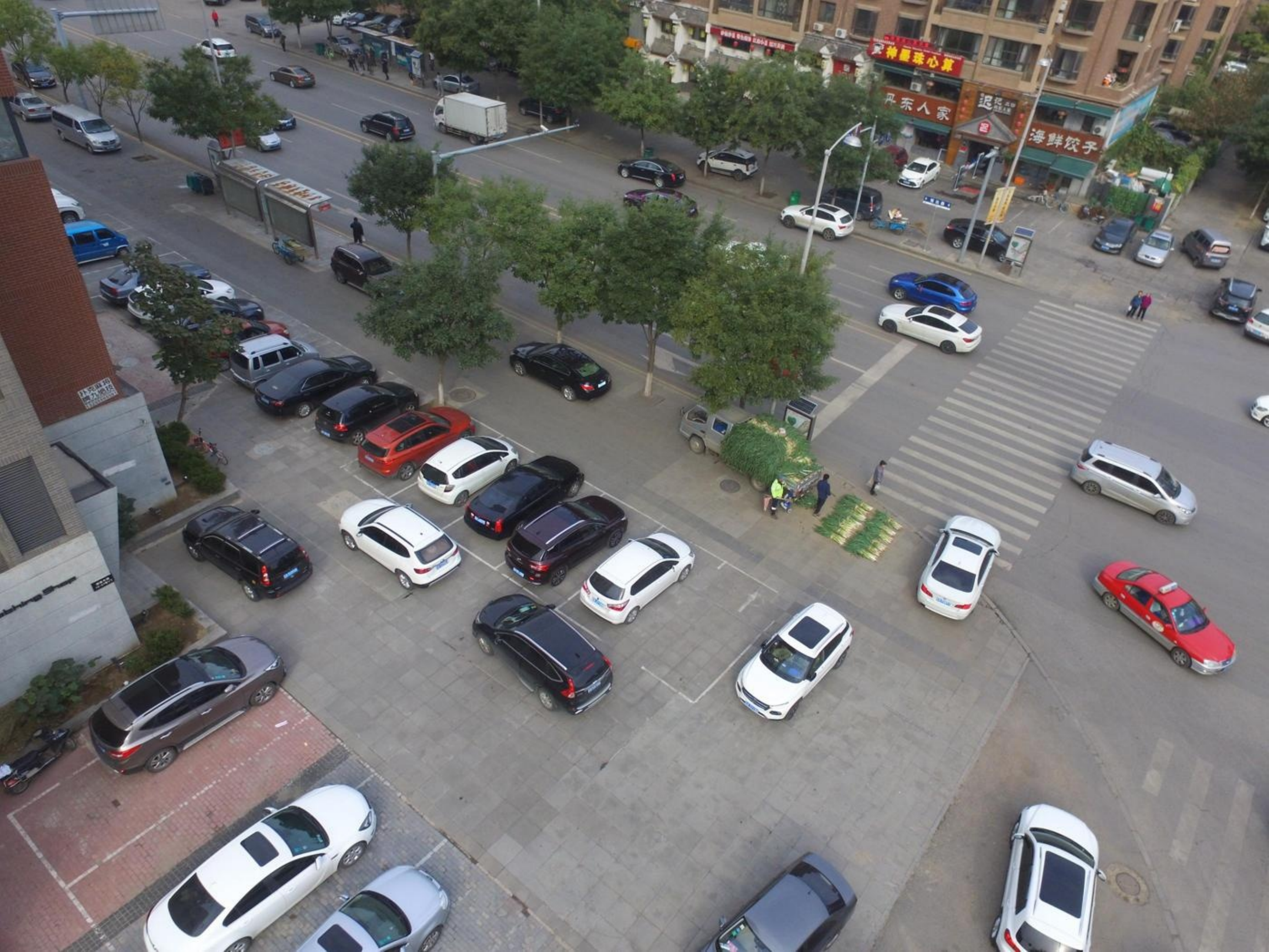}}
& Lens   & normal     & \cgood{normal}     & \cgood{normal} \\
& Sensor & rgb        & \cgood{rgb}        & \cgood{rgb} \\
& Domain & real       & \cbad{simulation}  & \cgood{real} \\
& View   & drone      & \cgood{drone}      & \cgood{drone} \\

\bottomrule
\end{tabular}
\end{subtable}

\end{minipage}
}
\end{table*}








\begin{figure}[t]
\centering
  \begin{subfigure}[b]{0.23\textwidth}
    \centering
    \includegraphics[width=\linewidth]{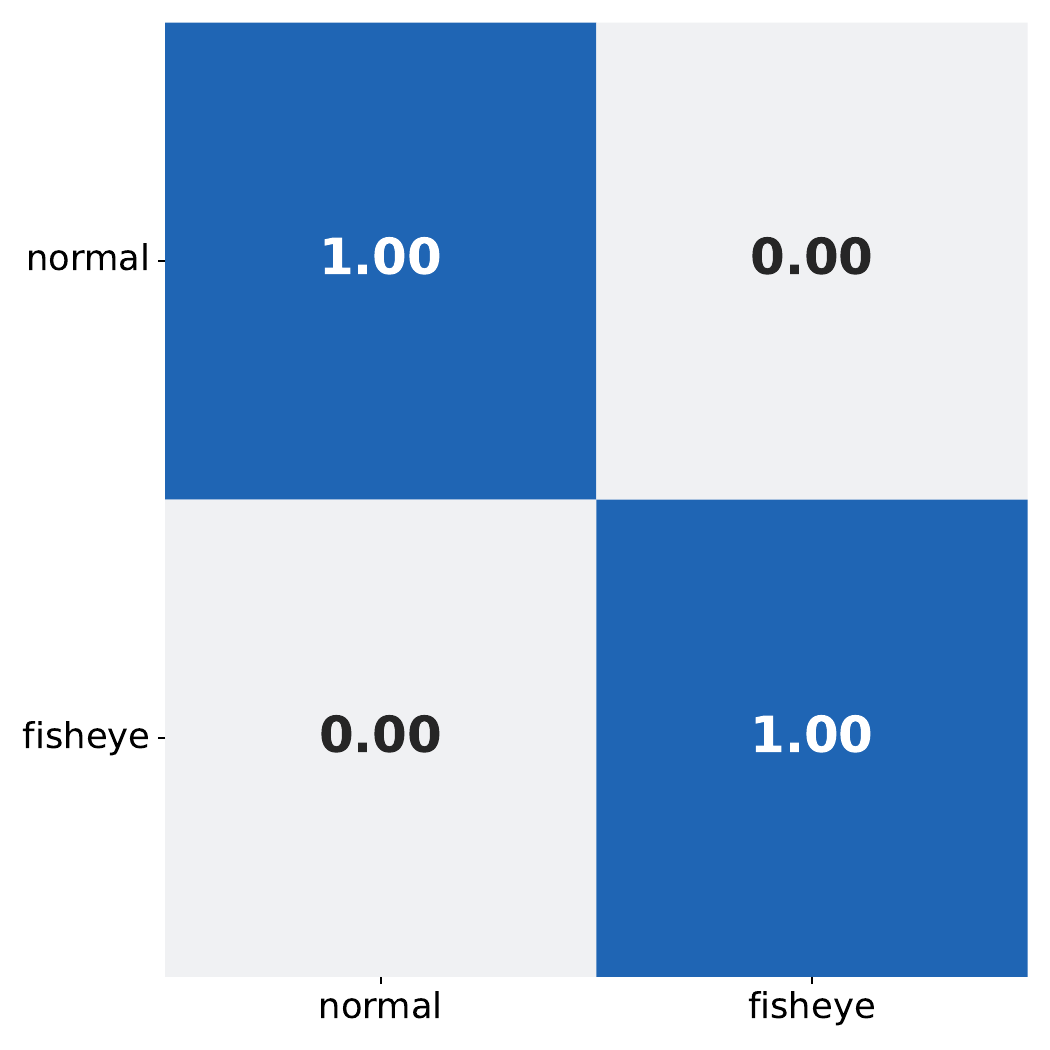}
    \caption{Lens}
  \end{subfigure}
  \hfill
  \begin{subfigure}[b]{0.23\textwidth}
    \centering
    \includegraphics[width=\linewidth]{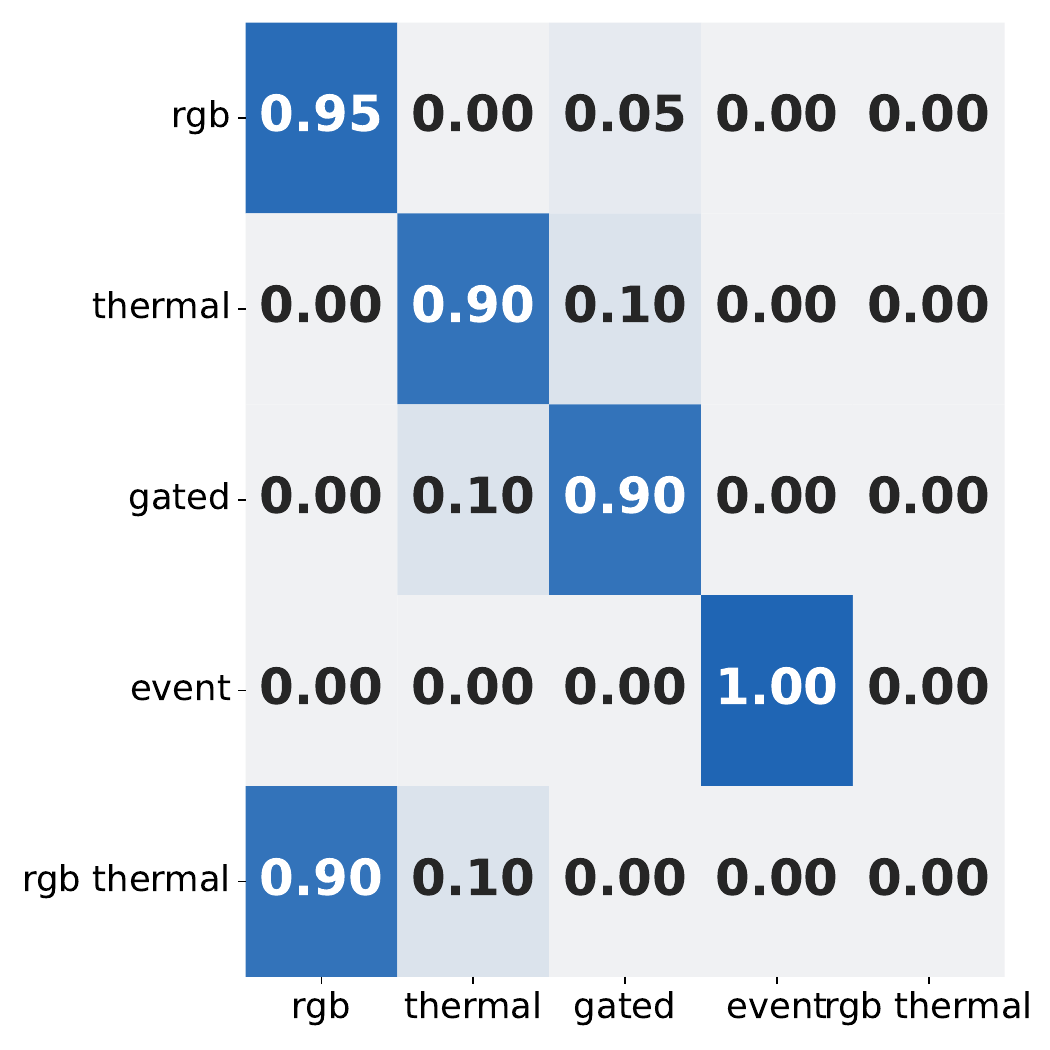}
    \caption{Sensor}
  \end{subfigure}
  \hfill
  \begin{subfigure}[b]{0.23\textwidth}
    \centering
    \includegraphics[width=\linewidth]{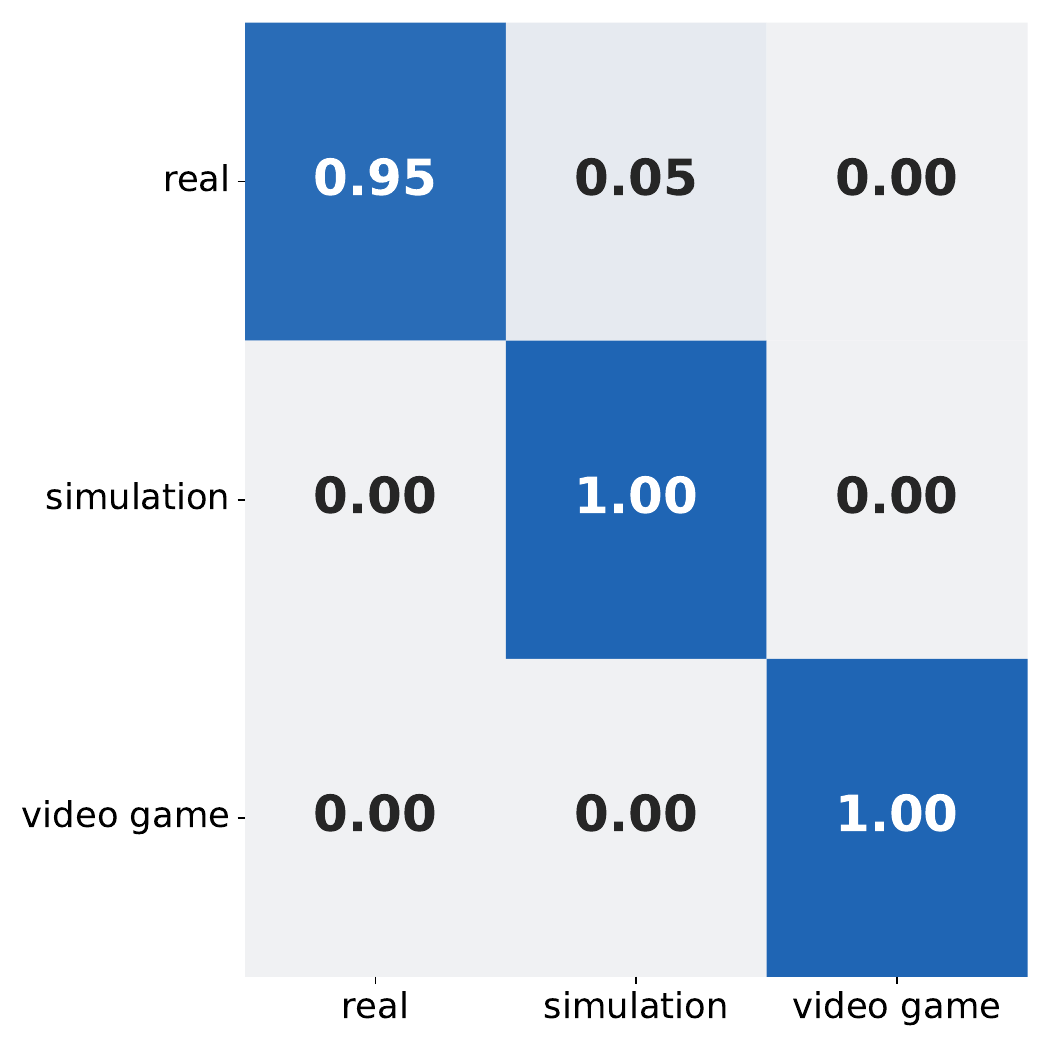}
    \caption{Domain}
  \end{subfigure}
  \hfill
  \begin{subfigure}[b]{0.23\textwidth}
    \centering
    \includegraphics[width=\linewidth]{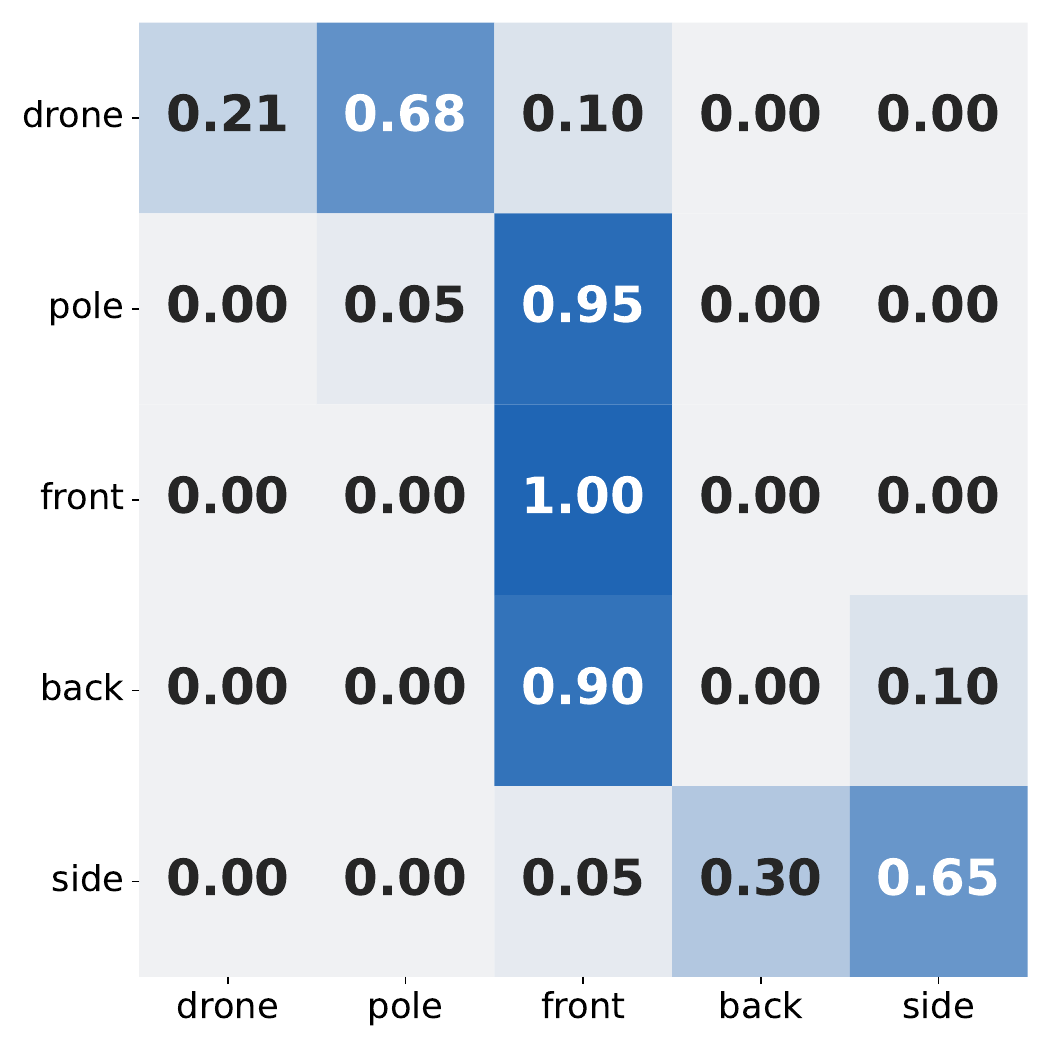}
    \caption{Viewpoint}
  \end{subfigure}
  
  \caption{\textbf{Confusion matrices of the zero-shot VLM for factor classification:} (a) Lens, (b) Sensor, (c) Domain, and (d) Viewpoint. Rows correspond to ground-truth labels and columns to predicted labels.}
  \label{fig:qwen_confusion_matrices}
\end{figure}

\section{Experimental Results}
 
\subsection{Implementation Details}
We utilize the DF-RICO benchmark from MULTI~\cite{godavarthy2026multi}, spanning $15$ autonomous driving and surveillance datasets annotated with four factor categories: (1) camera lens (\textit{normal, fisheye}), (2) source/domain (\textit{real, simulation, video-game}), (3) viewpoint (\textit{front, back, side, drone, pole}), and (4) sensor modality (\textit{rgb, thermal, rgb-thermal, gated, event}).

\noindent \textbf{X-MULTI:} We follow the implementation setup of MULTI~\cite{godavarthy2026multi}. We use Stable Diffusion XL (SDXL)~\cite{podell2023sdxl} as our generative backbone, where each factor is represented by $n = 15$ vectors optimized via AdamW~\cite{Loshchilov2017DecoupledWD} (weight decay: $10^{-2}$) using a linear warmup and cosine annealing learning rate schedule (maximum: $10^{-4}$). Training is conducted for $10$ epochs with a total batch size of $4$.

For the VLM-based supervision branch, we employ Qwen2-VL-7B-Instruct~\cite{Qwen2VL} as a zero-shot VLM classifier starting from epoch $3$ with supervision strength of $10^{-6}$. Each mini-batch consists of $3$ real samples and $1$ synthetic sample. We formulate structured factor-specific prompts as shown in Table~\ref{tab:factor_prompts} that independently query the VLM for each imaging factor: sensor, lens, domain, and viewpoint. To prevent incorrect supervision of unreliable VLM predictions which is discussed in Section~\ref{sssec:vlm_analysis}, rgb-thermal sensor and non-front viewpoints are masked to 0.

\noindent \textbf{I-FAA:} We employ DINOv3~\cite{simeoni2025dinov3} as the vision backbone for all factor classifiers. We train the linear prediction heads using the AdamW optimizer~\cite{Loshchilov2017DecoupledWD} with a learning rate of $10^{-4}$ and a cosine annealing learning rate schedule~\cite{liu2022super}, and a batch size of $32$. Table~\ref{tab:classifier_augmentations} shows data augmentations carried out during training of each factor classifier. 

\noindent \textbf{Evaluation Metrics:}
We employ standard image generation metrics: Fréchet Inception Distance (FID)~\cite{heusel2017gans} to measure distributional similarity between real and generated images, CLIP Score~\cite{hessel2021clipscore} to assess semantic alignment with factor-based prompts (evaluated on three components: factors only, content description only, and full prompt), Inception Score (IS)~\cite{salimans2016improved} for image quality, and Diversity Score (DS)~\cite{zhang2018unreasonable} for perceptual diversity. To measure factor disentanglement, we employ our introduced metric, I-FAA, and analyze its reliability in Sec.~\ref{sec:metric_analysis}.

\noindent \textbf{Baselines:} We compare against: (1) SDXL Zeroshot~\cite{podell2023sdxl}, the pre-trained model without adaptation; (2) DreamBooth~\cite{ruiz2023dreambooth}, fine-tuning-based personalization; (3) Inspiration Tree~\cite{vinker2023concept}, hierarchical disentanglement of stylistic components (evaluated only on existing combinations, as its learned embeddings do not correspond to independent factor categories that can be recombined into novel combinations); and (4) MULTI~\cite{godavarthy2026multi}, a prior factor disentanglement method.

\subsection{Imaging Factor Disentanglement Results}
The primary goal is to evaluate the quality of the synthesis of images with unseen novel factor combinations. We show both qualitative and quantitative results along with the analysis of the choice of VLM and prompt designs for VLM-based supervision. 

\subsubsection{Novel Factor Combinations:}
Table~\ref{tab:qual_novel} shows qualitative inspection of images generated with novel factor combinations. SDXL Zeroshot fails to reflect target factors, while MULTI, DreamBooth, and X-MULTI produce more plausible combinations. X-MULTI shows clearer factor adherence particularly for viewpoint, lens, and domain than other baselines. X-MULTI shows less proficiency with thermal sensor but handled other sensors competently.

Table~\ref{tab:qual_novel_controlnet} shows the usage of ControlNets~\cite{zhang2023adding} for guiding the image generation process, and altering a factor (to respect control maps we use sensor or domain factors only). In this setting, our approach demonstrates good precision compared to MULTI and DreamBooth, which outperforms SDXL Zeroshot. Specifically, when substituting the event sensor for a gated one, DreamBooth retained event artifacts while our method executed the transformation cleanly. Similarly, shifting domains from real to video-game, our method completely eliminated real-world traces while MULTI and DreamBooth could not achieve this.

Table~\ref{tab:novel_quantitative} reports quantitative metrics for novel factor combinations with and without using ControlNets~\cite{zhang2023adding}. Under I-FAA, X-MULTI achieves the best overall factor disentanglement, CLIP alignment with factors, and DS while other metrics remain mixed across methods.

\subsubsection{Effect of VLM-based Supervision:}
We evaluate the contribution of VLM-based supervision by comparing MULTI (without VLM-based supervision) and X-MULTI (with supervision) on novel factor combinations. MULTI achieves an average I-FAA of 0.47, while X-MULTI achieves 0.53, representing a 0.06 (11\% relative) improvement. X-MULTI shows a best improvements in lens and viewpoint factors with 21\%, 18\% relative improvement respectively, while both sensor and domain show a relative improvement of 0.01. This demonstrates that supervision enhances factor disentanglement, especially for lens, and viewpoint.

\subsubsection{VLM Analysis:}\label{sssec:vlm_analysis}
Since VLM predictions provide the supervision in X-MULTI, we evaluate their reliability in a zero-shot setting using the prompts in Table~\ref{tab:factor_prompts}.
Figure~\ref{fig:qwen_confusion_matrices} presents the confusion matrices for Lens, Sensor, Domain, and Viewpoint classification. Across all factors, the VLM achieves an average classification accuracy of $\sim0.78$, demonstrating that the model captures meaningful semantic representations of imaging factors.
However, the analysis also reveals that certain factor categories are less reliably recognized. In particular, the rgb-thermal sensor exhibits strong confusion with rgb images, and several viewpoint categories other than front show inconsistent predictions. Since these inaccurate classification signals can negatively affect disentanglement learning in X-MULTI, we disable the VLM supervision loss for these categories during training. Specifically, the supervision strength is set to zero for the rgb-thermal sensor and for all viewpoints except front.

\subsubsection{Ablation Studies:} We carry out different ablation studies to evaluate the design of core components of X-MULTI. 
\begin{itemize} 
\item \textbf{(1) VLM Selection and Prompt Design:} Since factor disentanglement depends on VLM feedback quality, we evaluate the zero-shot classification performance of VLM on ground-truth data.
First, comparing VLM backbones shows that Qwen2-VL-7B-Instruct~\cite{Qwen2VL} gives 0.78 average accuracy and 0.51 for LLaVA-1.6~\cite{liu2024improved}, indicating Qwen provides much stronger visual grounding. 
Second, comparing prompt designs using Qwen2-VL shows that detailed factor-specific prompts with per-class descriptions achieve 0.78 accuracy, simple prompts without descriptions achieve 0.70, and a single unified prompt classifying all factors jointly drops to 0.63, demonstrating that detailed, isolated prompts provide the best signal for X-MULTI.

\item \textbf{(2) Supervision strength:} We vary the supervision strength ($\lambda$) and evaluate I-FAA on novel factor combinations. Moderate ($10^{-6}$) and weak ($10^{-8}$) supervision perform best, both reaching an I-FAA of 0.53. In contrast, overly strong supervision ($10^{-4}$) degrades performance to 0.47, indicating that excessive supervision over-constrains factor adaptation.
\end{itemize}

\subsection{Analysis of Metrics}\label{sec:metric_analysis}
To validate the reliability of FAA and I-FAA, we test whether their underlying classifiers isolate imaging factors independently rather than exploiting dataset-specific shortcuts. Specifically, we analyze cross-factor alignment using Grad-CAM attention maps to visually verify semantic localization, Cramer’s V correlation~\cite{cramer1999mathematical} to statistically detect information leakage, and metric scores alongside factor predictions on ground-truth DF-RICO~\cite{godavarthy2026multi} images to establish performance. 

\noindent \textbf{Attention Map Visualization:} Table~\ref{tab:attention} visualizes Grad-CAM~\cite{selvaraju2017grad} attention maps for FAA and I-FAA across imaging factors. It shows that I-FAA learns more localized and semantically meaningful attention regions for each factor category, whereas FAA often exhibits entangled attention patterns. For example, in the fisheye lens samples, I-FAA strongly attends to image boundary curvature and geometric distortions, while FAA focuses on unrelated scene regions. Similarly, for viewpoint prediction, I-FAA concentrates on perspective-specific regions, whereas FAA frequently attends to the image center irrespective of viewpoint semantics. For sensor-related factors, I-FAA captures modality-specific characteristics scattered across regions more distinctly than FAA. Domain attention maps also demonstrate that I-FAA focuses on scene appearance cues by attending on objects, while FAA exhibits more inconsistent activations. 

\noindent \textbf{Cramér's V Correlation Analysis:} Figure~\ref{fig:cramersv} presents the pairwise Cramér's V correlations~\cite{cramer1999mathematical} between predicted imaging factors for FAA and I-FAA. Ideally, a factor-isolated evaluation metric should produce low inter-factor correlations, since each factor prediction should depend primarily on the target factor rather than on co-occurring factors. FAA exhibits strong coupling between multiple factor pairs, particularly lens-viewpoint, domain-viewpoint, and lens-domain. In contrast, I-FAA reduces most of these correlations, indicating lower cross-factor leakage. However, residual correlations remain, especially for factor pairs that are strongly coupled or sparsely represented in DF-RICO. This highlights that augmentation in I-FAA classifier training mitigates shortcut learning but cannot fully compensate for correlations in the underlying data.

\noindent \textbf{Metric and Factor Prediction Comparison:} Since FAA was originally proposed as an evaluation metric without independently validating the accuracy of its underlying classifiers on ground-truth images, we first evaluate both FAA and I-FAA classifiers on DF-RICO. Table~\ref{tab:metrics_evaluation_gt} shows that I-FAA improves the average classification accuracy from 0.43 to 0.95. This improvement stems from the class-balanced training and factor-specific augmentations introduced in I-FAA, which reduce shortcut learning from correlated imaging factors. The largest gains are observed for viewpoint and domain, where FAA frequently confuses factor predictions. Table~\ref{tab:metric_prediction_comparison_gt} further illustrates that I-FAA produces more reliable factor-wise predictions, while FAA is more affected by inter-factor interference.



\section{Discussion}
Our numerical results lead to the following main findings:

\noindent \textbf{VLM-Guided Supervision:} Zero-shot VLM supervision provides additional training signals by predicting factors from synthetically generated images during training. This significantly enhances factor alignment, particularly for geometric attributes. $\blacktriangleright$ \textit{Zero-shot supervision enhances disentanglement.}


\noindent \textbf{VLM Reliability and Prompt Context:} X-MULTI relies on a VLM to provide supervision. Our VLM analysis shows that some factor values are less reliably recognized, so we mask unreliable predictions during training. This indicates that performance depends on both the VLM choice and the design of factor-specific prompts. Providing richer visual context in the prompts may further improve supervision. $\blacktriangleright$ \textit{VLM-based supervision is useful, but its reliability depends on prompt design and VLM factor recognition.}

\noindent \textbf{Reliability of I-FAA:} FAA exhibits cross-factor correlations consistent with shared representations entangling factor predictions, while I-FAA reduces this leakage through factor-specific augmentations. However, residual correlations remain and could be addressed with stronger augmentations or correlation-reduction methods~\cite{lee2023towards,gui2025mitigating,zheng2024learning,yang2023mitigating,chew2024understanding}. $\blacktriangleright$ \textit{I-FAA reduces correlations for imaging factor disentanglement, but they could be reduced further.}

\section{Conclusion}
In this work, we addressed imaging factor disentanglement for controllable image generation of unseen novel combinations of imaging factors. We first identify that MULTI's pixel-level reconstruction focuses only on available training data leading to no factor-level supervision for novel combinations. We therefore proposed \textbf{X-MULTI}, which augments MULTI with zero-shot VLM supervision for unseen factor combinations to enforce semantic factor identity. We also identified a limitation in the FAA metric, i.e., it produces strongly entangled factor predictions, evidenced by high cross-factor correlation, and attention maps. To address this, we introduced \textbf{I-FAA}, an evaluation framework with targeted augmentation strategies that reduces these correlations and provides improved disentanglement assessment. Experiments demonstrated that X-MULTI achieves stronger factor disentanglement than baselines, particularly for novel factor combinations, and I-FAA substantially reduces correlations and establishes a more robust metric for factor disentanglement evaluation. Despite promising results, imaging factor disentanglement remains challenging.

\bibliographystyle{splncs04}
\bibliography{main}

\begin{thebibliography}{10}
\providecommand{\url}[1]{\texttt{#1}}
\providecommand{\urlprefix}{URL }
\providecommand{\doi}[1]{https://doi.org/#1}

\bibitem{avrahami2023break}
Avrahami, O., Aberman, K., Fried, O., Cohen-Or, D., Lischinski, D.: Break-a-scene: Extracting multiple concepts from a single image. In: SIGGRAPH Asia. pp. 1--12 (2023)

\bibitem{bai2025qwen3}
Bai, S., Cai, Y., Chen, R., Chen, K., Chen, X., Cheng, Z., Deng, L., Ding, W., Gao, C., Ge, C., et~al.: Qwen3-vl technical report. arXiv preprint arXiv:2511.21631  (2025)

\bibitem{butt2024colorpeel}
Butt, M.A., Wang, K., Vazquez-Corral, J., van~de Weijer, J.: Colorpeel: Color prompt learning with diffusion models via color and shape disentanglement. In: ECCV. pp. 456--472. Springer (2024)

\bibitem{cai2024vip}
Cai, M., Liu, H., Mustikovela, S.K., Meyer, G.P., Chai, Y., Park, D., Lee, Y.J.: Vip-llava: Making large multimodal models understand arbitrary visual prompts. In: Proceedings of the IEEE/CVF Conference on Computer Vision and Pattern Recognition. pp. 12914--12923 (2024)

\bibitem{chefer2023attend}
Chefer, H., Alaluf, Y., Vinker, Y., Wolf, L., Cohen-Or, D.: Attend-and-excite: Attention-based semantic guidance for text-to-image diffusion models. ACM transactions on Graphics (TOG)  \textbf{42}(4),  1--10 (2023)

\bibitem{chew2024understanding}
Chew, O., Lin, H.T., Chang, K.W., Huang, K.H.: Understanding and mitigating spurious correlations in text classification with neighborhood analysis. In: Findings of the association for computational linguistics (EACL). pp. 1013--1025 (2024)

\bibitem{cho2023dall}
Cho, J., Zala, A., Bansal, M.: Dall-eval: Probing the reasoning skills and social biases of text-to-image generation models. In: Proceedings of the IEEE/CVF international conference on computer vision. pp. 3043--3054 (2023)

\bibitem{cramer1999mathematical}
Cram{\'e}r, H.: Mathematical methods of statistics, vol.~9. Princeton university press (1999)

\bibitem{esser2024scaling}
Esser, P., Kulal, S., Blattmann, A., Entezari, R., M{\"u}ller, J., Saini, H., Levi, Y., Lorenz, D., Sauer, A., Boesel, F., et~al.: Scaling rectified flow transformers for high-resolution image synthesis. In: ICML. pp. 12606--12633. PMLR (2024)

\bibitem{frenkel2024implicit}
Frenkel, Y., Vinker, Y., Shamir, A., Cohen-Or, D.: Implicit style-content separation using b-lora. In: ECCV. pp. 181--198. Springer (2024)

\bibitem{gal2022image}
Gal, R., Alaluf, Y., Atzmon, Y., Patashnik, O., Bermano, A.H., Chechik, G., Cohen-or, D.: An image is worth one word: Personalizing text-to-image generation using textual inversion. In: ICLR (2022)

\bibitem{garibi2025tokenverse}
Garibi, D., Yadin, S., Paiss, R., Tov, O., Zada, S., Ephrat, A., Michaeli, T., Mosseri, I., Dekel, T.: Tokenverse: Versatile multi-concept personalization in token modulation space. ACM Transactions On Graphics (TOG)  \textbf{44}(4),  1--11 (2025)

\bibitem{godavarthy2026multi}
Godavarthy, S., Neuwirth-Trapp, M., Faasch, T.F., Bieshaar, M., Moeller, M., Paudel, D.P.: Multi: Disentangling camera lens, sensor, view, and domain for novel image generation. arXiv preprint arXiv:2605.12134  (2026)

\bibitem{gui2025mitigating}
Gui, S., Ji, S.: Mitigating spurious correlations in llms via causality-aware post-training. arXiv preprint arXiv:2506.09433  (2025)

\bibitem{hertz2022prompt}
Hertz, A., Mokady, R., Tenenbaum, J., Aberman, K., Pritch, Y., Cohen-or, D.: Prompt-to-prompt image editing with cross-attention control. In: ICLR (2022)

\bibitem{hessel2021clipscore}
Hessel, J., Holtzman, A., Forbes, M., Le~Bras, R., Choi, Y.: Clipscore: A reference-free evaluation metric for image captioning. In: Proceedings of the 2021 conference on empirical methods in natural language processing. pp. 7514--7528 (2021)

\bibitem{heusel2017gans}
Heusel, M., Ramsauer, H., Unterthiner, T., Nessler, B., Hochreiter, S.: Gans trained by a two time-scale update rule converge to a local nash equilibrium. Advances in neural information processing systems  \textbf{30} (2017)

\bibitem{jin2024image}
Jin, C., Tanno, R., Saseendran, A., Diethe, T., Teare, P.A.: An image is worth multiple words: Discovering object level concepts using multi-concept prompt learning. In: International Conference on Machine Learning (ICML). PMLR (2024)

\bibitem{kwon2024concept}
Kwon, G., Jenni, S., Li, D., Lee, J.Y., Ye, J.C., Heilbron, F.C.: Concept weaver: Enabling multi-concept fusion in text-to-image models. In: CVPR. pp. 8880--8889 (2024)

\bibitem{laurenccon2024matters}
Lauren{\c{c}}on, H., Tronchon, L., Cord, M., Sanh, V.: What matters when building vision-language models? Advances in Neural Information Processing Systems  \textbf{37},  87874--87907 (2024)

\bibitem{lee2023towards}
Lee, S., Payani, A., Chau, D.H.P.: Towards mitigating spurious correlations in image classifiers with simple yes-no feedback. In: AI and HCI Workshop at the International Conference on Machine Learning (ICML Workshop) (2023)

\bibitem{li2023layerdiffusion}
Li, P., Huang, Q., Ding, Y., Li, Z.: Layerdiffusion: Layered controlled image editing with diffusion models. In: SIGGRAPH Asia 2023 Technical Communications (SIGGRAPH Asia), pp.~1--4 (2023)

\bibitem{liu2025unziplora}
Liu, C., Shah, V., Cui, A., Lazebnik, S.: Unziplora: Separating content and style from a single image. In: ICCV. pp. 16776--16785 (2025)

\bibitem{liu2024improved}
Liu, H., Li, C., Li, Y., Lee, Y.J.: Improved baselines with visual instruction tuning. In: Proceedings of the IEEE/CVF Conference on Computer Vision and Pattern Recognition (CVPR). pp. 26296--26306 (2024)

\bibitem{liu2022super}
Liu, Z.: Super convergence cosine annealing with warm-up learning rate. In: CAIBDA 2022; 2nd International Conference on Artificial Intelligence, Big Data and Algorithms. pp.~1--7. VDE (2022)

\bibitem{Loshchilov2017DecoupledWD}
Loshchilov, I., Hutter, F.: Decoupled weight decay regularization. In: ICLR (2017)

\bibitem{luo2025dual}
Luo, G., Granskog, J., Holynski, A., Darrell, T.: Dual-process image generation. In: Proceedings of the IEEE/CVF International Conference on Computer Vision. pp. 17972--17983 (2025)

\bibitem{motamed2023lego}
Motamed, S., Paudel, D.P., Van~Gool, L.: Lego: Learning to disentangle and invert personalized concepts beyond object appearance in text-to-image diffusion models. In: European Conference on Computer Vision (ECCV). pp. 116--133. Springer (2024)

\bibitem{nichol2022glide}
Nichol, A.Q., Dhariwal, P., Ramesh, A., Shyam, P., Mishkin, P., Mcgrew, B., Sutskever, I., Chen, M.: Glide: Towards photorealistic image generation and editing with text-guided diffusion models. In: ICML. pp. 16784--16804. PMLR (2022)

\bibitem{podell2023sdxl}
Podell, D., English, Z., Lacey, K., Blattmann, A., Dockhorn, T., M{\"u}ller, J., Penna, J., Rombach, R.: Sdxl: Improving latent diffusion models for high-resolution image synthesis. In: ICLR (2023)

\bibitem{rombach2022high}
Rombach, R., Blattmann, A., Lorenz, D., Esser, P., Ommer, B.: High-resolution image synthesis with latent diffusion models. In: CVPR. pp. 10684--10695 (2022)

\bibitem{ruiz2023dreambooth}
Ruiz, N., Li, Y., Jampani, V., Pritch, Y., Rubinstein, M., Aberman, K.: Dreambooth: Fine tuning text-to-image diffusion models for subject-driven generation. In: CVPR. pp. 22500--22510 (2023)

\bibitem{saharia2022photorealistic}
Saharia, C., Chan, W., Saxena, S., Li, L., Whang, J., Denton, E.L., Ghasemipour, K., Gontijo~Lopes, R., Karagol~Ayan, B., Salimans, T., et~al.: Photorealistic text-to-image diffusion models with deep language understanding. Advances in neural information processing systems  \textbf{35},  36479--36494 (2022)

\bibitem{salimans2016improved}
Salimans, T., Goodfellow, I., Zaremba, W., Cheung, V., Radford, A., Chen, X.: Improved techniques for training gans. Advances in neural information processing systems  \textbf{29} (2016)

\bibitem{selvaraju2017grad}
Selvaraju, R.R., Cogswell, M., Das, A., Vedantam, R., Parikh, D., Batra, D.: Grad-cam: Visual explanations from deep networks via gradient-based localization. In: Proceedings of the IEEE international conference on computer vision. pp. 618--626 (2017)

\bibitem{shentu2024attencraft}
Shentu, J., Watson, M., Al~Moubayed, N.: Attencraft: Attention-guided disentanglement of multiple concepts for text-to-image customization. CoRR  (2024)

\bibitem{simeoni2025dinov3}
Sim{\'e}oni, O., Vo, H.V., Seitzer, M., Baldassarre, F., Oquab, M., Jose, C., Khalidov, V., Szafraniec, M., Yi, S., Ramamonjisoa, M., et~al.: Dinov3. arXiv preprint arXiv:2508.10104  (2025)

\bibitem{sohn2023styledrop}
Sohn, K., Ruiz, N., Lee, K., Chin, D.C., Blok, I., Chang, H., Barber, J., Jiang, L., Entis, G., Li, Y., et~al.: Styledrop: text-to-image generation in any style. In: Proceedings of the 37th International Conference on Neural Information Processing Systems. pp. 66860--66889 (2023)

\bibitem{vinker2023concept}
Vinker, Y., Voynov, A., Cohen-Or, D., Shamir, A.: Concept decomposition for visual exploration and inspiration. ACM Transactions on Graphics (TOG)  \textbf{42}(6),  1--13 (2023)

\bibitem{Qwen2VL}
Wang, P., Bai, S., Tan, S., Wang, S., Fan, Z., Bai, J., Chen, K., Liu, X., Wang, J., Ge, W., Fan, Y., Dang, K., Du, M., Ren, X., Men, R., Liu, D., Zhou, C., Zhou, J., Lin, J.: Qwen2-vl: Enhancing vision-language model's perception of the world at any resolution. arXiv preprint arXiv:2409.12191  (2024)

\bibitem{wen2023improving}
Wen, S., Fang, G., Zhang, R., Gao, P., Dong, H., Metaxas, D.: Improving compositional text-to-image generation with large vision-language models. arXiv preprint arXiv:2310.06311  (2023)

\bibitem{wu2024v}
Wu, P., Xie, S.: V?: Guided visual search as a core mechanism in multimodal llms. In: Proceedings of the IEEE/CVF Conference on Computer Vision and Pattern Recognition. pp. 13084--13094 (2024)

\bibitem{xu2024dreamanime}
Xu, C., Xu, Y., Zhang, H., Xu, X., He, S.: Dreamanime: Learning style-identity textual disentanglement for anime and beyond. IEEE Transactions on Visualization and Computer Graphics  (2024)

\bibitem{yang2023mitigating}
Yang, Y., Nushi, B., Palangi, H., Mirzasoleiman, B.: Mitigating spurious correlations in multi-modal models during fine-tuning. In: International Conference on Machine Learning (ICML). pp. 39365--39379. PMLR (2023)

\bibitem{zhang2023adding}
Zhang, L., Rao, A., Agrawala, M.: Adding conditional control to text-to-image diffusion models. In: ICCV. pp. 3836--3847 (2023)

\bibitem{zhang2018unreasonable}
Zhang, R., Isola, P., Efros, A.A., Shechtman, E., Wang, O.: The unreasonable effectiveness of deep features as a perceptual metric. In: Proceedings of the IEEE conference on computer vision and pattern recognition. pp. 586--595 (2018)

\bibitem{zhang2024attention}
Zhang, Y., Yang, M., Zhou, Q., Wang, Z.: Attention calibration for disentangled text-to-image personalization. In: CVPR. pp. 4764--4774 (2024)

\bibitem{zheng2024learning}
Zheng, G., Ye, W., Zhang, A.: Learning robust classifiers with self-guided spurious correlation mitigation. In: Proceedings of the International Joint Conference on Artificial Intelligence (IJCAI). pp. 5599--5607 (2024)

\bibitem{zhong2025mod}
Zhong, W., Yang, H., Liu, Z., He, H., He, Z., Niu, X., Zhang, D., Li, G.: Mod-adapter: Tuning-free and versatile multi-concept personalization via modulation adapter. arXiv preprint arXiv:2505.18612  (2025)

\end{thebibliography}

\clearpage

\appendix

  \begin{center}
    \vspace{0.5em}
    {\Large\normalfont Supplementary Material\par} 
    \vspace{1.0em}
  \end{center}

\renewcommand{\thesection}{S.\arabic{section}}
\renewcommand{\thefigure}{S.\arabic{figure}}
\renewcommand{\thetable}{S.\arabic{table}}
\setcounter{figure}{0}
\setcounter{table}{0}

\section{Experimental Results on X-MULTI: Existing Factor Combinations}

\FloatBarrier
\begin{table*}[t]
\centering
\caption{\textbf{Qualitative comparison of existing combinations.} X-MULTI (ours) faithfully reproduces all target sensor, lens, and domain modalities present in the training distribution.}
\scriptsize
\setlength{\tabcolsep}{2pt}
\renewcommand{\arraystretch}{1.05}
\resizebox{\textwidth}{!}{
\begin{tabular}{lcccccccc}
\toprule

\textbf{Domain} 
& real & simulation & real & real & real & real & real & real\\
\textbf{Sensor} 
& event & rgb & rgb & thermal & gated & thermal & rgb & rgb \\
\textbf{Viewpoint} 
& normal & normal & fisheye & fisheye & normal & normal & fisheye & fisheye \\
\textbf{Lens} 
& front & front & drone & pole & front & front & side & pole \\

\midrule

SDXL Zeroshot~\cite{podell2023sdxl}
& \includegraphics[width=1.7cm,height=1.7cm]{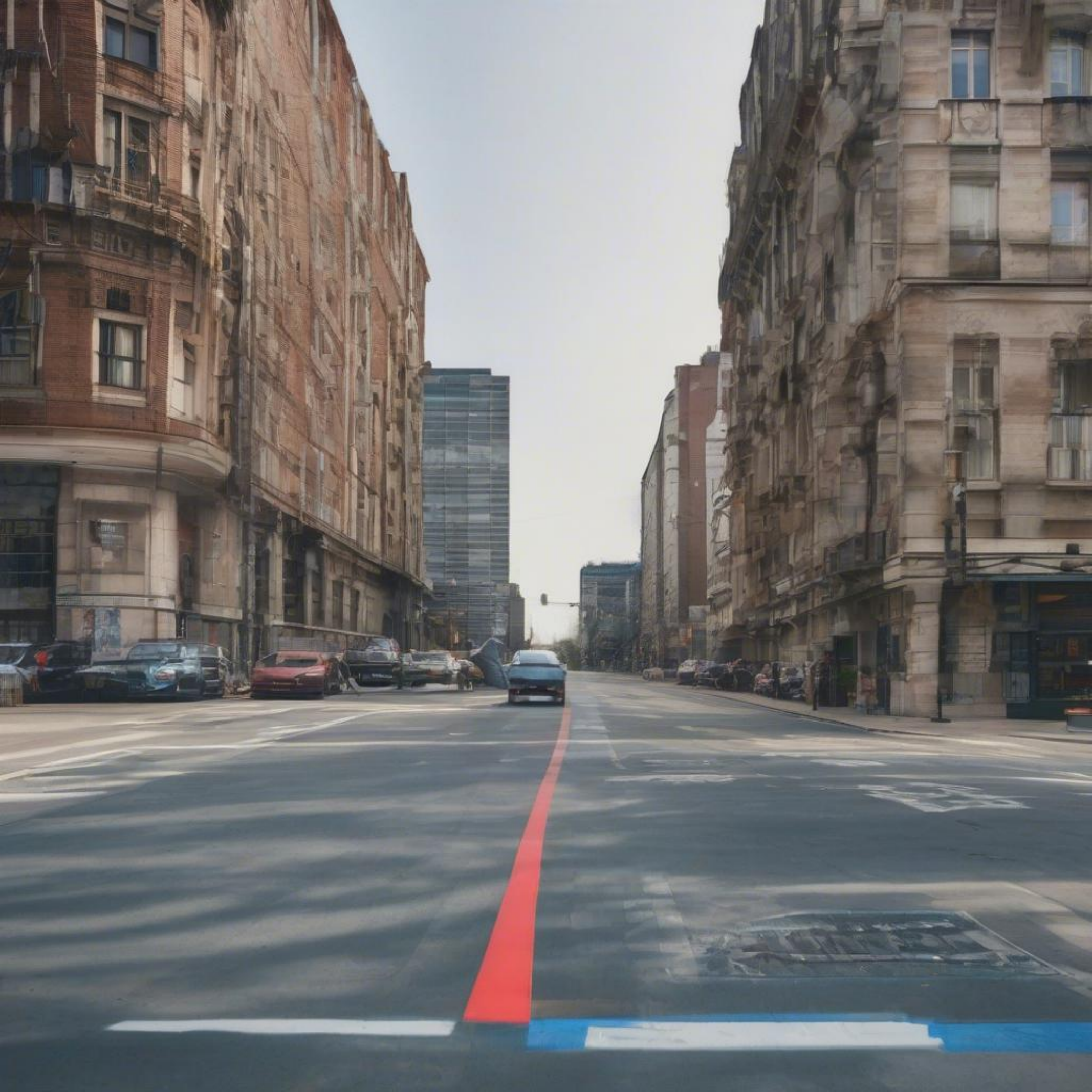}
& \includegraphics[width=1.7cm,height=1.7cm]{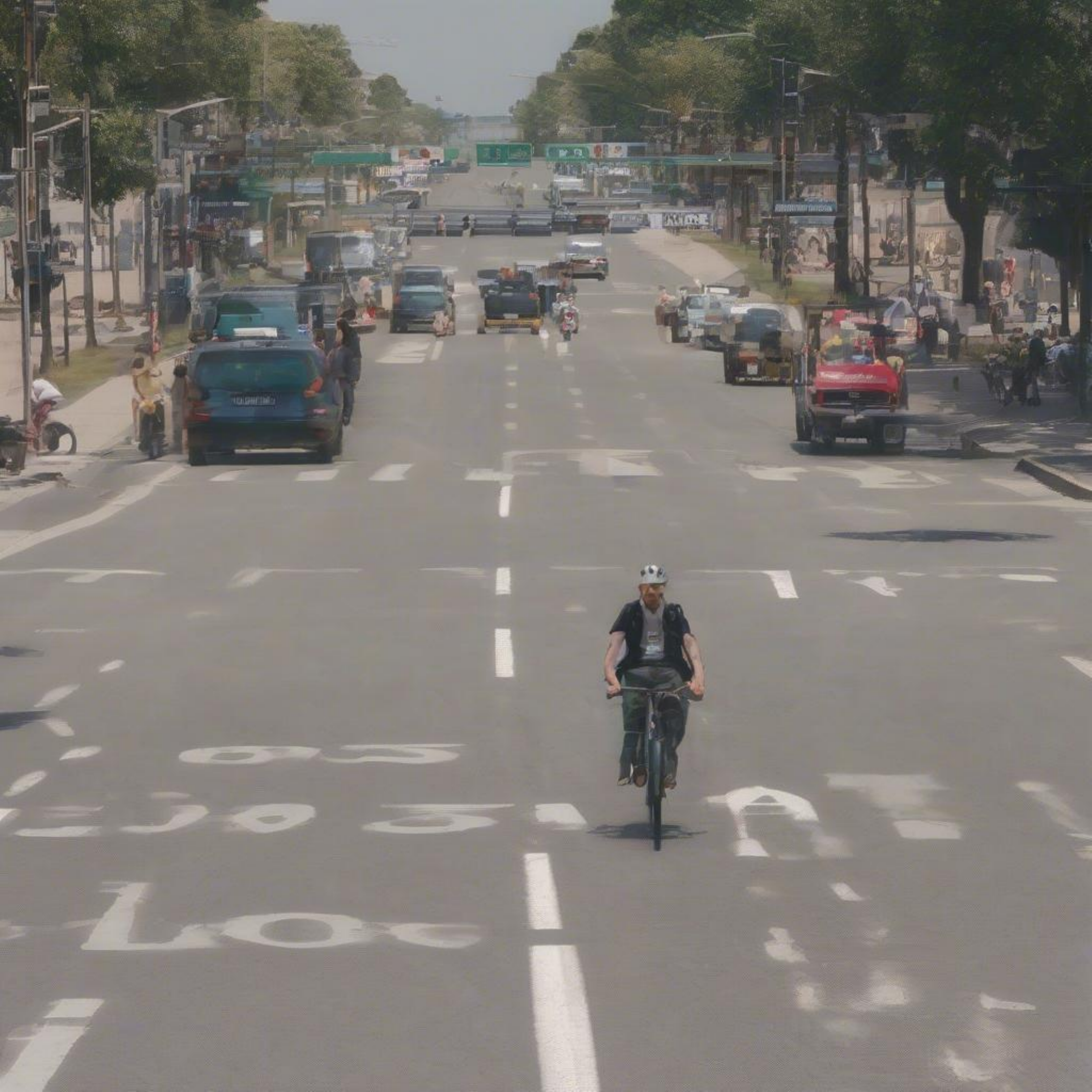}
& \includegraphics[width=1.7cm,height=1.7cm]{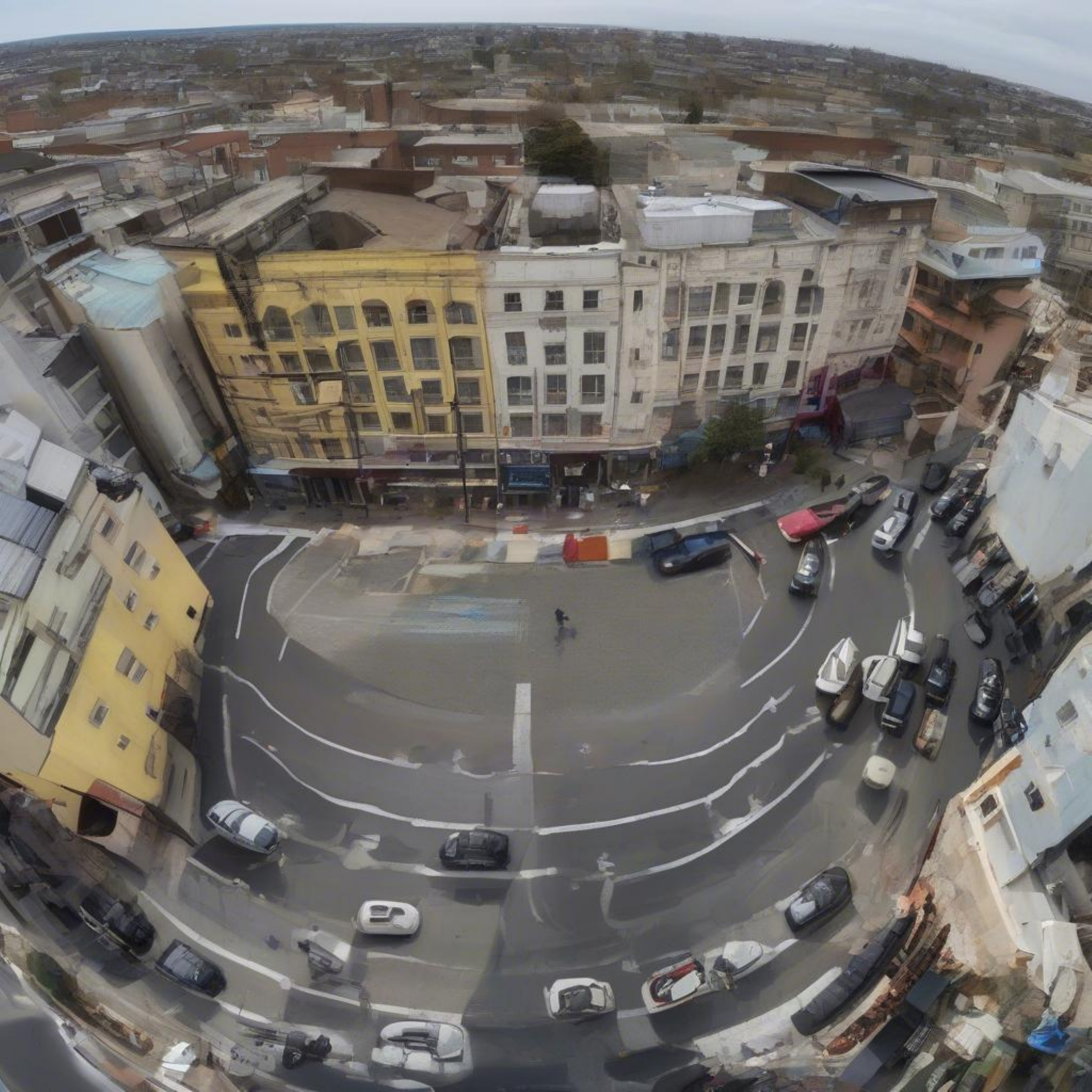}
& \includegraphics[width=1.7cm,height=1.7cm]{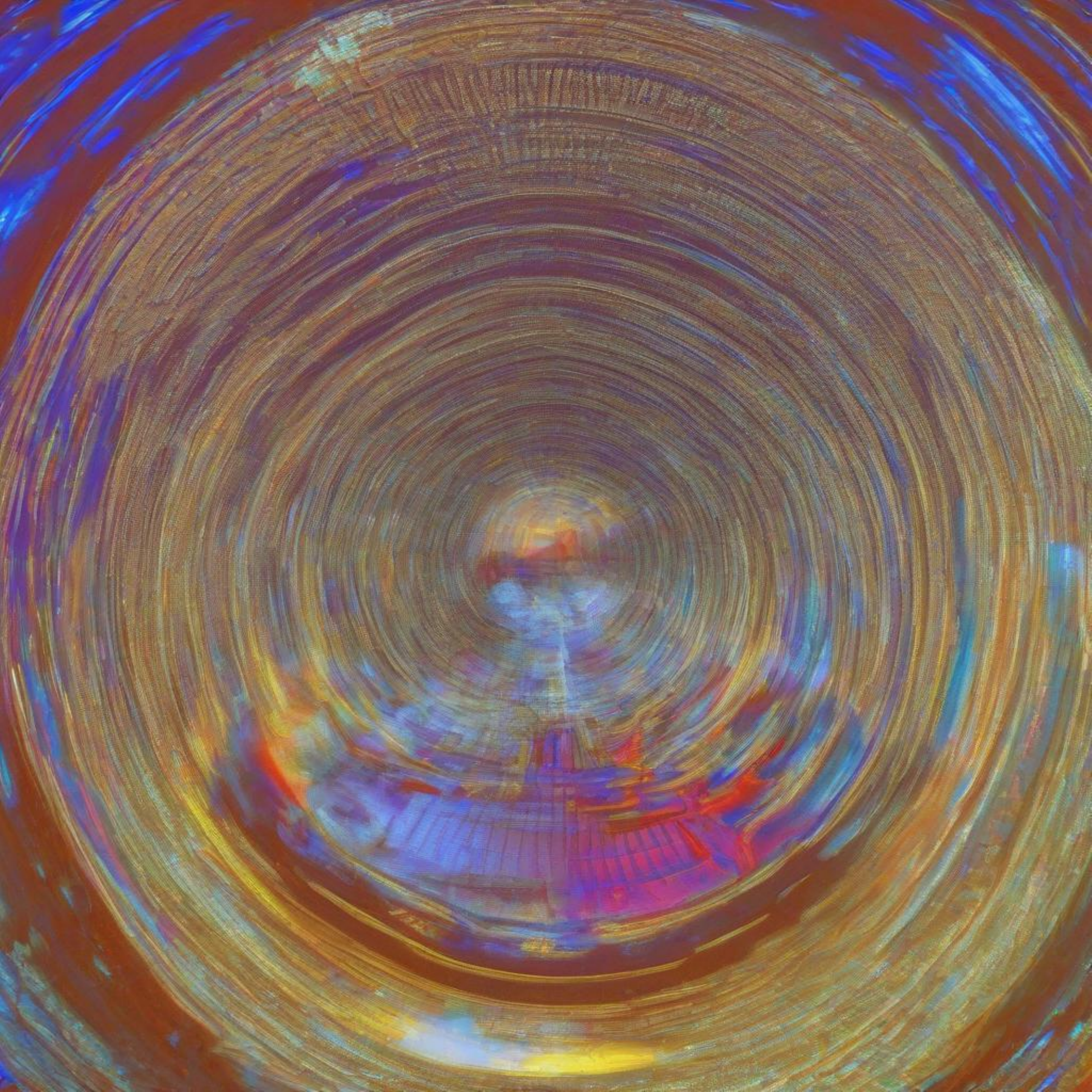}
& \includegraphics[width=1.7cm,height=1.7cm]{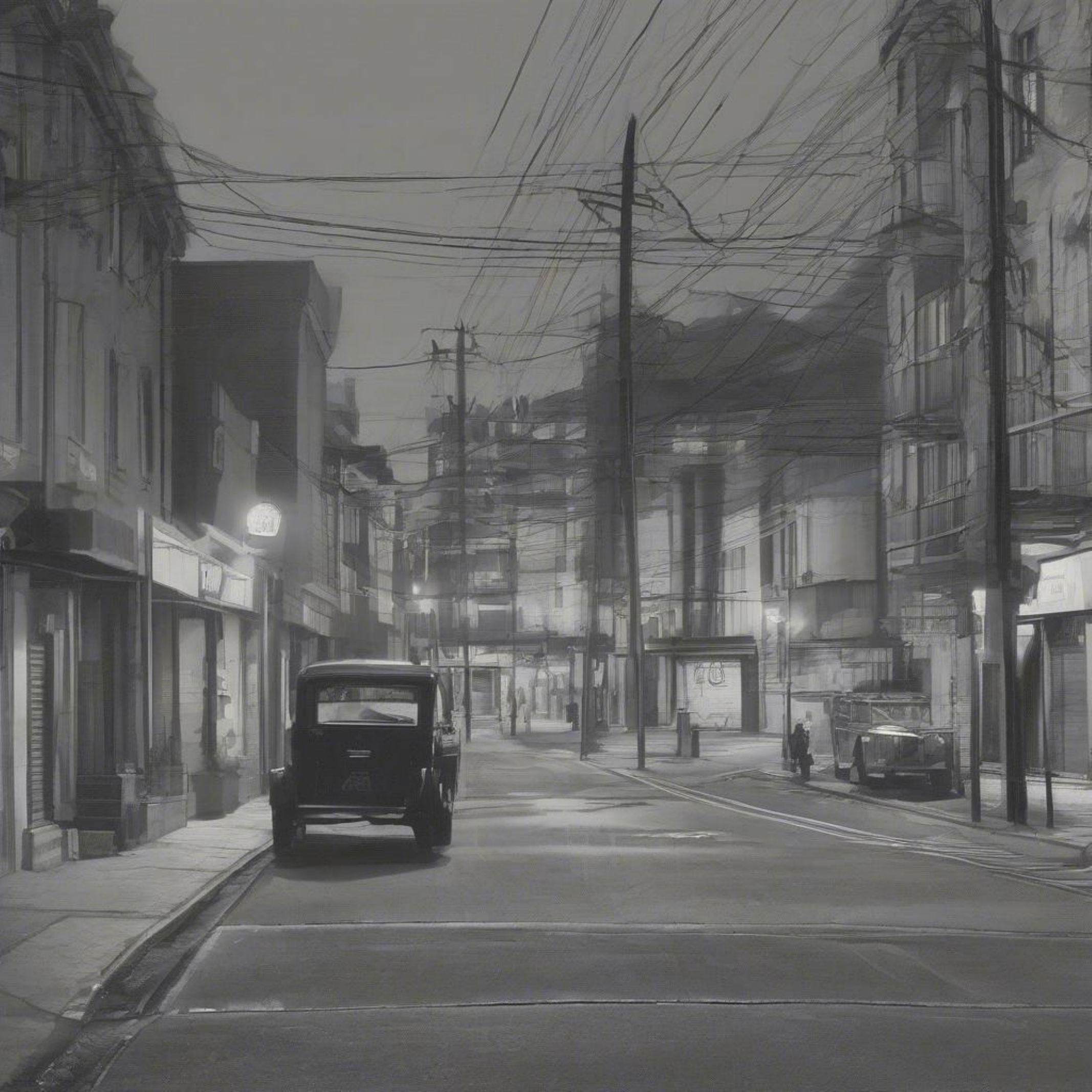}
& \includegraphics[width=1.7cm,height=1.7cm]{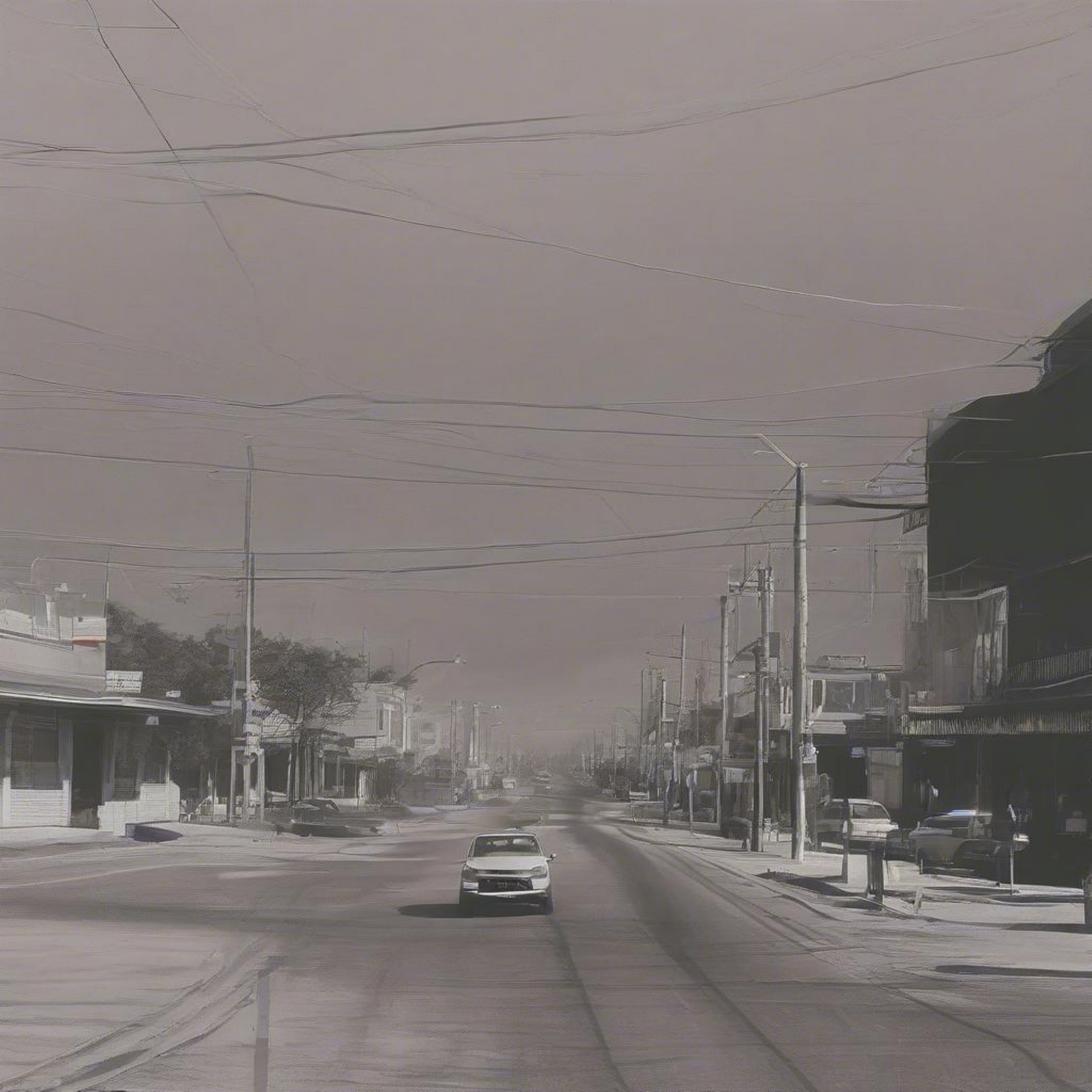}
& \includegraphics[width=1.7cm,height=1.7cm]{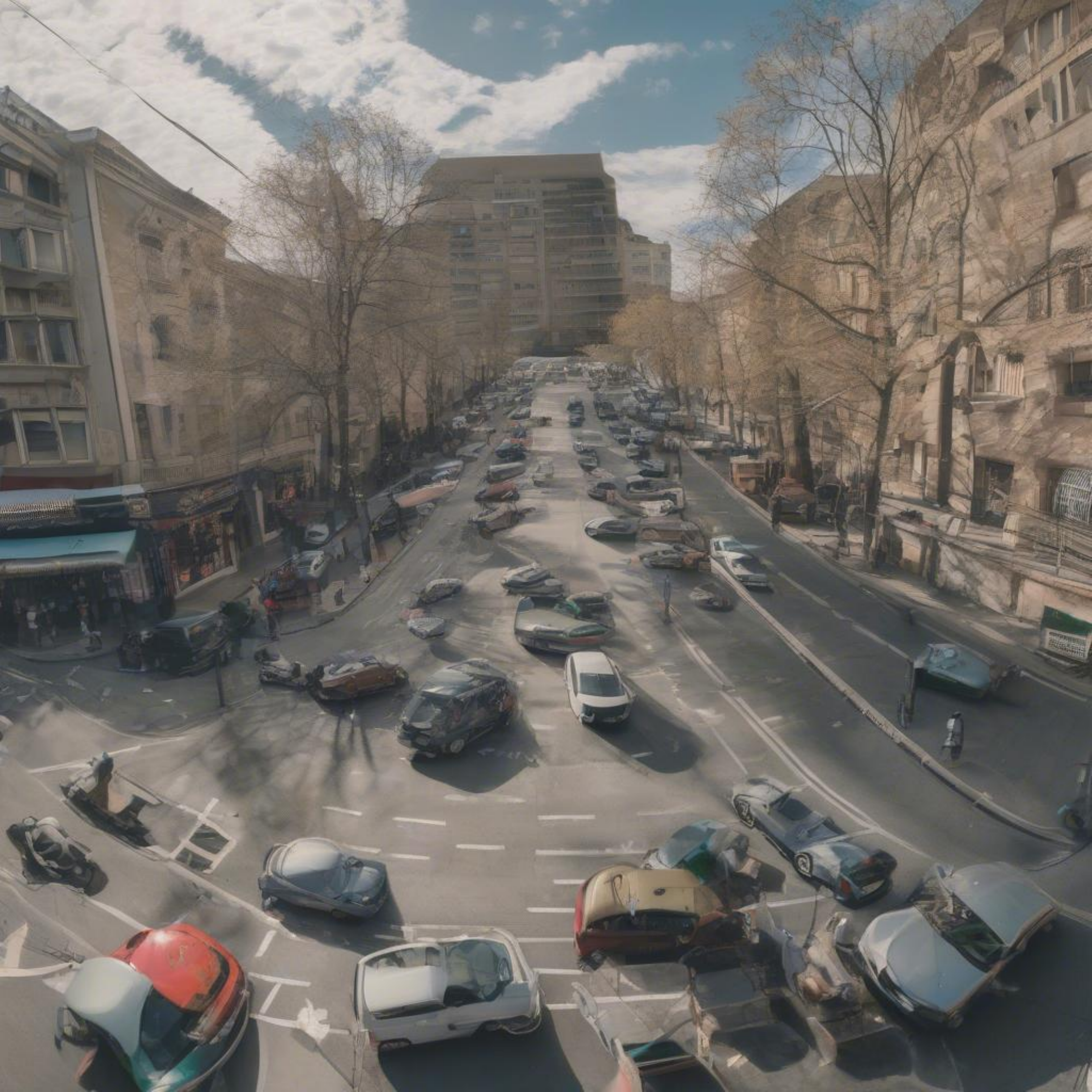}
& \includegraphics[width=1.7cm,height=1.7cm]{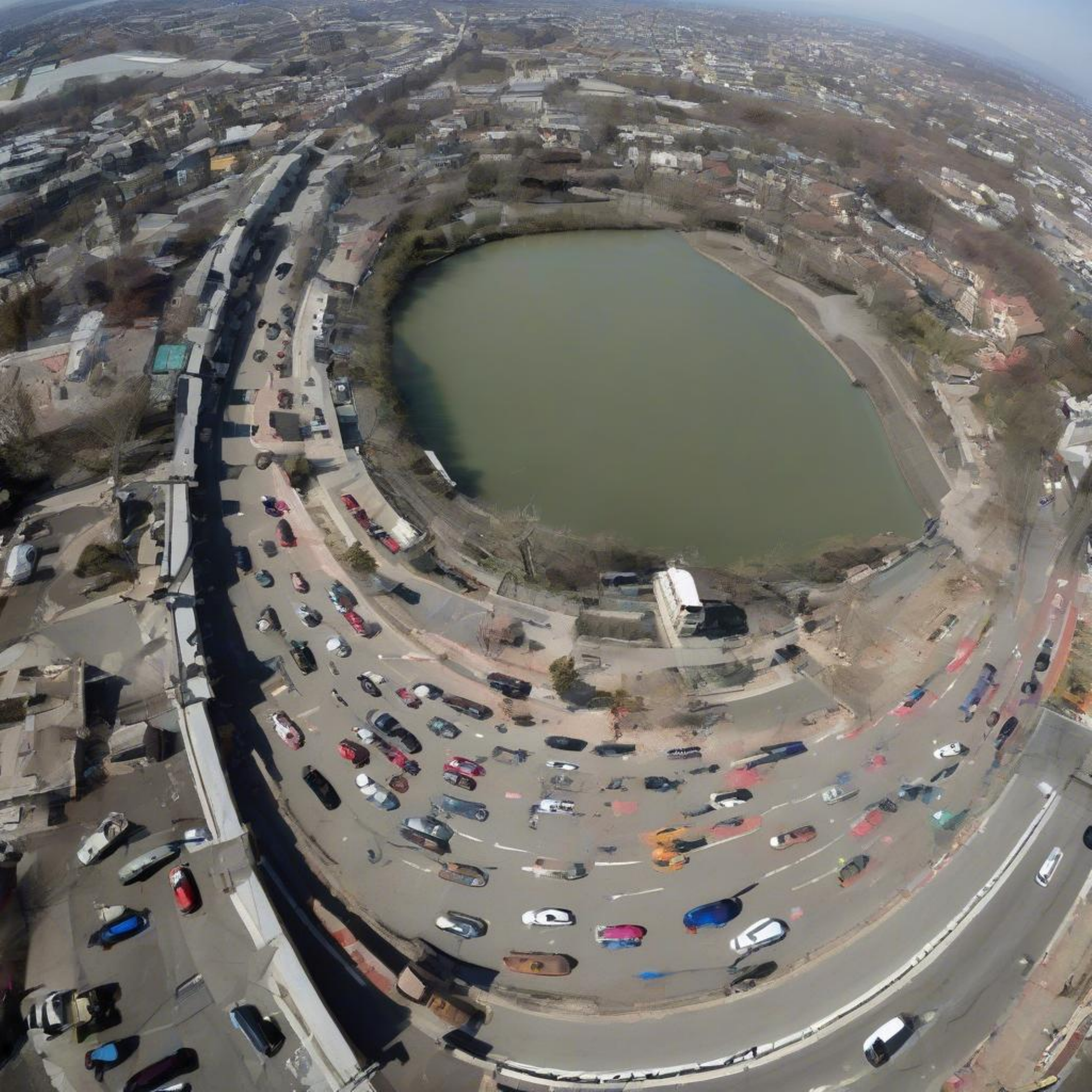} \\

DreamBooth~\cite{ruiz2023dreambooth}
& \includegraphics[width=1.7cm,height=1.7cm]{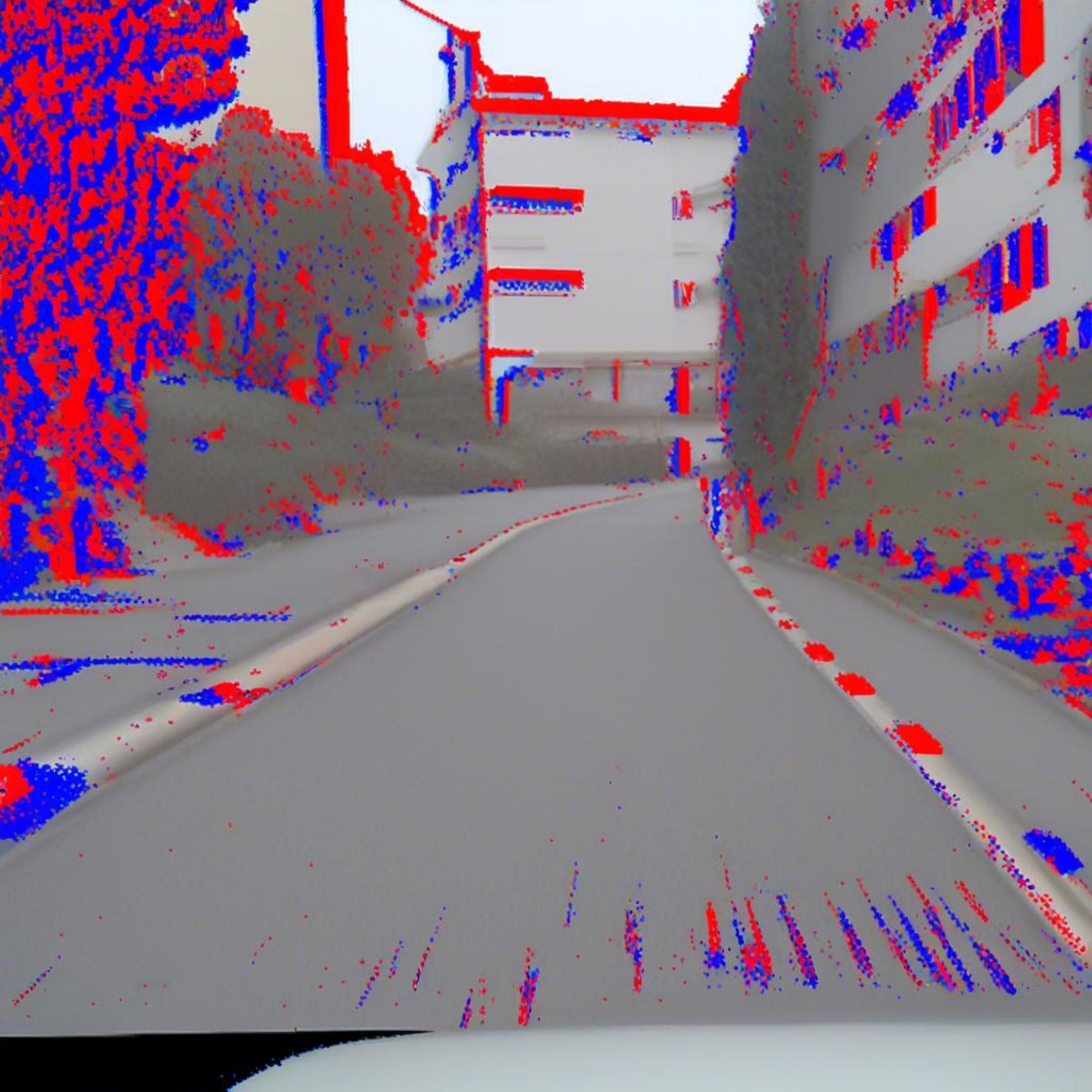}
& \includegraphics[width=1.7cm,height=1.7cm]{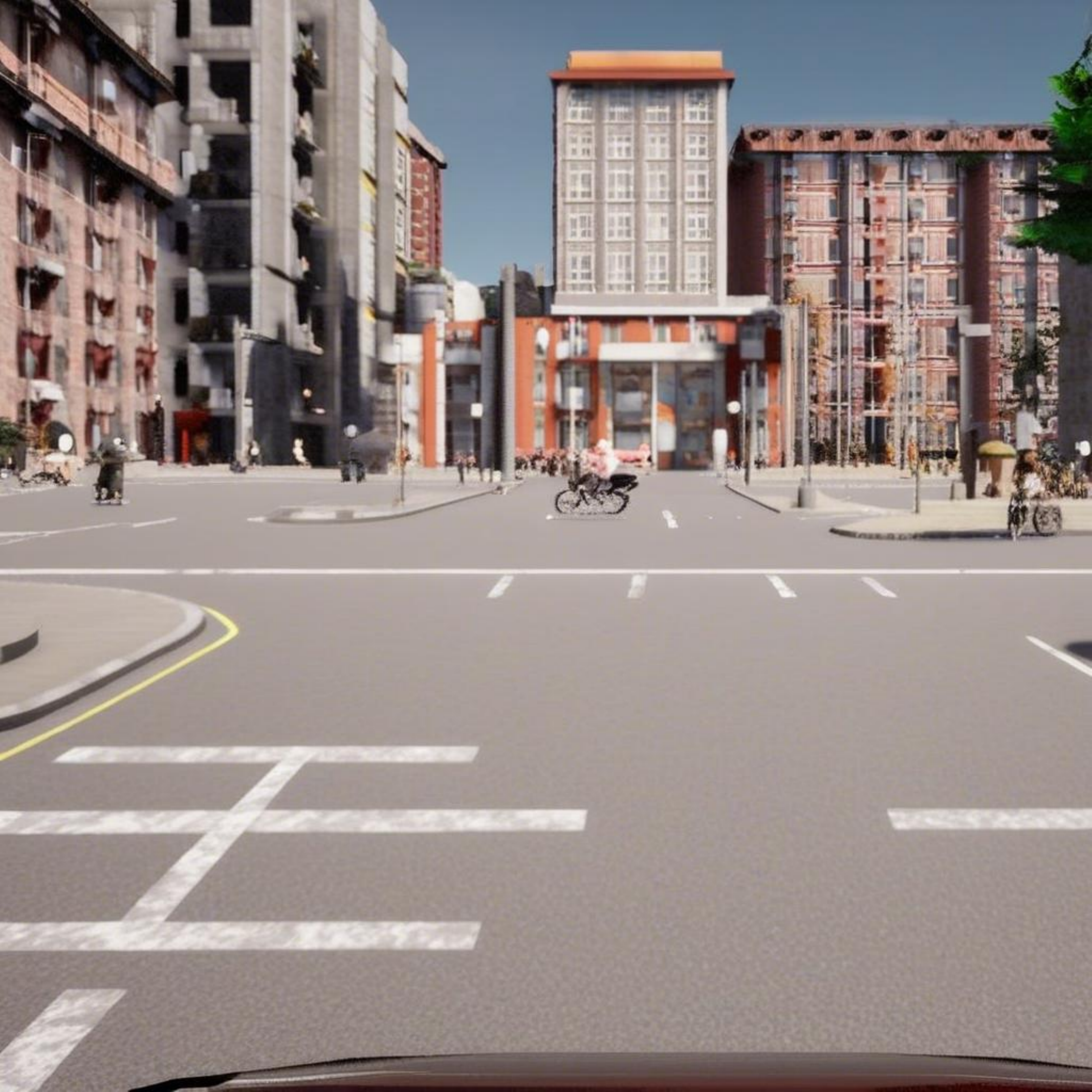}
& \includegraphics[width=1.7cm,height=1.7cm]{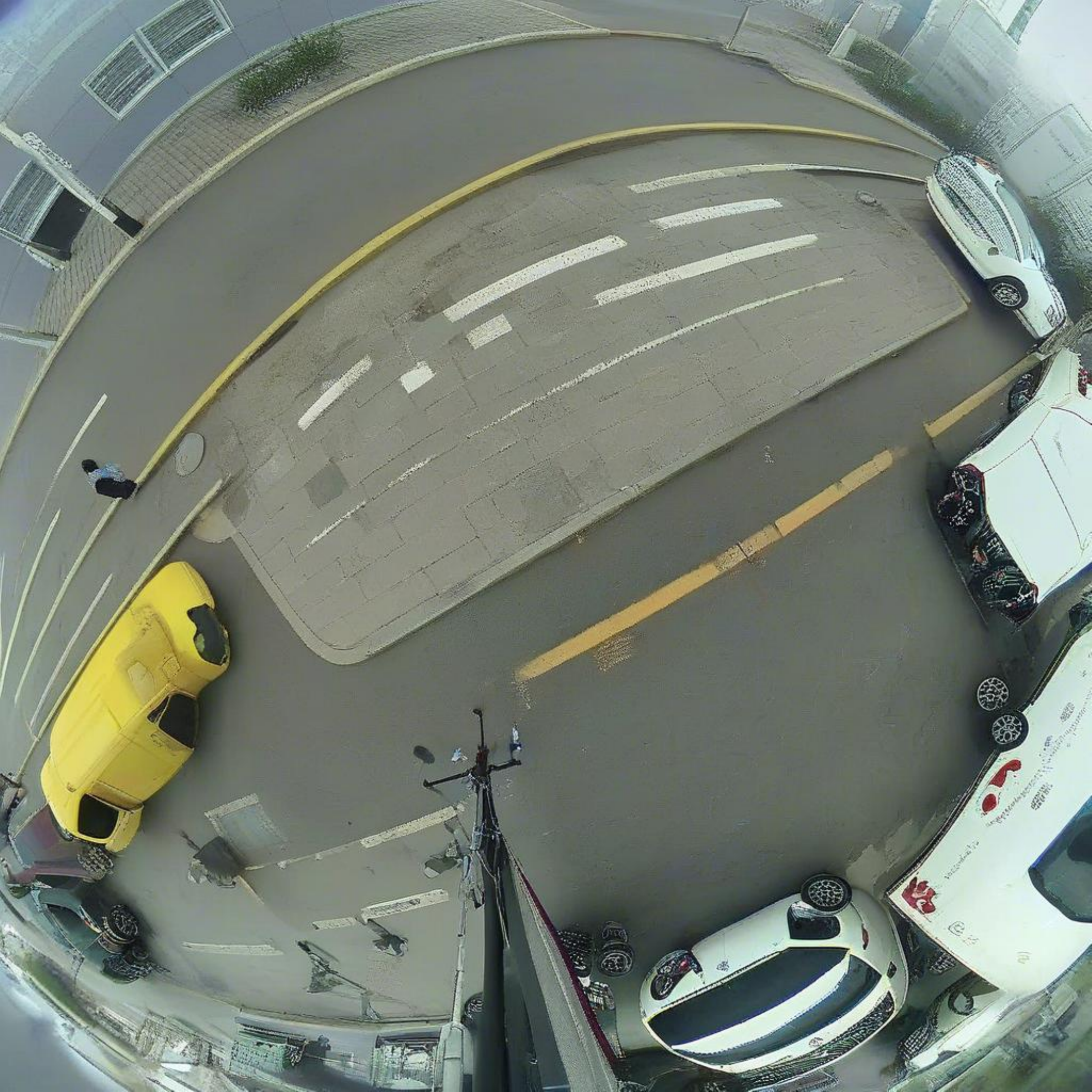}
& \includegraphics[width=1.7cm,height=1.7cm]{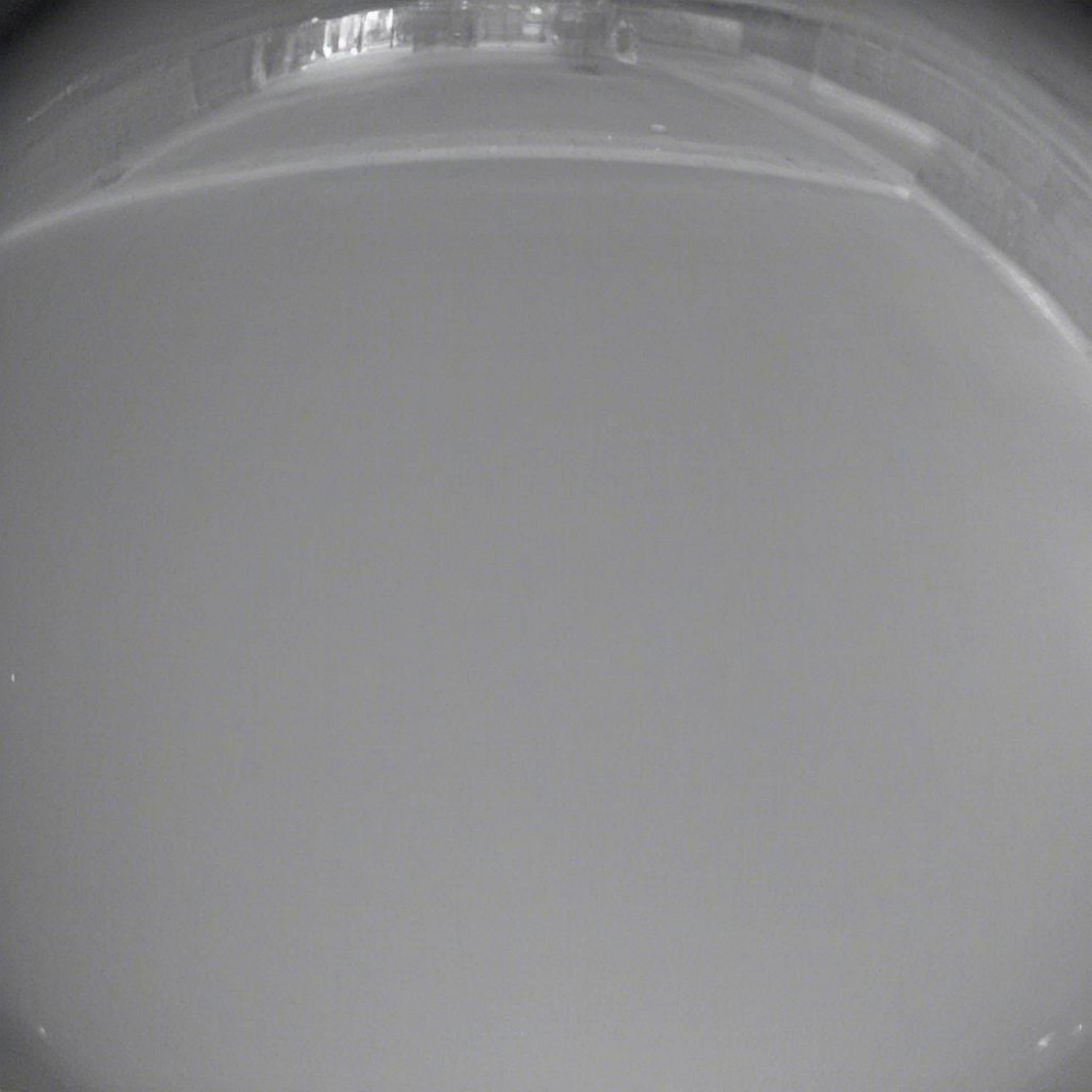}
& \includegraphics[width=1.7cm,height=1.7cm]{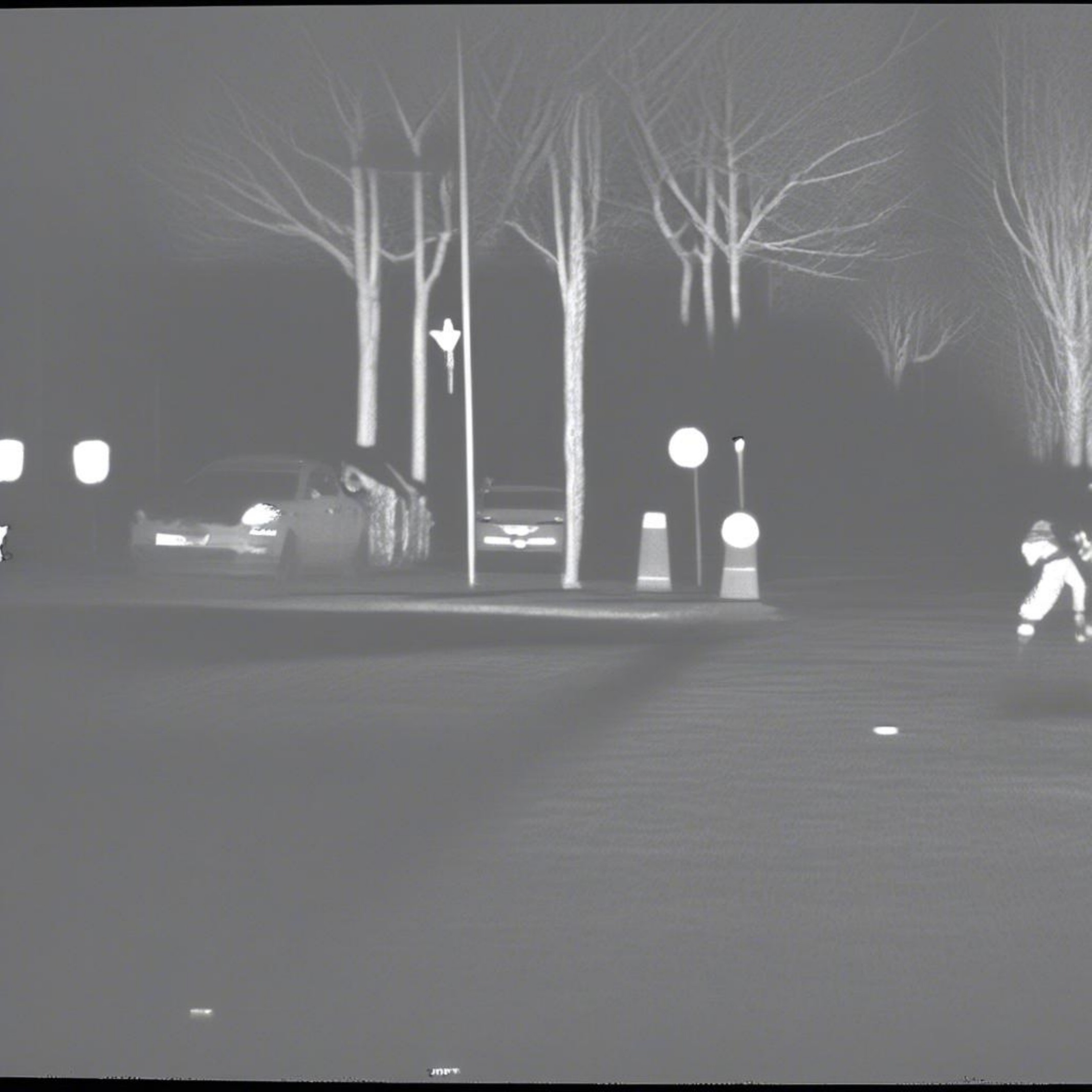}
& \includegraphics[width=1.7cm,height=1.7cm]{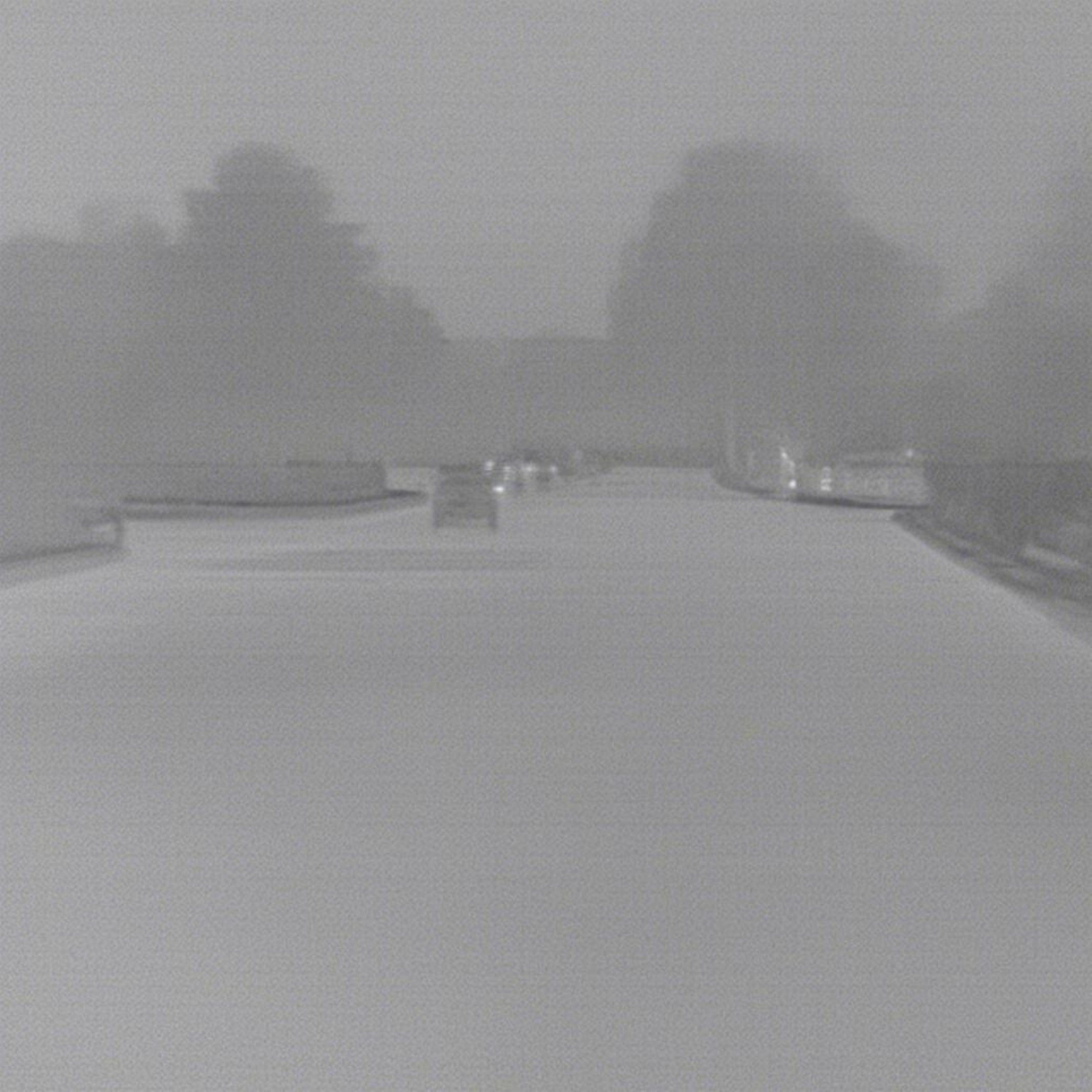}
& \includegraphics[width=1.7cm,height=1.7cm]{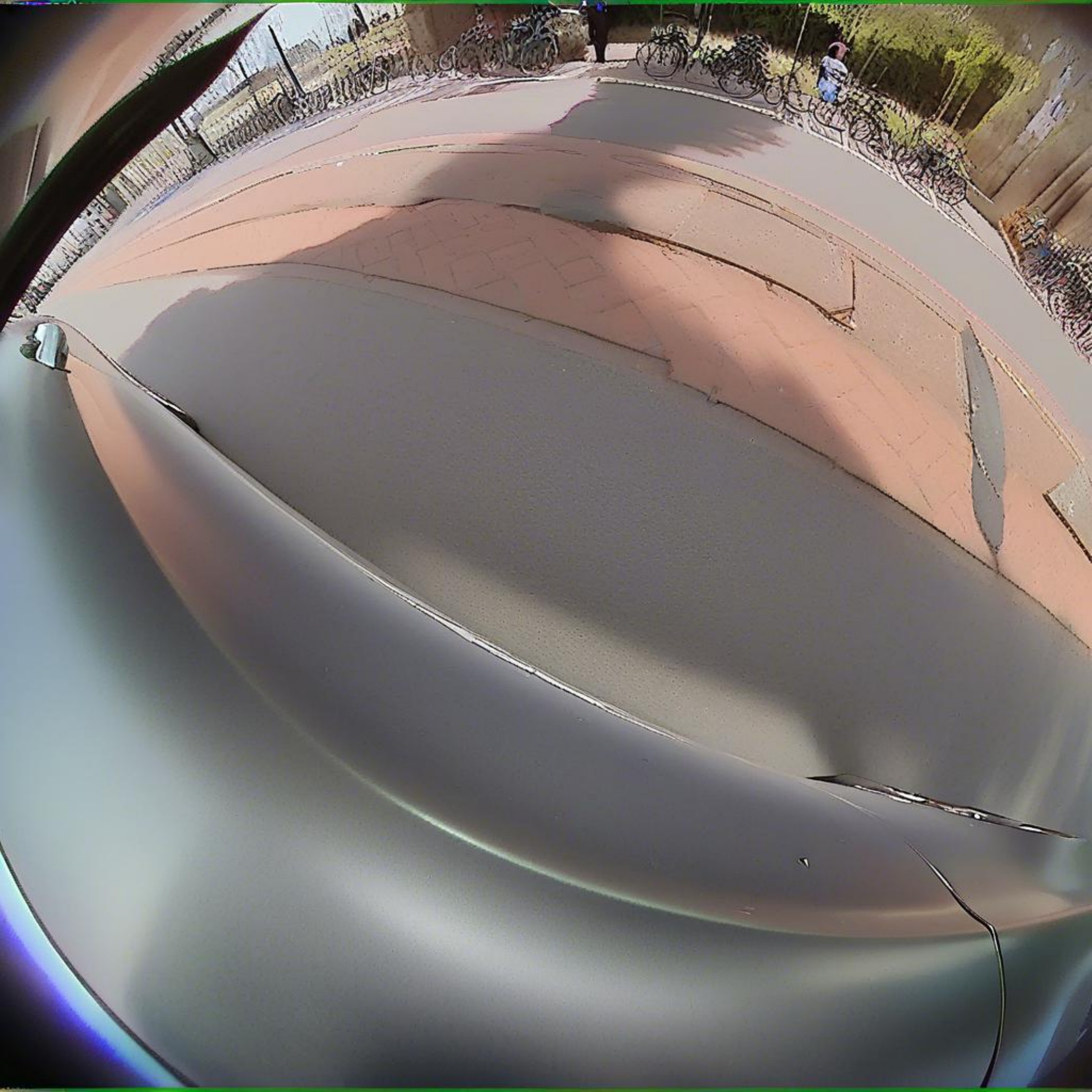}
& \includegraphics[width=1.7cm,height=1.7cm]{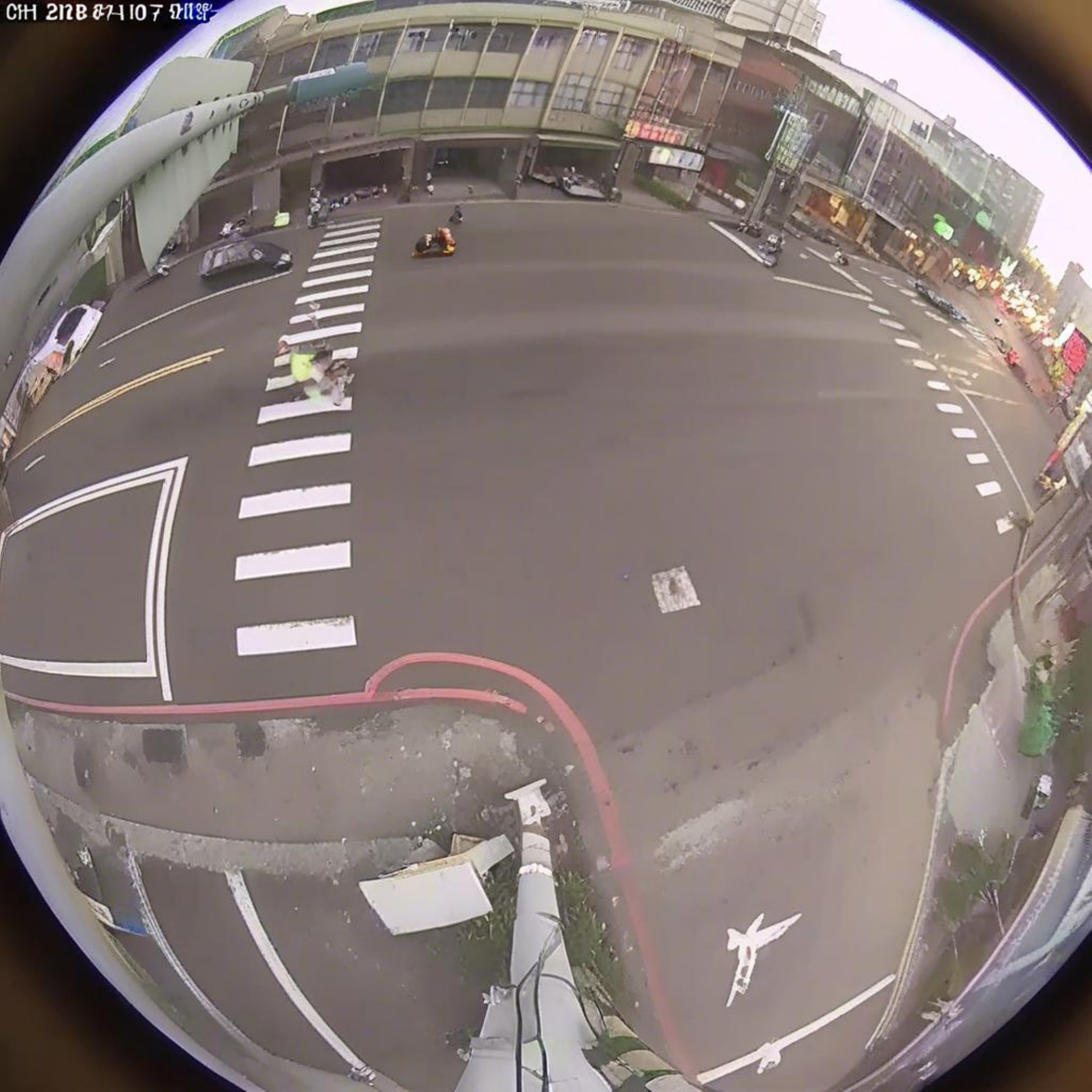}  \\

Inspiration Tree~\cite{vinker2023concept}
& \includegraphics[width=1.7cm,height=1.7cm]{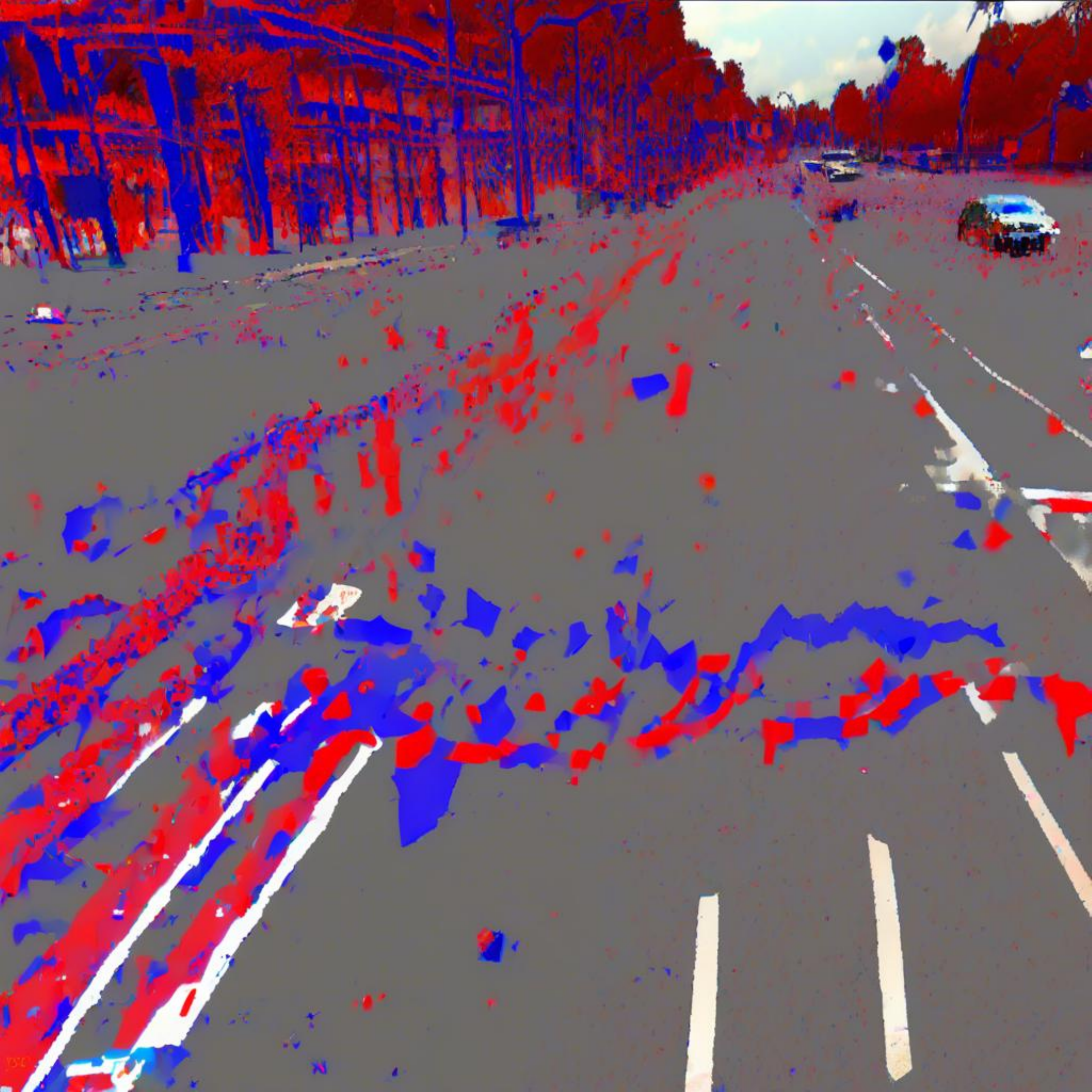}
& \includegraphics[width=1.7cm,height=1.7cm]{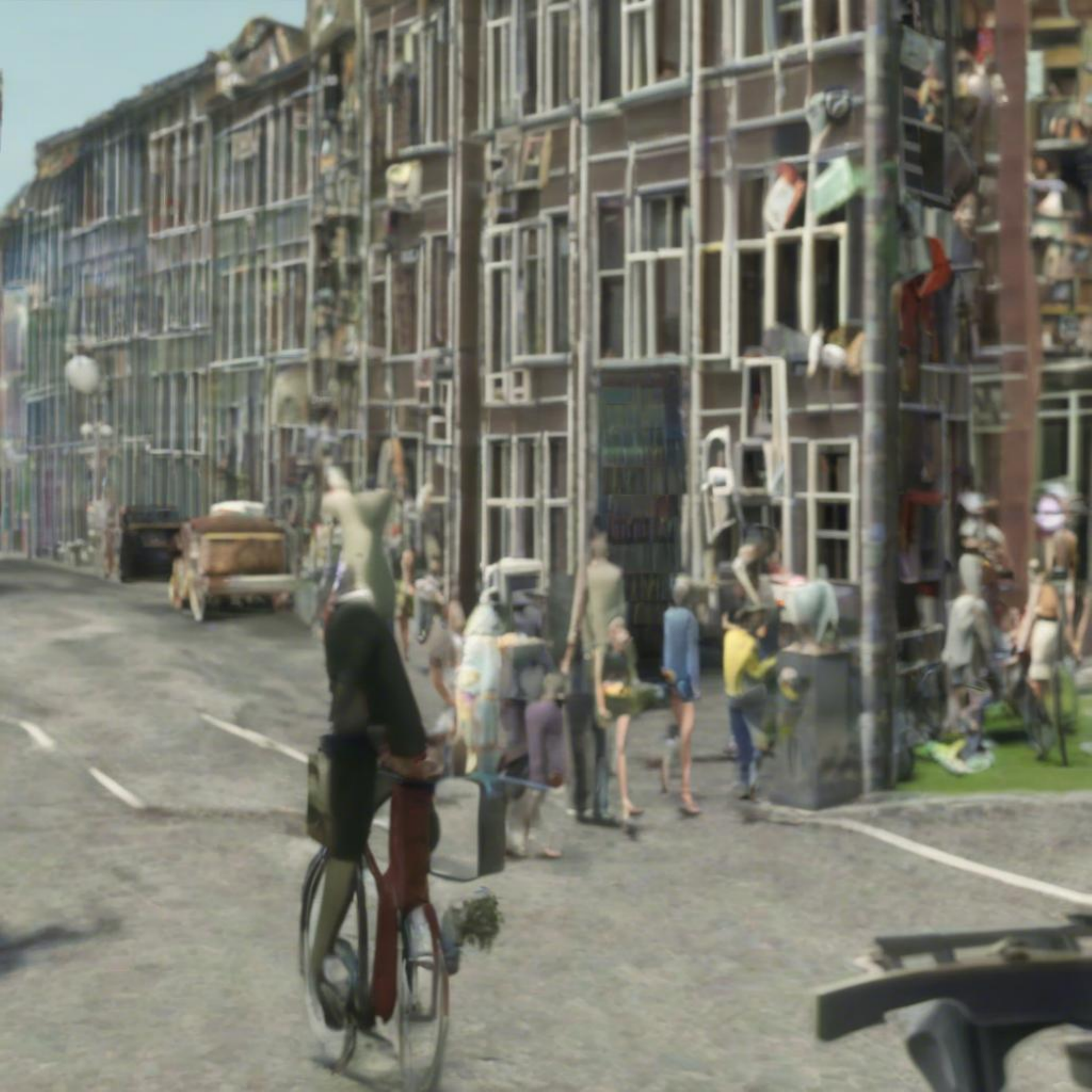}
& \includegraphics[width=1.7cm,height=1.7cm]{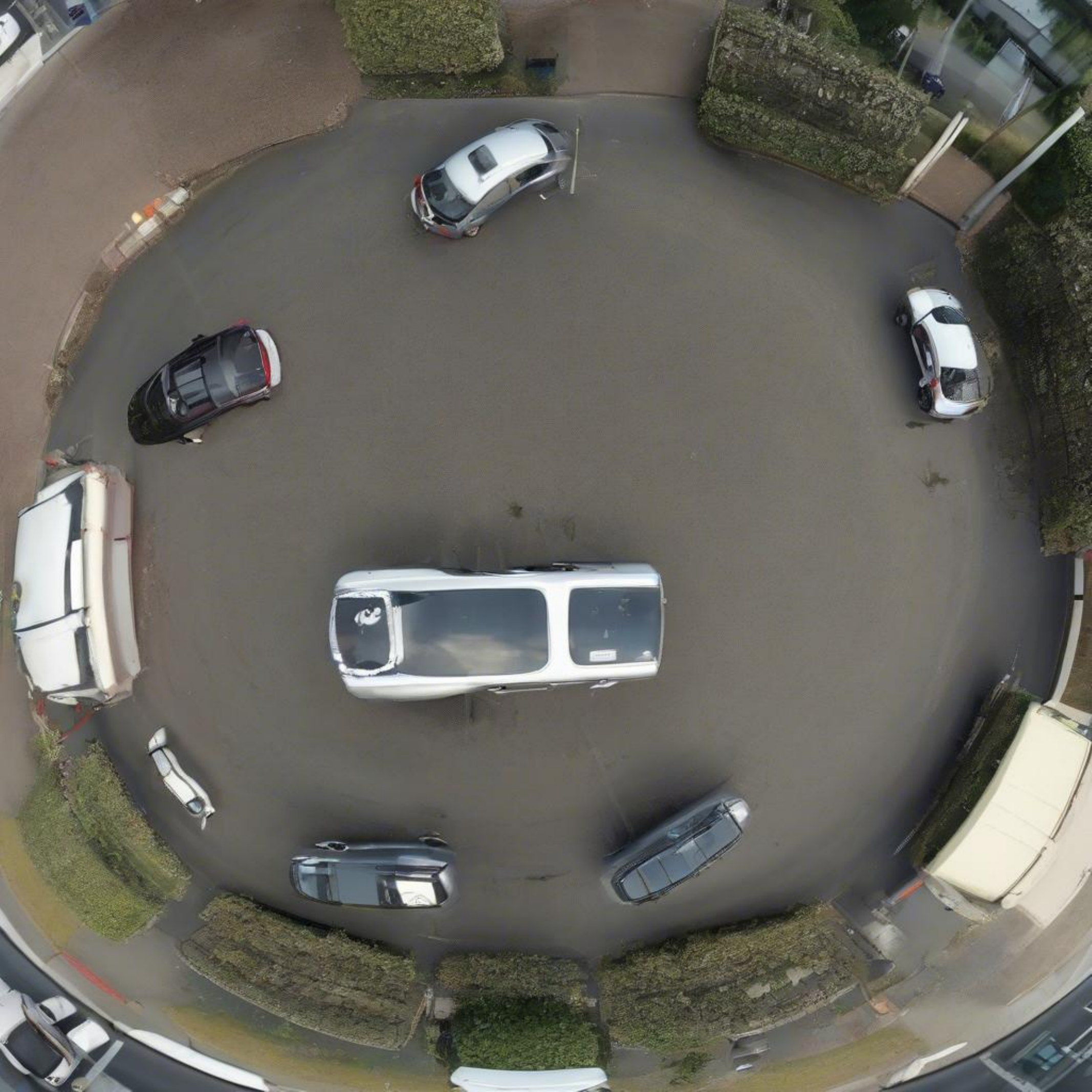}
& \includegraphics[width=1.7cm,height=1.7cm]{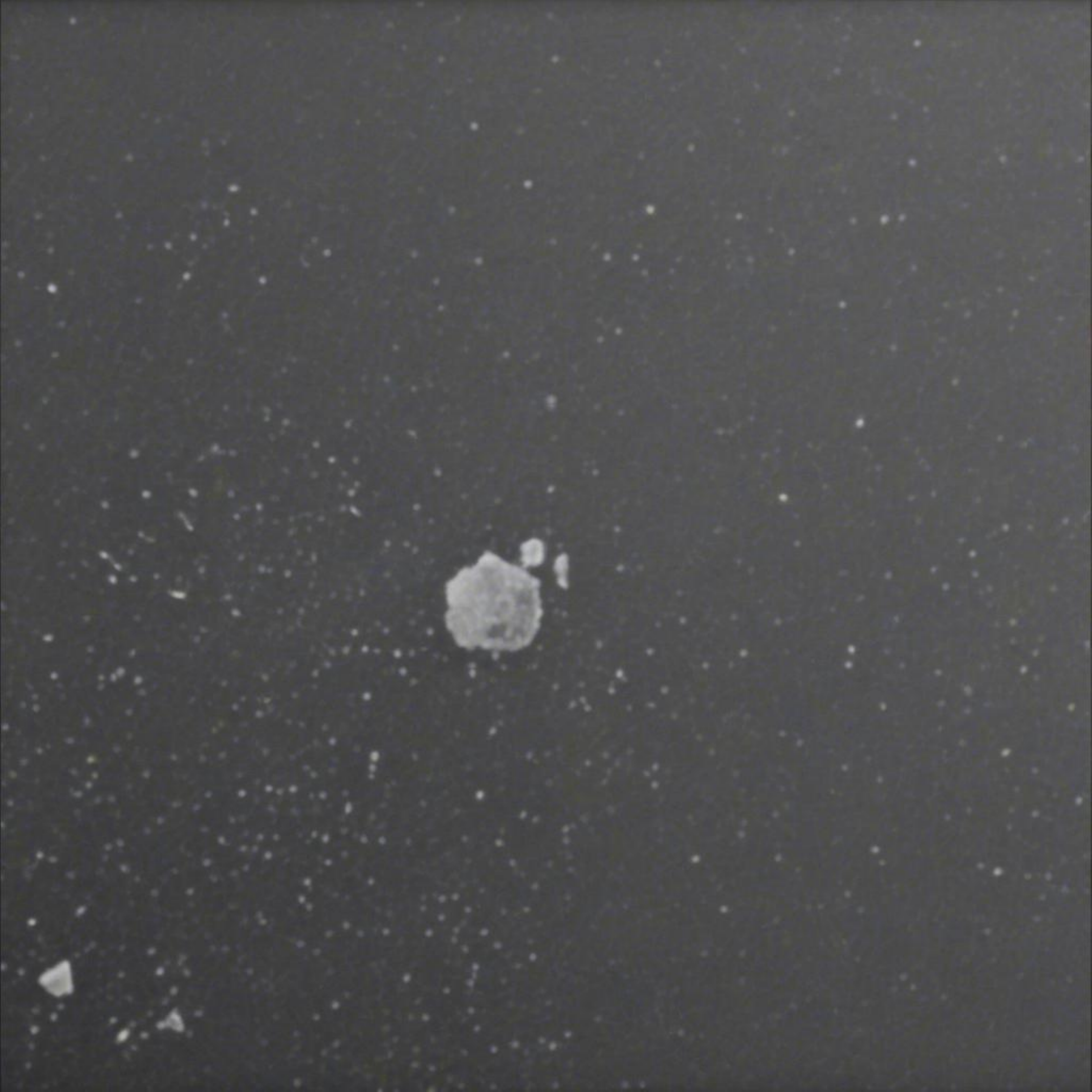}
& \includegraphics[width=1.7cm,height=1.7cm]{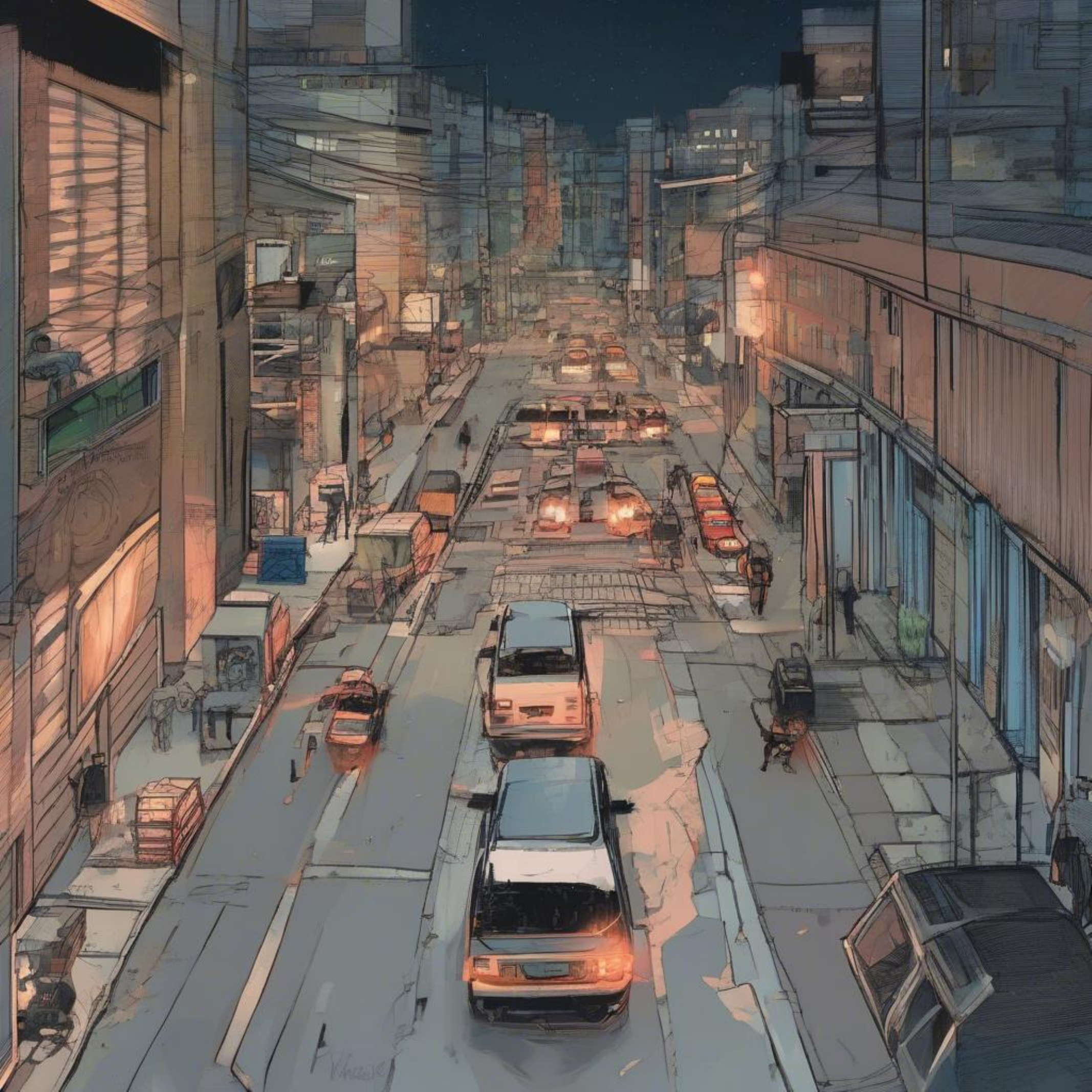}
& \includegraphics[width=1.7cm,height=1.7cm]{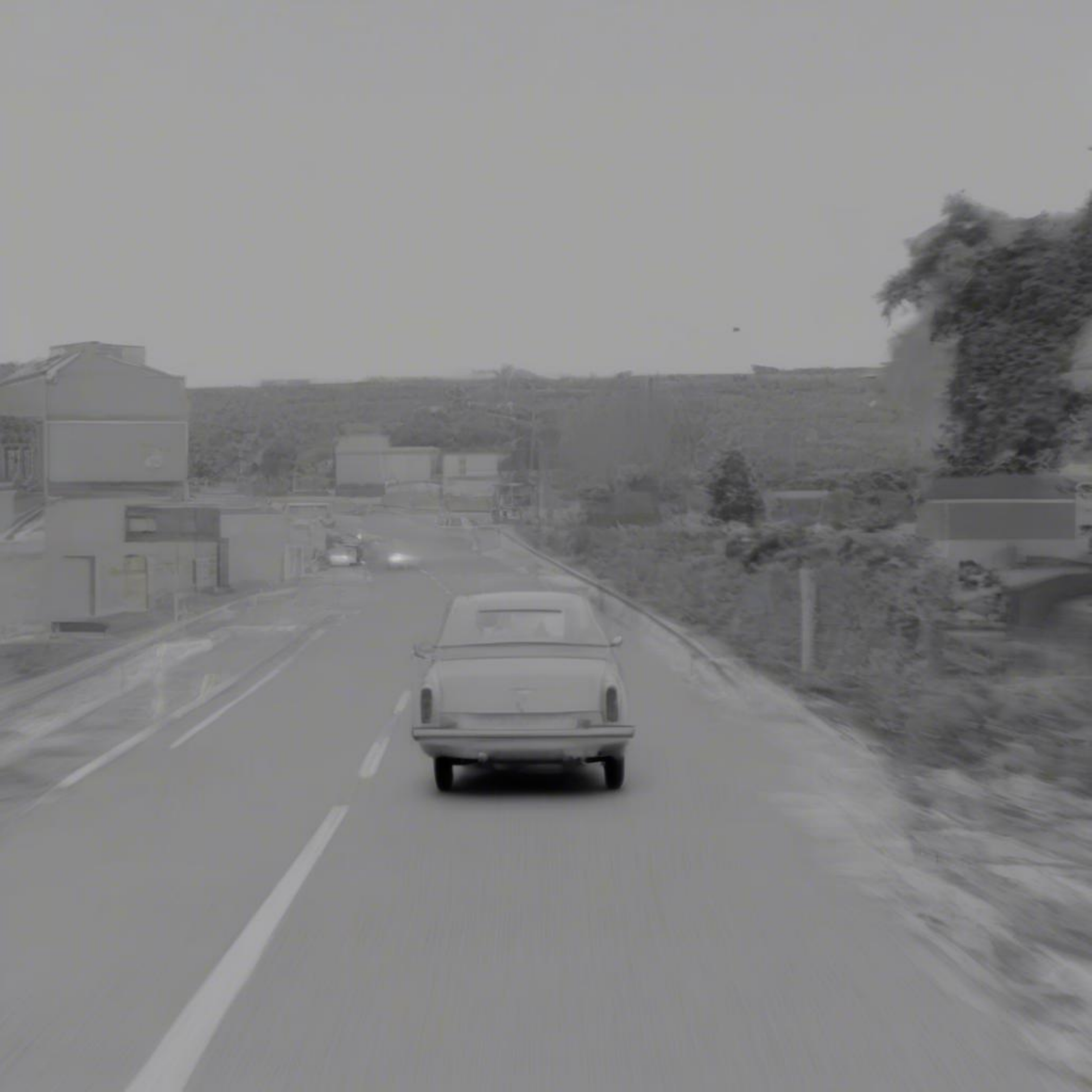}
& \includegraphics[width=1.7cm,height=1.7cm]{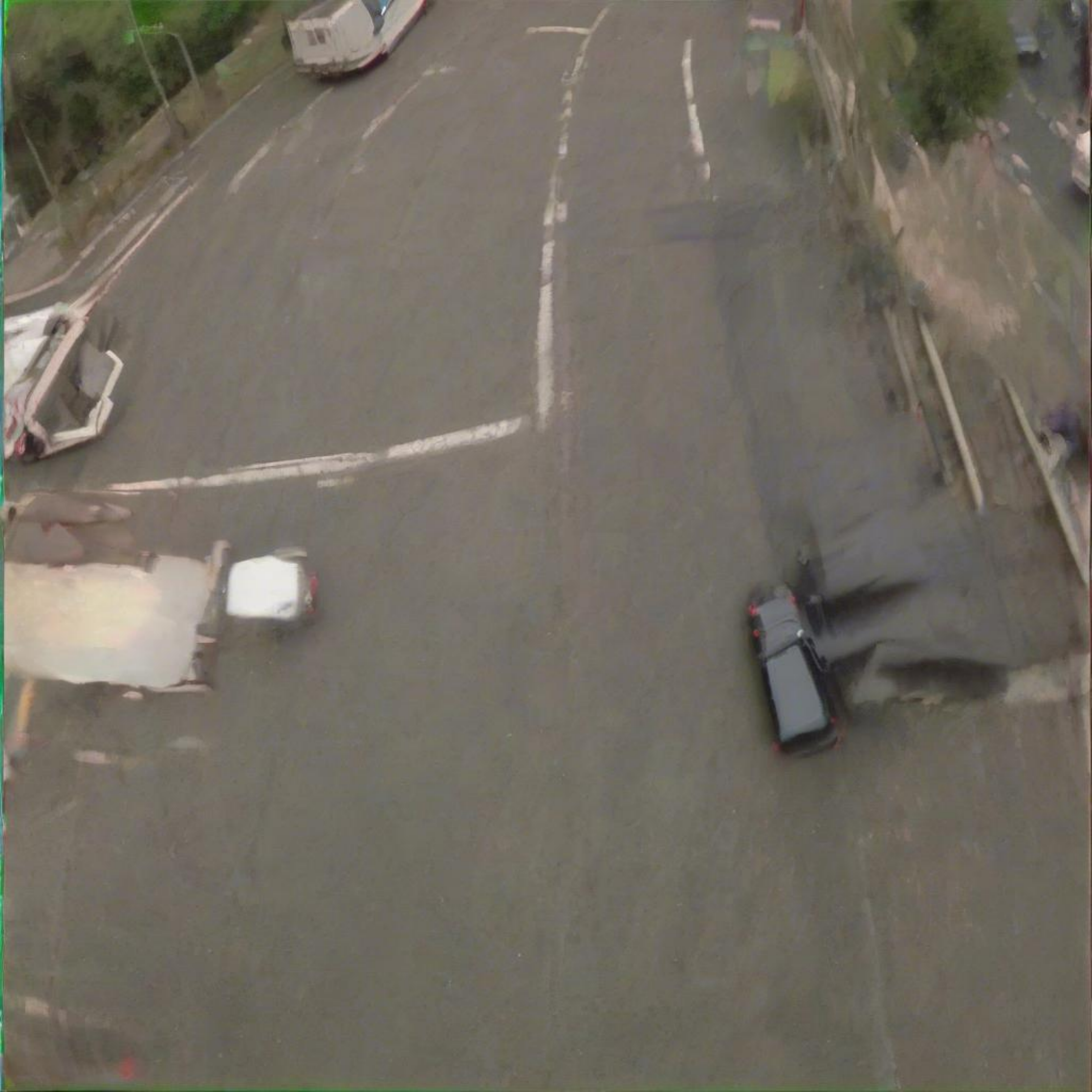}
& \includegraphics[width=1.7cm,height=1.7cm]{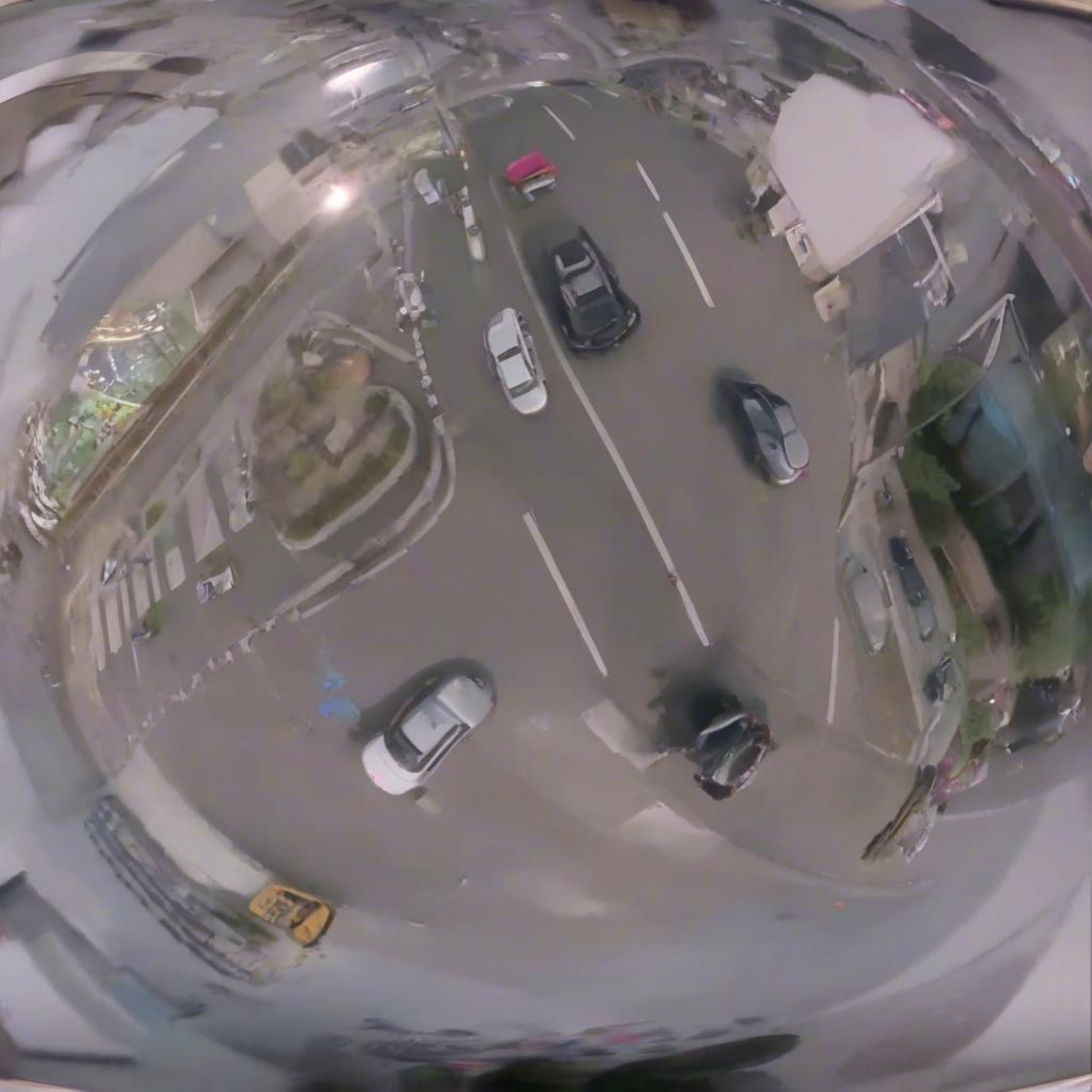} \\

MULTI~\cite{godavarthy2026multi}
& \includegraphics[width=1.7cm,height=1.7cm]{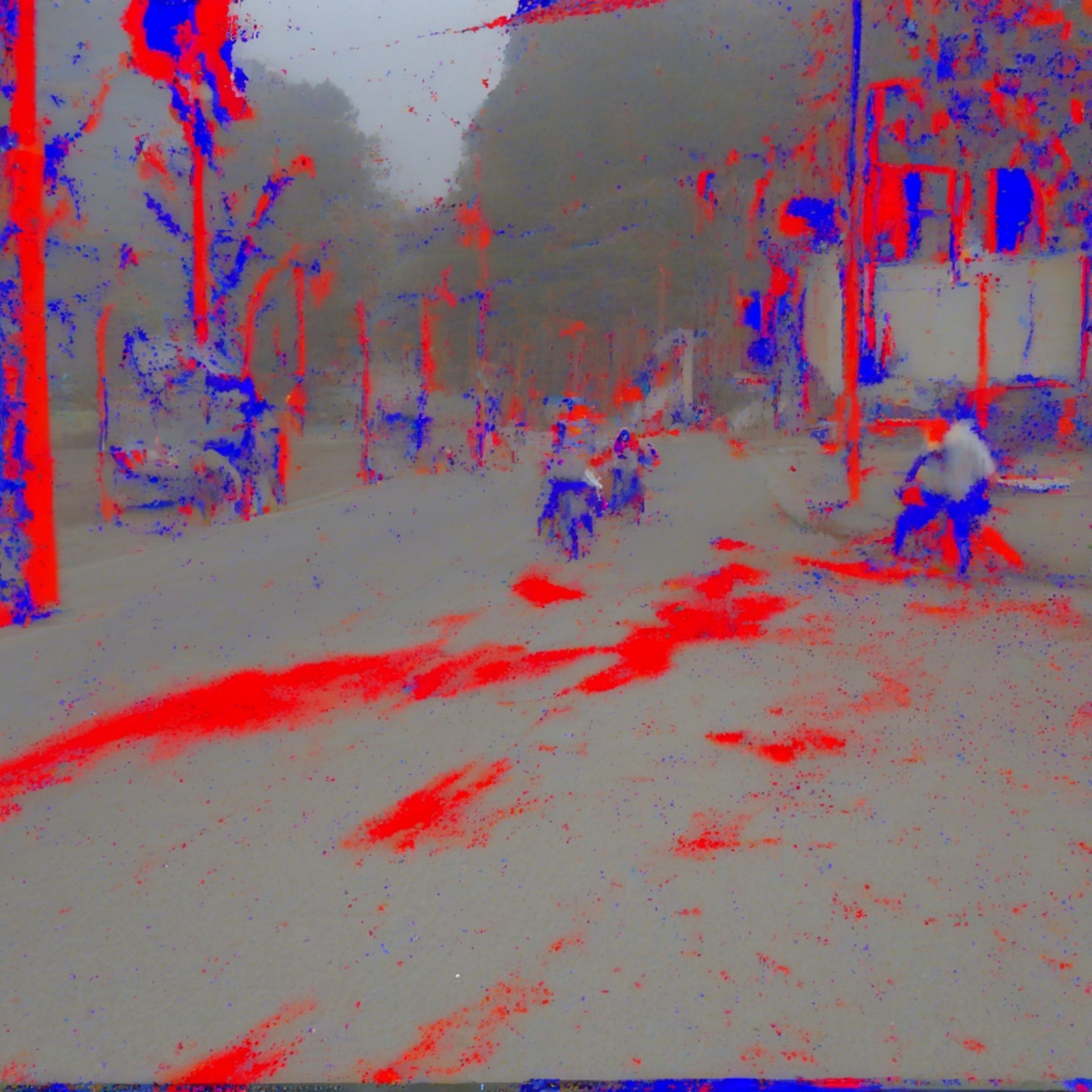}
& \includegraphics[width=1.7cm,height=1.7cm]{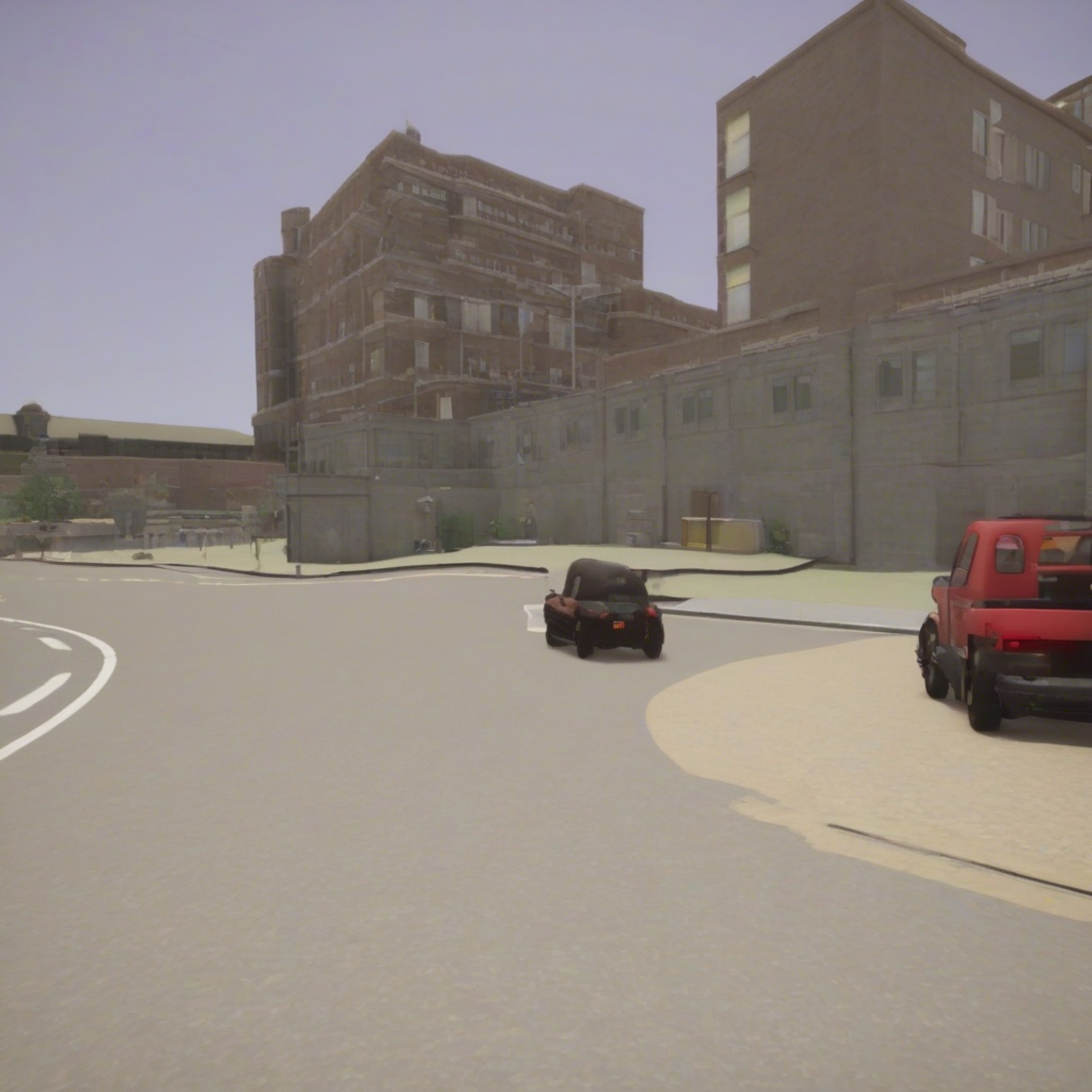}
& \includegraphics[width=1.7cm,height=1.7cm]{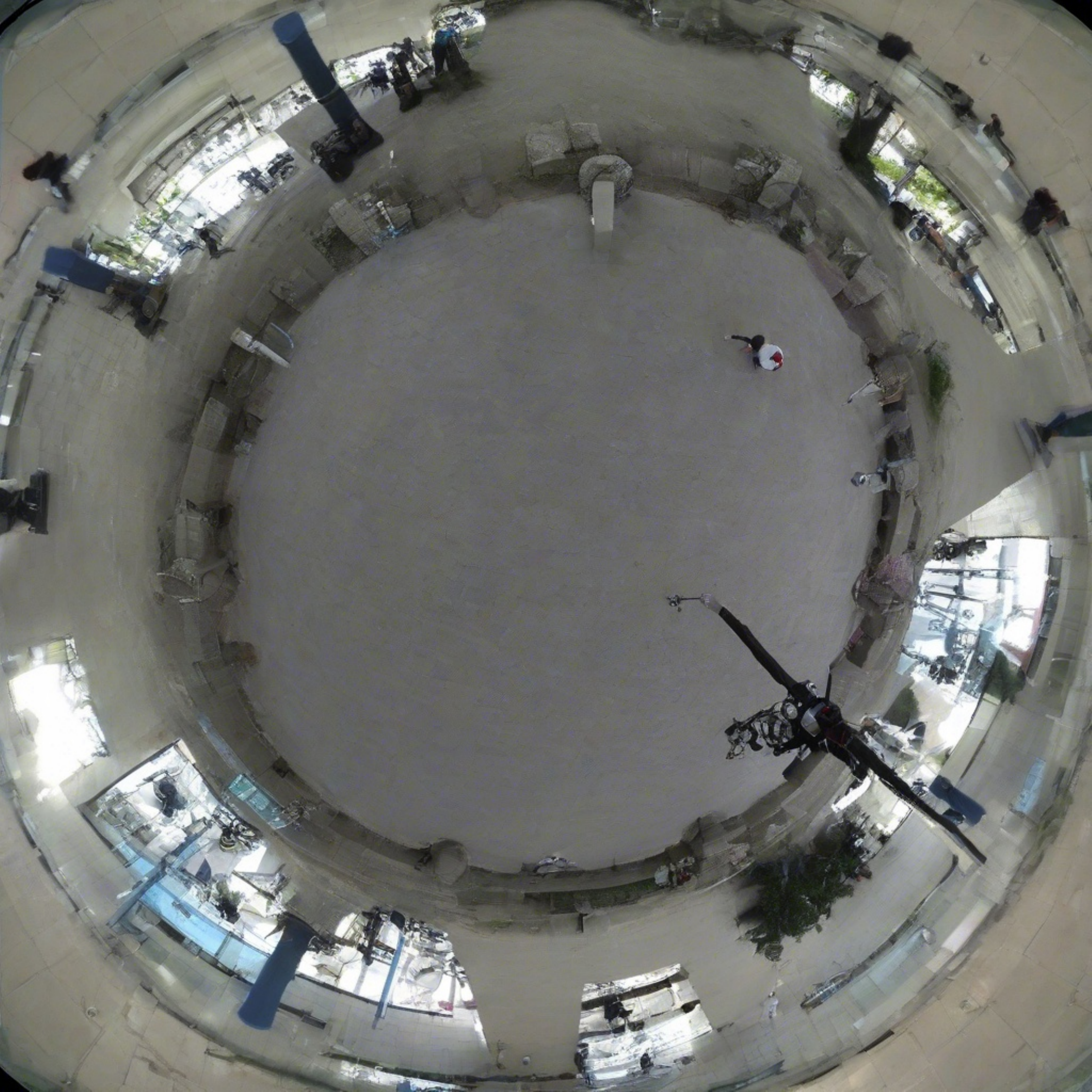}
& \includegraphics[width=1.7cm,height=1.7cm]{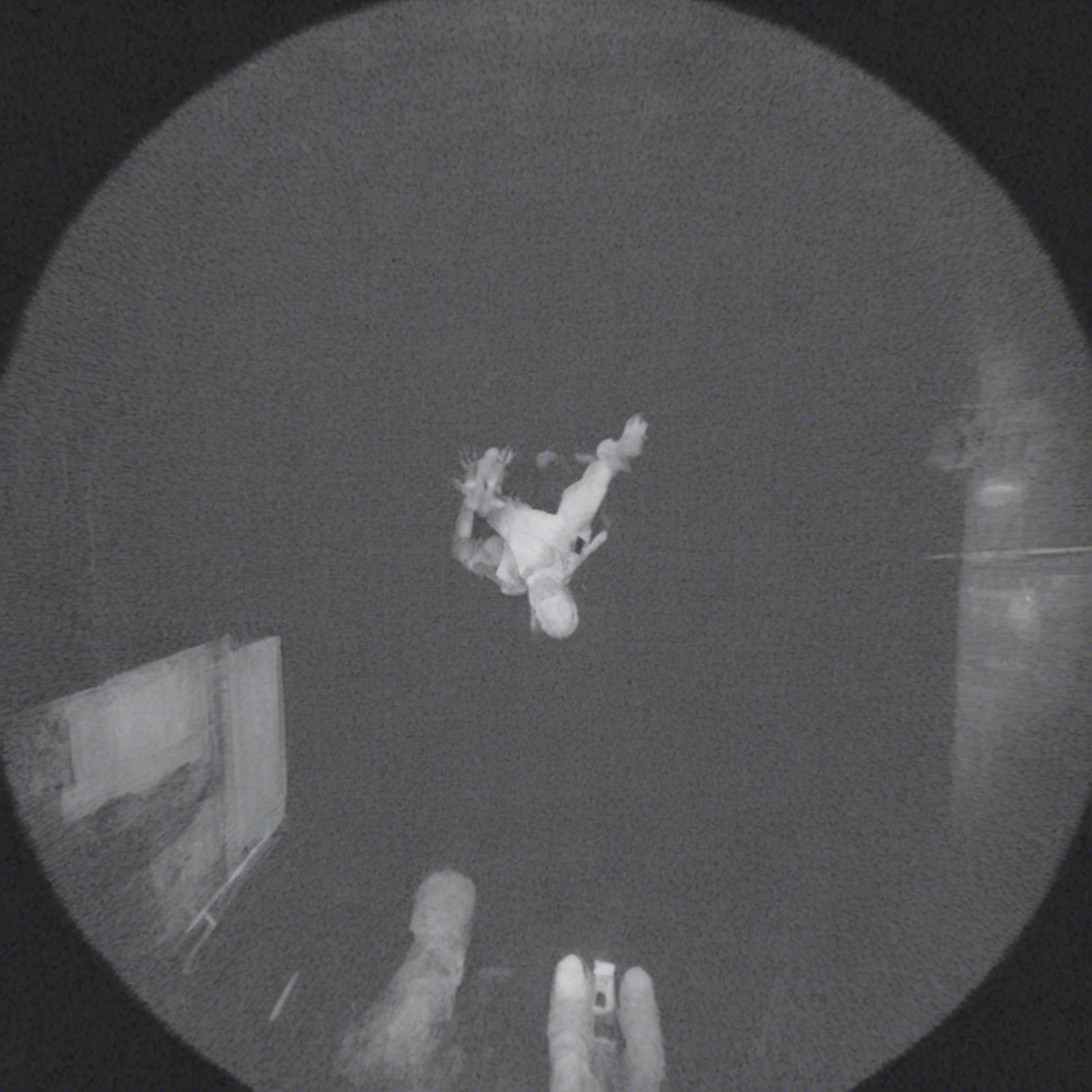}
& \includegraphics[width=1.7cm,height=1.7cm]{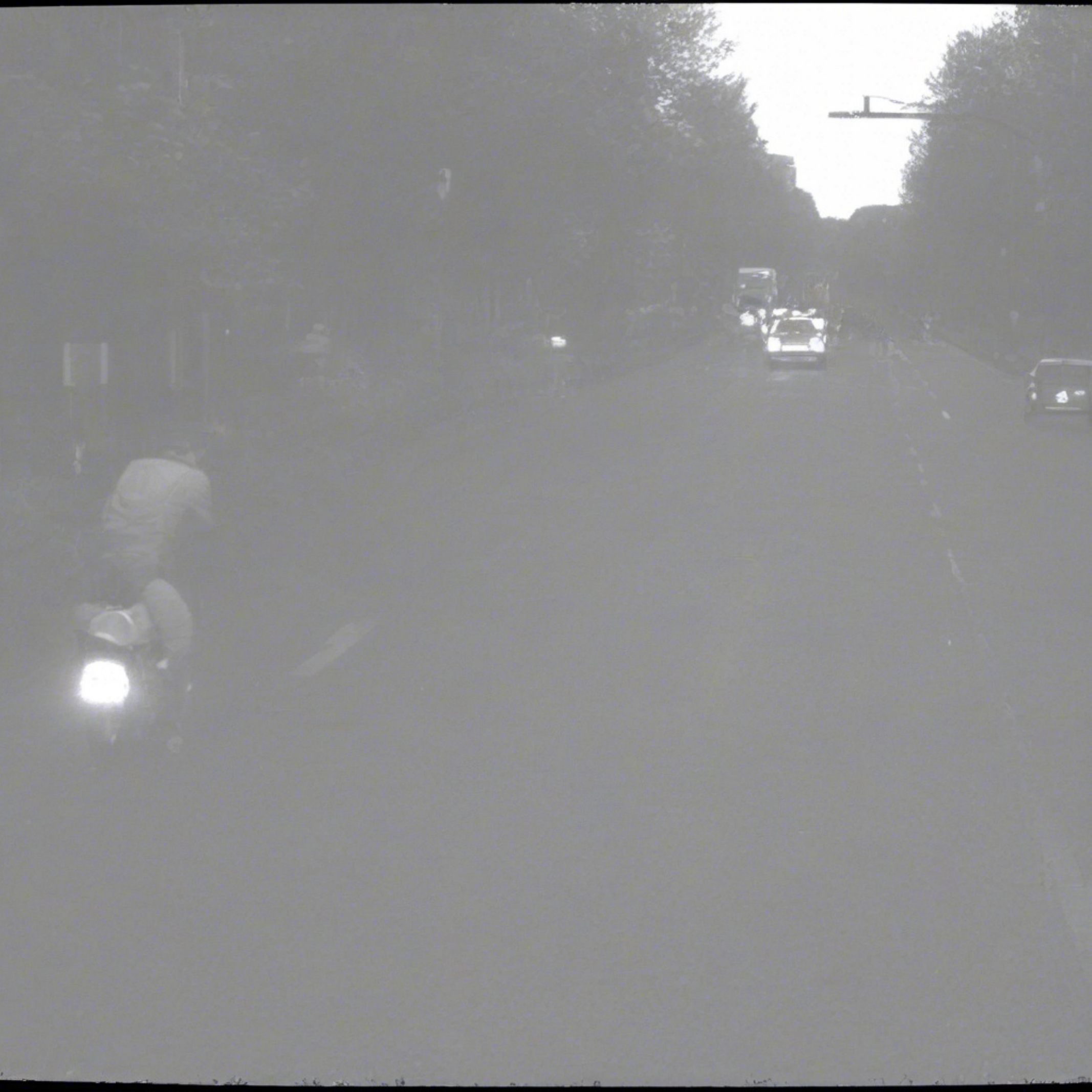}
& \includegraphics[width=1.7cm,height=1.7cm]{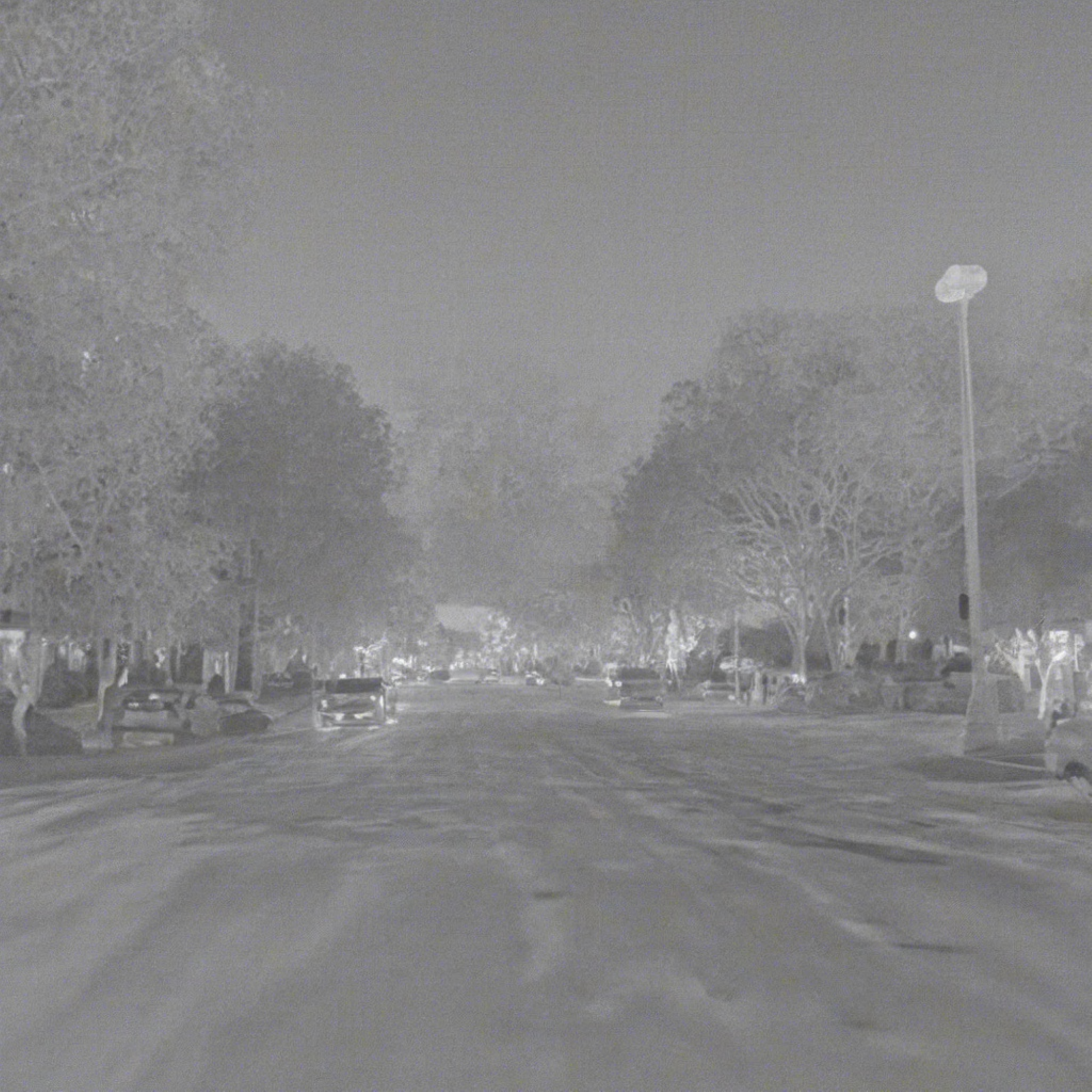}
& \includegraphics[width=1.7cm,height=1.7cm]{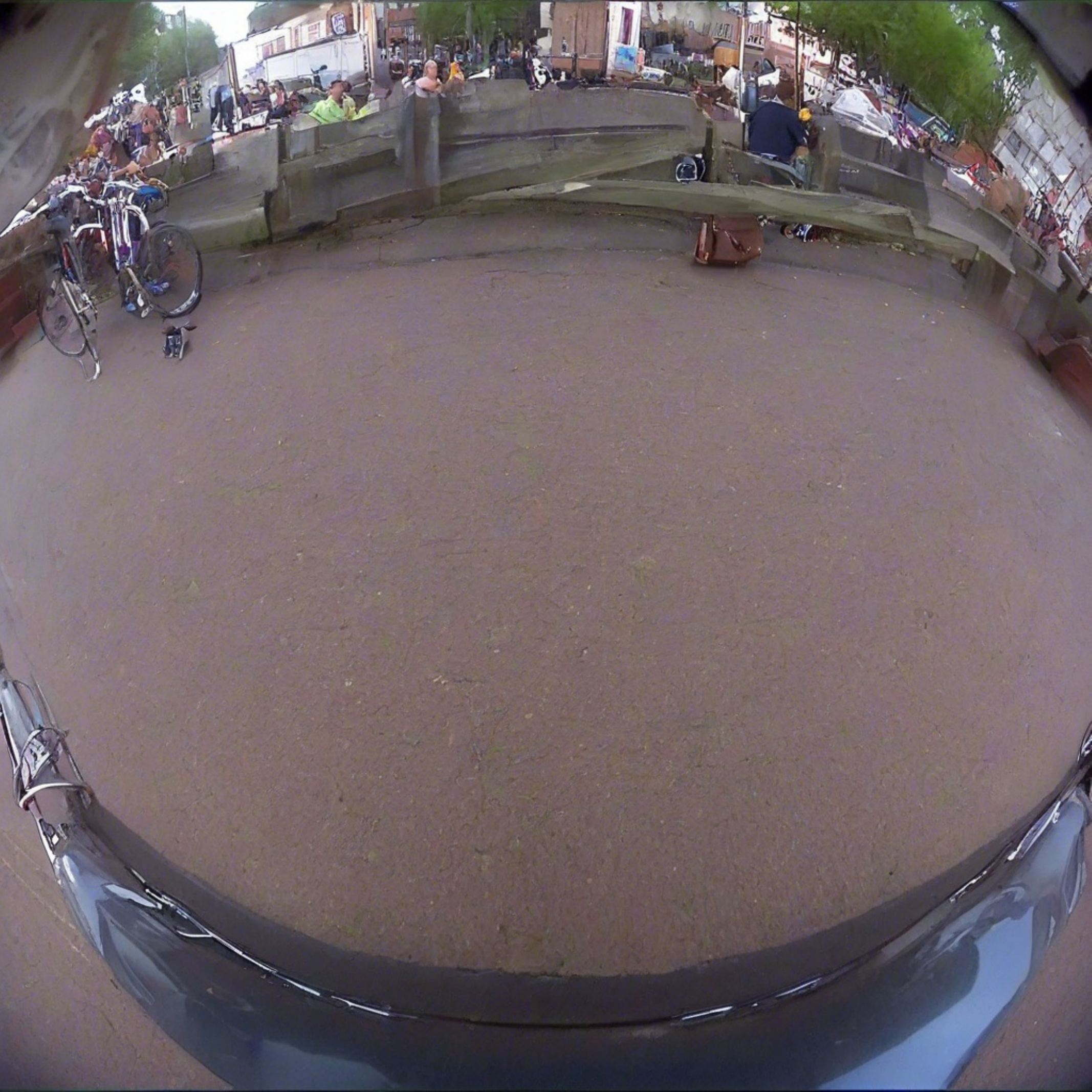}
& \includegraphics[width=1.7cm,height=1.7cm]{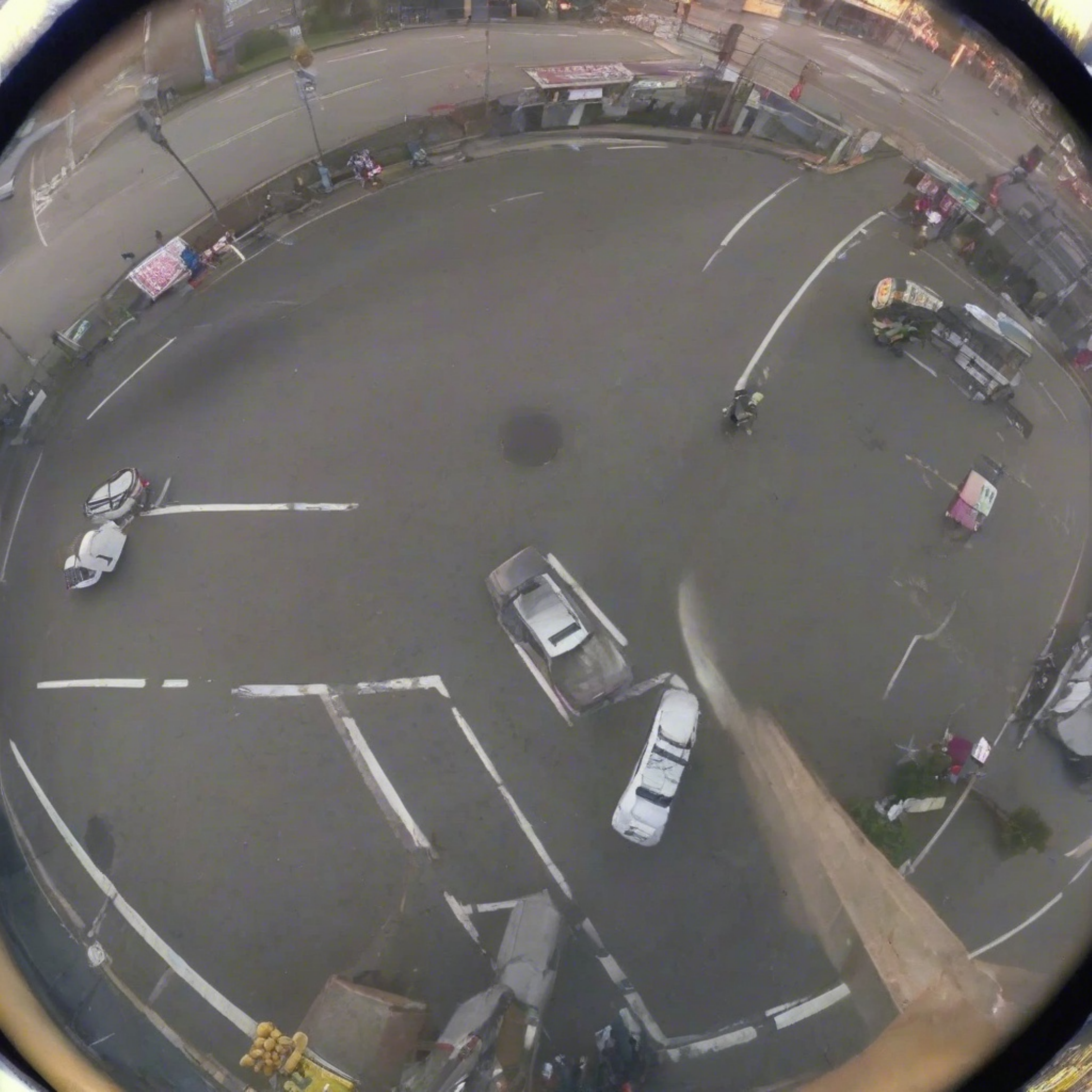} \\

X-MULTI
& \includegraphics[width=1.7cm,height=1.7cm]{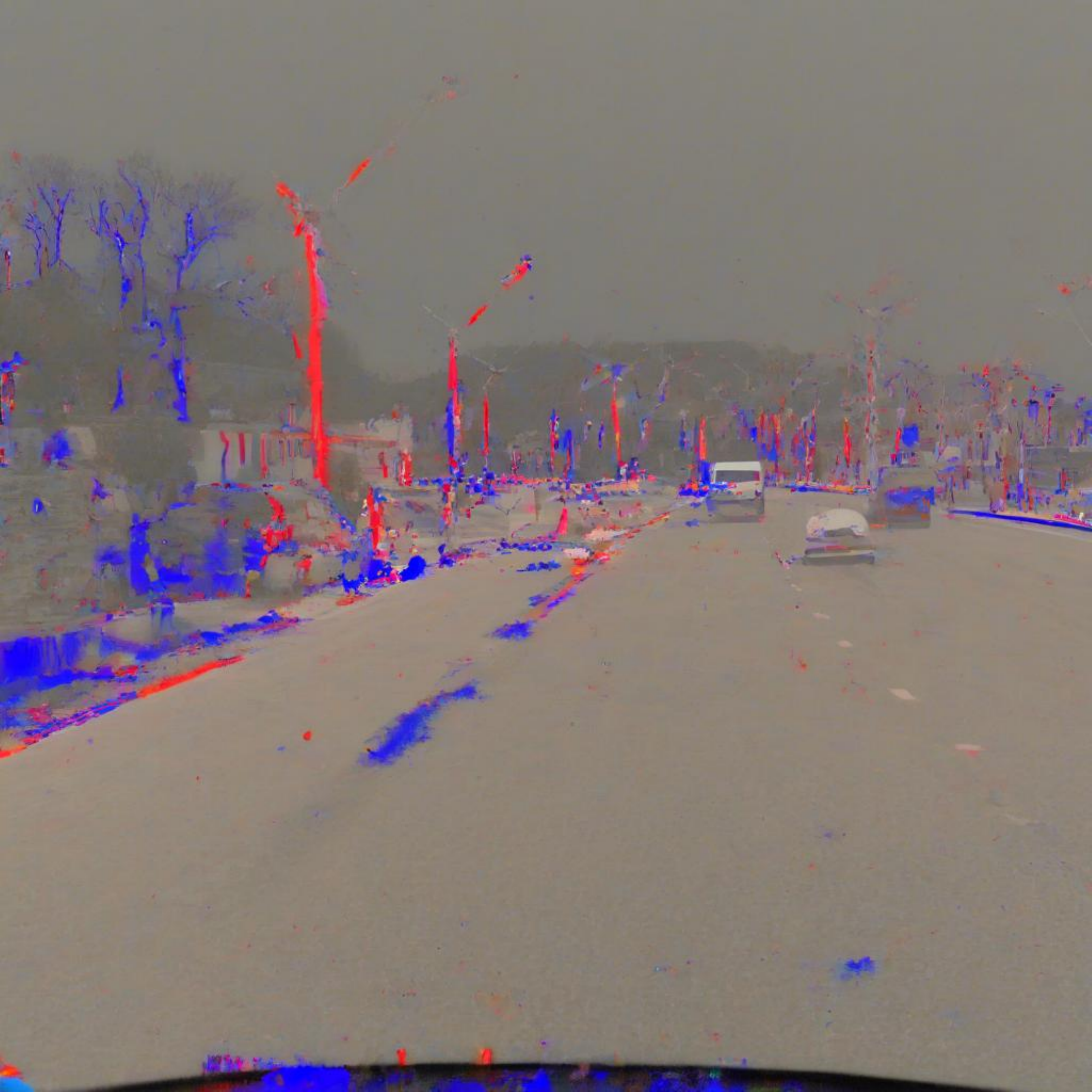}
& \includegraphics[width=1.7cm,height=1.7cm]{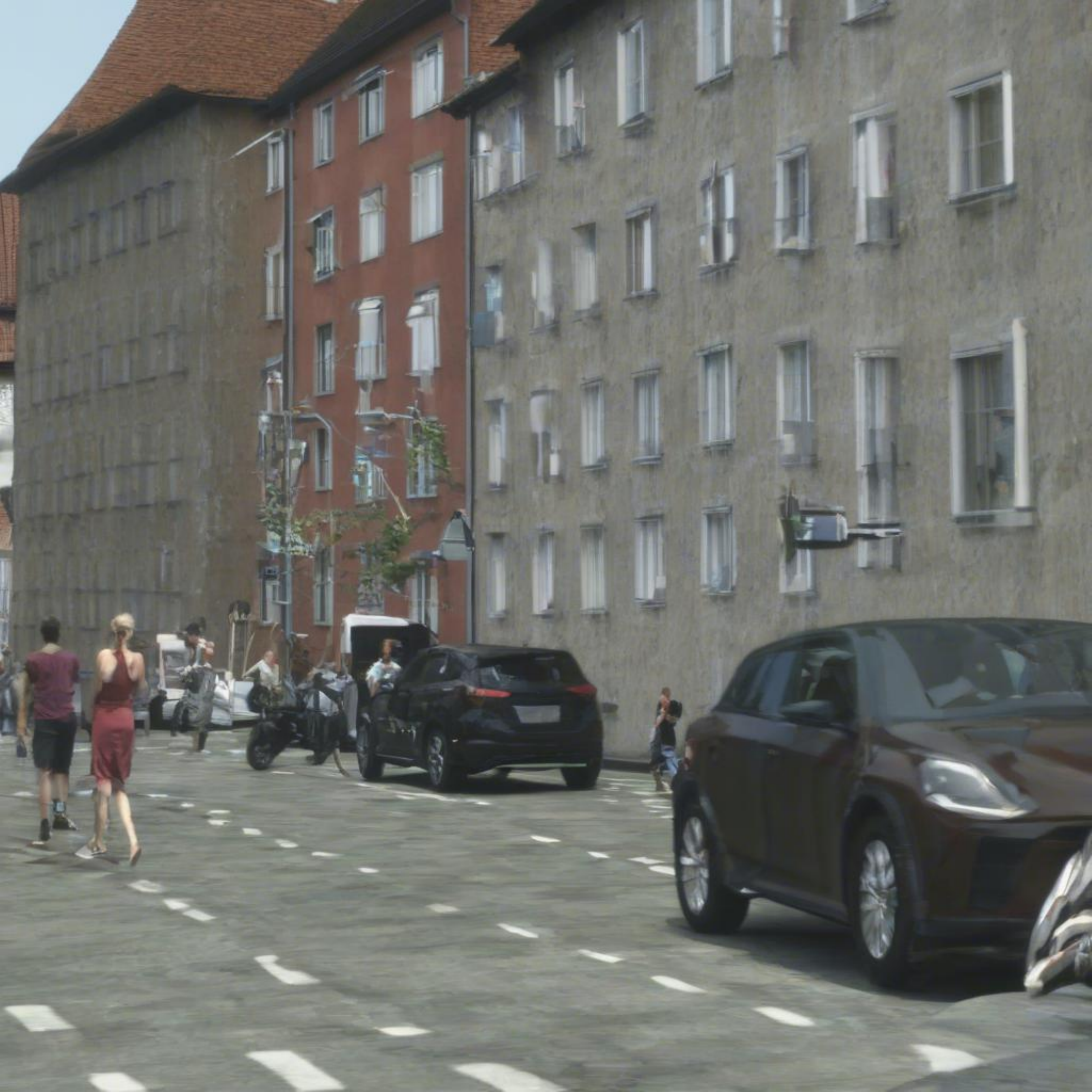}
& \includegraphics[width=1.7cm,height=1.7cm]{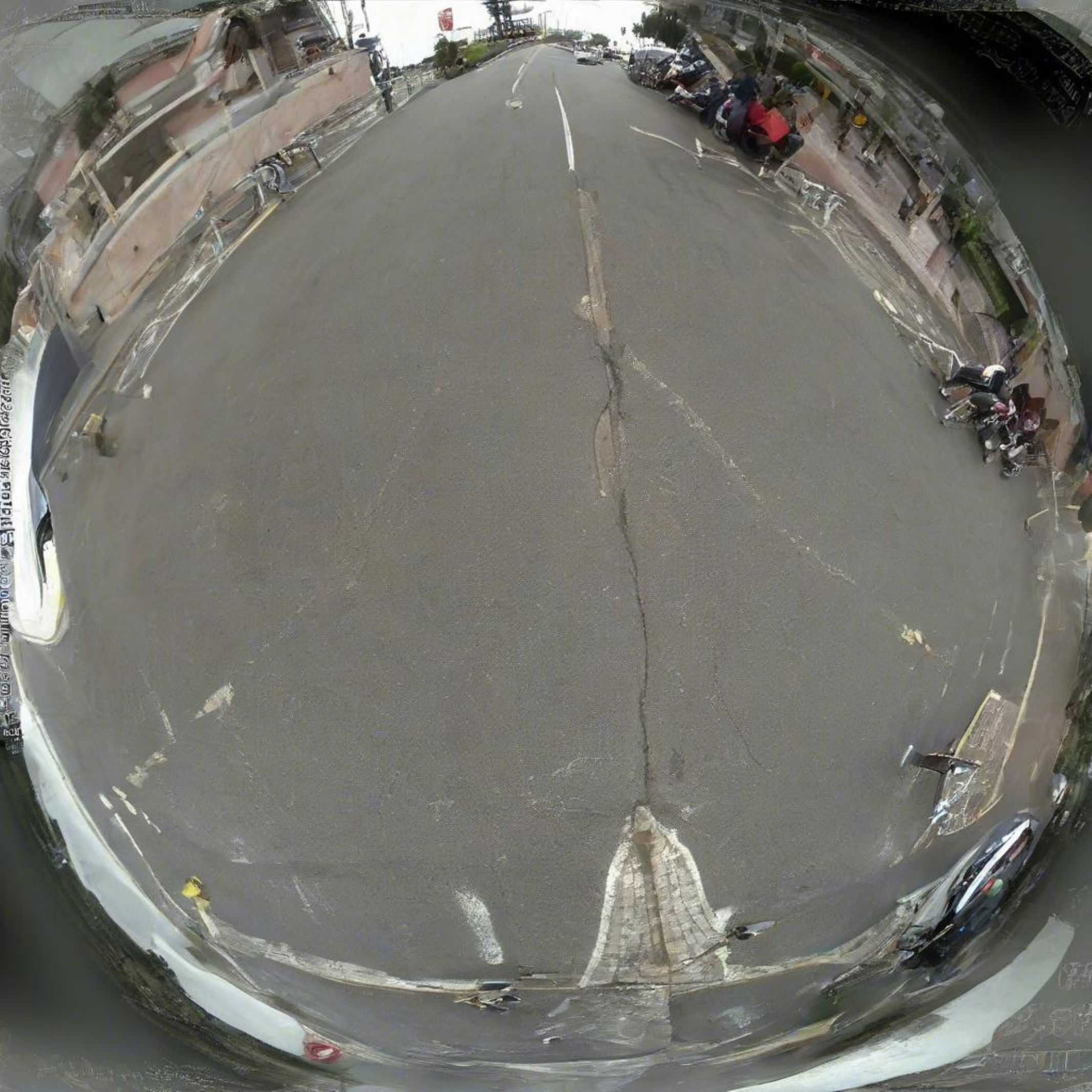}
& \includegraphics[width=1.7cm,height=1.7cm]{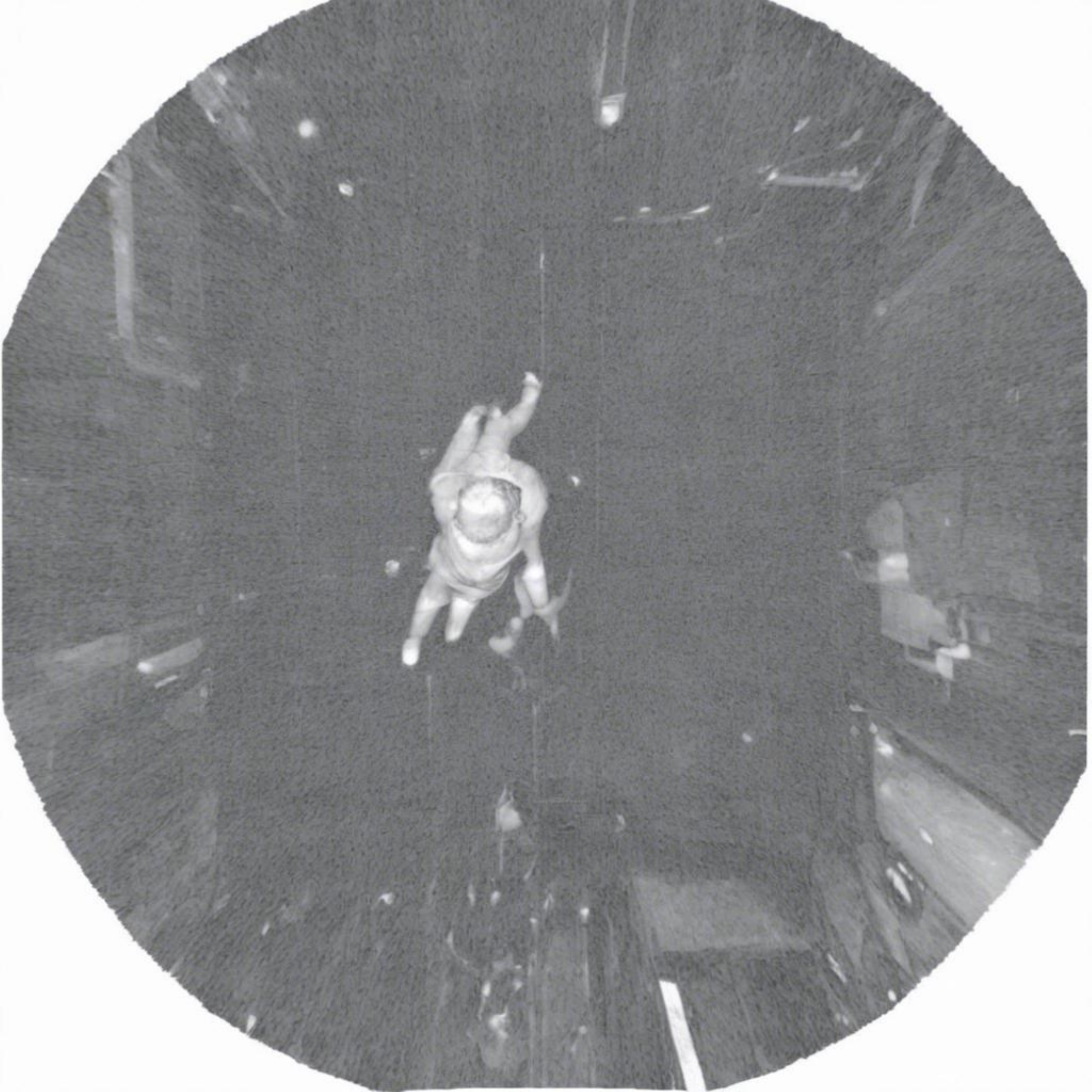}
& \includegraphics[width=1.7cm,height=1.7cm]{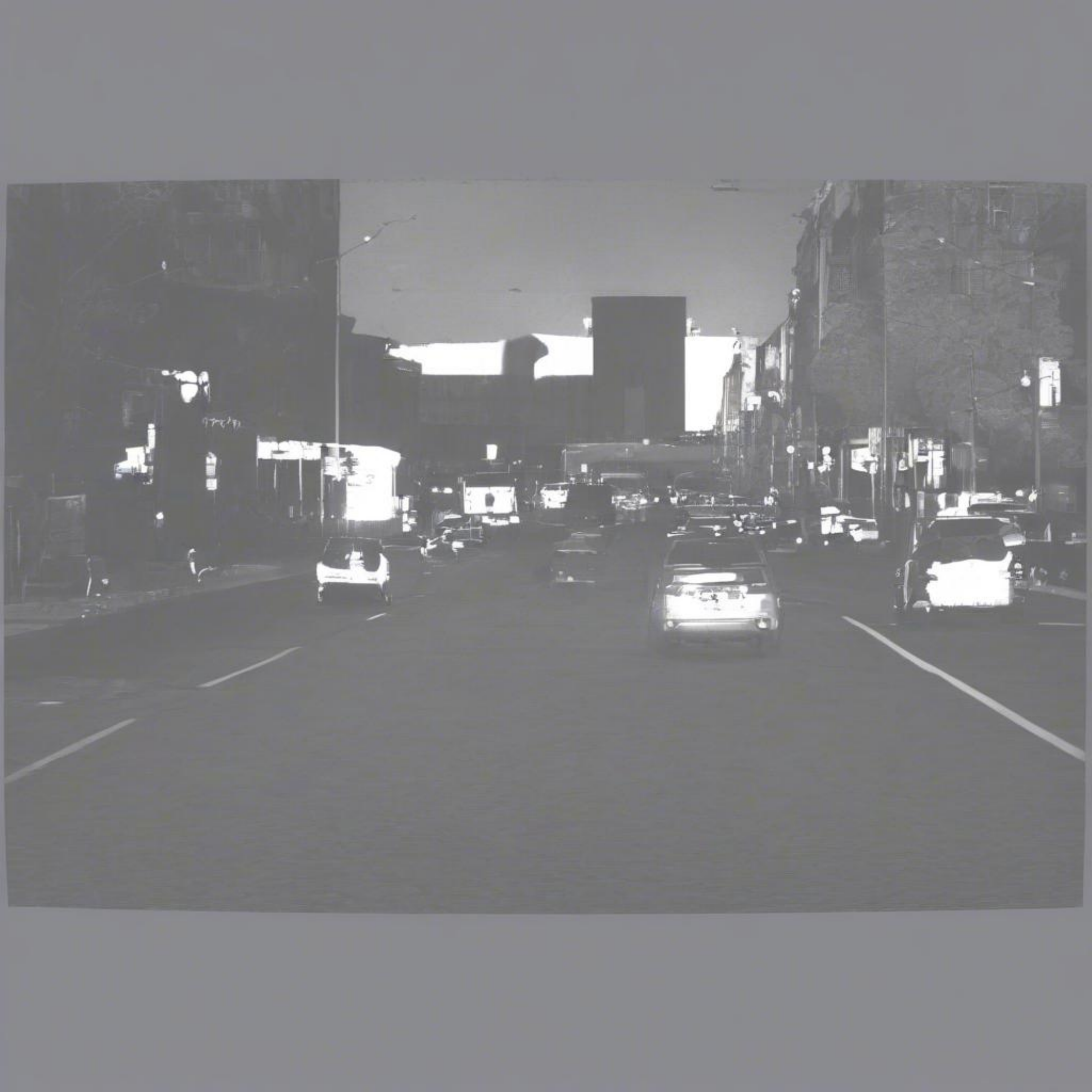}
& \includegraphics[width=1.7cm,height=1.7cm]{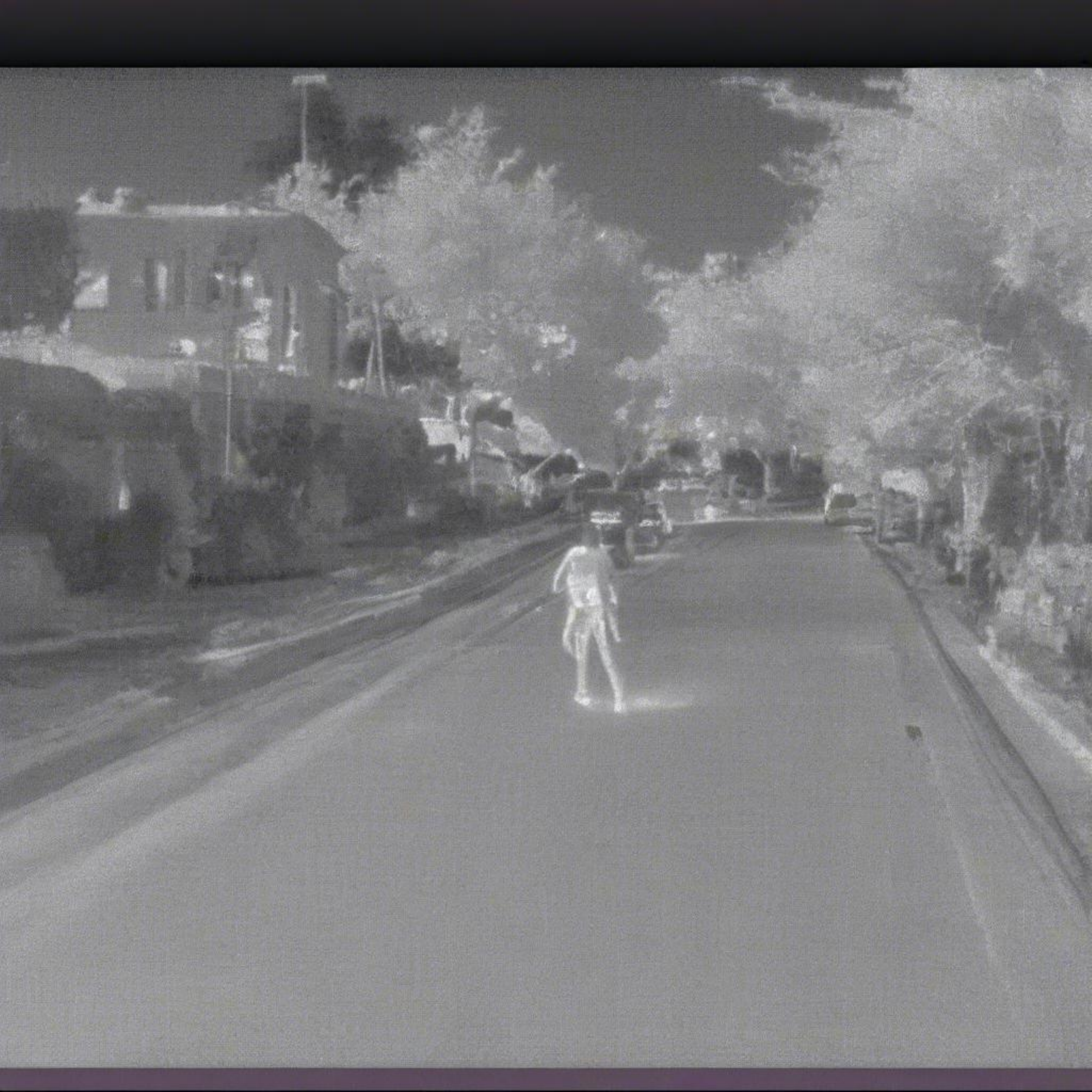}
& \includegraphics[width=1.7cm,height=1.7cm]{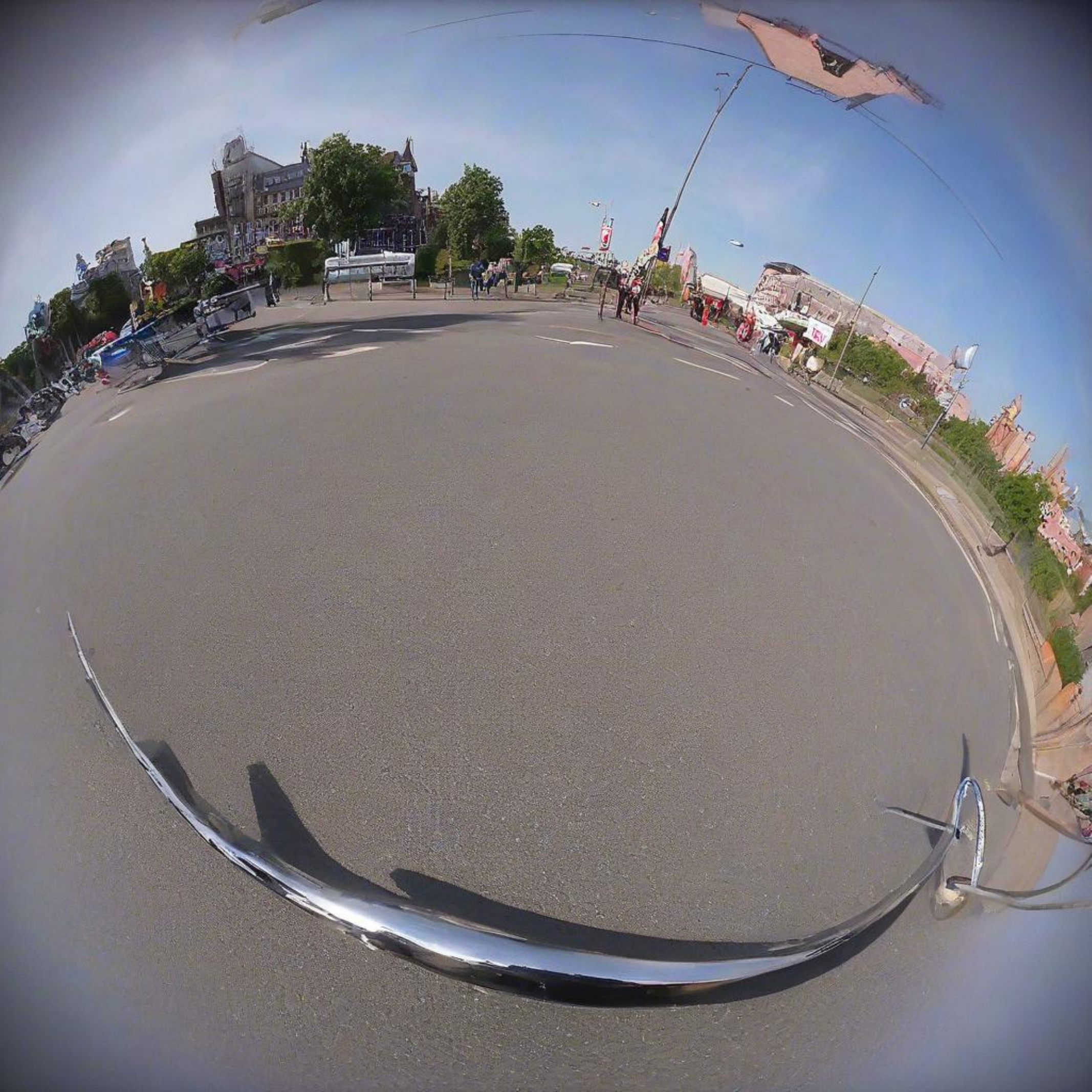}
& \includegraphics[width=1.7cm,height=1.7cm]{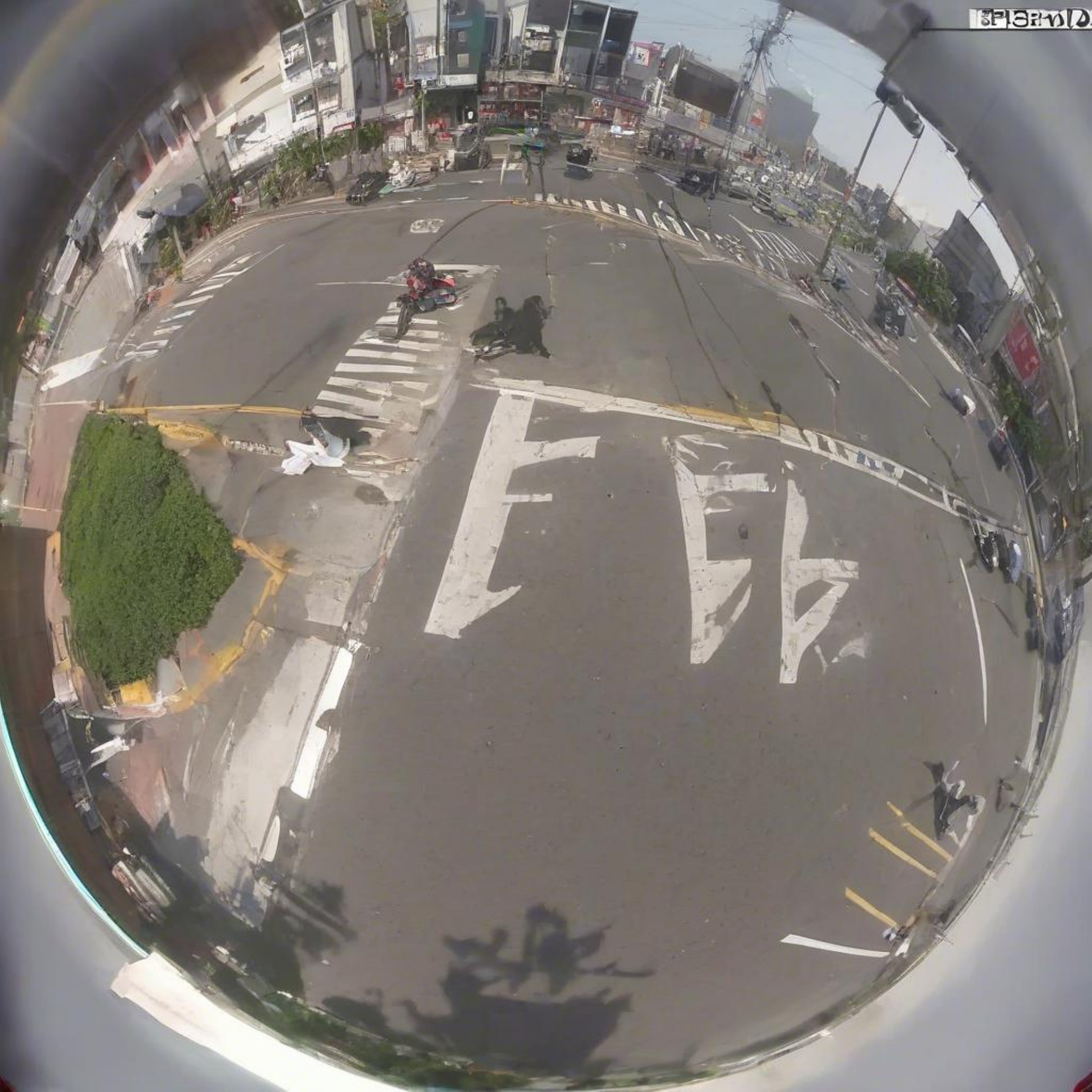} \\

\bottomrule
\end{tabular}}
\label{tab:qual_existing}
\end{table*}

\begin{table*}[t]
\centering
\caption{\textbf{Quantitative comparison on existing factor combinations.} Best in bold. X-MULTI (ours) achieves the highest I-FAA metric while maintaining highly competitive underlying image generation quality scores.}
\scriptsize
\setlength{\tabcolsep}{3.2pt}
\renewcommand{\arraystretch}{1.08}
\resizebox{\textwidth}{!}{
\begin{tabular}{lcccccccccccc}
\toprule
\multirow{3}{*}{Method}
& \multicolumn{7}{c}{Image Generation Metrics}
& \multicolumn{5}{c}{Improved-Factor Alignment Accuracy (I-FAA)} \\
\cmidrule(lr){2-8} \cmidrule(lr){9-13}
& \multirow{2}{*}{FID~\cite{heusel2017gans} $\downarrow$}
& \multirow{2}{*}{IS~\cite{salimans2016improved} $\uparrow$}
& \multicolumn{4}{c}{CLIP Score~\cite{hessel2021clipscore} $\uparrow$}
& \multirow{2}{*}{DS~\cite{zhang2018unreasonable} $\uparrow$}
& \multirow{2}{*}{Lens $\uparrow$}
& \multirow{2}{*}{Sensor $\uparrow$}
& \multirow{2}{*}{Domain $\uparrow$}
& \multirow{2}{*}{View $\uparrow$}
& \multirow{2}{*}{Avg. $\uparrow$} \\
\cmidrule(lr){4-7}
& & & factor & context & average & full-prompt & & & & & & \\
\midrule

DreamBooth~\cite{ruiz2023dreambooth} & 58.27 & 5.02 & 24.52 & 23.75 & 24.64 & 24.86 & 0.57 & \textbf{1.00} & \textbf{0.91} & \textbf{0.96} & 0.80 & 0.91 \\
SDXL Zeroshot~\cite{podell2023sdxl} & 60.19 & 4.36 & 20.70 & 24.68 & 21.19 & 25.13 & \textbf{0.59} & 0.65 & 0.71 & 0.65 & 0.36 & 0.59 \\
Inspiration Tree~\cite{vinker2023concept} & 55.35 & \textbf{5.08} & 19.78 & 21.89 & 20.84 & 23.78 & \textbf{0.59} & 0.60 & 0.64 & 0.68 & 0.42 & 0.58 \\
MULTI~\cite{godavarthy2026multi} & 60.13 & 4.68 & 23.82 & 24.77 & 24.29 & 27.30 & 0.56 & 0.98 & 0.90 & 0.94 & 0.82 & 0.91 \\
X-MULTI (ours) & \textbf{54.48} & 4.90 & \textbf{24.37} & \textbf{24.97} & \textbf{24.66} & \textbf{28.08} & \textbf{0.59} & \textbf{1.00} & \textbf{0.91} & \textbf{0.96} & \textbf{0.82} & \textbf{0.92} \\

\midrule
DreamBooth-Canny & 58.39 & 4.54 & 24.13 & 25.64 & 25.39 & 28.27 
& 0.57 & 0.99 & \textbf{0.92} & \textbf{0.98} & 0.84 & 0.93 \\
SDXL Zeroshot-Canny & 63.24 & 4.76 & 23.96 & \textbf{27.34} & 25.65 & 28.92 & \textbf{0.59} & \textbf{1.00} & 0.73 & 0.93 & 0.76 & 0.85 \\
Inspiration Tree-Canny & 48.45 & \textbf{5.21} & 20.81 & 26.80 & 23.80 & 29.53 & \textbf{0.59} & 0.95 & 0.78 & 0.81 & 0.75 & 0.82 \\
MULTI-Canny & \textbf{45.39} & 4.69 & 24.64 & 25.90 & 25.27 & 28.33 & 0.55 & 0.99 & 0.91 & 0.97 & 0.85 & 0.93 \\
X-MULTI-Canny (ours) & 47.72 & 4.67 & \textbf{24.81} & 26.52 & \textbf{25.67} & \textbf{29.68} & 0.57 & \textbf{1.00} & 0.91 & \textbf{0.98} & \textbf{0.85} & \textbf{0.94} \\

\midrule
DreamBooth-Depth & 59.35 & 4.87 & 24.06 & 25.17 & 25.11 & 27.18 & 0.57 & \textbf{1.00} & \textbf{0.92} & \textbf{0.97} & 0.82 & 0.92 \\
SDXL Zeroshot-Depth & 65.39 & 4.75 & 23.59 & \textbf{27.42} & \textbf{25.50} & 28.23 & \textbf{0.59} & \textbf{1.00} & 0.75 & 0.92 & 0.73 & 0.85 \\
Inspiration Tree-Depth & \textbf{55.15} & \textbf{5.15} & 20.62 & 26.38 & 23.50 & 28.20 & \textbf{0.59} & 0.95 & 0.79 & 0.81 & 0.75 & 0.82 \\
MULTI-Depth & 60.60 & 4.79 & 24.38 & 25.75 & 25.06 & 27.93 & 0.55 & 0.99 & \textbf{0.92} & \textbf{0.97} & \textbf{0.85} & \textbf{0.93} \\
X-MULTI-Depth (ours) & 55.58 & 4.66 & \textbf{24.63} & 26.13 & 25.38 & \textbf{28.81} & 0.53 & \textbf{1.00} & \textbf{0.92} & \textbf{0.97} & \textbf{0.85} & \textbf{0.93} \\

\bottomrule
\end{tabular}}
\label{tab:existing_quantitative}
\end{table*}


Having evaluated X-MULTI on the core objective of novel factor combinations, this section provides an extended evaluation of X-MULTI on existing factor combinations that are combinations present in the training data to demonstrate that our semantic supervision does not compromise performance on the training distributions. 

Table~\ref{tab:qual_existing} demonstrates that when testing existing factor combinations, SDXL Zeroshot struggled to represent imaging factors accurately. While MULTI, DreamBooth, and Inspiration Tree showed competence, our method excelled at generating images that most faithfully matched their source datasets. Notably, X-MULTI captured specialized sensors like gated, rgb-thermal, and event well, yet our approach's representations were superior.
Table~\ref{tab:existing_quantitative} compares baseline methods across image generation metrics for existing combinations. The results show that our method has best I-FAA, indicating that it achieves good factor disentanglement, while it performs comparable across other metrics.

\section{Augmentations of Factor Classifiers for I-FAA}
To prevent the factor classifiers from relying on spurious correlation shortcuts inherent in the DF-RICO benchmark, we enforce strict, non-overlapping boundaries for the data transformations across each individual head. Table~\ref{tab:augmentation_hyperparameters} details the exact operational parameter bounds utilized during training. For instance, color deviations are completely restricted from the Sensor classifier to ensure it focuses strictly on physical noise distributions and exposure characteristics, while geometric radial limits are maximized on the Lens classifier to highlight fisheye edge-distortions without confounding viewpoint perspectives.

\begin{table*}[t]
\centering
\caption{\textbf{Detailed operational parameter ranges for factor classifiers of I-FAA (ours).}}
\label{tab:augmentation_hyperparameters}
\tiny
\setlength{\tabcolsep}{2pt}
\renewcommand{\arraystretch}{1.2}
\begin{tabularx}{\textwidth}{l p{2.8cm} X}
\toprule
\textbf{Factor} & \textbf{Augmentation Operator} & \textbf{Exact Transformation} \\
\midrule
\multirow{6}{*}{\textbf{Lens}} 
  & Random Rotation & Max $\pm 5^\circ$ (Probability $p=0.5$) \\
  & Random Affine Translation & Max $\pm 3\%$ horizontally/vertically ($p=0.5$) \\
  & Color Jittering & Brightness/Contrast/Saturation: $\pm 0.2$, Hue: $\pm 0.1$ ($p=0.5$) \\
  & Grayscale Conversion & Probability $p=0.1$ \\
  & Gaussian Blur & Kernel Size: $5 \times 5$, Sigma Range: $[0.1, 2.0]$ ($p=0.5$) \\
  & Noise \& Occlusions & Gaussian Noise (Std: $0.05$), Random Erasing ($p=0.1$), Cutout ($p=0.1$), HideAndSeek ($6\times6$ grid, $p=0.1$) \\
\midrule
\multirow{5}{*}{\textbf{Sensor}} 
  & Random Perspective & Distortion Scale: $0.35$ (Probability $p=0.4$) \\
  & Optical Distortion & Distortion Limit: $\pm 0.12$, Shift Limit: $\pm 0.03$ ($p=0.3$, applied inside Albumentations transform with $p=0.5$) \\
  & Spatial Adjustments & Rotation Max: $\pm 5^\circ$, Affine Translation Max: $\pm 10\%$ ($p=0.5$) \\
  & Spatial Cropping & Cropped to square via \texttt{RandomCropToSquare} \\
  & Occlusions \& Erasures & Cutout Length: $6$\:px ($p=0.05$), Random Erasing Scale: $[0.005, 0.03]$ ($p=0.05$), HideAndSeek ($8\times8$ grid, $p=0.05$) \\
\midrule
\multirow{5}{*}{\textbf{Domain}} 
  & Color Modification & Brightness/Contrast/Saturation: $\pm 0.2$, Hue: $\pm 0.1$ ($p=0.5$) \\
  & Grayscale Conversion & Probability $p=0.1$ \\
  & Gaussian Blur & Heavy Blur (Kernel Size: $7 \times 7$, Sigma Range: $[1.0, 3.0]$, $p=0.5$) \\
  & Geometric Distortions & Perspective Scale: $0.35$ ($p=0.4$), Albumentations Optical Distortion: $\pm 0.12$ ($p=0.3$) \\
  & Spatial Scaling \& Noise & Affine Translation: $\pm 10\%$, Gaussian Noise Std: $0.05$ ($p=0.5$), Random Erasing ($p=0.1$), Cutout ($p=0.1$), HideAndSeek ($p=0.1$) \\
\midrule
\multirow{4}{*}{\textbf{Viewpoint}} 
  & Horizontal Flipping & Restricted to conservative probability $p=0.1$ \\
  & Random Rotation \& Affine & Rotation Max: $\pm 5^\circ$, Affine Translation Max: $\pm 3\%$ ($p=0.5$) \\
  & Color Modification & Brightness/Contrast/Saturation: $\pm 0.2$, Hue: $\pm 0.1$ ($p=0.5$), Grayscale ($p=0.1$) \\
  & Spatial Blurring \& Distortion & Gaussian Blur ($p=0.2$, Sigma: $[0.1, 1.0]$), Albumentations Optical Distortion: $\pm 0.03$, Shift Limit: $\pm 0.01$ ($p=0.15$), Gaussian Noise Std: $0.05$ ($p=0.5$) \\
\bottomrule
\end{tabularx}
\end{table*}

\section{Additional Correlation Analysis on FAA and I-FAA}

\begin{figure*}[t]
    \centering
    \includegraphics[width=\textwidth]{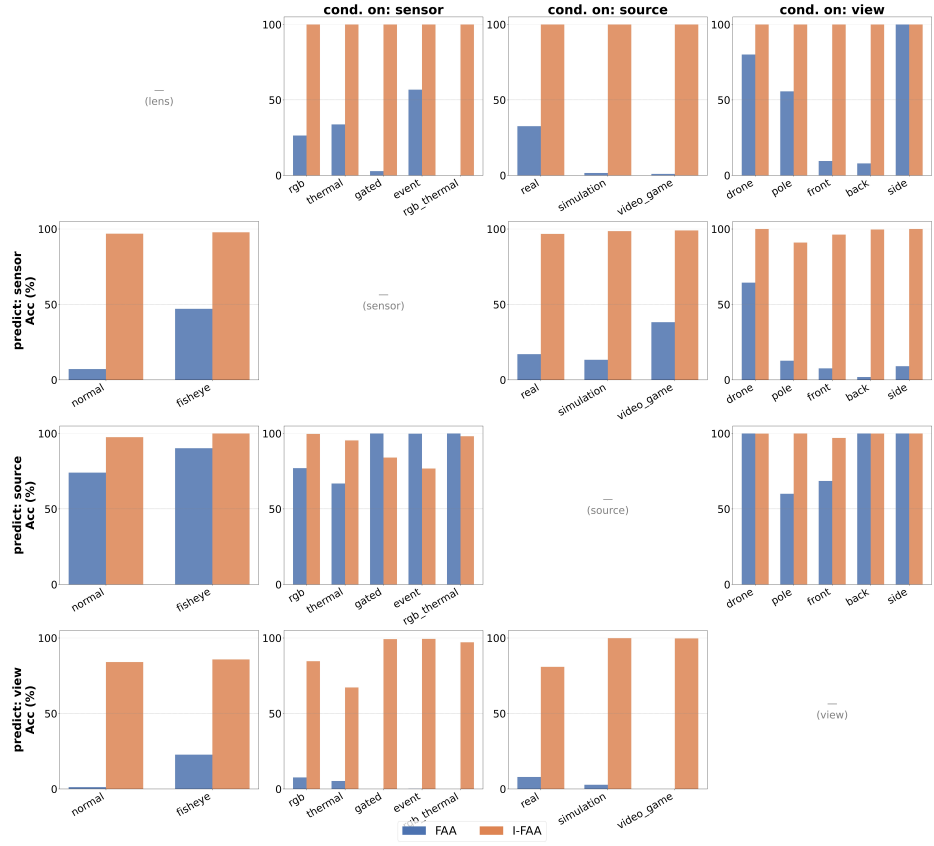}
    \caption{\textbf{Cross-factor accuracy variance analysis.} Individual factor prediction accuracies evaluated under ground-truth class conditioning. I-FAA (ours) has less cross-factor variance.}
    \label{fig:cross_factor_accuracy}
\end{figure*}

\begin{figure*}[t]
    \centering
    \includegraphics[width=\textwidth]{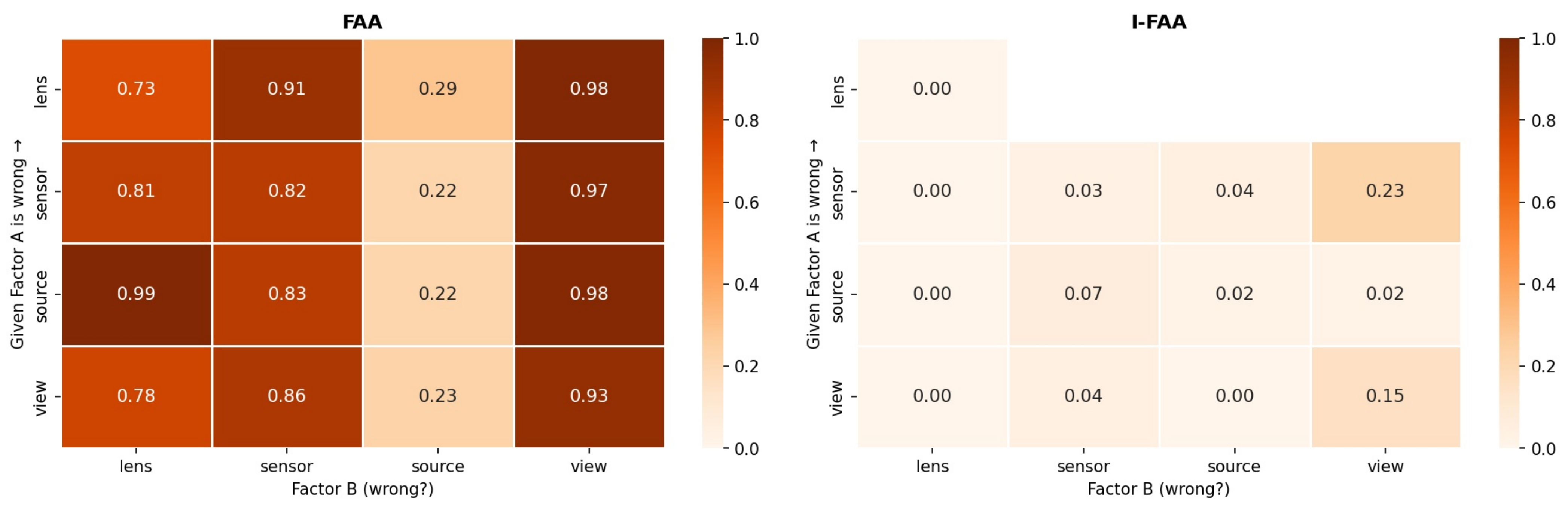}
    \caption{\textbf{Error co-occurrence patterns.} Conditional joint error probabilities map the independence of classification failures, showing I-FAA (ours) has low co-occurrence.}
    \label{fig:error_cooccurrence}
\end{figure*}

To provide a deeper diagnostic evaluation of the structural limitations of the baseline Factor Alignment Accuracy (FAA) and validate the independence of our proposed Improved-FAA (I-FAA) metric, we conduct an extended correlation analysis across two distinct paradigms. Specifically, we examine the resilience of the classifiers against structural domain changes via cross-factor accuracy variance, and inspect the independence of prediction failures through error co-occurrence patterns.

\subsection{Cross-factor accuracy variance} 
Figure~\ref{fig:cross_factor_accuracy} shows prediction accuracy of one factor conditioned on the ground-truth class of another. I-FAA maintains near-100\% accuracy regardless of conditioning class, while FAA degrades sharply. Lens accuracy conditioned on viewpoint ranges from $\sim$0\% to 75\% in FAA, yet stays at 100\% under I-FAA. Similar instability appears for viewpoint and sensor, where FAA drops below 5\% for certain classes. FAA shows standard deviations as high as 37.0\%.

\subsection{Error co-occurrence} Figure~\ref{fig:error_cooccurrence} shows the conditional probability 
$P(\text{factor } B \text{ wrong} \mid \text{factor } A \text{ wrong})$, 
revealing whether prediction failures are shared or isolated. FAA prediction failures are almost never isolated: $P(\text{view wrong} \mid \text{lens wrong}) = 0.98$, $P(\text{sensor wrong} \mid \text{lens wrong}) = 0.91$, and $P(\text{lens wrong} \mid \text{domain wrong}) = 0.99$, indicating that factor predictions are not independent. I-FAA errors are nearly fully isolated, with all probabilities at or near 0.

\begin{table}[t]
\centering
\caption{\textbf{Extended qualitative comparison of attention maps between FAA and I-FAA.}}
\label{tab:attention_additional}
\scriptsize
\setlength{\tabcolsep}{4pt} 
\renewcommand{\arraystretch}{0.5}

\begin{tabular}{cccccc}
\toprule
\textbf{Image} & \textbf{Metric} & \textbf{Lens} & \textbf{Sensor} & \textbf{Domain} & \textbf{View} \\
\midrule

\multirow{2}{*}{\scalebox{1.0}{\begin{tabular}[c]{@{}c@{}}\includegraphics[width=0.090\textwidth]{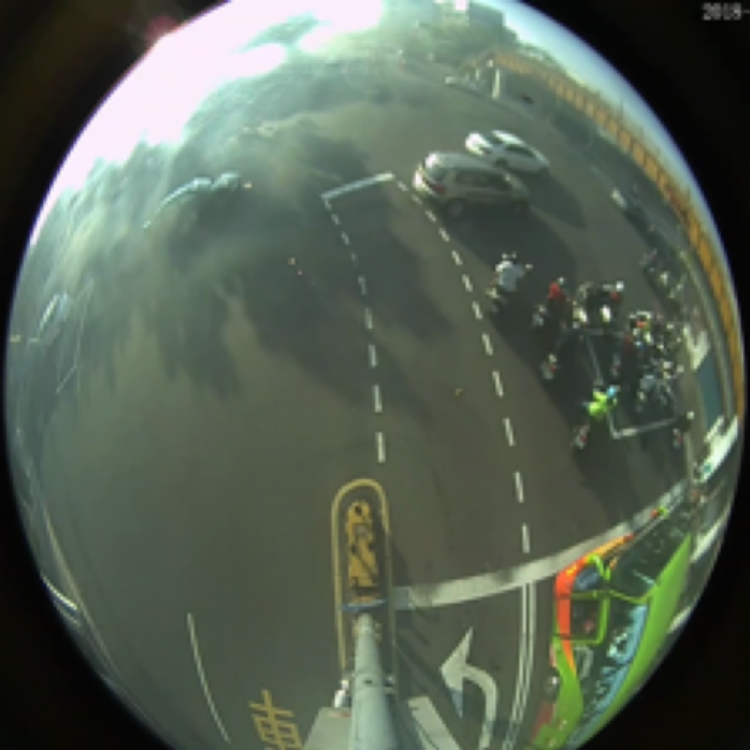}\\[-0.2ex]{\tiny fisheye/rgb/real/pole}\end{tabular}}}
& \textbf{FAA}~\cite{godavarthy2026multi}
& \includegraphics[width=0.075\textwidth]{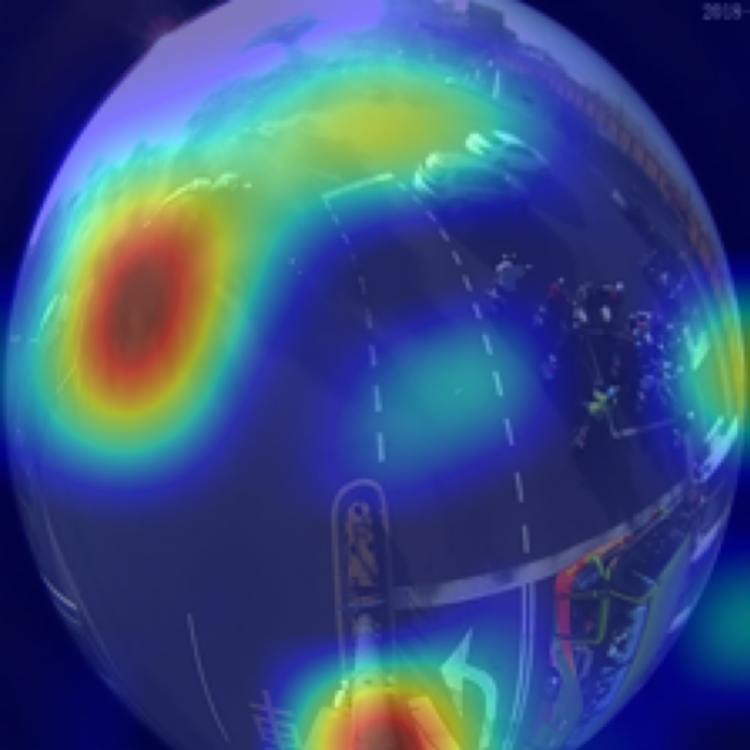}
& \includegraphics[width=0.075\textwidth]{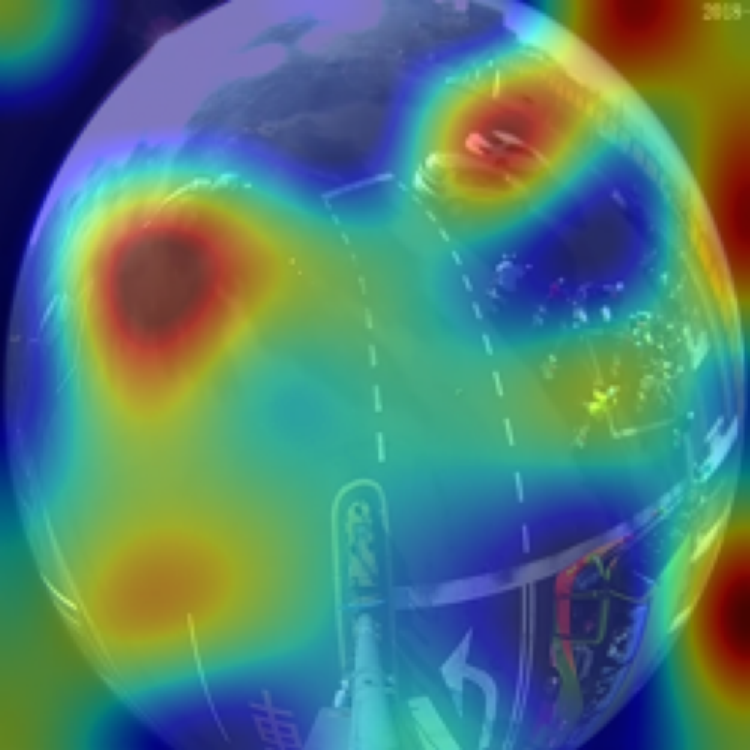}
& \includegraphics[width=0.075\textwidth]{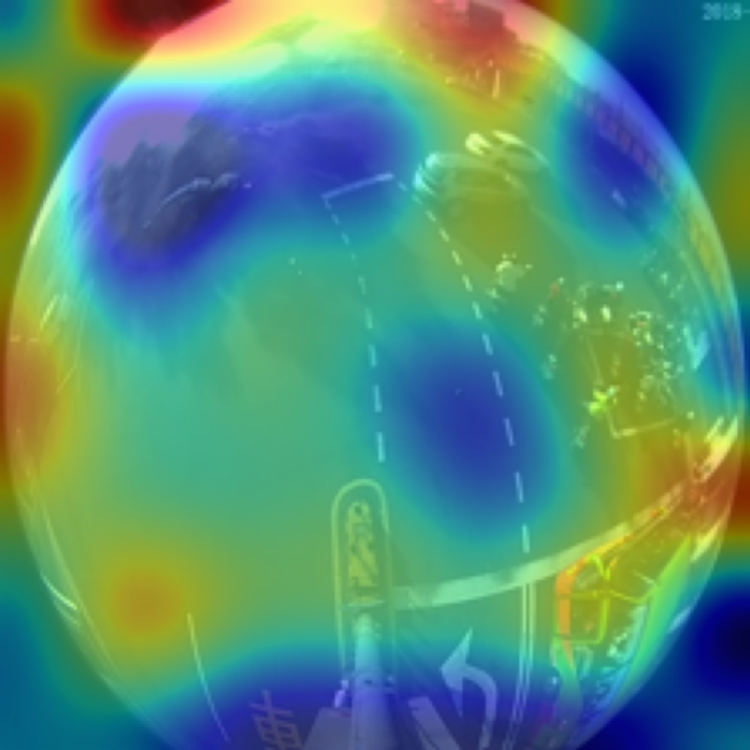}
& \includegraphics[width=0.075\textwidth]{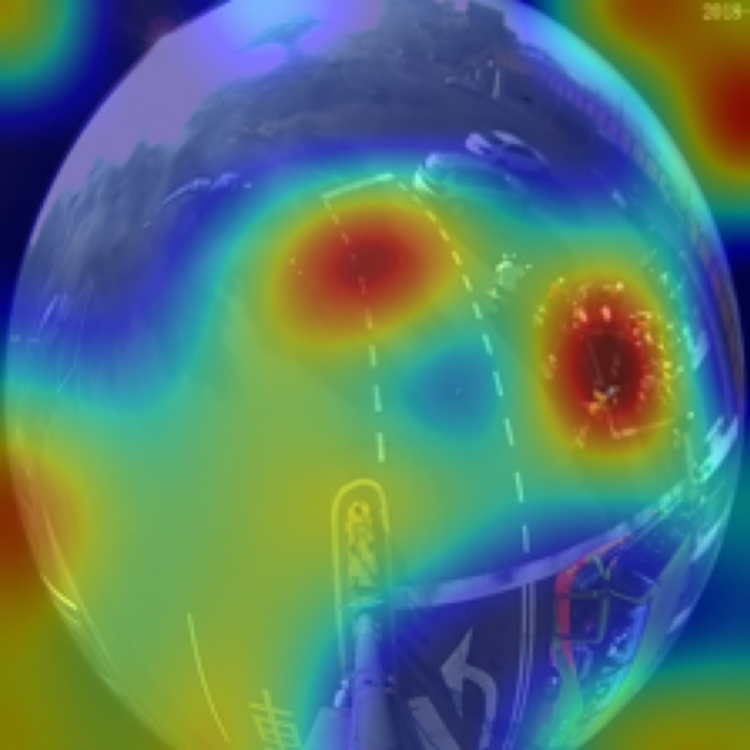} \\
\cmidrule(lr){2-6}
& \textbf{I-FAA}
& \includegraphics[width=0.075\textwidth]{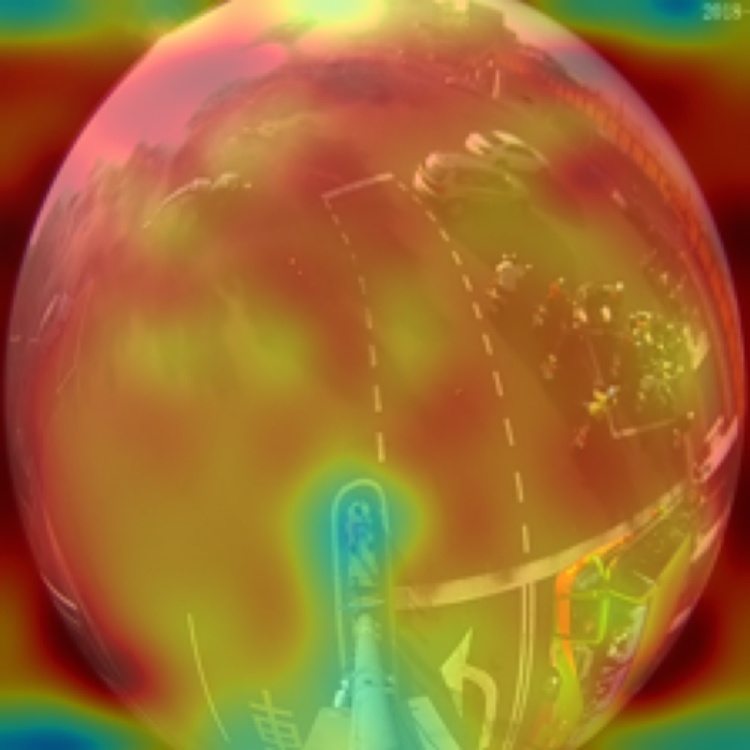}
& \includegraphics[width=0.075\textwidth]{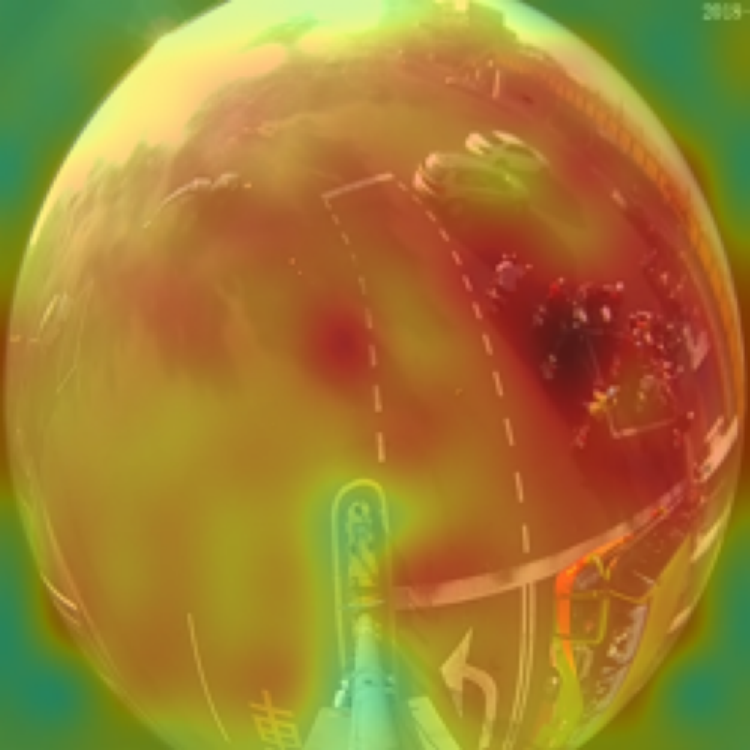}
& \includegraphics[width=0.075\textwidth]{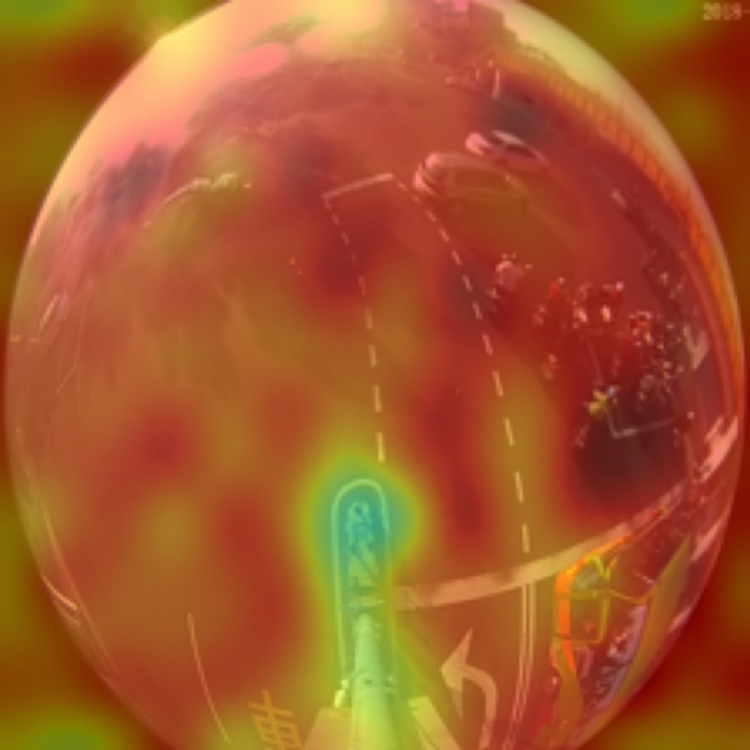}
& \includegraphics[width=0.075\textwidth]{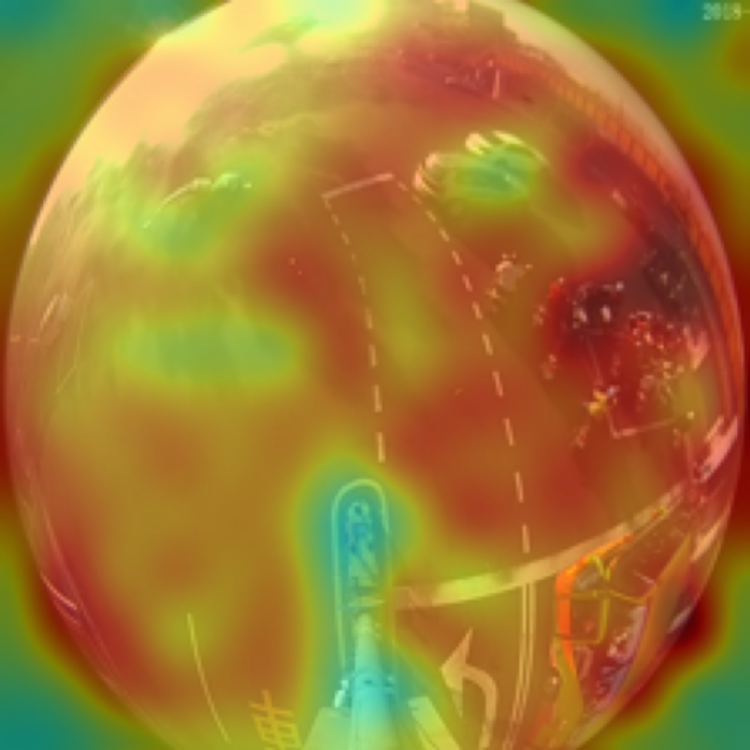} \\

\midrule

\multirow{2}{*}{\scalebox{1.0}{\begin{tabular}[c]{@{}c@{}}\includegraphics[width=0.090\textwidth]{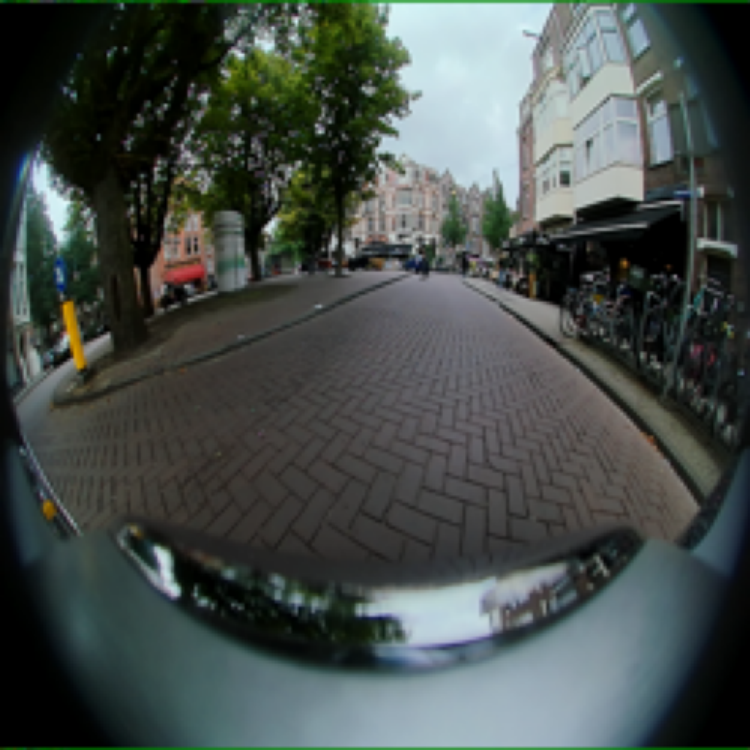}\\[-0.2ex]{\tiny fisheye/rgb/real/front}\end{tabular}}}
& \textbf{FAA}~\cite{godavarthy2026multi}
& \includegraphics[width=0.075\textwidth]{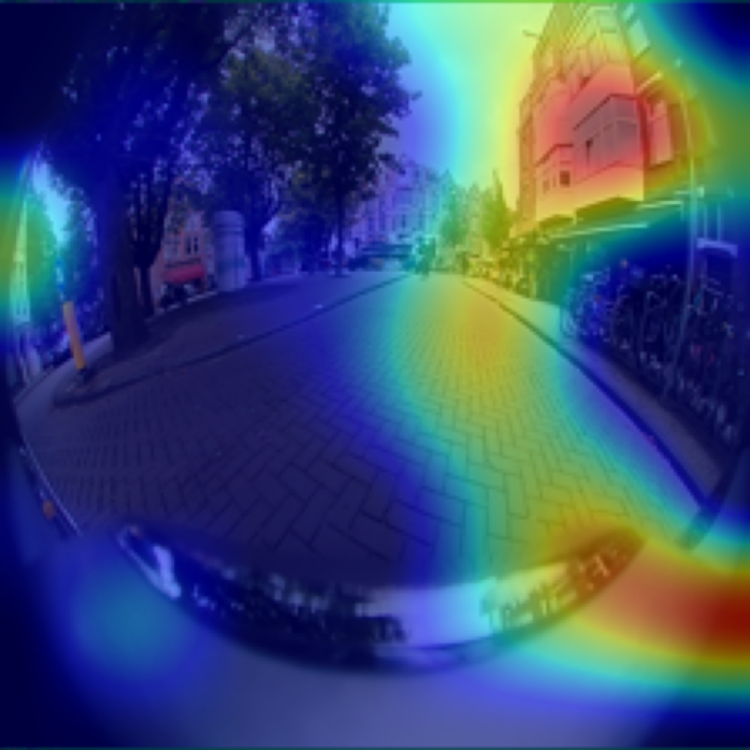}
& \includegraphics[width=0.075\textwidth]{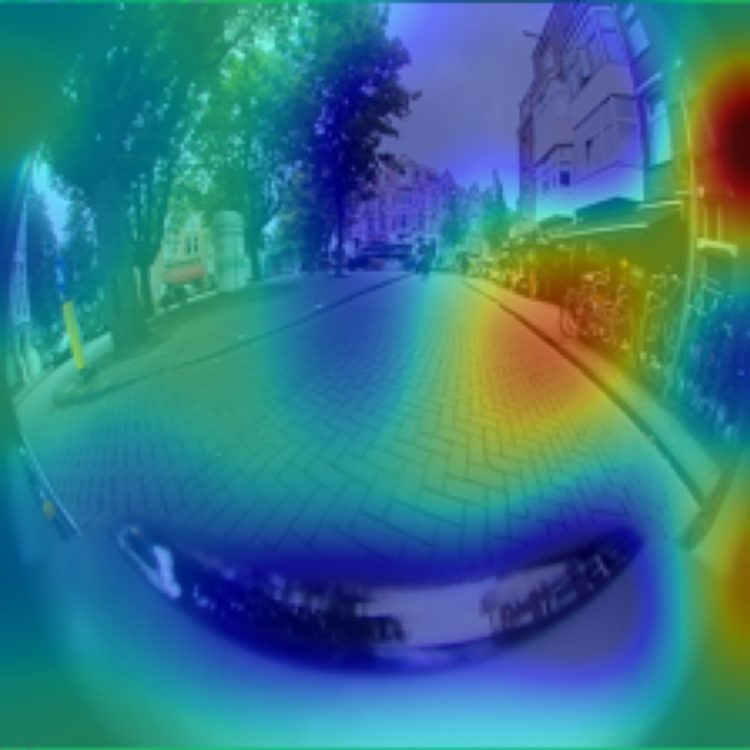}
& \includegraphics[width=0.075\textwidth]{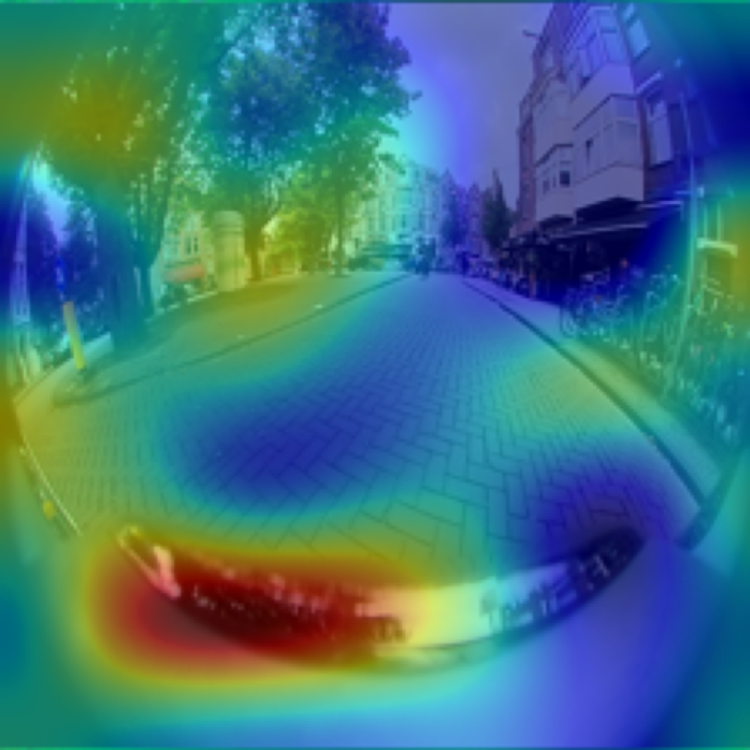}
& \includegraphics[width=0.075\textwidth]{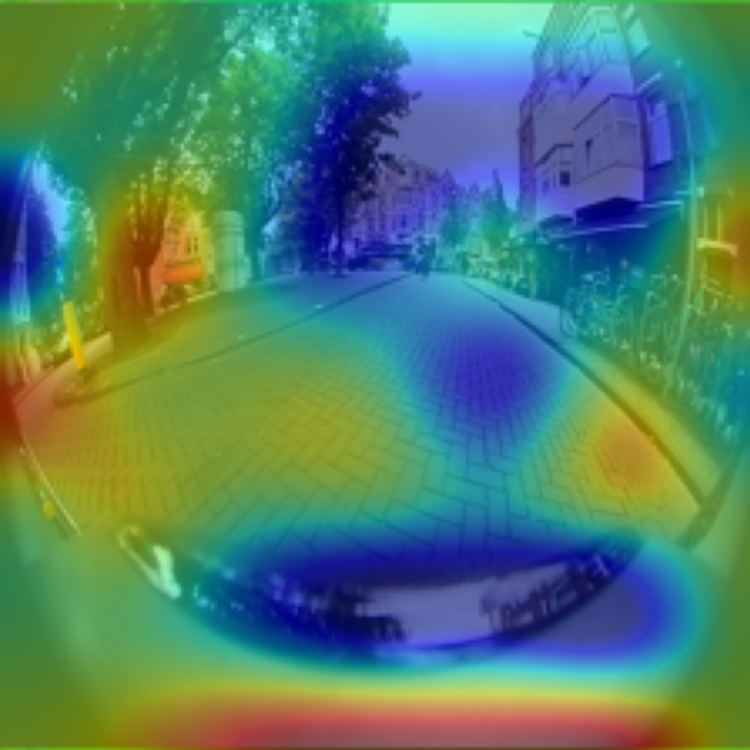} \\
\cmidrule(lr){2-6}
& \textbf{I-FAA}
& \includegraphics[width=0.075\textwidth]{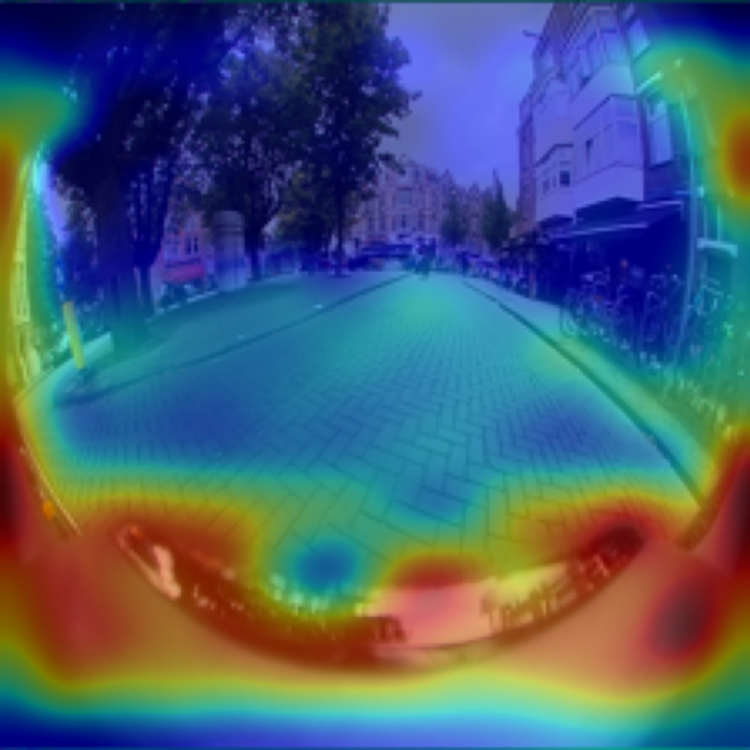}
& \includegraphics[width=0.075\textwidth]{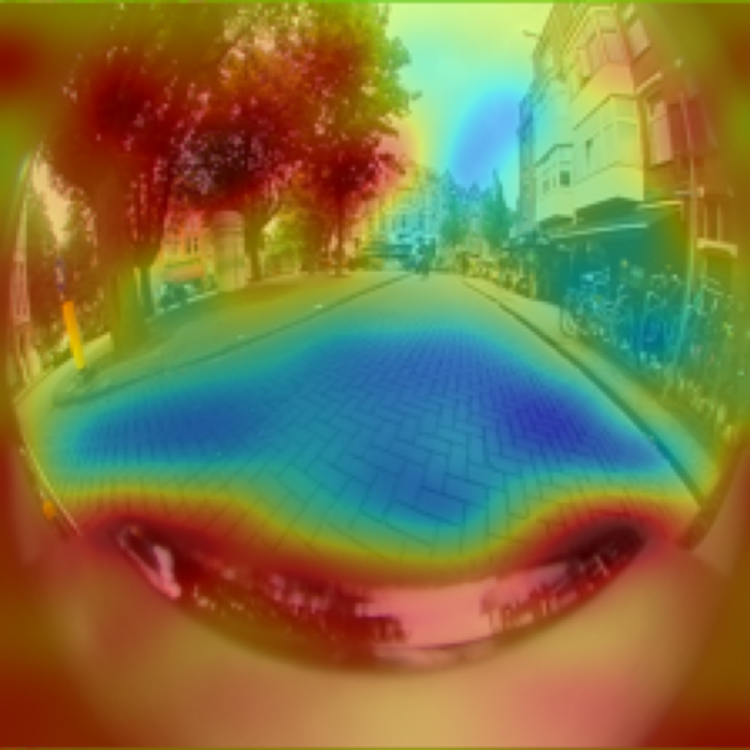}
& \includegraphics[width=0.075\textwidth]{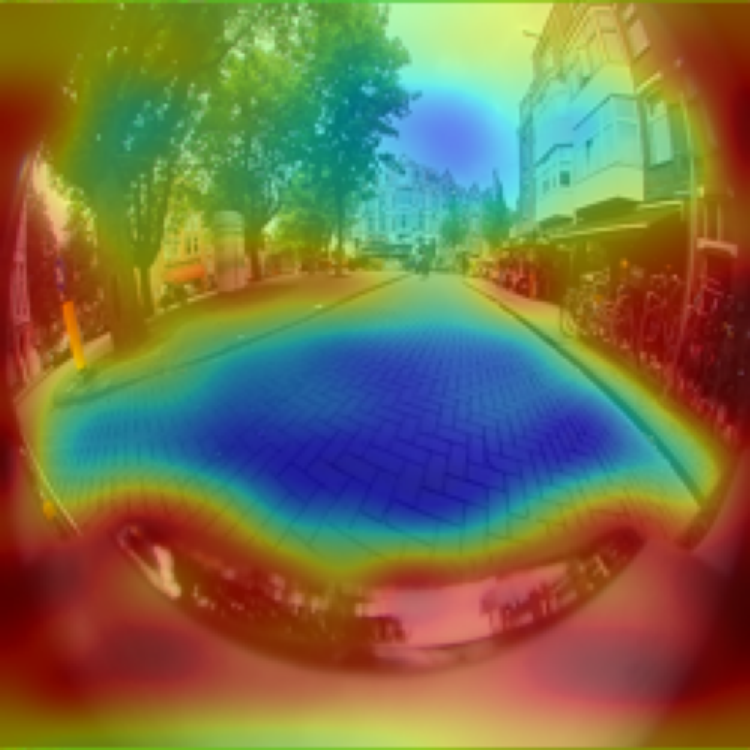}
& \includegraphics[width=0.075\textwidth]{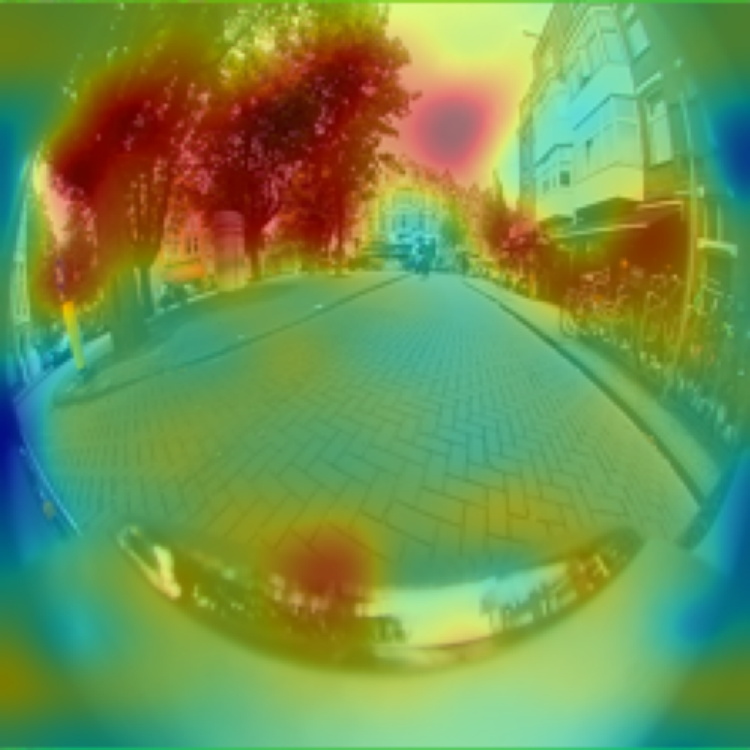} \\

\midrule

\multirow{2}{*}{\scalebox{1.0}{\begin{tabular}[c]{@{}c@{}}\includegraphics[width=0.090\textwidth]{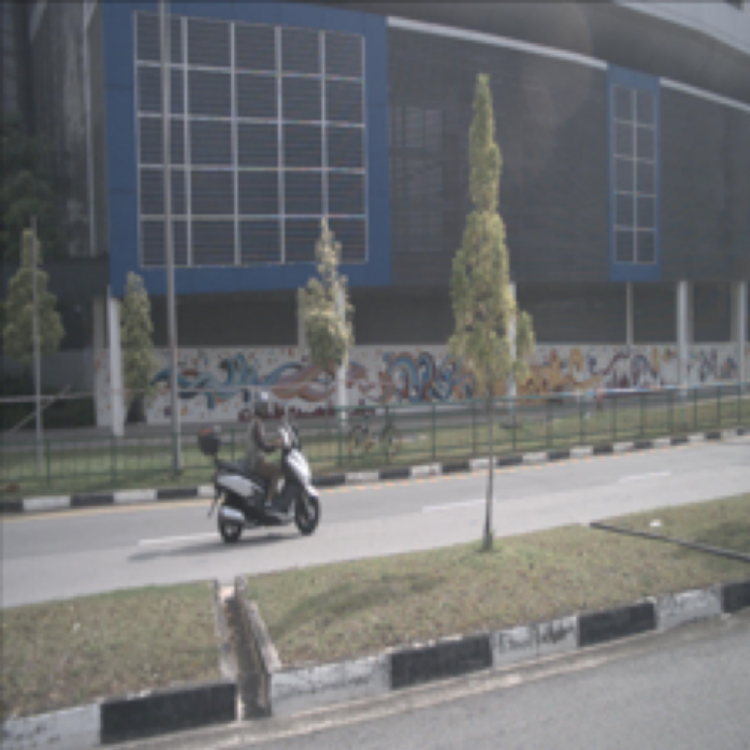}\\[-0.2ex]{\tiny normal/rgb/real/front}\end{tabular}}}
& \textbf{FAA}~\cite{godavarthy2026multi}
& \includegraphics[width=0.075\textwidth]{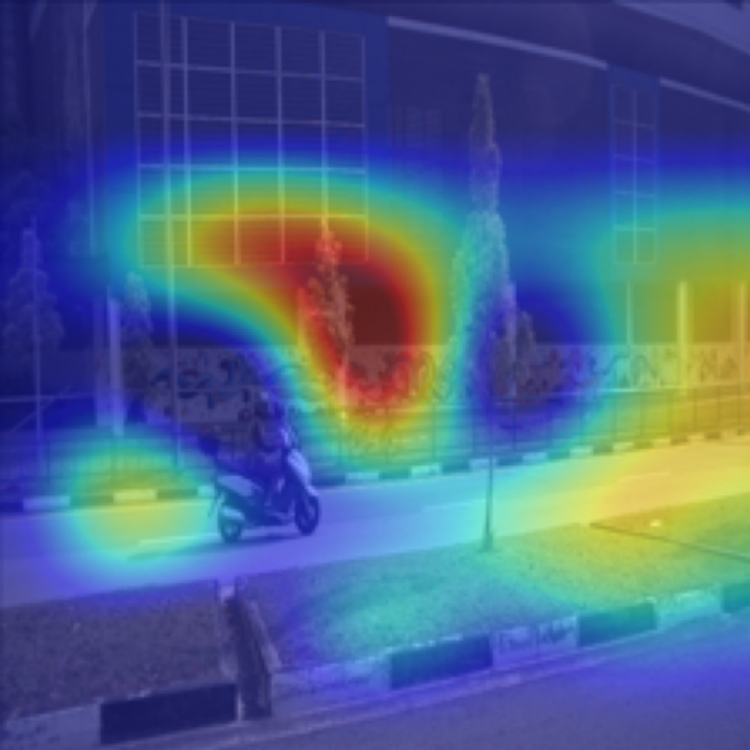}
& \includegraphics[width=0.075\textwidth]{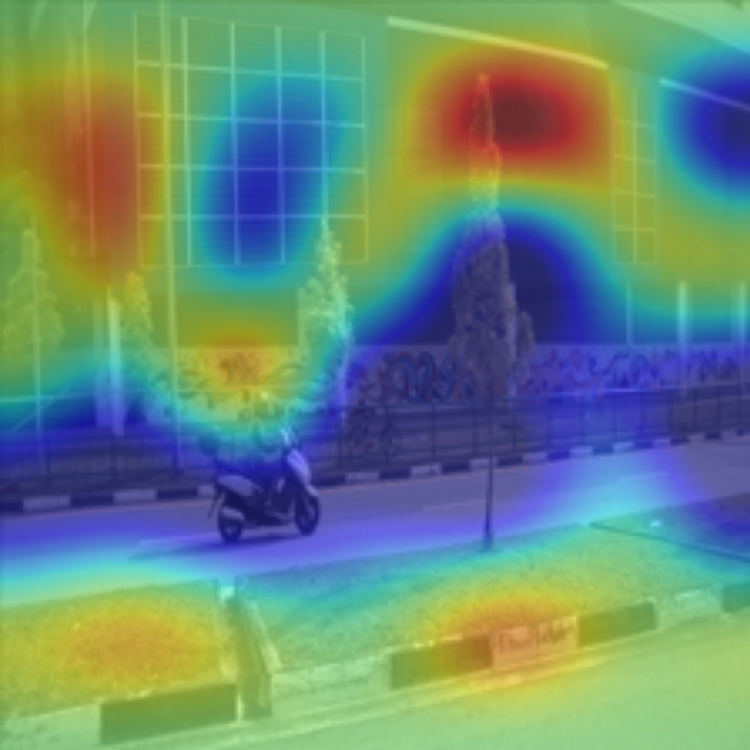}
& \includegraphics[width=0.075\textwidth]{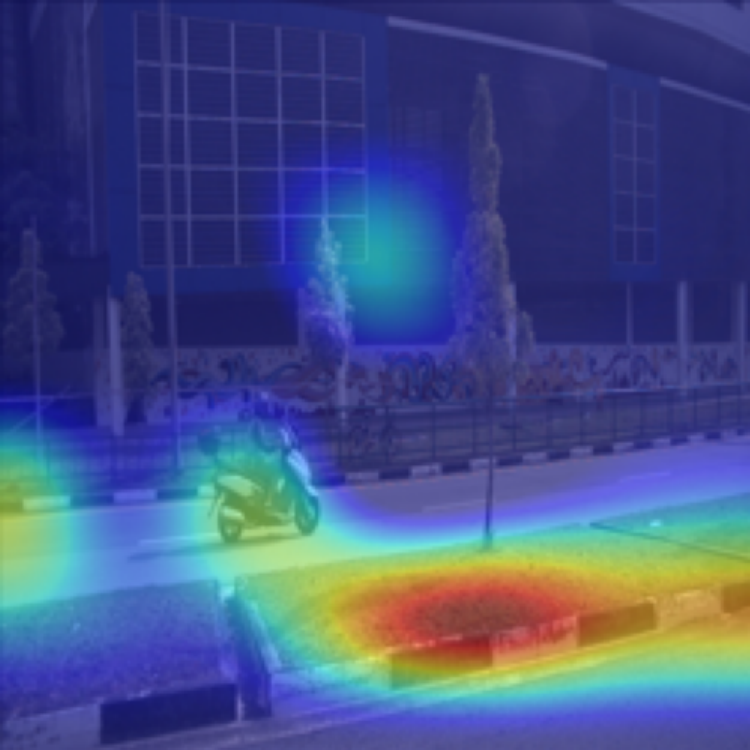}
& \includegraphics[width=0.075\textwidth]{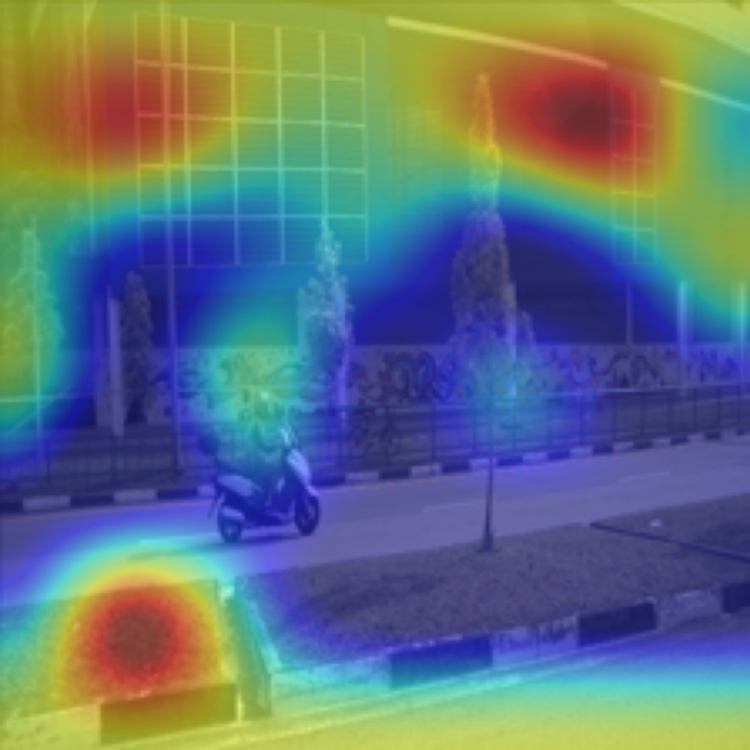} \\
\cmidrule(lr){2-6}
& \textbf{I-FAA}
& \includegraphics[width=0.075\textwidth]{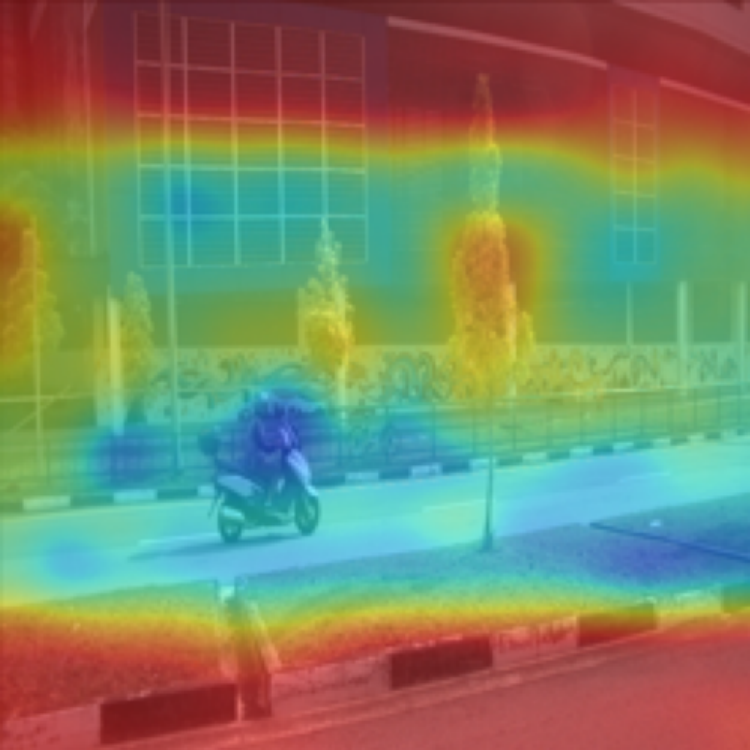}
& \includegraphics[width=0.075\textwidth]{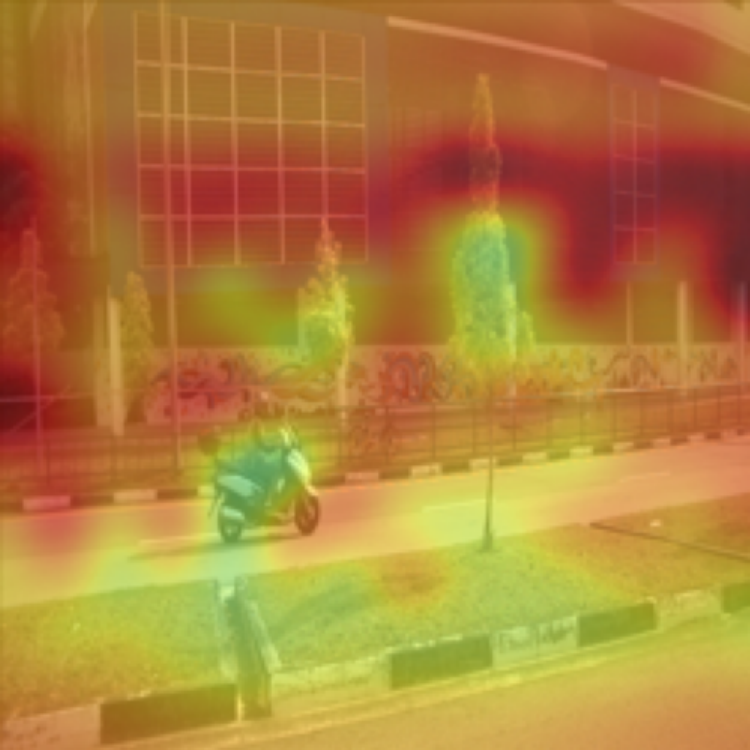}
& \includegraphics[width=0.075\textwidth]{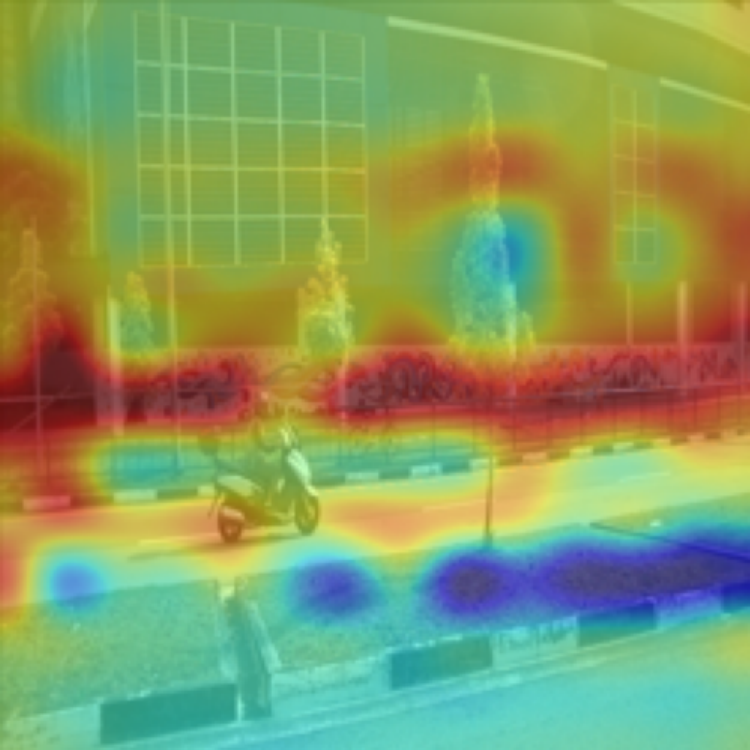}
& \includegraphics[width=0.075\textwidth]{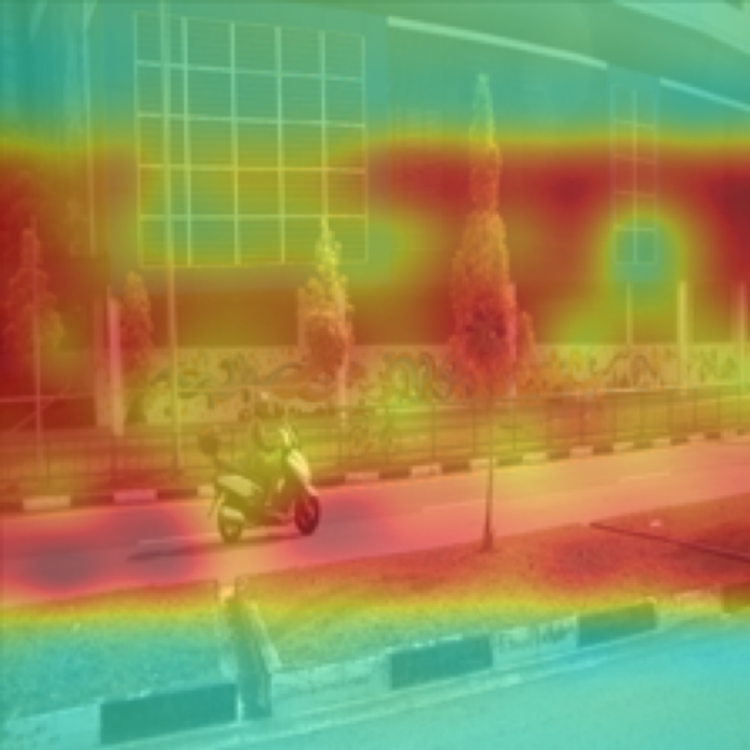} \\

\midrule

\multirow{2}{*}{\scalebox{1.0}{\begin{tabular}[c]{@{}c@{}}\includegraphics[width=0.090\textwidth]{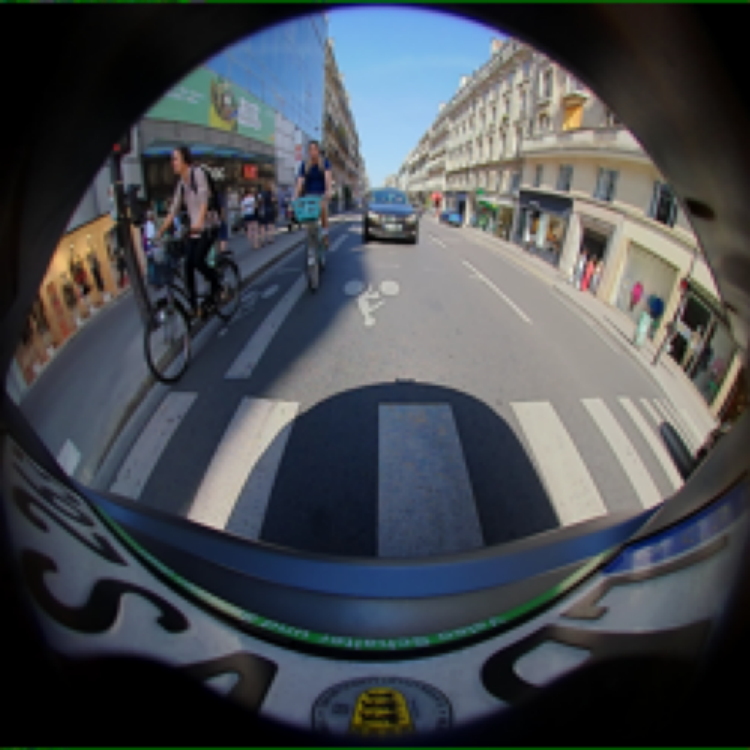}\\[-0.2ex]{\tiny fisheye/rgb/real/back}\end{tabular}}}
& \textbf{FAA}~\cite{godavarthy2026multi}
& \includegraphics[width=0.075\textwidth]{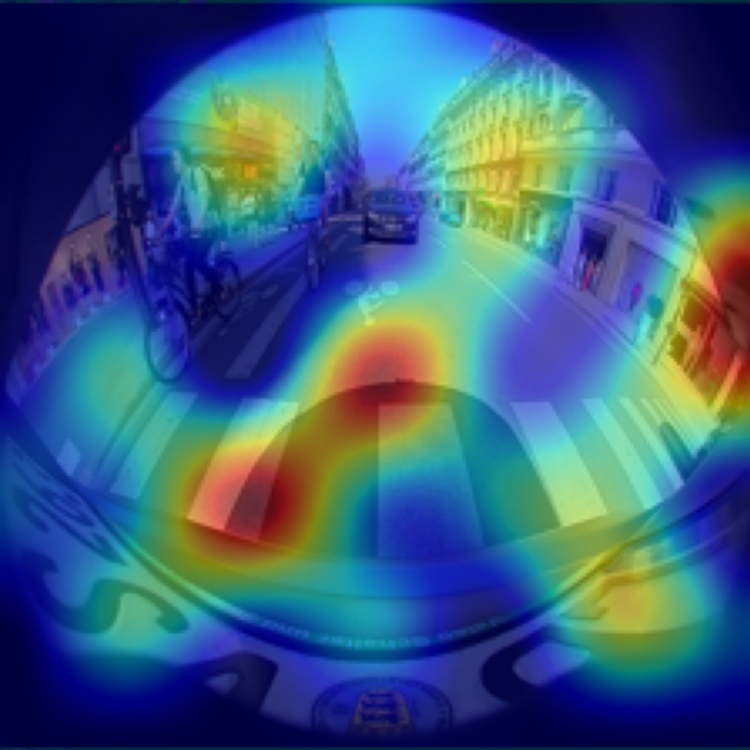}
& \includegraphics[width=0.075\textwidth]{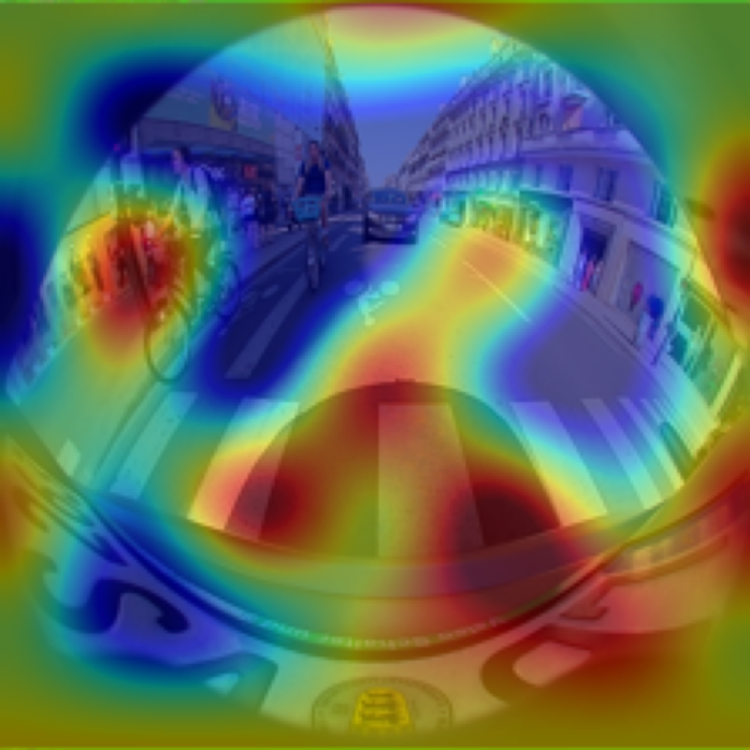}
& \includegraphics[width=0.075\textwidth]{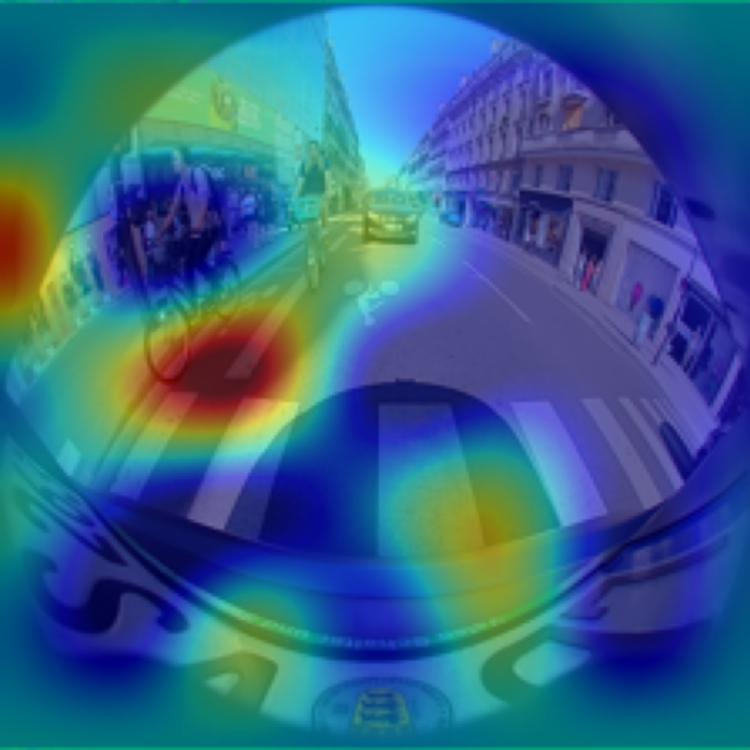}
& \includegraphics[width=0.075\textwidth]{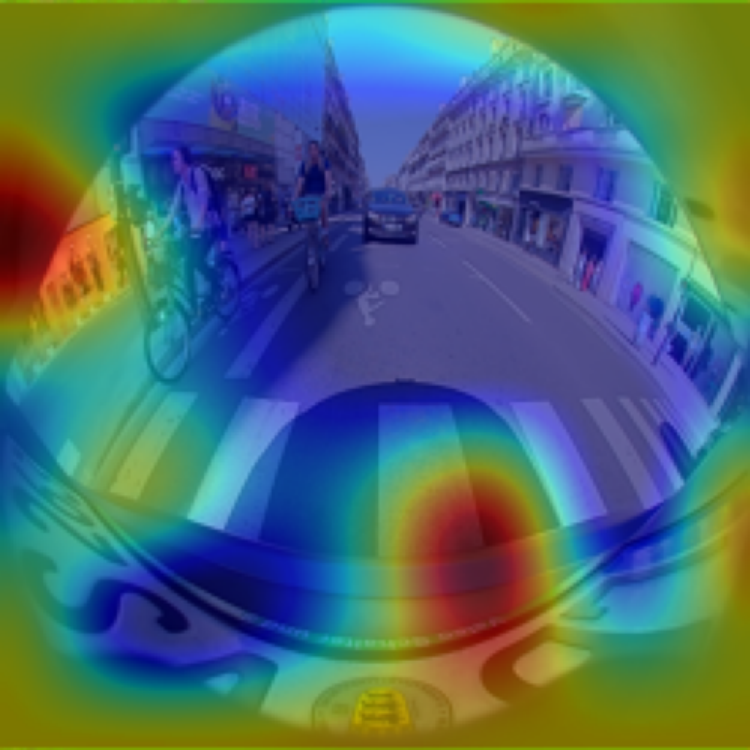} \\
\cmidrule(lr){2-6}
& \textbf{I-FAA}
& \includegraphics[width=0.075\textwidth]{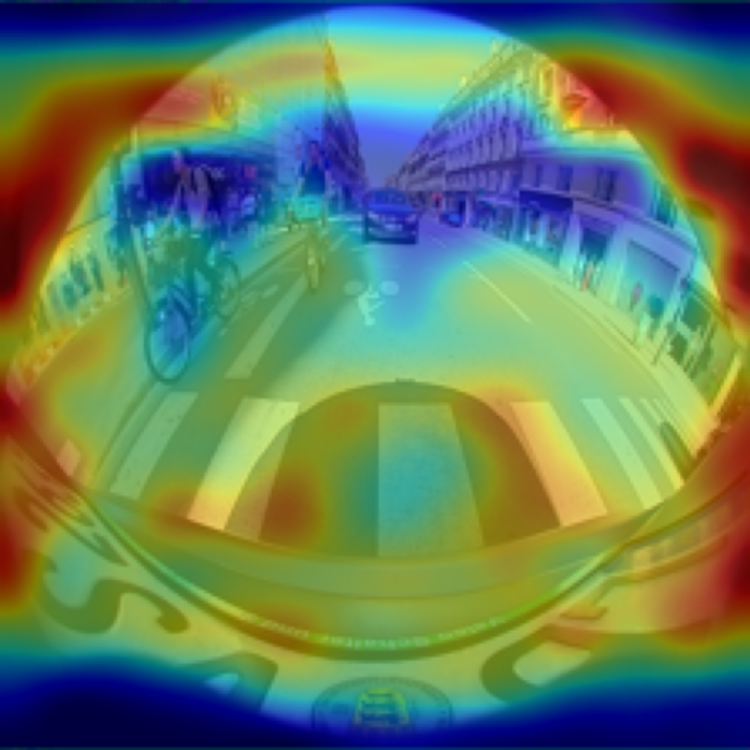}
& \includegraphics[width=0.075\textwidth]{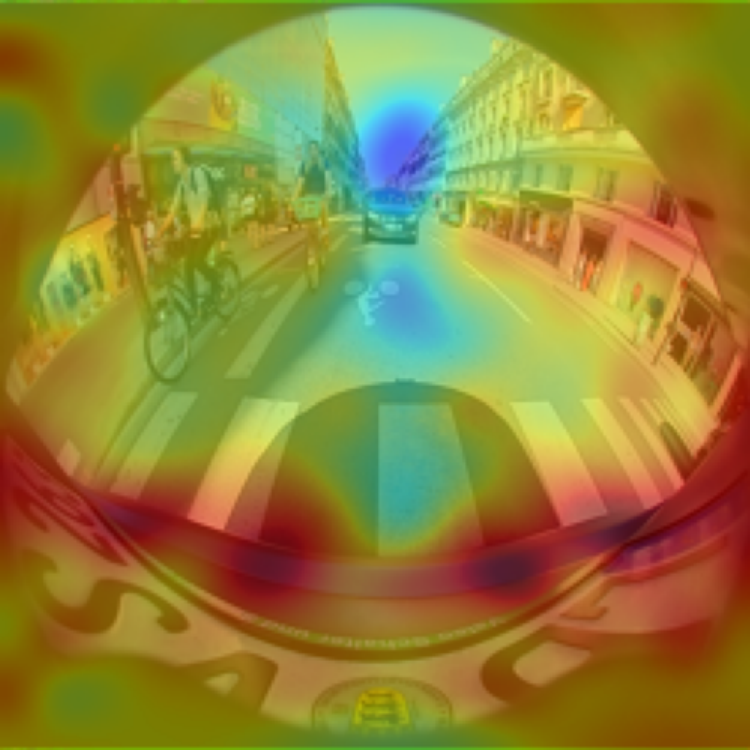}
& \includegraphics[width=0.075\textwidth]{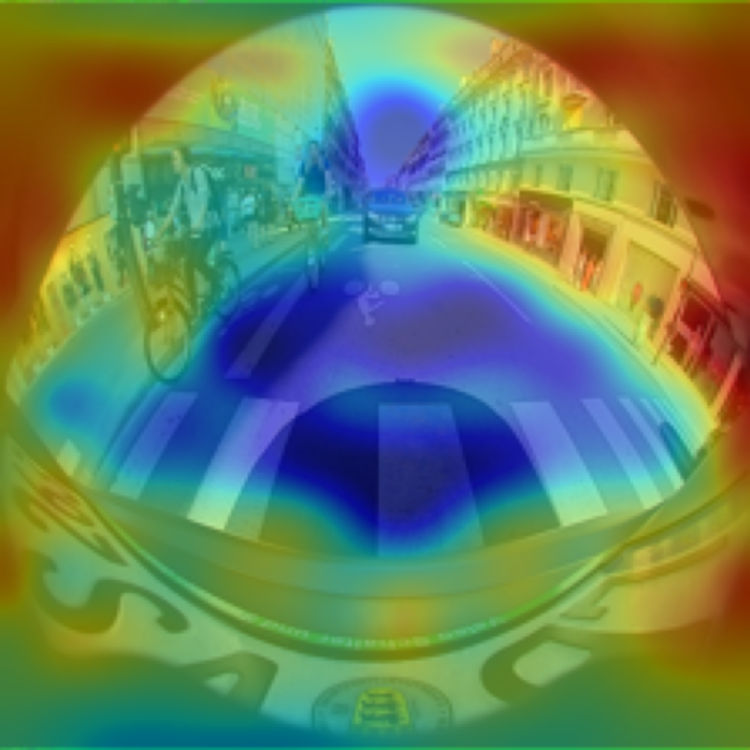}
& \includegraphics[width=0.075\textwidth]{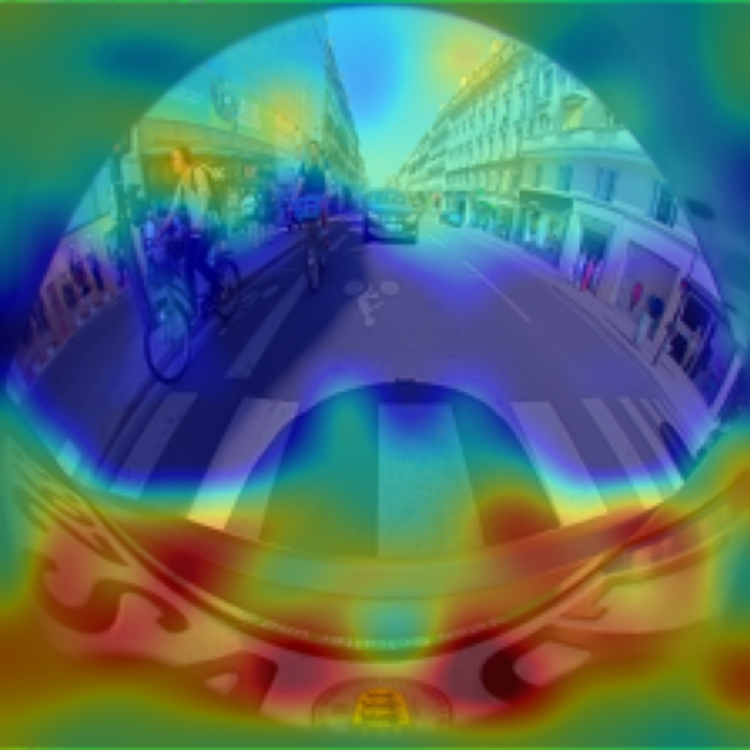} \\

\midrule

\multirow{2}{*}{\scalebox{1.0}{\begin{tabular}[c]{@{}c@{}}\includegraphics[width=0.090\textwidth]{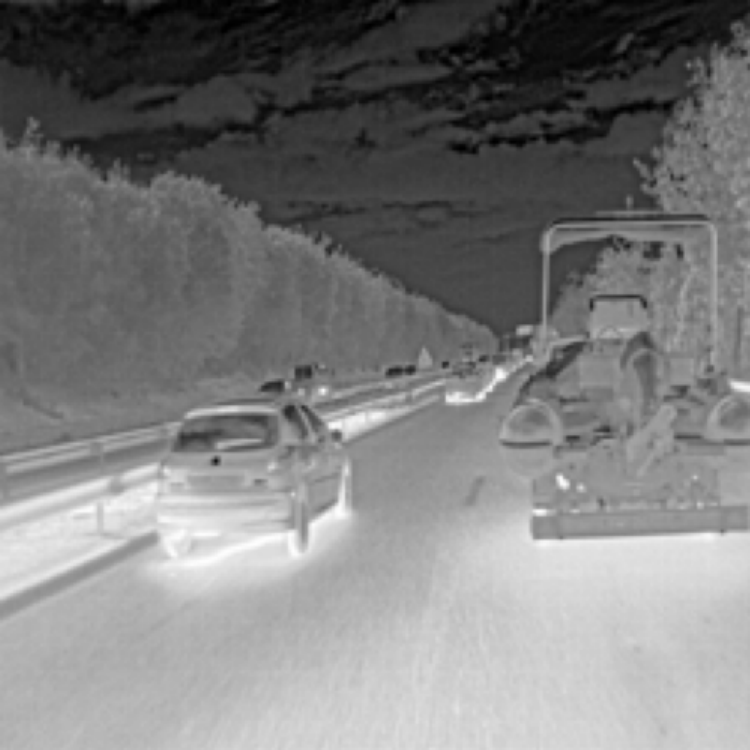}\\[-0.2ex]{\tiny normal/thermal/real/front}\end{tabular}}}
& \textbf{FAA}~\cite{godavarthy2026multi}
& \includegraphics[width=0.075\textwidth]{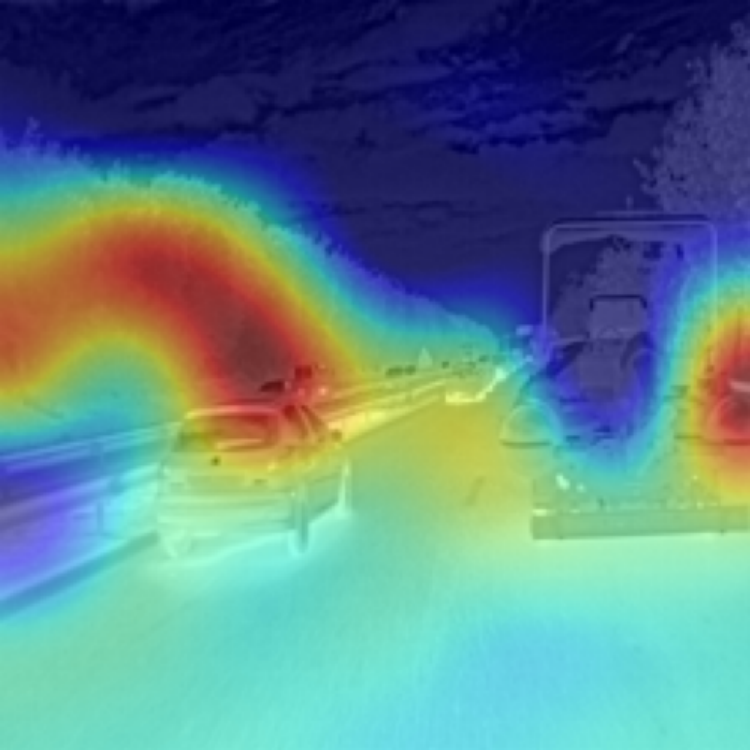}
& \includegraphics[width=0.075\textwidth]{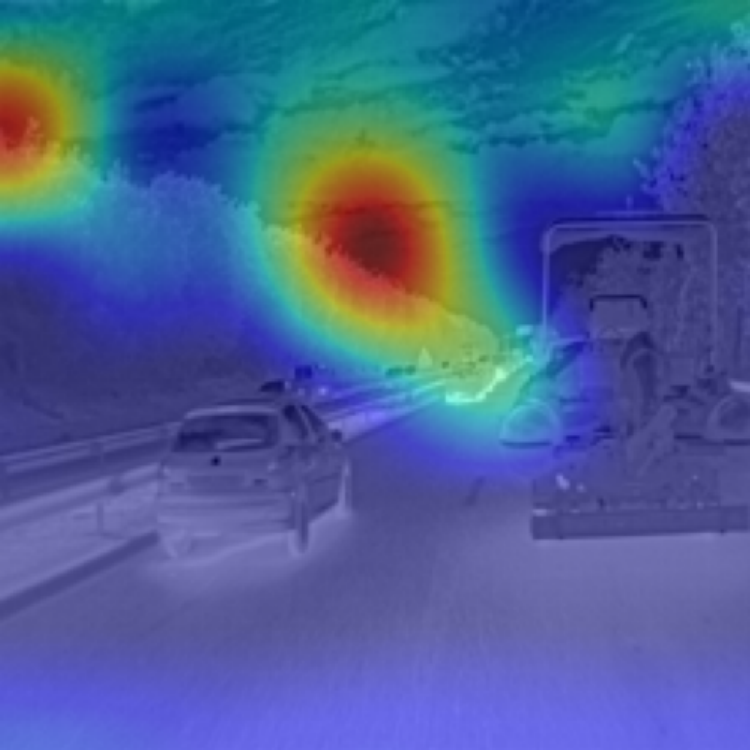}
& \includegraphics[width=0.075\textwidth]{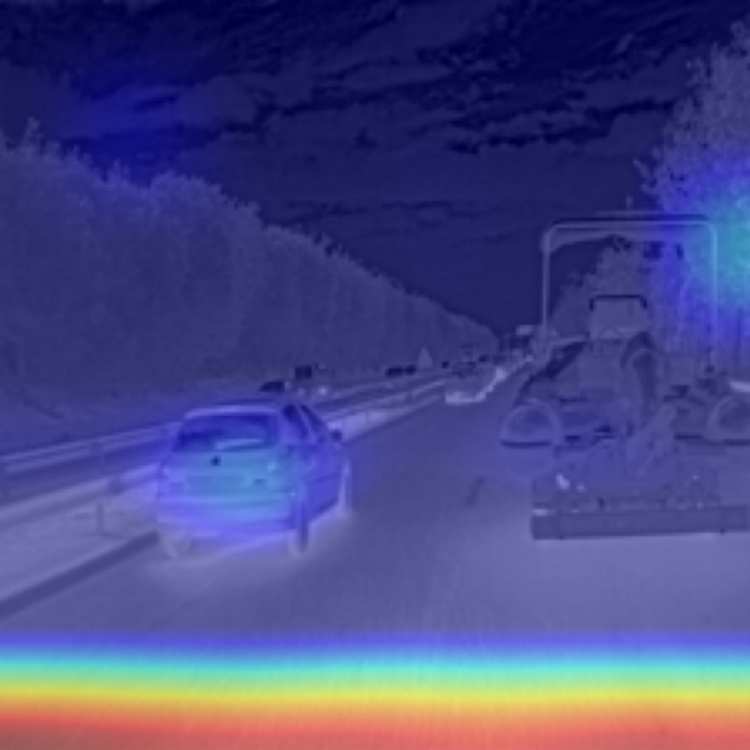}
& \includegraphics[width=0.075\textwidth]{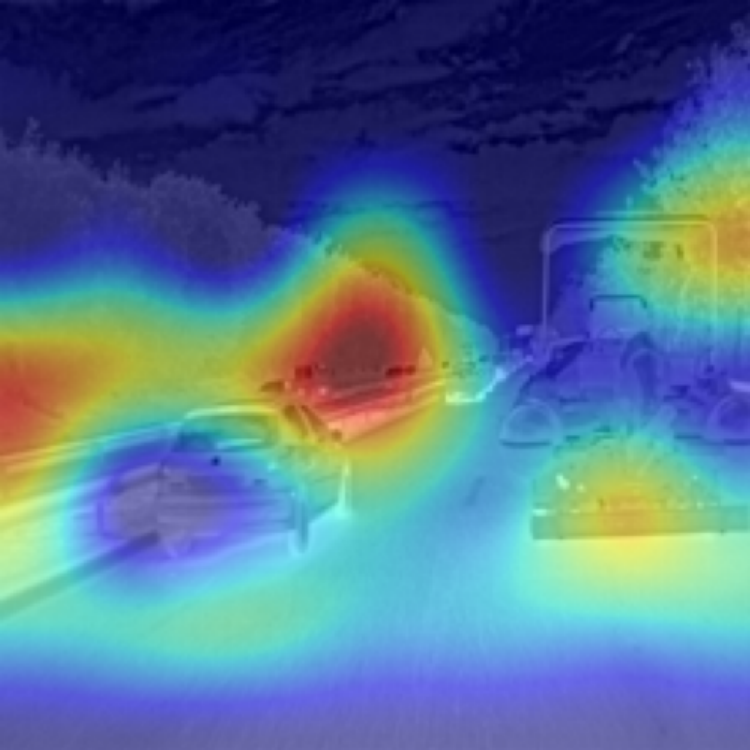} \\
\cmidrule(lr){2-6}
& \textbf{I-FAA}
& \includegraphics[width=0.075\textwidth]{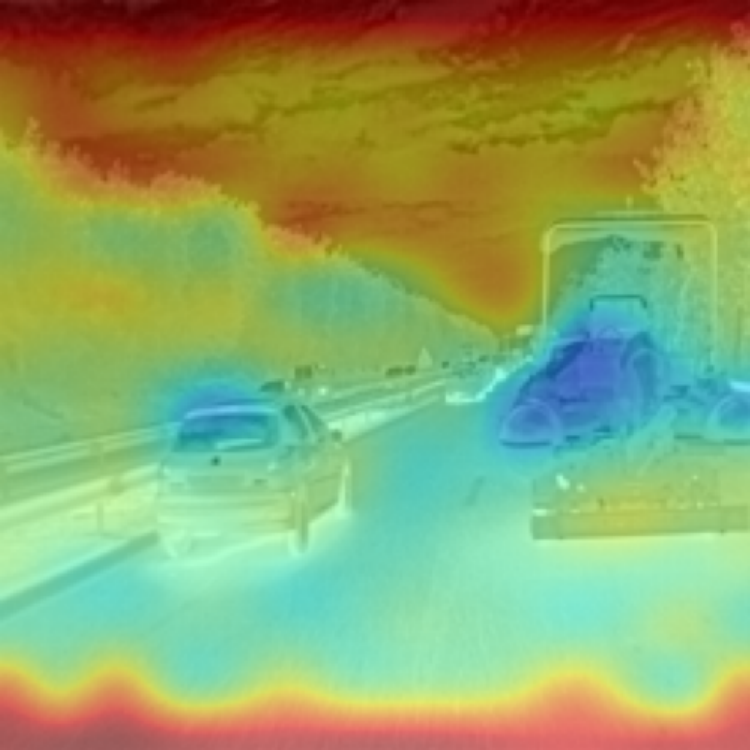}
& \includegraphics[width=0.075\textwidth]{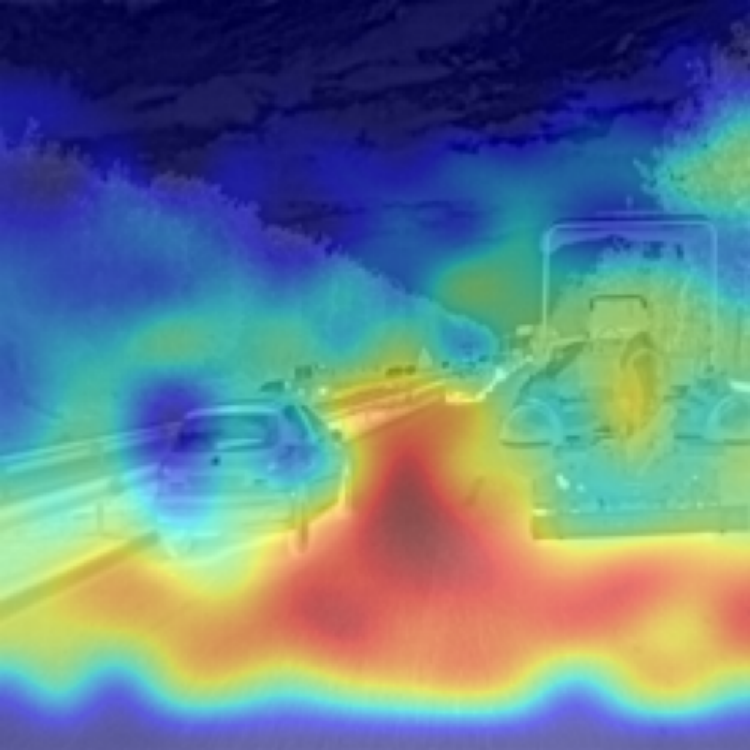}
& \includegraphics[width=0.075\textwidth]{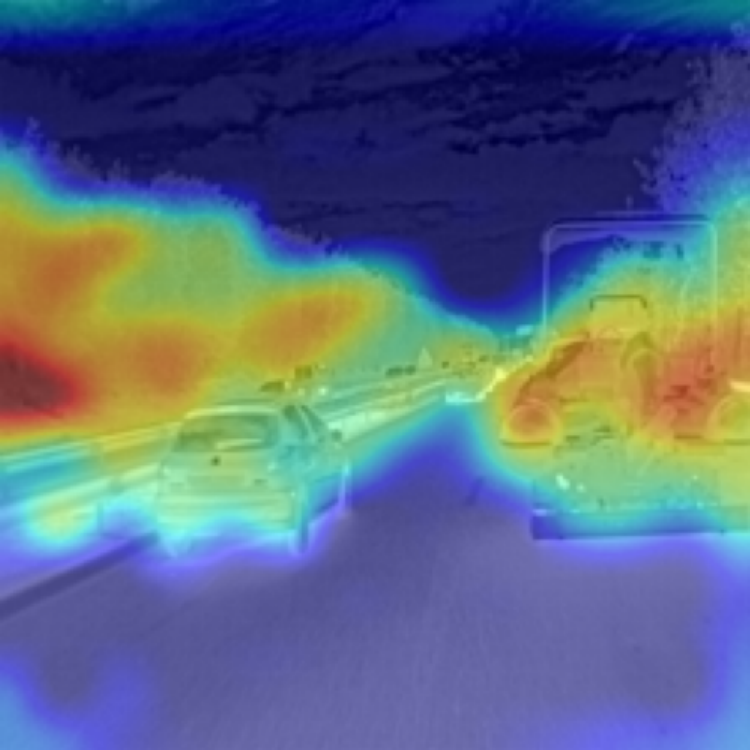}
& \includegraphics[width=0.075\textwidth]{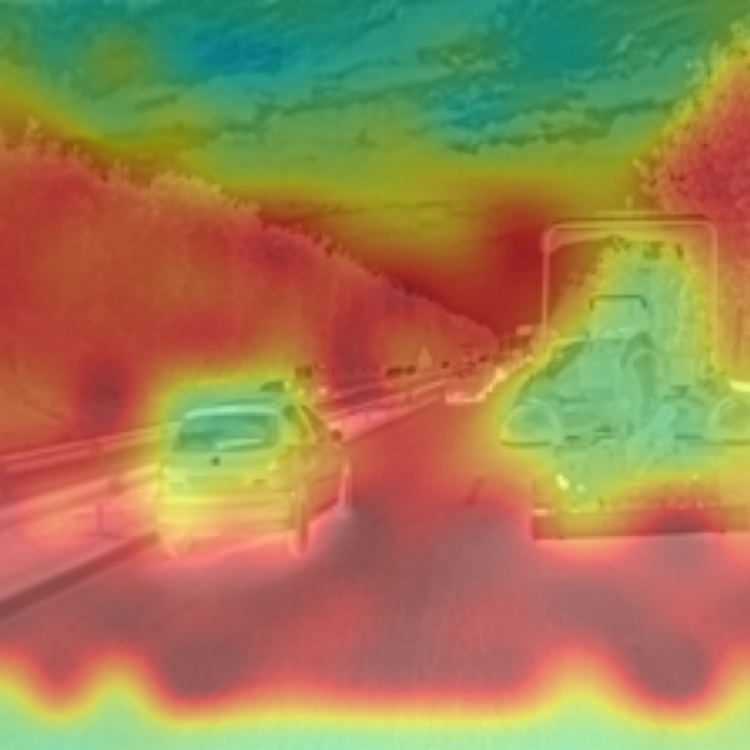} \\

\bottomrule
\end{tabular}
\end{table}

\end{document}